\PassOptionsToPackage{dvipsnames}{xcolor}
\documentclass[acmtog,nonacm]{acmart}

\setcopyright{none}
\makeatletter
\let\@authorsaddresses\@empty
\makeatother

\usepackage[ruled]{algorithm2e} %

\SetAlFnt{\small}
\SetAlCapFnt{\small}
\SetAlCapNameFnt{\small}
\SetAlCapHSkip{0pt}

\usepackage{multirow}
\usepackage{pifont}
\usepackage{xspace}
\usepackage{cleveref}
\usepackage{makecell}
\usepackage{subcaption}

\usepackage{natbib}
\usepackage{array}
\usepackage{enumitem}

\usepackage[breakable]{tcolorbox}
\newtcolorbox{promptbox}[1][]{%
  colback=gray!5,
  colframe=gray!60,
  fonttitle=\bfseries\small,
  title={#1},
  breakable,
  fontupper=\footnotesize,
  left=4pt, right=4pt, top=2pt, bottom=2pt,
  before skip=6pt, after skip=6pt,
}

\usepackage{changepage}

\usepackage{xcolor}

\definecolor{darkgreen}{HTML}{008000}
\definecolor{orange}{RGB}{255,165,0}

\newcommand{\ap}[1]{``#1''}

\newcommand{\lefttxt}[1]{%
  \raisebox{0.75cm}{%
    \parbox{0.125\linewidth}{\centering\emph{#1}}%
  }%
}

\makeatletter
\DeclareRobustCommand\onedot{\futurelet\@let@token\@onedot}
\def\@onedot{\ifx\@let@token.\else.\null\fi\xspace}
\def\eg{\emph{e.g}\onedot}

\makeatother

\begin{document}
\title{VideoSketcher: Sequential Sketch Generation Using Video Model Priors}

\author{Hui Ren}
\affiliation{
\institution{UIUC}
\country{USA}
}

\author{Yuval Alaluf}
\affiliation{
\institution{Runway}
\country{USA}
}

\author{Omer Bar Tal}
\affiliation{
\institution{Runway}
\country{USA}
}

\author{Alexander Schwing}
\affiliation{
\institution{UIUC}
\country{USA}
}

\author{Antonio Torralba}
\affiliation{
\institution{MIT}
\country{USA}
}

\author{Yael Vinker}
\affiliation{
\institution{MIT}
\country{USA}
}

\begin{teaserfigure}
\vspace{-0.45cm}
    \centering
    \includegraphics[width=0.98\linewidth]{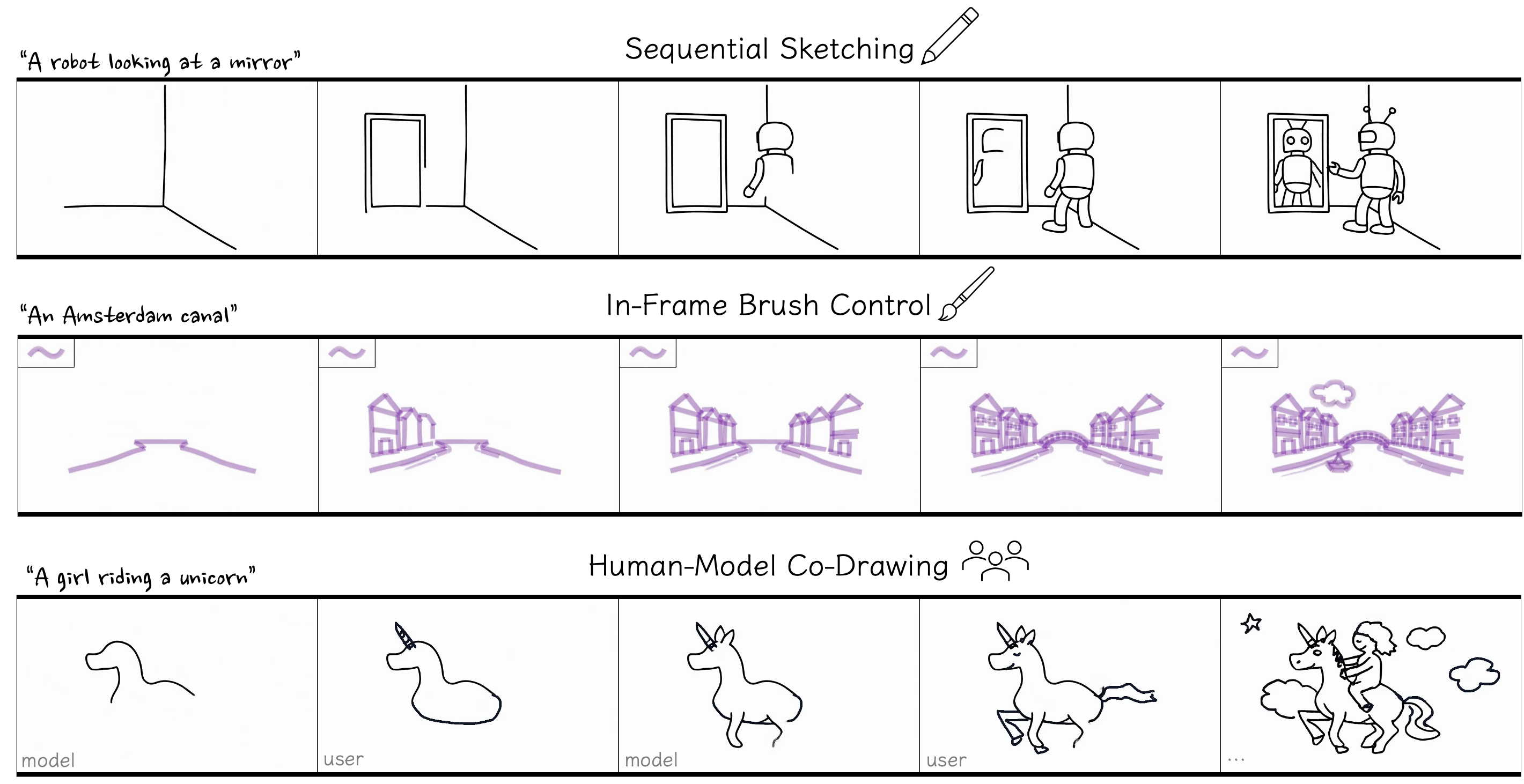}
    \\[-8pt]
    \caption{VideoSketcher enables sequential sketch generation in pixel space via video diffusion priors. Given a text prompt, our method generates a step-by-step drawing process that follows natural stroke order with high visual quality (top). Our approach further supports user-specified brush styles (middle) and real-time human-model co-drawing through an autoregressive framework (bottom).
    Please see the full videos in our project page: \textcolor{MidnightBlue}{\url{https://videosketcher.github.io/}} 
    }
    \vspace{0.15cm}
    \label{fig:teaser}
\end{teaserfigure}

\addtocontents{toc}{\protect\setcounter{tocdepth}{-10}}

\begin{abstract}
Sketching is inherently sequential: strokes are drawn progressively to explore and refine ideas. Yet most generative approaches treat sketches as static images, ignoring the temporal process underlying creative exploration. Modeling this sequential structure remains challenging: prior methods either rely on large-scale human-drawn datasets with limited diversity, or use large language models (LLMs) to produce drawing instructions, often at the cost of visual fidelity.
We present \emph{VideoSketcher}, a method for generating high-quality sketching processes by adapting pretrained text-to-video diffusion models to the sparse, continuous nature of sketch formation. Our key insight is that LLMs and video diffusion models offer complementary strengths: LLMs act as semantic planners that decompose concepts into step-by-step instructions, while video diffusion models serve as powerful ``renderers'' that translate them into temporally coherent sketch sequences.
We introduce a two-stage fine-tuning strategy that decouples temporal structure from visual appearance: stroke ordering is learned from synthetic shape compositions, while style is distilled from as few as seven hand-drawn examples. Despite minimal supervision, our method can generate diverse, high-quality sequential sketches that faithfully follow specified drawing orders. Our framework naturally extends to brush style control and autoregressive generation, supporting artistic applications.
\end{abstract}

\maketitle

\section{Introduction}\label{sec:introduction}
Sketches are a fundamental medium for exploring, communicating, and refining ideas~\cite{fan2023drawing,tversky2013visualizing}. Their expressive power lies not only in the final result, but also in the \emph{process} of drawing itself: through the gradual accumulation of strokes, creators externalize thoughts, explore alternatives, and iteratively refine emerging concepts~\cite{goldschmidt1992serial,tversky2003sketches}. 

Generative models that capture this sequential nature of sketching, rather than treating sketches as static images, could enable richer human-machine interaction, including visual brainstorming, real-time feedback, and collaborative prototyping through a natural visual medium.
Yet modeling the drawing \emph{process} remains an open challenge: the goal is not simply to produce strokes incrementally, but to do so in a \emph{meaningful order}, building structure through semantically coherent progressions.

Prior methods learn stroke ordering directly from large-scale human-drawn datasets~\cite{SketchRNN,Bhunia2020PixelorAC,ge2020creative,Deformable_Stroke,bhunia2021doodleformer,das2020beziersketch}, often requiring millions of sketches~\cite{quickDrawData}. While effective within their training domain, these approaches are fundamentally limited to a small set of predefined categories and the amateur style typical of crowd-sourced data.

More recently, SketchAgent~\cite{SketchAgent_Vinker2025} showed that Large Language Models (LLMs) can be used to overcome the data bottleneck by leveraging their rich knowledge of diverse concepts and their compositional structure. Specifically, they prompt an LLM \cite{claude} to directly output stroke-by-stroke drawing instructions as SVG-like code, enabling generalization far beyond predefined categories. However, while LLMs excel at deciding \emph{what} to draw and in what \emph{order}, they struggle with \emph{how} to draw it — their semantic prior is far stronger than their visual reasoning. As a result, SketchAgent's outputs exhibit limited visual fidelity and detail.

In this work, we introduce \emph{VideoSketcher}, a data-effecient method for generating high-fidelity sequential sketches of diverse concepts by leveraging the strong visual and temporal priors of video diffusion models.
Our key insight is that video diffusion models and LLMs offer complementary strengths: LLMs provide semantic understanding that enables meaningful planning, but are limited in visual fidelity, whereas video diffusion models excel at high-quality visual synthesis but lack an intrinsic notion of drawing order. 
VideoSketcher unifies these strengths, using an LLM as the planner and a video diffusion model as the ``renderer''.

We represent a sketch sequence as a short video in pixel space, in which strokes are progressively drawn on a blank canvas. Despite the apparent gap between photorealistic video and abstract hand-drawn sketches, we show that video diffusion models can be effectively adapted to sketch generation using only a \emph{handful} of carefully constructed examples.

A key challenge is teaching the model not only what sketches should look like, but also how they should unfold over time, following the drawing order specified by an LLM. To address this, we introduce a two-stage fine-tuning strategy that decouples temporal structure from visual appearance. In the first stage, stroke ordering is learned from synthetic compositions of geometric primitives inspired by Gestalt principles~\cite{koffka1999principles}. Each composition is rendered with multiple drawing orders, teaching the model both a “visual vocabulary” of primitives and the ability to follow text-specified sequences. In the second stage, we adapt the model to the visual style of hand-drawn sketches using as few as seven examples. Despite this minimal supervision, our method generates high-quality sequential sketches across diverse concepts, faithfully following specified orderings while maintaining rich visual detail and temporal coherence (see~\Cref{fig:teaser}).

Building on our data-efficient, model-agnostic fine-tuning strategy, we introduce a mechanism for brush-level control in generated videos. We embed a target brush directly within each frame as a visual cue (\Cref{fig:teaser}, middle), and fine-tune an image-to-video model such that the first frame serves as a conditioning signal, enabling the use of new brushes at inference time. The model learns to infer stroke appearance from this visual cue, generalizing to unseen brushes while capturing fine-grained stylistic properties. This extends capabilities typically associated with parametric representations to a purely pixel-based setting.

We further demonstrate interactive co-drawing by applying our training strategy to an autoregressive video model. As these models are less mature and require more data, we leverage our adapted diffusion model to synthesize additional training examples, effectively bootstrapping their performance. 
Together, these results suggest that pretrained video diffusion models offer a powerful and flexible prior for modeling drawing processes, providing a new perspective on sequential sketch generation that does not rely on large-scale sketch datasets or explicit parametric stroke representations.

\begin{figure*}
    \centering
    \includegraphics[width=1\linewidth]{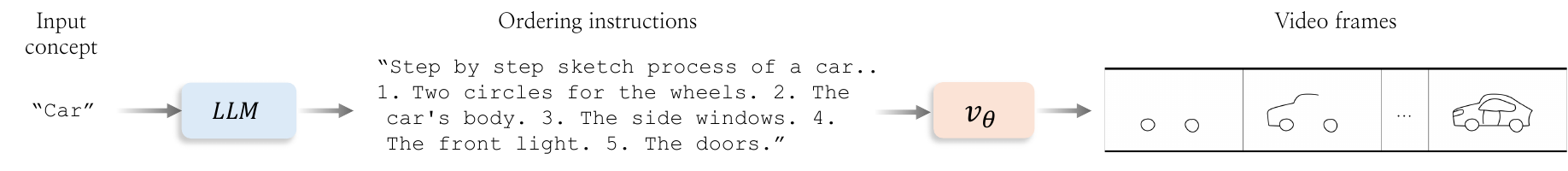}
    \vspace{-0.6cm}
    \caption{\textbf{Complementary strengths of LLMs and video diffusion models.} Given a high-level concept to be drawn, the LLM acts as a planner, decomposing it into an ordered sequence of drawing steps. The video model, trained to generate sketch-like outputs, acts as a soft "renderer" — translating the given textual instructions into a sketch-progression video. By bypassing explicit parametric representations, we leverage the rich visual prior of video models: a high-level cue like ``wheels'' is sufficient for the model to produce faithful visual strokes, without requiring large-scale parametric data.}
    \label{fig:high-level-pipe}
\end{figure*}

\vspace{-0.2cm}
\section{Related Work}\label{sec:prev_work}

\paragraph{Sequential sketch generation.}
A common approach to sequential sketch generation is to represent sketches as parametric stroke sequences and train generative models on large collections of human-drawn sketches~\cite{song2018learning,SketchRNN,qi2021sketchlattice,Deformable_Stroke,liu2021neural,bhunia2021doodleformer}.
SketchRNN~\cite{SketchRNN} pioneered this direction by introducing the QuickDraw dataset~\shortcite{quickDrawData}, the largest collection of parametric human-drawn sequential sketches. 
However, these methods are inherently constrained to a small set of predefined categories (e.g., 340 in QuickDraw), and to the amateur style typical to croud-sourced datasets.

The rise of large vision-language models (VLMs) \cite{Radfordclip,rombach2022highresolution,saharia2022photorealistic} has enabled a fundamentally different approach to sketch generation \cite{vinker2022clipasso,Vinker_2023_ICCV,NEURIPS2023_DiffSketcher,CLIPDraw, arar2025swiftsketch, gal2023breathing}. Rather than relying on human-drawn sketch datasets, these methods generate sketches through direct paranetric optimization guided by pretrained VLMs. However, they do not model drawing as a \emph{process}, resulting in unstructured sets of strokes.

More recently, SketchAgent~\cite{SketchAgent_Vinker2025} proposed a sequential sketching approach by leveraging large language models (LLMs) \cite{claude}. Recognizing that LLMs encode strong priors over objects, their parts, and their relative importance, it generates sketches as sequences of textual drawing commands rendered on a canvas. This enables versatile sketch generation beyond predefined categories.
However, while LLMs are strong semantic planners, they lack strong visual priors, leading to overly simple, child-like sketches with limited visual fidelity.
To overcome this bottleneck, we leverage video diffusion models as visual renderers of LLM commands, enabling substantially richer visual detail while preserving text-specified stroke ordering.

Recent work \cite{paintsalter,paintsundo} apply video models to painting reconstruction -- recovering or reversing the creation process of a \emph{given} painting. To achieve this, they rely on $20K$ Procreate time-lapse recordings for training, producing relatively coarse, frame-level progressions.
In contrast, we pursue a fundamentally different goal: generating new sketches from given textual descriptions, with local stroke-by-stroke progression.
Crucially, we show that this can be achieved using only a \emph{handful} of training examples.

Finally, a related line of work uses reinforcement learning to train painting agents that produce sequential paintings for various tasks~\cite{ganin2018synthesizing,mellor2019unsupervised,mihai2021learning,zhou2018learning}. However, these methods are not designed to model semantically meaningful stroke ordering and are typically restricted to narrow domains, such as faces or a small set of predefined objects.

\vspace{-0.2cm}
\paragraph{Video priors and interactive video generation}
Recent advances in video generation show that models trained on large-scale video data capture strong temporal structure and can serve as effective priors for new visual tasks~\cite{Wan2.1,HunyuanVideo,LTX-Video,openai2025sora2,veo3technicalreport,video-zero-shot-reasoners,gal2023breathing}. We build on this insight by adapting pretrained video generation models to learn sketching behavior in a few-shot setting.

Standard video diffusion models generate entire video clips jointly, making inference computationally expensive and limiting interactivity. Recent work, therefore, explores causal or autoregressive video models that use temporally causal attention to generate frames sequentially~\cite{Causvid,SelfForcing,LongLive,chen2024diffusion,yin2024improved,yin2024one,cui2025self,yin2025slow}. While these models may trade some visual fidelity for efficiency, they are better suited for human-in-the-loop applications. We adopt such an autoregressive model for sequential sketch generation, enabling collaborative sketching.

\setlength{\abovedisplayskip}{4pt}
\setlength{\belowdisplayskip}{4pt}
\setlength{\abovedisplayshortskip}{2pt}
\setlength{\belowdisplayshortskip}{2pt}

\section{Preliminaries}\label{sec:preliminaries}
Text-to-video diffusion models generate video frames conditioned on a given text prompt describing the desired scene. Standard models generate all frames jointly via a gradual denoising process.

Given a video $V \in \mathbb{R}^{K \times H \times W \times 3}$, a spatio-temporal variational autoencoder (VAE) encodes it into a latent representation $x_0$. 
Recent video diffusion models adopt rectified flow matching~\cite{liu2022flow,lipman2022flow}, which defines a linear interpolation path between clean data $x_0$ and Gaussian noise $\epsilon \sim \mathcal{N}(0, I)$:
\begin{equation}
x_t = (1 - t)\, x_0 + t\, \epsilon, \quad t \in [0, 1],
\end{equation}
where $t = 0$ corresponds to clean data and $t = 1$ corresponds to pure noise. 
The diffusion model, $v_\theta$ (commonly implemented as a Diffusion Transformer~\shortcite{peebles2023scalable}) 
is trained to predict the corresponding velocity field $v = \epsilon - x_0$ by minimizing the following objective:
\begin{equation}
\mathcal{L} = \mathbb{E}_{x_0, \epsilon, t} \left[ \left\| v_\theta(x_t, t, y) - (\epsilon - x_0) \right\|^2 \right],
\label{eq:loss}
\end{equation}
where $y$ denotes the text conditioning embedding.

At inference time, generation begins from noise $x_T \sim \mathcal{N}(0, I)$. Samples are obtained by integrating the differential equation
$
\frac{dx_t}{dt} = v_\theta(x_t, t, y)
$
from $t = 1$ to $t = 0$, yielding a clean latent representation $x_0$, which is then decoded into a video via the VAE decoder $\mathcal{D}(x_0)$.
In this work, we build on Wan~2.1~\shortcite{Wan2.1}, a pretrained open-source text-to-video diffusion model.

\section{Method}\label{sec:method}
Our goal is to generate high-quality sketching processes of given textual concepts, generalizing to diverse, unseen inputs. To accomplish this, rather than relying on large-scale human-drawn sketch datasets, we leverage the strong visual and temporal priors of a pretrained video diffusion model $v_\theta$, along with the seamntic priors of LLMs. The video model serves as a powerful ``renderer'' of sketch progressions, while the LLM serves as the backbone ``planner'', determining the semantic parts composing the concept and their drawing order (illustrated in~\Cref{fig:high-level-pipe}). 
The key challenge lies in adapting the video model, trained on photorealistic content, to the sparse, structured appearance of sketches, while conditioning it to faithfully follow the ordering specified by the LLM.

\subsection{Data Representation}
\paragraph{Sketch Representation.} We represent a sketch sequence as a short video $V \in \mathbb{R}^{K \times H \times W \times 3}$ in pixel space, depicting black strokes progressively drawn on a blank canvas.
Our goal is to model two levels of temporal structure: the global ordering of strokes, and the local, continuous progression \emph{within} each stroke.
To achieve this, we first capture the sketching process as SVG files~\shortcite{w3c_svg_2011}, which preserve both the sequence of strokes and the path of each individual stroke (unlike, e.g., Procreate time-lapse recordings, which do not retain per-stroke trajectory information), and then parse and render them into pixel videos by animating each stroke gradually along its recorded path at a consistent visual pace.

\paragraph{Text Prompt Construction.} Each training video is paired with a text description specifying the subject and the order in which its parts should be drawn. We use a structured format consisting of a brief subject description followed by a numbered sequence of drawing steps. For example: \textit{``Step-by-step sketch process of a desk lamp, following this drawing order: 1. Lampshade – a cone-shaped top part that directs the light downward. 2. Light bulb \dots''}. At training time, we manually write these descriptions for each sketch. At inference, we use an LLM~\cite{OpenAI2026GPT5} to automatically produce them for new concepts, leveraging its strength in semantic decomposition and planning (details in supplementary). 

\begin{figure}
    \centering
    \includegraphics[width=1\linewidth]{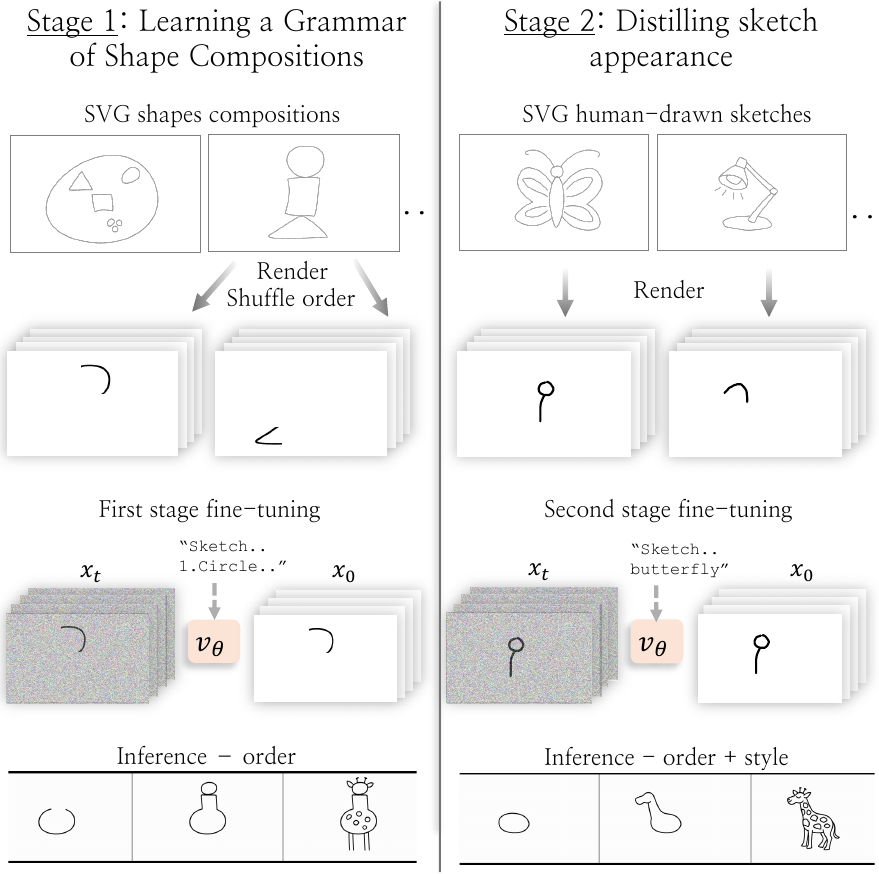} 
    \vspace{-0.6cm}\caption{\textbf{Two-stage fine-tuning scheme.} Left: synthetic sketches composed of simple geometric primitives teach drawing ``grammar'' and stroke ordering, independently of appearance. Right: a small set of human-drawn sketches of real-world objects adapts the model to a target visual style.}
    \vspace{-0.4cm}
    \label{fig:shapes}
\end{figure}

\subsection{Two-Stage Finetuning}
\label{sec:twostage}
While video diffusion models encode strong visual priors, they lack an intrinsic notion of meaningful drawing order. As we demonstrate, naive fine-tuning on sketch videos can produce sketch-like appearance but often yields arbitrary or inconsistent stroke sequences. The challenge is to achieve both high visual fidelity and explicit temporal control over the sketching process. We address this with a two-stage fine-tuning strategy that disentangles these two objectives:

\paragraph{(1) Learning a ``Grammar'' of Shape Compositions}

Before training on real-world sketches, we first teach the model a basic drawing ``grammar'': simple shapes, their spatial relationships, and how to follow ordering instructions. This design is inspired by how people learn to draw, starting with simple shapes and compositional rules before progressing to more complex subjects~\cite{kellog1969analyzing,edwards1989drawing}.
This process is illustrated in~\Cref{fig:shapes}, left.
We construct a small dataset of simple geometric primitives, including circles, ellipses, triangles, rectangles, polygons, curves, and lines.
These primitives are arranged in diverse spatial configurations inspired by Gestalt principles \cite{koffka1999principles}, such as containment, adjacency, overlap, and grouping, reflecting the compositional building blocks underlying more complex sketches. 
For each shapes configuration, we vary the order in which the shapes are drawn, along with the respective prompt, producing three distinct temporal variations, yielding 15 training videos in total. 

We then fine-tune the video diffusion model following standard practice (as described in \Cref{sec:preliminaries}). The model is trained to denoise the sketching videos at different noise levels $t$, conditioned on our step-by-step textual drawing instructions.
Because the shapes are visually simple and semantically neutral, this dataset minimizes appearance-related variability and encourages the model to focus on learning temporal stroke ordering rather than object-specific visual details.

\paragraph{(2) Distilling Sketch Appearance}
While training on primitive shape compositions is effective for learning stroke ordering, models trained only on such data tend to compose drawings directly from these primitives, resulting in sketches that lack the desired visual appearance (see \Cref{fig:shapes}, bottom left). To bridge this gap, we perform a second fine-tuning stage using a small set of seven real-world sketches drawn by an artist: a lamp, car, chair, tree, cup, butterfly, and flower. Each sketch preserves the natural drawing order as captured from the artist's original SVG recording.
This stage adapts the model to the target visual aesthetic and level of abstraction that we aim to produce at inference time. Because the model has already learned to follow explicit ordering instructions during the synthetic pretraining stage, this fine-tuning primarily transfers appearance information.

\begin{figure}[t]
    \centering
    \vspace{-0.1cm}
    \includegraphics[width=1\linewidth]{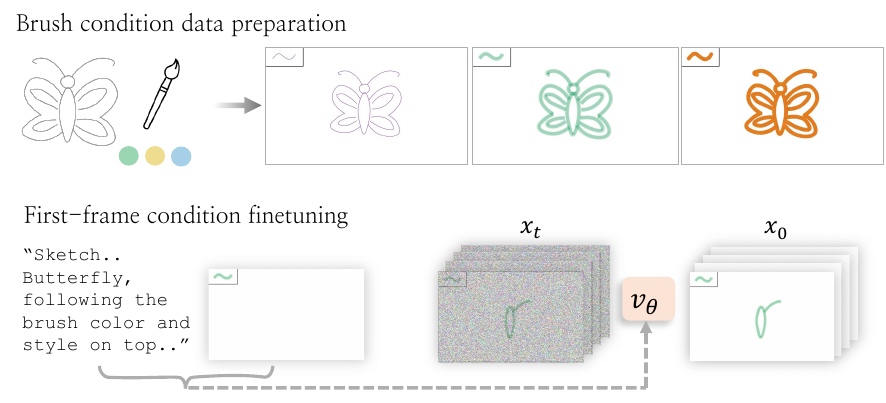}\vspace{-0.2cm}
    \caption{ \textbf{Brush conditioning.} Top: Data preparation — each SVG sketch is rendered with diverse brush styles and colors. Bottom: We fine-tune an Image-to-Video model where the first frame, containing a brush exemplar in the corner, conditions the generated stroke appearance. }\vspace{-0.4cm}
    \label{fig:brush_control_pipe}
\end{figure}

\begin{figure*}[t]
\centering
\setlength{\tabcolsep}{0pt}
\renewcommand{\arraystretch}{0.2}
\small

\begin{tabular}{c c}
  \parbox{0.95\linewidth}{\scriptsize\emph{``Step by step sketch process of an Amsterdam canal, following this drawing order: 1. Canal water shape running through the scene. 2. Narrow buildings along the canal edge. 3. A bridge arch crossing the water. 4. Bicycles along the railing. 5. A small boat floating in the canal.''}} \\[6pt]
  \includegraphics[width=0.95\linewidth]{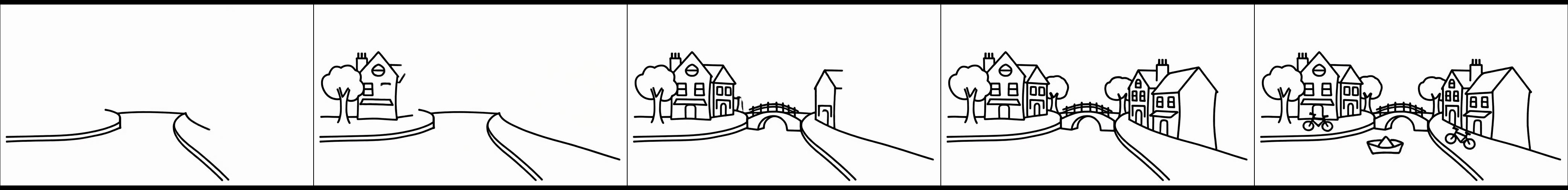} \\[6pt]

  \parbox{0.95\linewidth}{\scriptsize\emph{``Step by step sketch process of a Rome street with a cat, following this drawing order: 1. Street ground plane and vanishing point. 2. Uneven stone building facades. 3. Arched windows, shutters, and doorways. 4. Steps and ledges near the foreground. 5. A relaxed cat sitting on the steps.''}} \\[6pt]
  \includegraphics[width=0.95\linewidth]{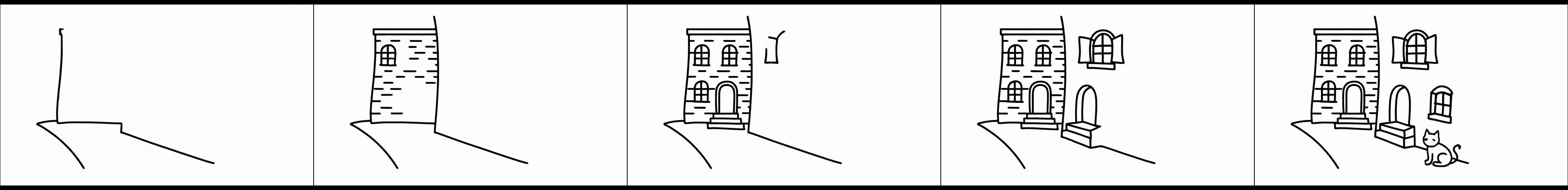} \\[6pt]

  \parbox{0.95\linewidth}{\scriptsize\emph{``Step by step sketch process of a Tokyo alley with lanterns, following this drawing order: 1. Narrow alley corridor with strong perspective. 2. Tall vertical building walls on each side. 3. Hanging rectangular signs and round lanterns at varied heights. 4. Shop doors and window frames. 5. Walking pedestrians. 6. Cables and glowing light accents.''}} \\[6pt]
  \includegraphics[width=0.95\linewidth]{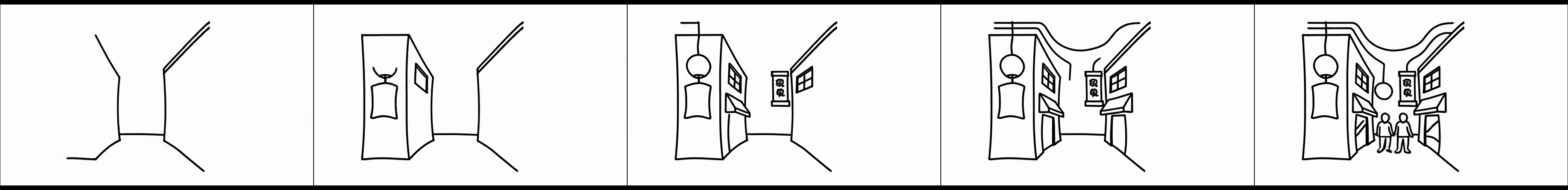} \\[6pt]

\end{tabular}

\vspace{-0.5cm}
\caption{\textbf{Qualitative results.} Results generated with our fine-tuned video model. Full video results are provided in the supplementary materials.}
\vspace{-0.2cm}
\label{fig:our_results}
\end{figure*}

\vspace{-0.2cm}
\subsection{Brush Conditioning}
A key advantage of our approach is that it is model-agnostic and can generalize across differnt video model architectures. We exploit this to introduce brush style control, enabled by adapting an Image-to-Video variant~\cite{Wan2.1}, which naturally conditions generation on a first frame. We repurpose this conditioning mechanism for brush style control: the first frame contains a blank canvas with a small brush exemplar in the corner, encoding the desired brush style and color. Crucially, we embed this exemplar across all frames, providing a consistent visual anchor that enables the model to reliably associate the cue with the generated stroke appearance throughout the sequence. The text prompt is augmented with ``follow the brush style and color on the top left'', reinforcing the visual cue. To train this variant, we render the same SVG sequences from~\Cref{sec:twostage} using six brush types and eight colors (demonstrated in \Cref{fig:brush_control_pipe}, top). Because the underlying sketch content remains fixed, the model learns to disentangle brush appearance from stroke structure — enabling generalization to unseen brush styles at inference time.

\subsection{Implementation details.}
\label{sec:implementaiton}
We build on the pretrained Wan2.1~14B~\shortcite{Wan2.1} model and fine-tune using LoRA adapters (rank 32) applied to the attention and the first two feed-forward layers. Training is conducted on 7 NVIDIA A100 GPUs with a learning rate of $1\mathrm{e}{-4}$. Stage~1 (synthetic shapes) runs for 700 epochs on 15 videos; Stage~2 (human sketches) runs for an additional 700 epochs on 7 videos. All training and inference use a resolution of $480 \times 832$ with 81 frames. Total training time is approximately 22 hours. At inference, we use 50 denoising steps, which takes $\sim16$ minutes per video on a single A100 GPU. We will release all code, data, and model weights upon publication. Additional training details are provided in the supplementary.

\section{Results}
We evaluate our method qualitatively and quantitatively across a range of sketching tasks.
Additional evaluations and full videos are provided in the supplementary.

\subsection{Text-Conditioned Sequential Generation}\label{sec:text_generation}

\paragraph{Qualitative.}
\Cref{fig:teaser,fig:our_results,fig:additional_our_results} demonstrate our method's ability to generate sequential sketches of diverse concepts, ranging from single objects to complex multi-element scenes requiring broad world knowledge — such as architectural styles in Amsterdam or Rome, and rendering diverse scene elements from high-level textual instructions (e.g., water canal, houses, bridge, steps, signs). Notably, this is achieved despite training on only 7 real sketches, leveraging the rich visual priors of video models combined with the strong semantic planning capabilities of LLMs. The generated sketches exhibit high visual fidelity, with clean, smooth progressions that appear parametric-like while following a natural drawing order. An additional advantage of the pixel-based representation is that strokes can be naturally erased and modified throughout the drawing (e.g., the steps in ``Rome street'', \Cref{fig:our_results} second row, are drawn in front of the already-rendered wall).
\Cref{fig:ordering_diversity} demonstrates the effect of altering ordering instructions for the same given concept — varying the text produces distinct sketching trajectories, demonstrating control over stroke ordering. Further diversity can be achieved by varying the initial noise or modifying the specified action, as shown in~\Cref{fig:additional_results_diversity} and in the supplementary.

\paragraph{Comparisons}
We compare our method against several baselines: naive prompting of Wan2.1~\shortcite{Wan2.1} (our backbone video model), PaintsUndo~\shortcite{paintsundo}, and SketchAgent~\shortcite{SketchAgent_Vinker2025}. We additionally include human-drawn sketches from the QuickDraw dataset~\shortcite{quickDrawData} as a reference. Following the evaluation protocol of SketchAgent, we randomly sample 50 categories from the QuickDraw dataset and generate two sketches per category using different random seeds, yielding 100 sketches in total. The same number of sketches is generated for all baselines, following their respective best practices. Since PaintsUndo requires a final frame as input, we provide final frames generated by FLUX.2~\shortcite{flux-2-2025}. Representative results with intermediate frames are shown in~\Cref{fig:qual_comparison} and in the supplamentary.

\begin{figure}[h]
    \centering
    \includegraphics[
        width=1\linewidth,
        trim=0pt 0pt 0pt 0.8cm,
        clip
    ]{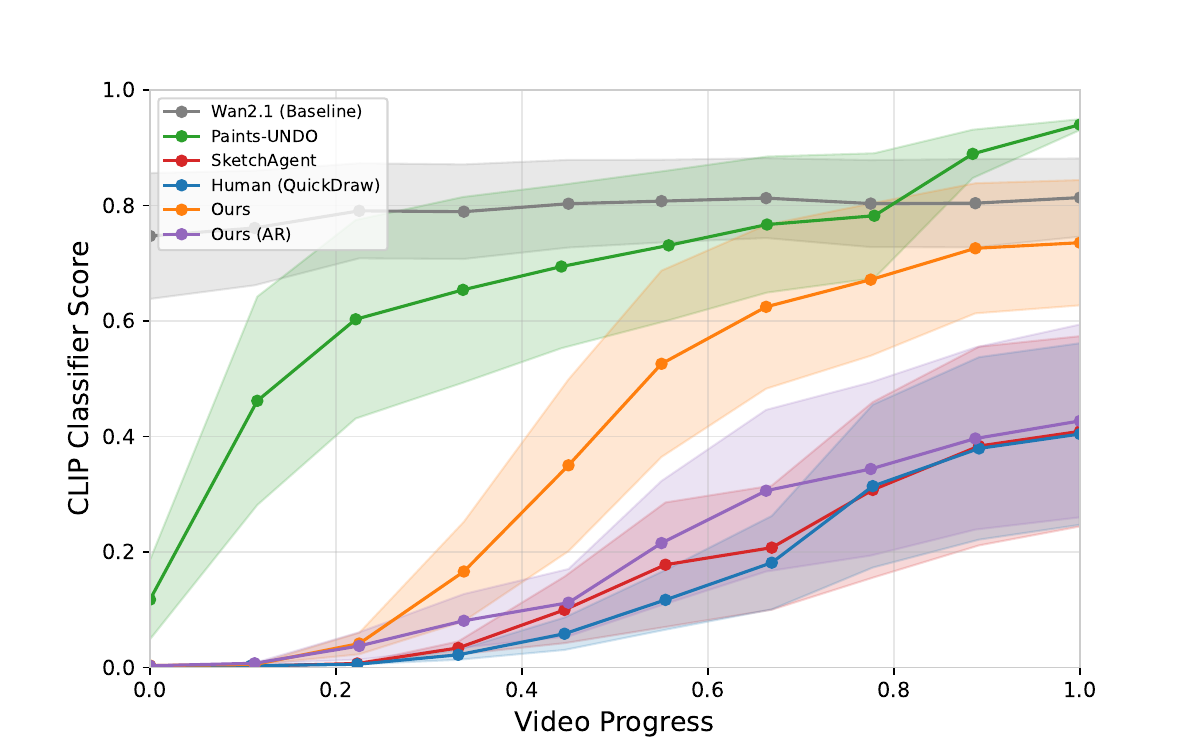} \\[-0.45cm]
    \caption{\textbf{CLIP-based recognition over the sketching process.} Our method gradually increases semantic recognizability as the sketch progresses, in contrast to baselines that either collapse temporal progression or attain lower recognition. Shaded areas indicate variance across samples.}
    \label{fig:clip_score_plot}
\end{figure}

Following common practice \shortcite{vinker2022clipasso,Vinker_2023_ICCV,SketchAgent_Vinker2025,DiffSketcher}, we use CLIP ViT-B/32 zero-shot classification to assess concept faithfulness, measuring both final-frame recognition (\Cref{tab:CLIPscore}) and recognition as a function of video progress (\Cref{fig:clip_score_plot}). 
As shown in~\Cref{tab:CLIPscore}, our method achieves 82\% Top-1 accuracy, substantially outperforming SketchAgent (48\%), whose outputs tend toward overly simplistic sketches that are often difficult to recognize (\Cref{fig:qual_comparison}, row 3) — comparable to the non-professional human sketches in QuickDraw (row 4). These results indicate that our fine-tuned video model generalizes effectively to a wide range of categories far beyond the seven sketches used during training.

However, high final-frame accuracy does not necessarily imply a meaningful drawing progression. Naive prompting of Wan~2.1 achieves high recognition, yet as the progression curves in~\Cref{fig:clip_score_plot} (gray) reveal, it produces nearly identical frames throughout the sequence, with no visible stroke-by-stroke evolution. PaintsUndo also attains high final-frame scores, but operates in a fundamentally different setting — it reverses a finished painting (produced by FLUX.2) rather than synthesizing a sketch from scratch. Accordingly, its recognition saturates within the first 20\% of the sequence (\Cref{fig:clip_score_plot}, green), resembling its undo-based formulation. Notably, PaintsUndo requires $20k$ human-drawn Procreate time-lapse recordings for training.
In contrast, our method is the only approach that combines high final recognition with a steady, gradual progression (\Cref{fig:clip_score_plot}, orange), where the concept emerges naturally over the course of the drawing — achieved using only seven human-drawn training sketches.

\begin{table}[t]
\caption{\textbf{CLIP-based sketch recognition.} Average Top-1 and Top-5 accuracy of a CLIP zero-shot classifier evaluated on the last frame of 100 sketches from 50 categories. Results are reported as mean ± std across categories.}
\vspace{-0.2cm}
\centering
\setlength{\tabcolsep}{6pt}
\resizebox{\columnwidth}{!}{
\small
\begin{tabular}{lcc}
\toprule
Method & Top-1 & Top-5 \\
\midrule
Naive Prompting (Wan 2.1)
& 0.92 $\pm$ 0.03 & 0.99 $\pm$ 0.01 \\

PaintsUndo (FLUX 2)~\cite{paintsundo}
& 1.00 $\pm$ 0.00 & 1.00 $\pm$ 0.00 \\

SketchAgent~\cite{SketchAgent_Vinker2025} 
& 0.48 $\pm$ 0.05 & 0.71 $\pm$ 0.05 \\

Human (QuickDraw~\cite{quickDrawData}) 
& 0.52 $\pm$ 0.05 & 0.70 $\pm$ 0.05 \\

\midrule
Ours
& 0.82 $\pm$ 0.04 & 0.96 $\pm$ 0.02 \\
\bottomrule
\end{tabular}
}
\label{tab:CLIPscore}
\vspace{-0.3cm}
\end{table}

\subsection{Brush Style Control}\label{sec:brush_control}
Our video-based formulation naturally supports versatile styles, including different brushes and general drawing styles.
Brush style can be controlled through a visual in-frame cue (with our first-frame condition video model variant).
\Cref{fig:additional_brush} shows results on novel concepts within familiar brushes (first row) and unseen brush styles (second and third rows). The model faithfully reproduces the provided brush exemplar, prducing parametric-like quality without requiring explicit brush parameters. Notably, even physical effects such as color variation from overlapping marker strokes are captured faithfully. This demonstrates the richness of video diffusion models prior, and their untapped potential for supporting artistic processes.
We quantitatively evaluate brush generalization by using five unseen brushes and five unseen colors to generate drawings from 30 unseen object categories, yielding 750 samples in total. For each of the 25 brush-color combinations, we manually draw a reference sketch in the target style, which serves as the ground-truth style exemplar for that combination. Following prior style transfer evaluation protocols~\cite{alaluf2024cross,gatys2015neural}, we measure style alignment between each generated sketch and its corresponding reference by computing the average $\ell_2$ distance between Gram matrices from five intermediate layers of a pretrained VGG-19 network over stroke regions. Our method achieves an average distance of 0.14, where 0 indicates perfect alignment and 0.33 corresponds to the expected distance when comparing against references from mismatched brush-color combinations. More details, resutls, and evaluations are provided in the supplamentary.

\begin{figure}
\centering
\setlength{\tabcolsep}{2pt} %
\renewcommand{\arraystretch}{0.2} %
\small

\newcommand{\rowlab}[1]{\rotatebox{0}{\strut #1}}

\begin{tabular}{r c c c}

\raisebox{0.175cm}{\rotatebox{90}{\begin{tabular}{c} Naive\\Wan2.1 \end{tabular}}} &
\includegraphics[width=0.92\linewidth]{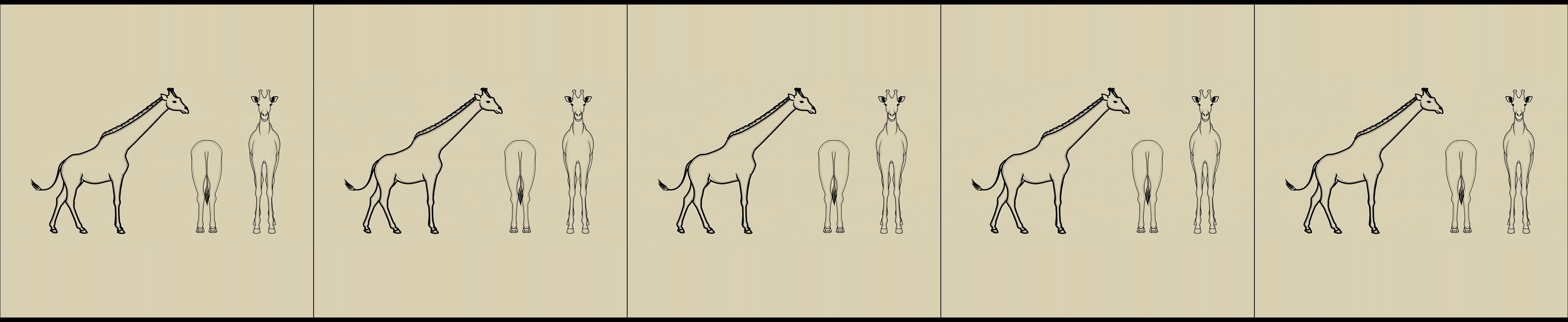} \\

\raisebox{0.175cm}{\rotatebox{90}{\begin{tabular}{c} Paints\\Undo \end{tabular}}} &
\includegraphics[width=0.92\linewidth]{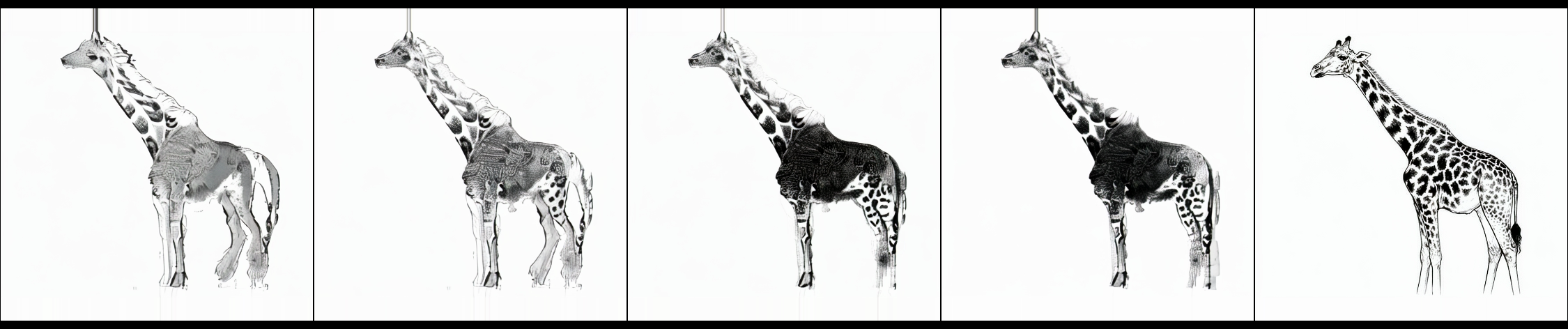} \\

\raisebox{0.175cm}{\rotatebox{90}{\begin{tabular}{c} Sketch\\Agent \end{tabular}}} &
\includegraphics[width=0.92\linewidth]{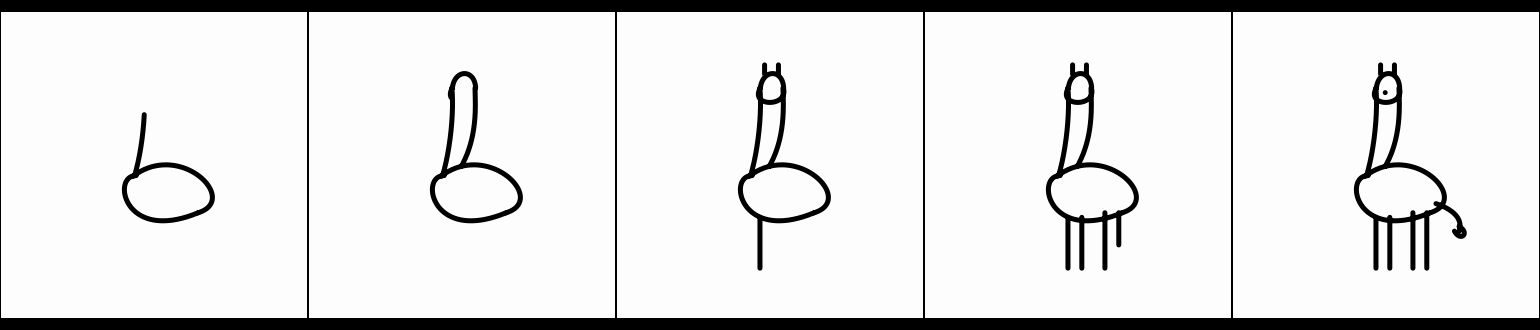} \\

\raisebox{0.0cm}{\rotatebox{90}{\begin{tabular}{c} Human\\(QuickDraw) \end{tabular}}} &
\includegraphics[width=0.92\linewidth]{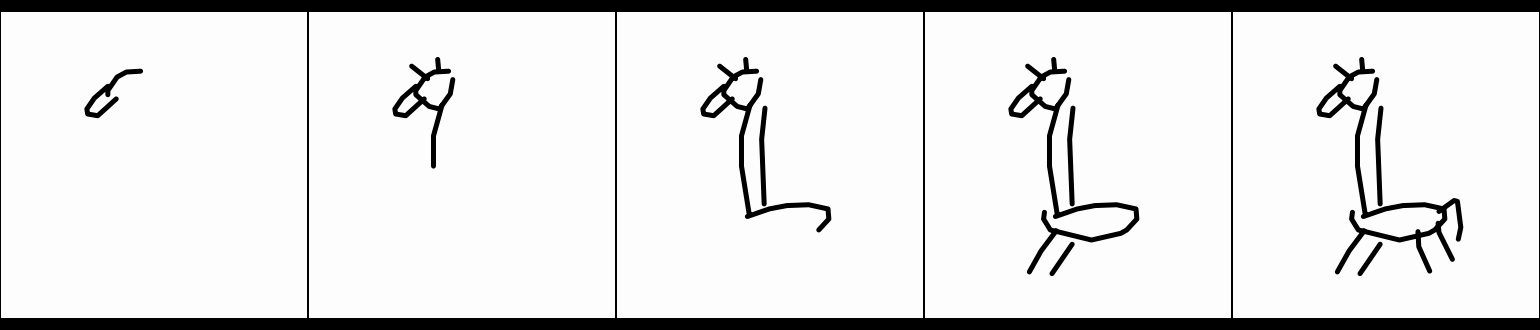} \\

\raisebox{0.3cm}{\rotatebox{90}{{\textbf{Ours}}}} &
\includegraphics[width=0.92\linewidth]{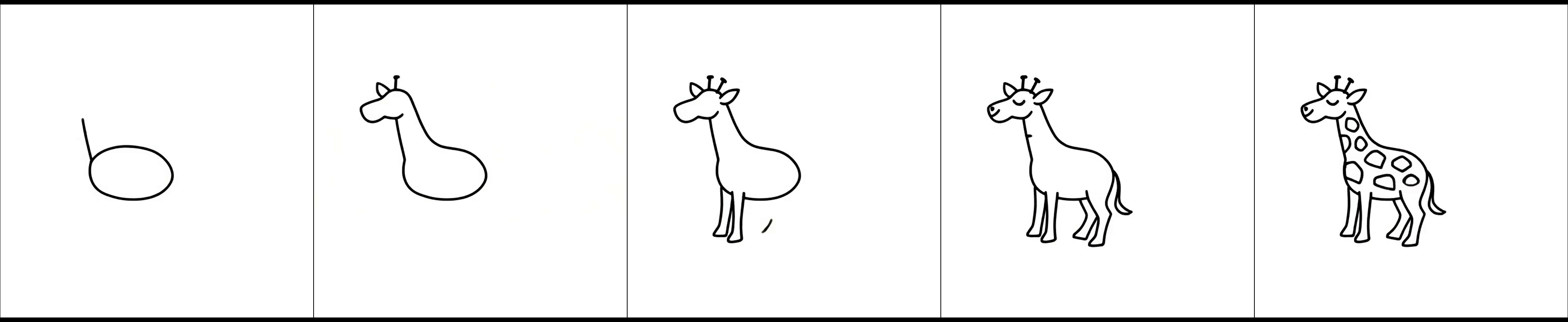} \\

\end{tabular}
\vspace{-0.25cm}
\caption{\textbf{Qualitative comparison} of sketch generation across methods. The concept depicted is ``a giraffe''.
}
\label{fig:qual_comparison}
\vspace{-10pt}
\end{figure}

\begin{figure*}[t]
\centering
\setlength{\tabcolsep}{0pt}
\renewcommand{\arraystretch}{0.7}
\small

\begin{tabular}{l@{\hspace{-5pt}} c@{\hspace{4pt}} c}

 & \parbox[c]{0.43\linewidth}{\centering \emph{Cat: 1. Body. 2. Head. 3. Ears. 4. Face details. 5. Legs. 6. Tail.}} 

 & \parbox[c]{0.43\linewidth}{\centering \emph{Cat: 1. Head. 2. Ears. 3. Body. 4. Legs. 5. Tail. 6. Face details.}}  \\[2pt]

\raisebox{0.3cm}{\parbox[c]{1.2cm}{\footnotesize \raggedright Primiti-\\ves Only}} &
\includegraphics[width=0.47\linewidth]{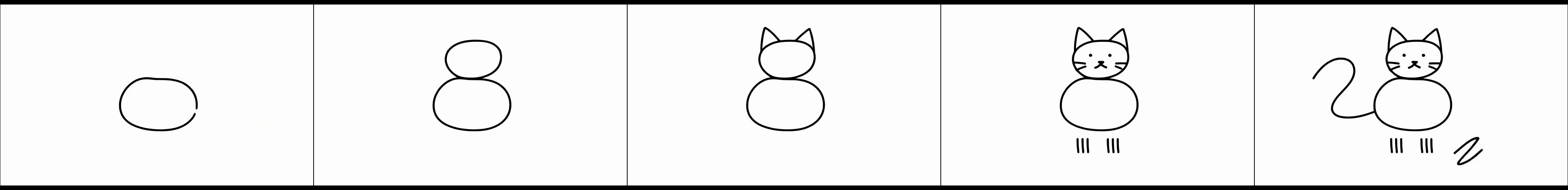} &
\includegraphics[width=0.47\linewidth]{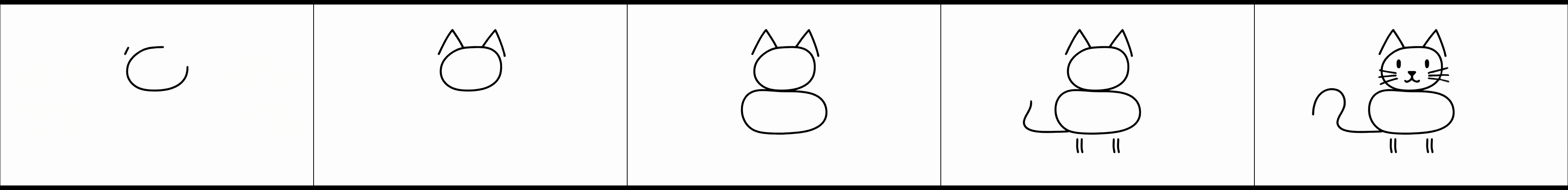} \\

\raisebox{0.3cm}{\parbox[c]{1.2cm}{\footnotesize \raggedright 7 Human\\Sketches\\Only}} &
\includegraphics[width=0.47\linewidth]{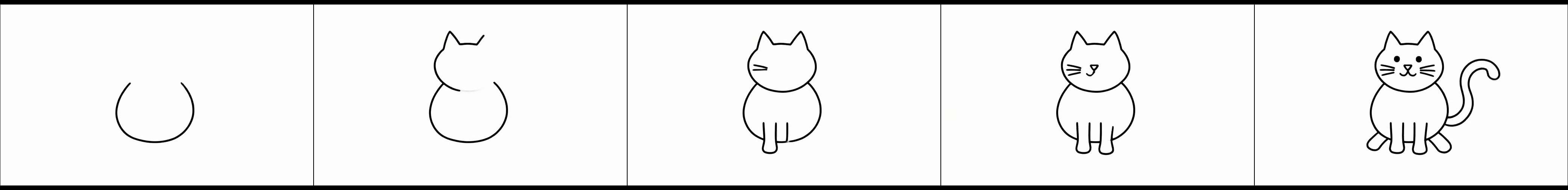} &
\includegraphics[width=0.47\linewidth]{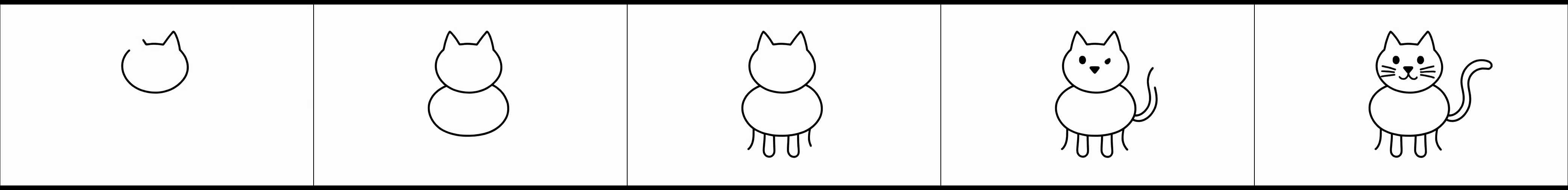} \\

\raisebox{0.3cm}{\footnotesize Full} &
\includegraphics[width=0.47\linewidth]{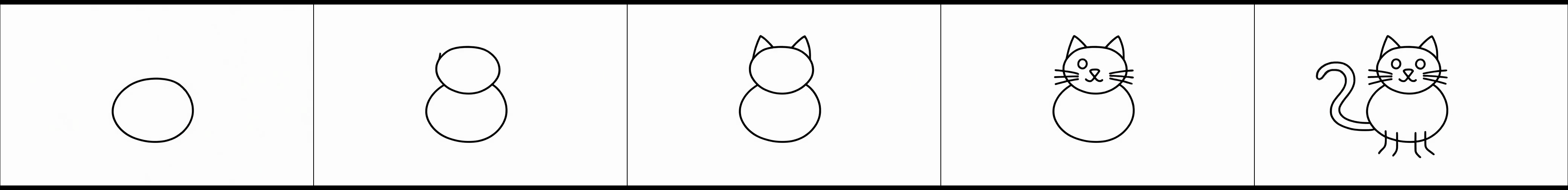} &
\includegraphics[width=0.47\linewidth]{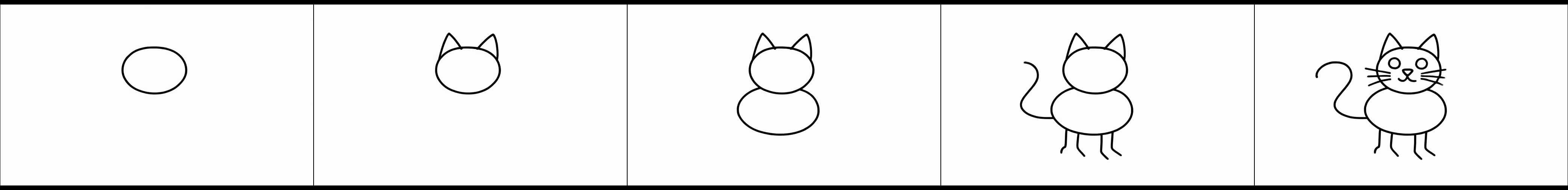}

\end{tabular}
\vspace{-10pt}
\caption{\textbf{Ablation study qualitative comparison.} We compare sketching processes from models trained only on geometric primitives, only on real sketches, and the full two-stage model. The full model combines the strengths of both baselines, producing visually appealing sketches that follow the specified order.}
\label{fig:ablation_comaprison}
\vspace{-10pt}
\end{figure*}

\subsection{Autoregressive Generation}
As our approach is data-centric and model-agnostic, it naturally extends beyond diffusion-based generation. We demonstrate this by adapting an autoregressive video model~\cite{Causvid}, which generates frames sequentially (rather than jointly, as performed in our base model), naturally enabling interactive scenarios such as human-model co-drawing. Autoregressive video models remain less mature than their diffusion counterparts, requiring more training data and producing lower visual quality even on natural videos. To address the data gap, we use our trained diffusion model as a teacher, synthesizing additional sketch sequences with LLM-generated prompts across 50 categories sampled from QuickDraw.
As shown in~\Cref{fig:additional_results_ar}, the autoregressive model produces visually coherent sketches with clear stroke-by-stroke progression, though with reduced fidelity compared to the full diffusion-based approach. 
We evaluate the autoregressive model using the same protocols as in \Cref{sec:text_generation}. For final-frame recognition, it achieves 51\% Top-1 and 73\% Top-5 accuracy — comparable to human drawings from QuickDraw (52\%/70\%) and SketchAgent (48\%/71\%), with the quality gap reflecting the current maturity of autoregressive video models.
To demonstrate interactivity, we built a prototype collaborative sketching interface where the user and model alternate adding strokes to a shared canvas. As shown in~\Cref{fig:ar_comparison}, users can co-draw with the model in real time, each adapting to the other's contributions. Since our approach is architecture-agnostic, improvements in autoregressive video models will directly benefit sketching quality, making this a promising direction for interactive sketch generation.

\section{Ablation Study}\label{sec:ablation}
We ablate a key design choice in our training procedure: separating the learning of stroke ordering from sketch appearance. We compare three variants: our full two-stage model, a model trained only on geometric primitives, and a model trained only on our seven human-drawn sketches.
We evaluate visual fidelity via CLIP-based recognition (following the protocol from \Cref{sec:text_generation}) and ordering fidelity.
Since no established metrics exist for evaluating ordering fidelity with respect to a text prompt, we adopt an LLM-guided evaluation protocol. We use Gemini-2.5-pro to extract a textual description of the drawing order from each generated video. We validate Gemini's extracted order with human annotations on 50 random samples, showing 92\% agreement. We then apply a head-to-head comparison between each combination of the three models, using an LLM to compare their extracted orderings against the ground-truth prompt and select the one that better adheres to it, with a ``Neither'' option when neither clearly fits.
As shown in~\Cref{tb:ablation_joined}, the two single-stage models exhibit a clear trade-off: the real-sketches-only model achieves high visual fidelity but poor ordering, while the primitives-only model follows orderings more faithfully but lacks visual quality. The full model combines the strengths of both, achieving strong ordering fidelity alongside high recognition rates.
We further illustrate this qualitatively in~\Cref{fig:ablation_comaprison}, showing the same concept drawn with two different orderings. The primitives-only model follows the specified order but produces geometric, low-quality outputs. Training only on real sketches improves appearance but often ignores the ordering. Combining both stages yields the best of both. Additional details and examples are provided in the supplementary.

\begin{table}[t]
\caption{\textbf{Ablation study.} (a) CLIP-based sketch recognition over our three model variants. (b) Ordering fidelity, where an LLM compares target stroke orderings with those inferred from two generated sketches and selects the closer match or ``Neither''.}
\vspace{-0.2cm}
\centering
\footnotesize
\begin{minipage}[c]{0.32\columnwidth}
    \centering
    \textbf{(a) Recognition}\\[4pt]
    \setlength{\tabcolsep}{3pt}
    \begin{tabular}{@{}lcc@{}}
    \toprule
    Method & Top-1 & Top-5 \\
    \midrule
    Only Prim.       & 0.73 & 0.86 \\
    Only 7-Human          & \textbf{0.88} & \textbf{0.96} \\
    Full  & 0.82 & 0.95 \\
    \bottomrule
    \end{tabular}
\end{minipage}%
\hspace{0.03\columnwidth}%
\begin{minipage}[c]{0.65\columnwidth}
    \centering
    \textbf{(b) Ordering Fidelity}\\[4pt]
    \setlength{\tabcolsep}{3pt}
    \begin{tabular}{@{}lccc@{}}
    \toprule
    Comparison & Model A & Model B & Neither \\
    \midrule
    Prim.\ vs.\ 7-Human & \textbf{50.0} & 37.2 & 12.8 \\
    Prim.\ vs.\ Full    & 26.9 & \textbf{53.4} & 19.7 \\
    7-Human vs.\ Full   & 29.6 & \textbf{48.3} & 22.1 \\
    \bottomrule
    \end{tabular}
\end{minipage}%
\label{tb:ablation_joined}
\vspace{-0.4cm}
\end{table}

\section{Limitations}
Our method has several limitations, illustrated in~\Cref{fig:limitations} and discussed further in the supplementary. Operating in pixel space provides less explicit structural control than parametric representations, occasionally leading to artifacts such as multiple strokes appearing within a single frame. Prompt adherence is also not guaranteed — when the model has a strong visual prior, it may deviate from instructions (e.g., altering an action or introducing color). Performance further depends on the video model's familiarity with a concept; compared to LLMs, video models are less knowledgeable about specialized domains such as mathematics, leading to failures on unusual concepts. 
Finally, autoregressive outputs do not yet match the visual quality of the diffusion-based model, reflecting the current maturity of autoregressive video generation. Since our approach is architecture-agnostic, we expect these results to improve as such models advance.
More broadly, operating in pixel space is computationally expensive (see~\Cref{sec:implementaiton}). We view this as an inherent trade-off for sidestepping key challenges of parametric approaches (data scarcity, discrete optimization, limited fidelity), and encourage future work to bridge this gap for example via hybrid pixel–vector models or maturing autoregressive architectures.

\begin{figure}[t]
\centering
\footnotesize

\begin{subfigure}[t]{\linewidth}
    \centering
    \caption{Multiple strokes per frame}
    \includegraphics[width=1\linewidth]{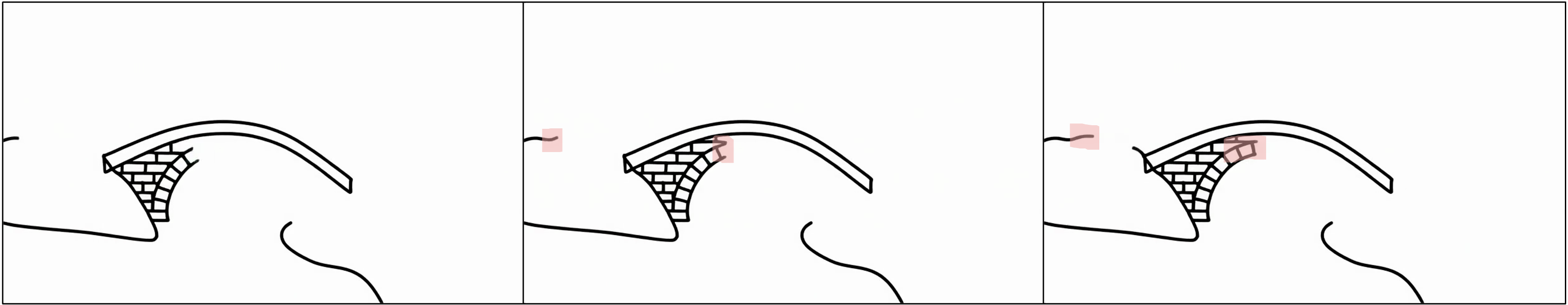}
\end{subfigure}

\vspace{-5pt}

\begin{subfigure}[t]{0.32\linewidth}
    \centering
    \caption{Prompt adherence}
    \includegraphics[width=\linewidth]{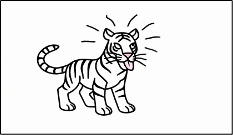}
    \emph{``A tiger roaring''}
\end{subfigure}
\hfill
\begin{subfigure}[t]{0.32\linewidth}
    \centering
    \caption{Limited knowledge}
    \includegraphics[width=\linewidth]{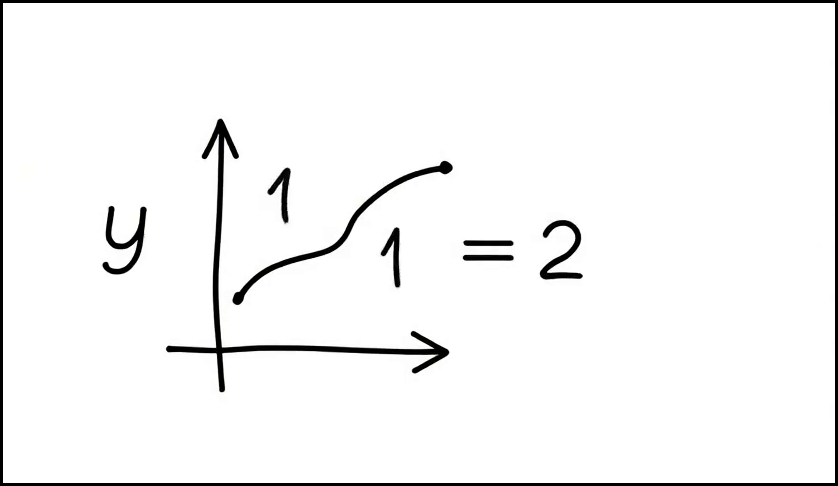}
    \emph{``$y=x^3$''}
\end{subfigure}
\hfill
\begin{subfigure}[t]{0.32\linewidth}
    \centering
    \caption{AR quality gap}
    \includegraphics[width=\linewidth]{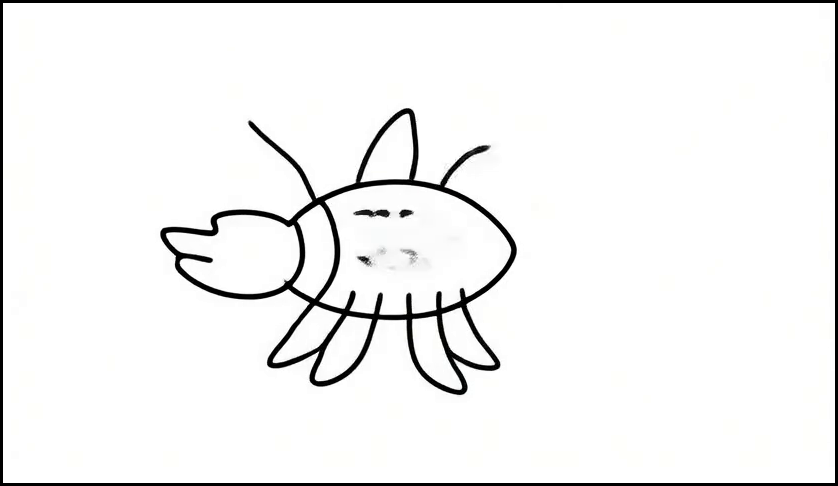}
    \emph{``A lobster''}
\end{subfigure}
\vspace{-8pt}
\caption{\textbf{Limitations.} (a) Multiple strokes may appear together (in red). (b) Model's prior can override prompt. (c) Concepts outside the video model's knowledge are incorrectly depicted. (d) Reduced quality for AR outputs.}
\label{fig:limitations}
\vspace{-6pt}
\end{figure}

\section{Conclusions}

We presented VideoSketcher, a data-efficient approach for sequential sketch generation that combines the semantic planning of LLMs with the visual and temporal priors of video diffusion models. Trained on only a handful of examples through a two-stage strategy that decouples ordering from appearance, our method generates high-quality, temporally coherent sketching processes and naturally extends to brush style control and interactive co-drawing.
Sequential sketch generation has traditionally relied on parametric stroke representations, which offer compactness and structural precision but face persistent challenges in data scarcity, discrete optimization, and limited visual fidelity. Our results demonstrate that video diffusion models, operating entirely in pixel space, encode a powerful and largely untapped prior for modeling drawing processes — achieving visual quality, generalization, and temporal coherence that sidestep these long-standing bottlenecks. 
Together, these results highlight the potential of pretrained video diffusion models as powerful priors for sequential sketch generation, pointing toward future approaches that bridge the parametric and pixel-based worlds.

\begin{acks}
We thank Moab Arar for his early insights and for engaging discussions.
We thank Anastasis Germanidis, Nick Petillo, Matt Kafonek, Qiming Fang, and the rest of the Runway team for their support and help throughout the project.
This work was partially supported by Hyundai Motor Co/MIT Agreement dated 2/22/2023, Hasso Plattner Foundation/MIT Agreement dated 11/02/2022, IBM/MIT Agreement No. W1771646, NSF grants 2008387, 2045586, 2106825, MRI 1725729, NIFA award 2020-67021-32799 and Amazon AI PhD Fellowship. The sponsors had no role in the experimental design or analysis, the decision to publish, or manuscript preparation. The authors have no competing interests to report.
\end{acks}

\bibliographystyle{ACM-Reference-Format}
\bibliography{main}

\begin{figure*}
\centering
\setlength{\tabcolsep}{2pt}
\renewcommand{\arraystretch}{0.2}
\small

\begin{tabular}{c c}
\parbox{0.95\linewidth}{\scriptsize\emph{``Step by step sketch process of a beach at sunset, following this drawing order: 1. Straight horizon line. 2. Ocean surface and shoreline curves. 3. Sun touching the horizon. 4. Layered sky bands or rays. 5. Small human silhouettes near the shore.''}} \\[6pt]
  \includegraphics[width=0.95\linewidth]{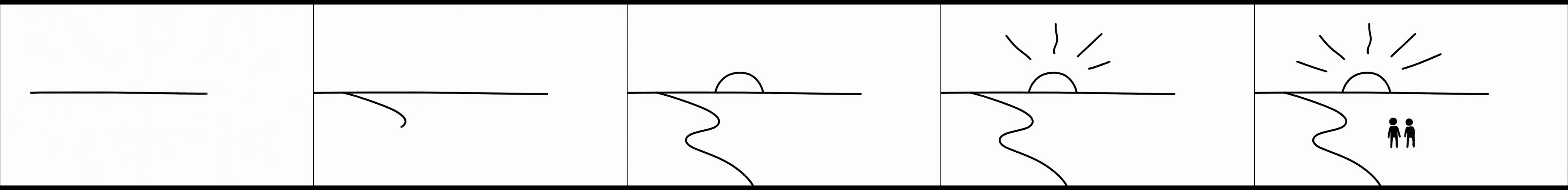} \\[6pt]

  \parbox{0.95\linewidth}{\scriptsize\emph{``Step by step sketch process of a lighthouse on a rocky shore, following this drawing order: 1. Curved shoreline. 2. Tall lighthouse tower. 3. Sweeping light beam. 4. Rocks and crashing waves. 5. Birds in flight.''}} \\[2pt]
  \includegraphics[width=0.95\linewidth]{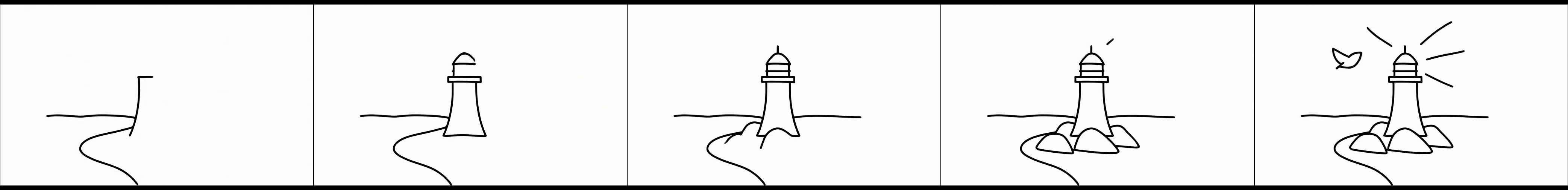} \\[6pt]

\end{tabular}

\vspace{-10pt}
\caption{\textbf{Qualitative results.} Additional results generated with our fine-tuned text-to-video model. Full video results are provided in the supplementary.}
\label{fig:additional_our_results}
\end{figure*}

\begin{figure*}[t]
\centering
\setlength{\tabcolsep}{0pt}
\renewcommand{\arraystretch}{0.6}
\small
\begin{tabular}{c}

    \emph{Concept: Elephant. Order: 1. Body. 2. Legs. 3. Head. 4. Ears. 5. Trunk. 6. Tail.} \\[2pt]
    \includegraphics[width=0.8\linewidth]{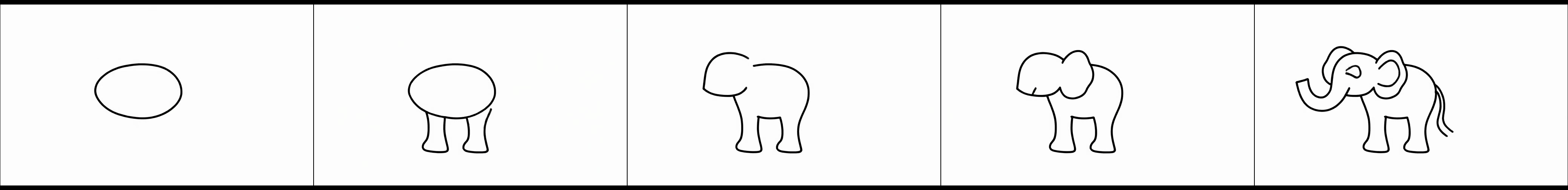} \\
    
    \emph{Concept: Elephant. Order: 1. Trunk. 2. Head. 3. Ears. 4. Body. 5. Legs. 6. Tail} \\[2pt]
    \includegraphics[width=0.8\linewidth]{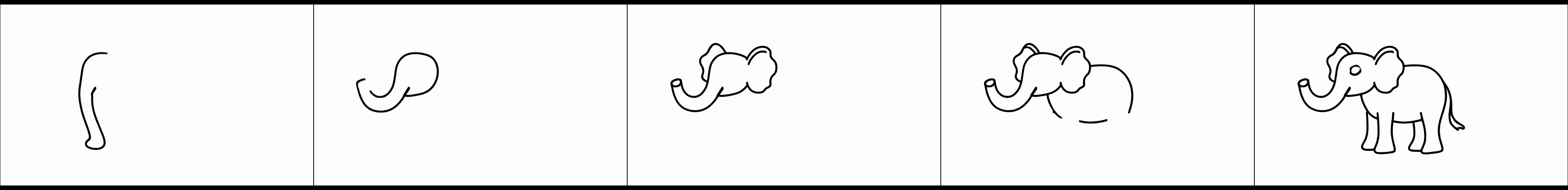} \\

    \emph{Concept: Robot. Order: 1. Torso. 2. Head. 3. Arms. 4. Legs. 5. Face} \\
    \includegraphics[width=0.8\linewidth]{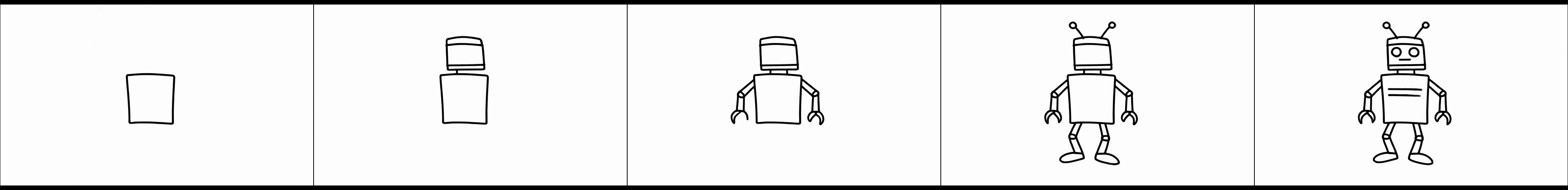} \\
    
    \emph{Concept: Robot. Order: 1. Head. 2. Face. 3. Torso. 4. Arms. 5. Legs.} \\
    \includegraphics[width=0.8\linewidth]{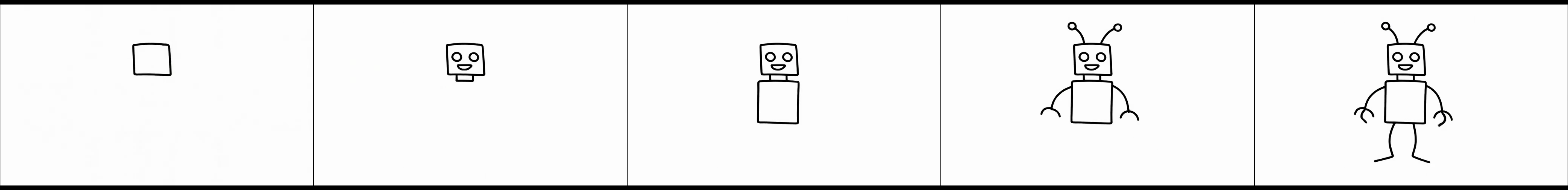} \\
    
\end{tabular}
\vspace{-8pt}
\caption{\textbf{Text-specified stroke ordering.} Each row shows the same concept generated using a different text prompt that specifies a distinct drawing order.}
\vspace{0.5cm}
\label{fig:ordering_diversity}
\end{figure*}

\begin{figure*}
\centering
\small

\begin{minipage}{0.58\linewidth}
    \centering
    \emph{``A traveler by a campfire''}\\
    \includegraphics[
        width=1\linewidth,
        trim=29cm 0pt 0pt 0cm,
        clip]{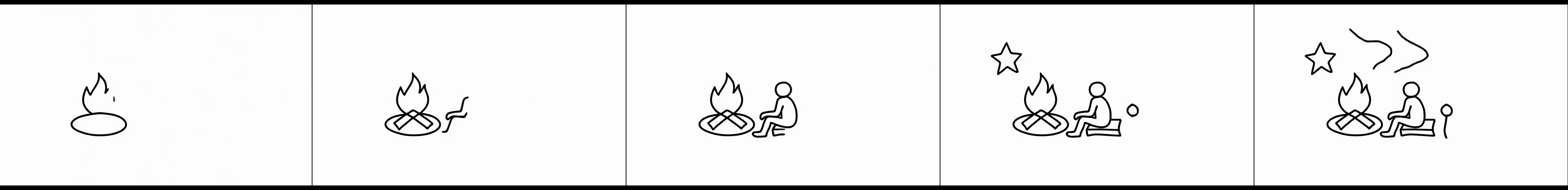}\\
    \includegraphics[width=1\linewidth,
        trim=29cm 0pt 0pt 0cm,
        clip]{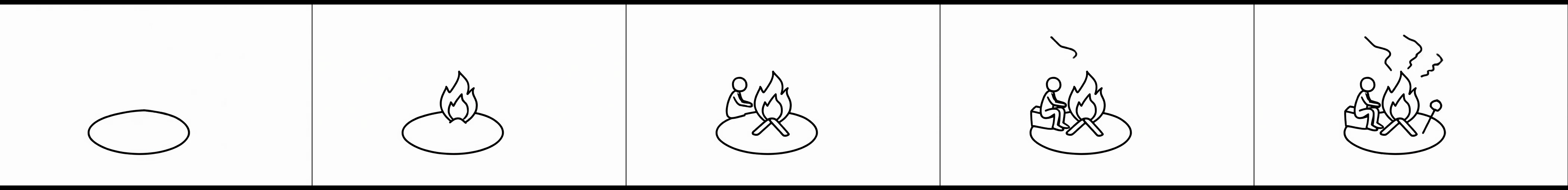}
\end{minipage}%
\hfill
\begin{minipage}{0.42\linewidth}
    \centering
    \begin{tabular}{@{}c @{\hspace{2pt}} c@{}}
        \raisebox{0.4cm}{\parbox[c]{0.15\linewidth}{\centering \emph{``Lego man\\waving''}}} &
        \includegraphics[width=0.75\linewidth]{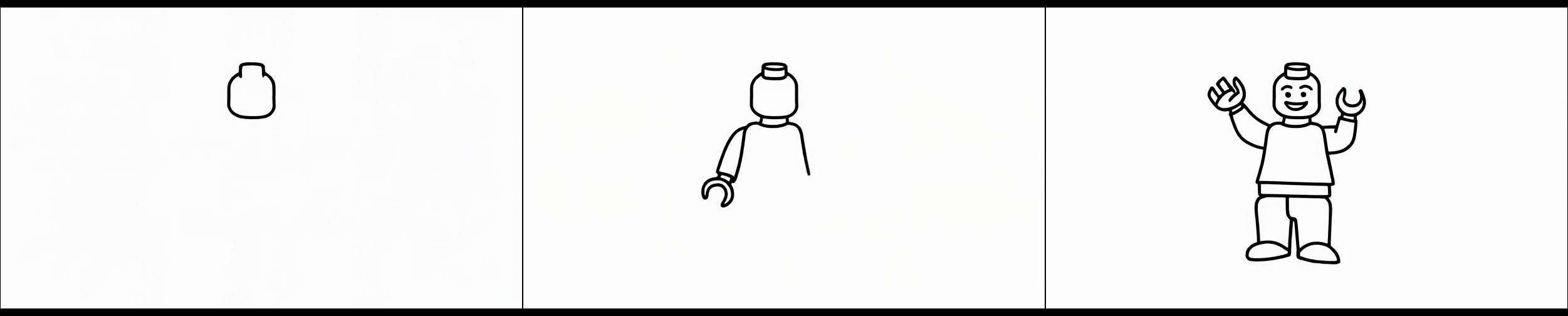}\\
        \raisebox{0.4cm}{\parbox[c]{0.15\linewidth}{\centering \emph{``Running''}}} &
        \includegraphics[width=0.75\linewidth]{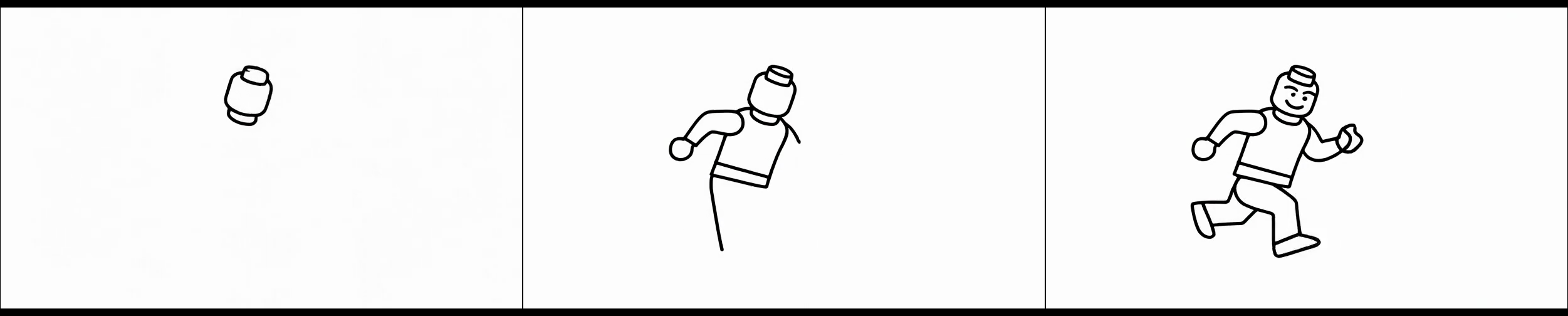}\\
        \raisebox{0.4cm}{\parbox[c]{0.15\linewidth}{\centering \emph{``Sitting''}}} &
        \includegraphics[width=0.75\linewidth]{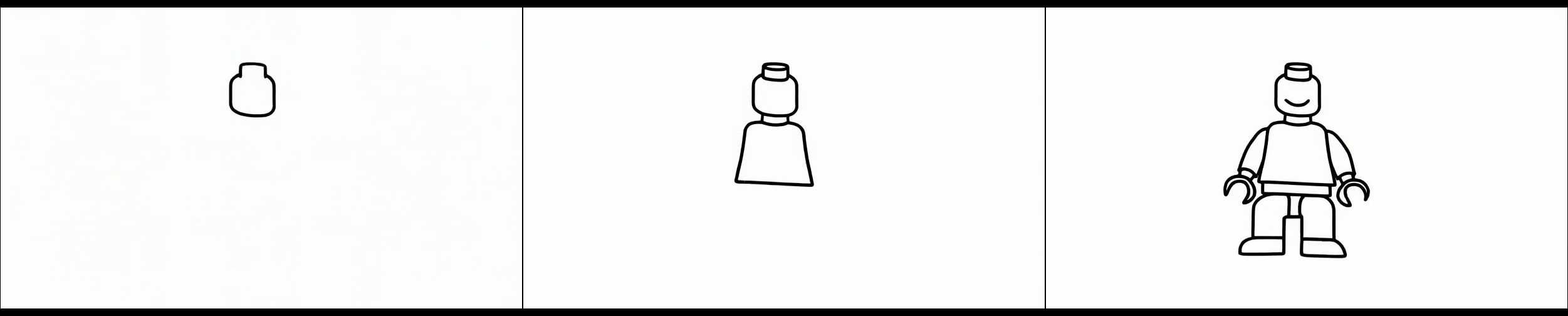}\\
    \end{tabular}
\end{minipage}

\vspace{-0.2cm}
\caption{\textbf{Diversity of sequential sketch generation.} Left: two seeds for the same prompt; Right: the same object in different settings/actions.}
\label{fig:additional_results_diversity}
\end{figure*}

\begin{figure*}[t]
\centering
\setlength{\tabcolsep}{0pt}
\renewcommand{\arraystretch}{0.6}
\small
\centering
\begin{tabular}{c}
\parbox{1\linewidth}{\scriptsize\emph{``Step by step sketch process of a child flying with a bundle of balloons over rooftops, using the color and style of the brush shown in the top-left corner, following this drawing order: 1. Child’s body – small floating figure. 2. Head, arms, and legs. 3. Balloon strings in hand. 4. Cluster of large balloons above. 5. Rooftops below. 6. Birds or clouds in the sky.''}} \\[4pt]
    \includegraphics[width=1\linewidth]{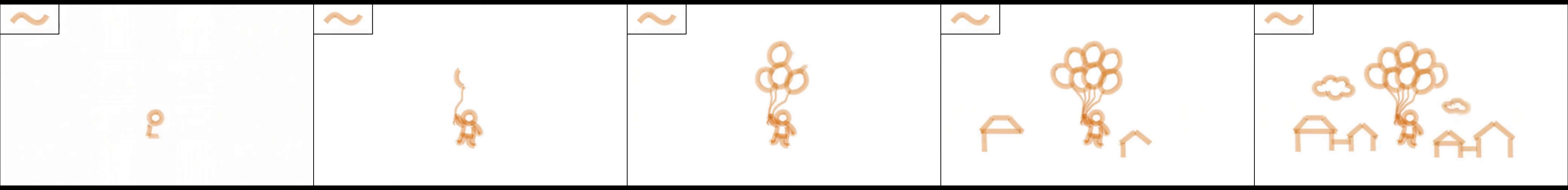}\\[4pt]

    \parbox{1\linewidth}{\scriptsize\emph{``Step by step sketch process of a laboratory bench, using the color and style of the brush shown in the top-left corner, following this drawing order: 1. Flat table surface. 2. Large glass containers. 3. Smaller tubes and instruments. 4. Stands, clamps, and cables. 5. Steam or glow effects. 6. Cluttered bench details.''}} \\[4pt]
    \includegraphics[width=1\linewidth]{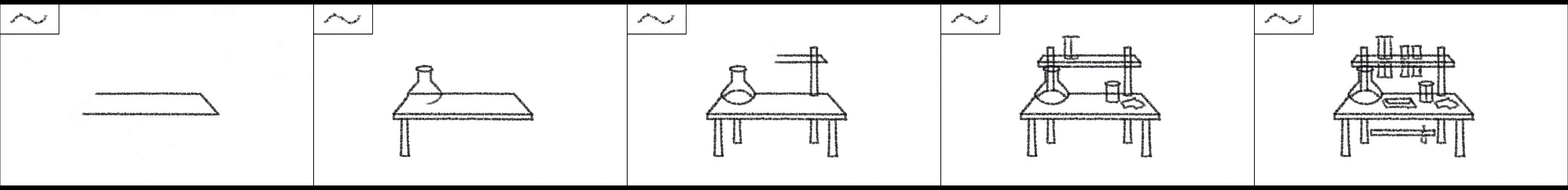}\\[4pt]

    \parbox{1\linewidth}{\scriptsize\emph{``Step by step sketch process of a squirrel, using the color and style of the brush shown in the top-left corner, following this drawing order: 1. Body – oval shape. 2. Head – smaller circle at the front. 3. Tail – big fluffy curve rising behind the body. 4. Legs and paws – small bent shapes under the body. 5. Face and ears – pointed ears, eye, nose, and mouth line. 6. Fur details – short strokes along the tail and chest.''}} \\[4pt]
    \includegraphics[width=1\linewidth]{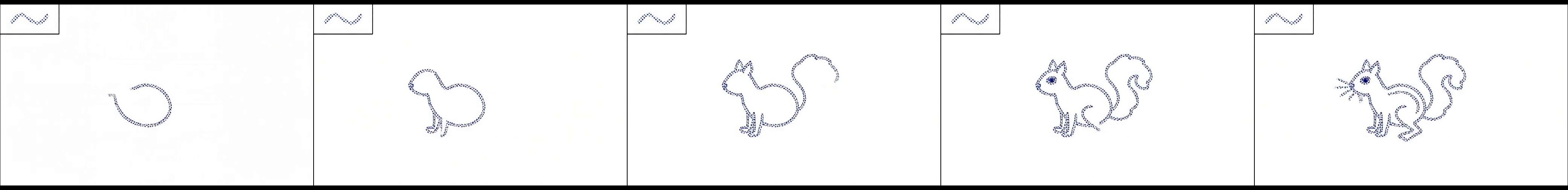}
\end{tabular}
\vspace{-0.3cm}
\caption{\textbf{Additional results for brush style control.} We show concepts drawn with seen (first row) and unseen (second, third rows) brush styles and colors.}
\label{fig:additional_brush}
\end{figure*}

\begin{figure*}[t]
\centering
\small
\setlength{\tabcolsep}{2pt}
\renewcommand{\arraystretch}{1}

\centering

\begin{tabular}{@{}c@{}}
\parbox{0.85\linewidth}{\scriptsize\emph{``Step by step sketch process of an octopus, following this drawing order: 1. The head – a large rounded dome. 2. The face – two eyes and a small mouth under them. 3. The front tentacles – draw two long curved arms flowing downward. 4. The remaining tentacles – add six more arms around, varying the curves. 5. The suckers – small circles along the undersides. ''}} \\[2pt]
    \includegraphics[width=0.85\linewidth]{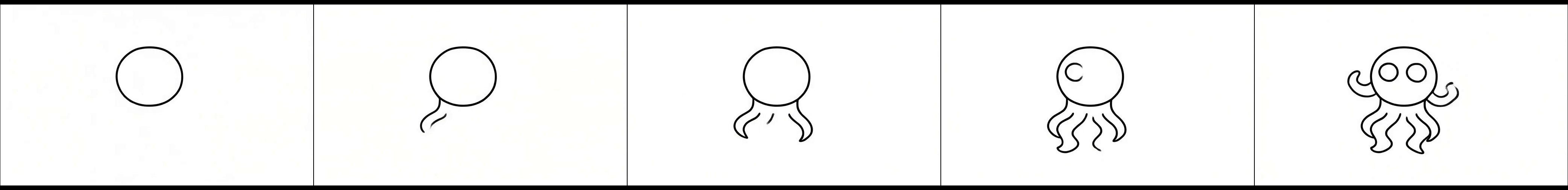}\\

\parbox{0.85\linewidth}{\scriptsize\emph{``Step by step sketch process of a squirrel, following this drawing order: 1. Body – oval shape. 2. Head – smaller circle at the front. 3. Tail – big fluffy curve rising behind the body. 4. Legs and paws – small bent shapes under the body. 5. Face and ears – pointed ears, eye, nose, and mouth line. 6. Fur details – short strokes along the tail and chest.''}} \\[2pt]
    \includegraphics[width=0.85\linewidth]{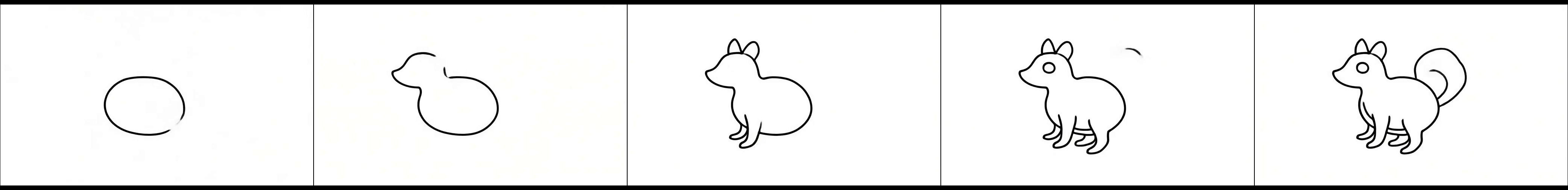}\\
\parbox{0.85\linewidth}{\scriptsize\emph{``Step by step sketch process of a penguin, following this drawing order: 1. The body as a tall oval. 2. The head as a smaller oval on top. 3.Add the belly patch shape. 4. Draw the beak and small eyes. 5. Add the flippers on the sides. 6. Draw the feet at the base.''}} \\[2pt]
    \includegraphics[width=0.85\linewidth]{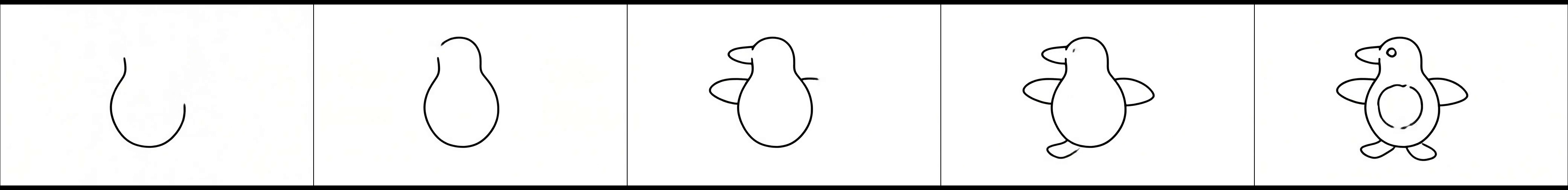}\\[-2pt]
\end{tabular}
\vspace{-0.1cm}
\caption{\textbf{Autoregressive results.} Additional results from our autoregressive model, which enables interactive generation while maintaining visual quality comparable to diffusion-based results.}
\label{fig:additional_results_ar}
\vspace{-8pt}
\end{figure*}
\hfill

\begin{figure*}[t]
\centering
\setlength{\tabcolsep}{0pt}
\renewcommand{\arraystretch}{0.9}
\small

\newcommand{\trimimg}[1]{%
  \frame{\includegraphics[width=0.19\linewidth, trim=3.5cm 0pt 3.5cm 0pt, clip]{#1}}%
}

\begin{minipage}{0.38\linewidth}
    \centering
    \includegraphics[width=1\linewidth]{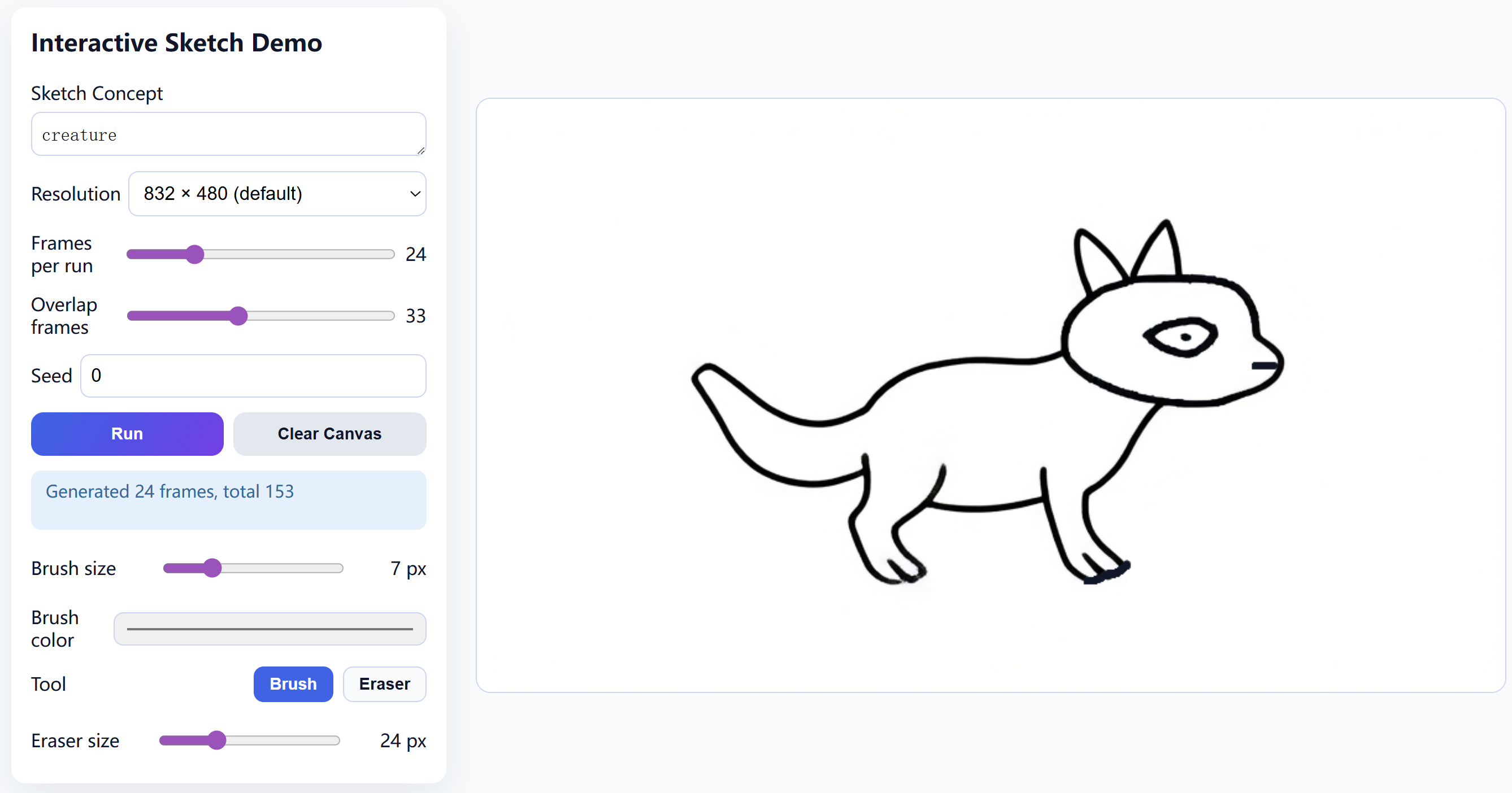}\\[10pt]

\end{minipage}
\hfill
\begin{minipage}{0.55\linewidth}
    \centering
\begin{tabular}{ccccc}

\multicolumn{1}{c}{User} &
\multicolumn{1}{c}{Model} &
\multicolumn{1}{c}{User} &
\multicolumn{1}{c}{Model} &
\multicolumn{1}{c}{User} \\

\trimimg{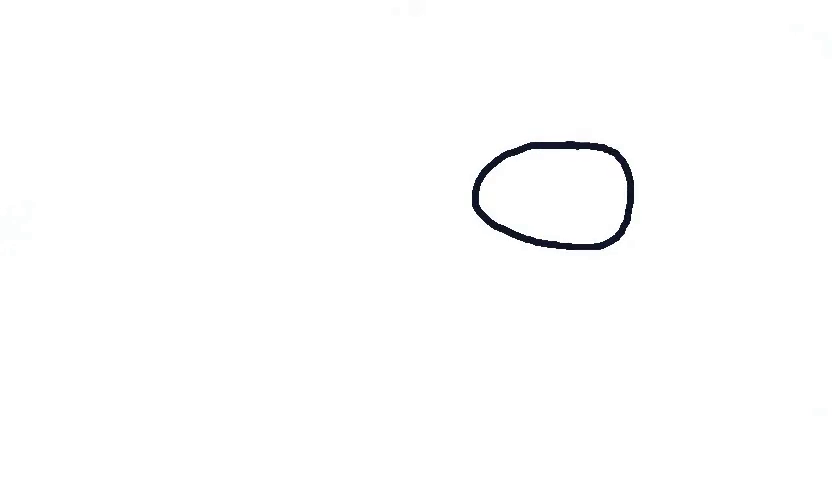} &
\trimimg{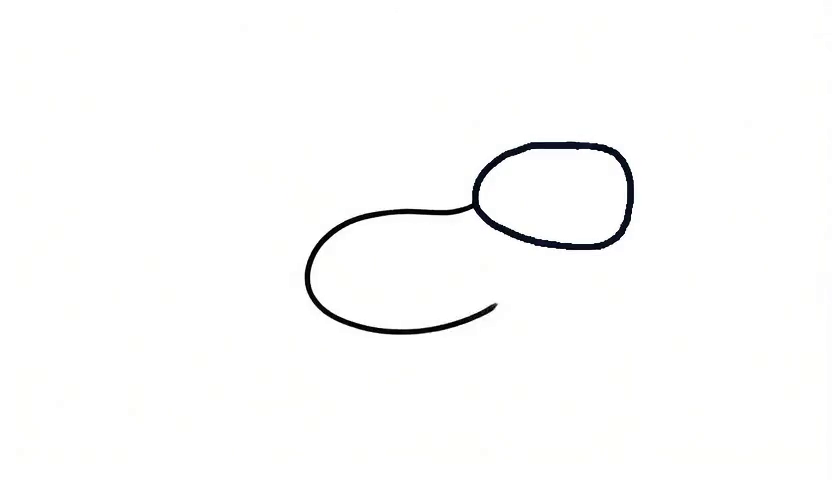} &
\trimimg{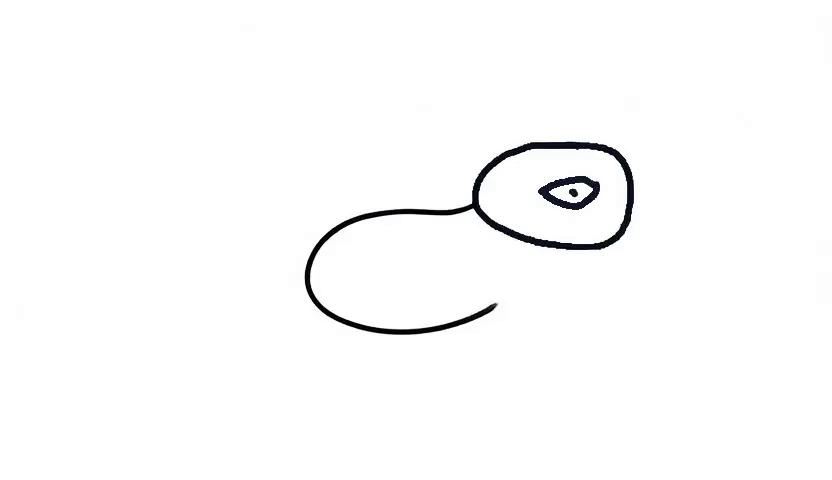} &
\trimimg{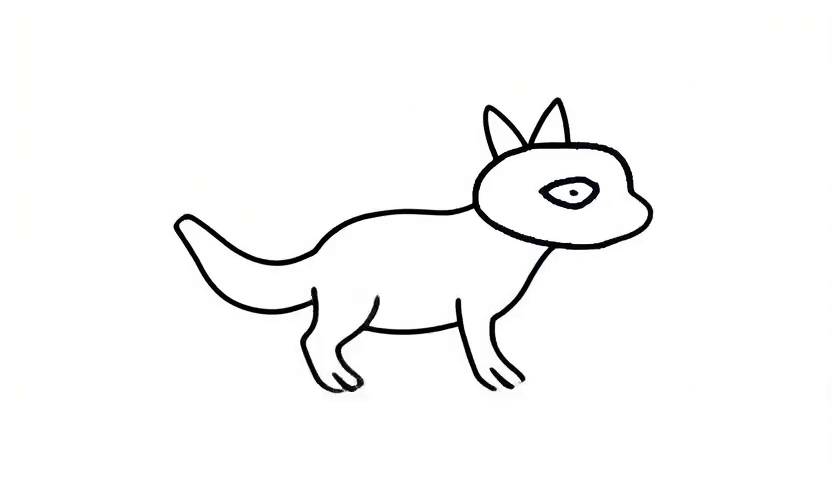} &
\trimimg{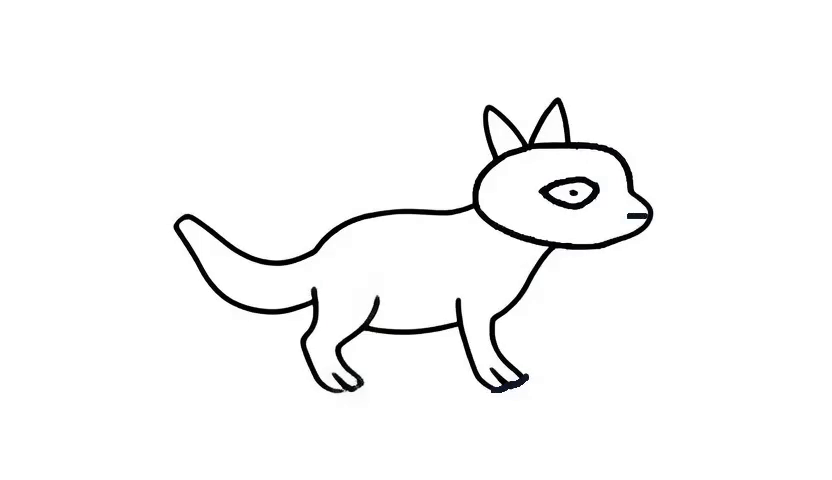} \\

\multicolumn{1}{c}{Model} &
\multicolumn{1}{c}{User} &
\multicolumn{1}{c}{Model} &
\multicolumn{1}{c}{User} & 
\multicolumn{1}{c}{Model}\\

\trimimg{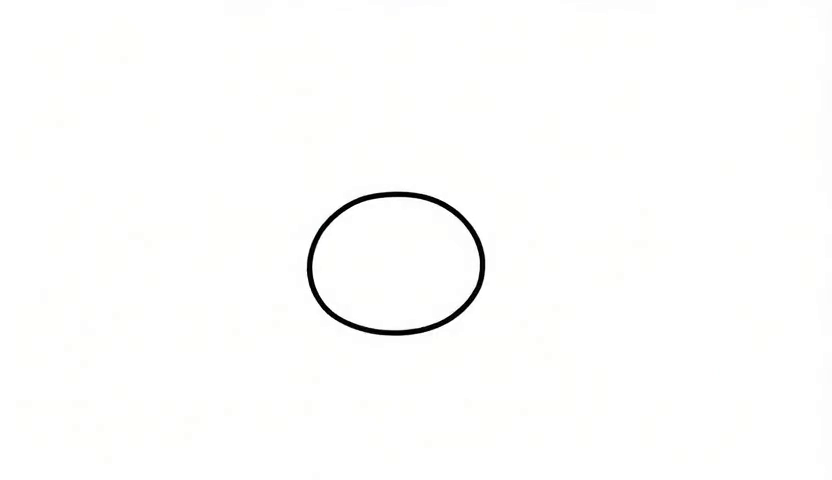} &
\trimimg{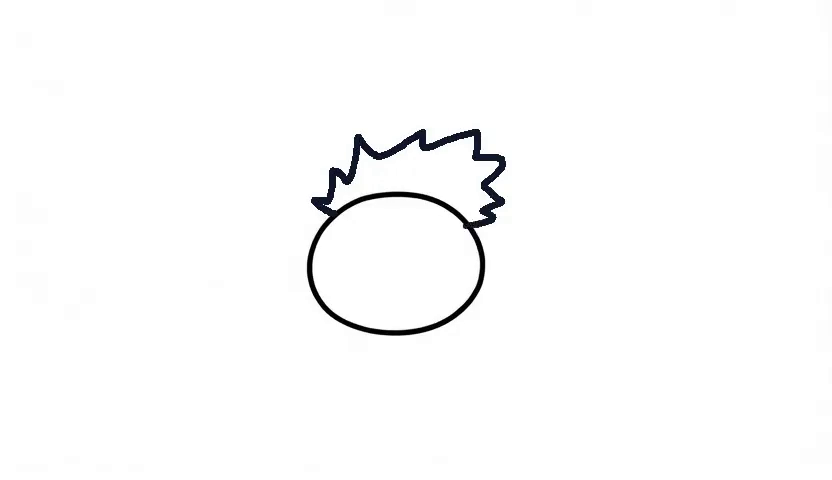} &
\trimimg{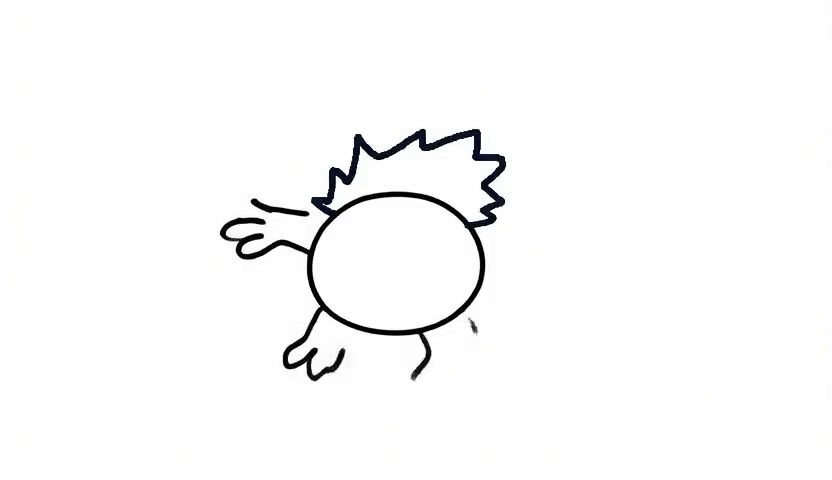} &
\trimimg{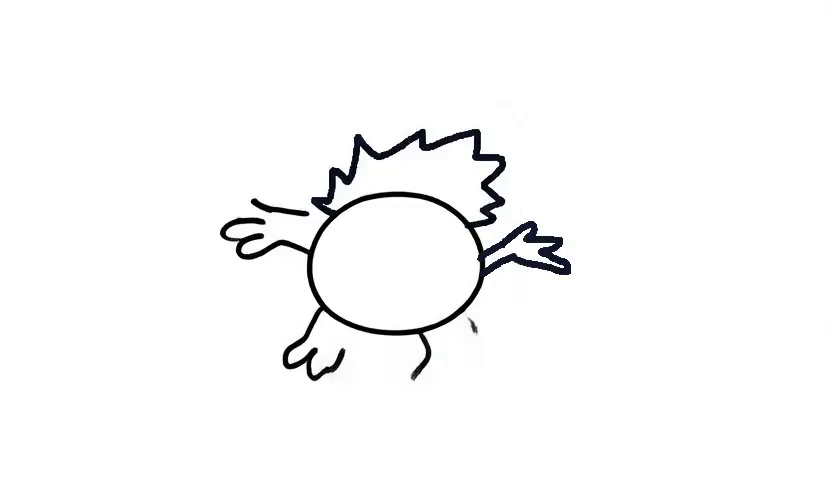} &
\trimimg{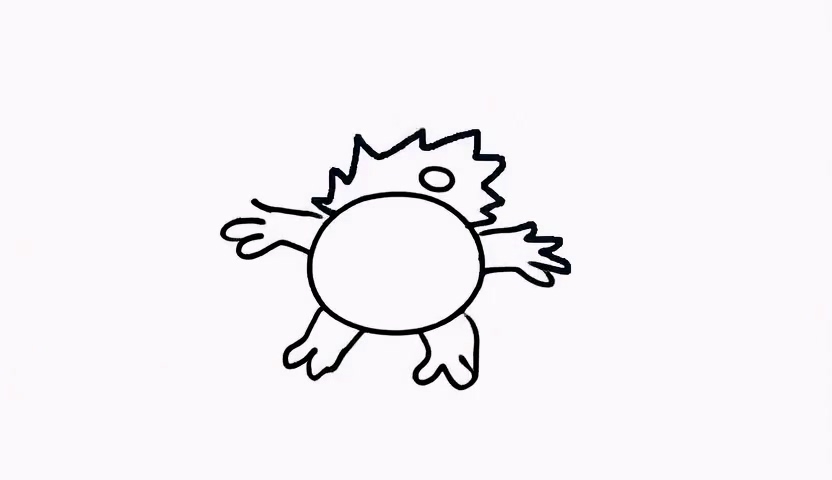} \\

\end{tabular}
\end{minipage}
\vspace{-0.4cm}
\caption{\textbf{Co-Drawing.}  Left: our interactive demo, where users draw alongside the model in real time. Right: turn-based co-drawing for ``a creature.''}

\label{fig:ar_comparison}
\end{figure*}

\clearpage

\twocolumn[%
  \begin{center}
    {\LARGE\bfseries VideoSketcher: Sequential Sketch Generation Using Video Model Priors \\ \vspace{0.3cm}Supplementary Material}
  \end{center}
  \vspace{0.5cm}
]

\appendix

\addtocontents{toc}{\protect\setcounter{tocdepth}{2}}

\makeatletter
\@starttoc{toc}
\makeatother

\section{Implementation Details}\label{sec:implementation}

\subsection{Training Details}
We use the pretrained Wan~2.1~14B~\cite{Wan2.1} diffusion model as our main base video backbone. Fine-tuning is performed using LoRA adapters applied to the attention layers and the first two layers of the feed-forward networks in the diffusion transformer. This design enables efficient adaptation while mitigating overfitting in the low-data regime. All diffusion models are trained using a LoRA rank of 32 and the standard rectified flow matching loss.

Training is conducted on 7 NVIDIA A100 GPUs with a batch size of 1 and a learning rate of $1\mathrm{e}{-4}$. The synthetic shape training stage runs for 700 epochs on 15 videos. The additional fine-tuning on the 7 human-drawn sketches is then applied for 700 additional epochs. 
Across all experiments, both training and inference are performed at a resolution of $480 \times 832$ with 81 frames.
Training overall takes about 22 hours.
At inference, for the best performance, we apply the trained model with 50 inference steps (though using 10 steps also produces plausible results and can save inference time). Inference of a video with 81 frames with 50 denoising steps and resolution $480 \times 832$ takes 16 minutes on a single A100 GPU.

\begin{figure}[t]
    \centering
    \includegraphics[width=1\linewidth]{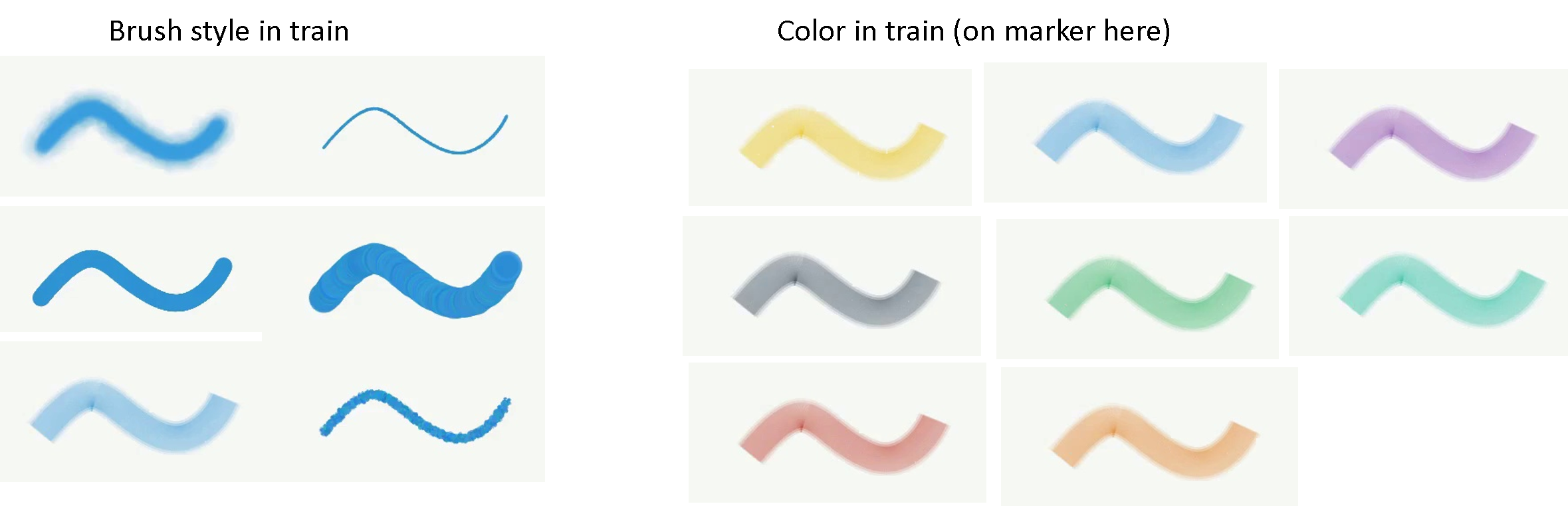} 
    \vspace{-0.5cm}
    \caption{Six brush styles (left) and 8 colors (right) used for training our brush-conditioned model.}
    \vspace{-0.7cm}
    \label{fig:brushes}
\end{figure}

\paragraph{Brush-Conditioned Image-to-Video Model.}
For brush style conditioning, we use the image-to-video variant of the Wan~2.1~14B model. The training data is augmented with 6 brush styles and 8 colors (shown in \Cref{fig:brushes}), resulting in 720 samples for the shape training stage and 336 samples for the human-drawn real sketches stage. The training process follows the same settings as described above for the text-to-video diffusion model. 

\paragraph{Autoregressive Model.}
For autoregressive sketch generation, we use the CausVid~\cite{Causvid} model as the base video model, which is a fine-tuned autoregressive variant of Wan~2.1~1.3B (a smaller model compared to the 14B model, with slightly reduced quality). We construct the training set using 43 videos generated by our 14B text-to-video diffusion model, together with 7 real-world sketches, yielding a total of 50 training videos. The 43 videos were generated by randomly sampling categories from the QuickDraw dataset (ones not used in our evaluation setup) and generating detailed prompts for them. We train the full diffusion transformer for 2700 epochs with a learning rate of $2\mathrm{e}{-6}$, using a regression loss on the ODE trajectory. Qualitative results on QuickDraw prompts are shown in~\Cref{fig:supp_ar_quickdraw_merged_0,fig:supp_ar_quickdraw_merged_2}.
Inference of a video with 81 frames and resolution $480 \times 832$ takes about 11 seconds on a single A100 GPU. This means it takes about 4 seconds to produce 24 frames, which is our default step size in the interactive demo, enabling real-time interaction.

\subsection{Training Data}
The seven human-drawn sketches were collected from an artist via Adobe Illustrator, which natively records both the global stroke order and per-stroke progression in the SVG format. The synthetic shape primitives were manually arranged into the desired compositions, with ordering variations introduced through manual shuffling. 

To produce a smooth animation, the total number of video frames is distributed across strokes proportionally to their geometric arc length. This ensures that longer strokes occupy more frames and shorter strokes fewer, keeping the apparent drawing speed roughly constant.
Each stroke segment is densely sampled into a polyline. At each animation frame, a partial prefix of the current stroke's polyline is rendered progressively onto a canvas using Matplotlib, accumulating all previously completed strokes, producing a frame-by-frame video of the drawing process.
The full-length video is then subsampled to a fixed count of 81 frames. Stroke-onset frames (the first frame of each new stroke) are treated as mandatory keyframes. The remaining frame budget is distributed proportionally across the intervals between keyframes using a Hamilton apportionment method, ensuring that denser or longer stroke sequences receive more intermediate samples.

The corresponding text prompts were created manually for this small dataset. The geometric primitive compositions used in the first training stage are visualized in~\Cref{fig:supp_trainset_shapes_1,fig:supp_trainset_shapes_2}, and the seven human-drawn sketches used in the second stage are shown in~\Cref{fig:supp_trainset_sketches}.

\subsection{Stroke Ordering Metric}
In the ablation study presented in the main paper, we report quantitative measurements of ordering fidelity for three model variants: a model trained solely on simple geometric primitives, a model trained on only seven human-drawn sketches, and our full model trained using the proposed two-stage approach.

Since no established metrics exist for evaluating ordering fidelity with respect to a text prompt, we adopt an LLM-guided evaluation protocol consisting of two stages. First, given a video generated by a model, we prompt an LLM to extract the sequence in which semantic parts are drawn (e.g., ``1. Body, 2. Head, 3. Face, \dots''). To ensure consistent terminology across models, we also provide the LLM with the target ordering, which constrains the vocabulary used to describe the parts (e.g., enforcing the term ``Body'' rather than alternatives such as ``Torso'').

Next, we perform pairwise, head-to-head comparisons between each combination of the three models. For each pair, the LLM compares the extracted orderings against the target ordering and selects the model that better adheres to it. When both models deviate from or match the target ordering to a similar extent, the LLM is allowed to return ``Neither,'' resulting in a tie. This evaluation is conducted over 100 prompts from QuickDraw and three random seeds, with the averaged results reported in Table 2 of the main paper. To validate the reliability of this LLM-based metric, we cross-referenced LLM judgments against human annotations on 50 randomly selected samples, finding 92\% agreement between the two.

\begin{figure}[t]
    \centering
    \small
    \setlength{\tabcolsep}{0pt}
    \begin{tabular}{c@{\hskip 10pt}ccc}
        & SketchAgent & Wan~2.1& Ours  \\
        \lefttxt{``Double-slit  \\ experiment''} & 
        \includegraphics[height=0.12\textwidth]{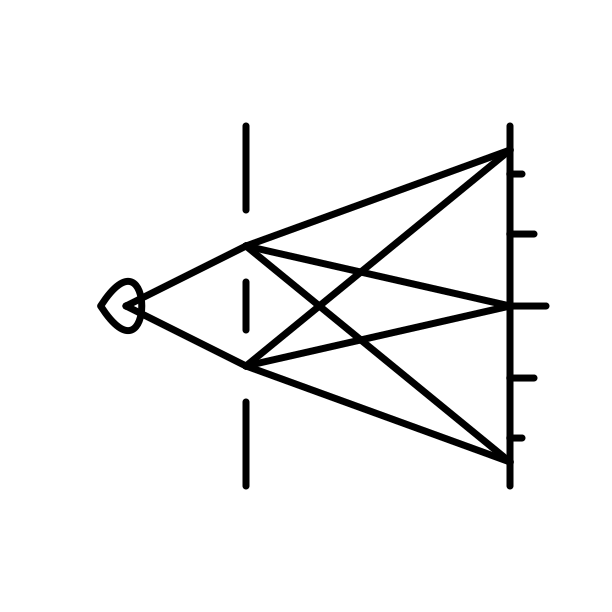} &
        \includegraphics[trim=4cm 0 4.5cm 0, clip, height=0.12\textwidth]{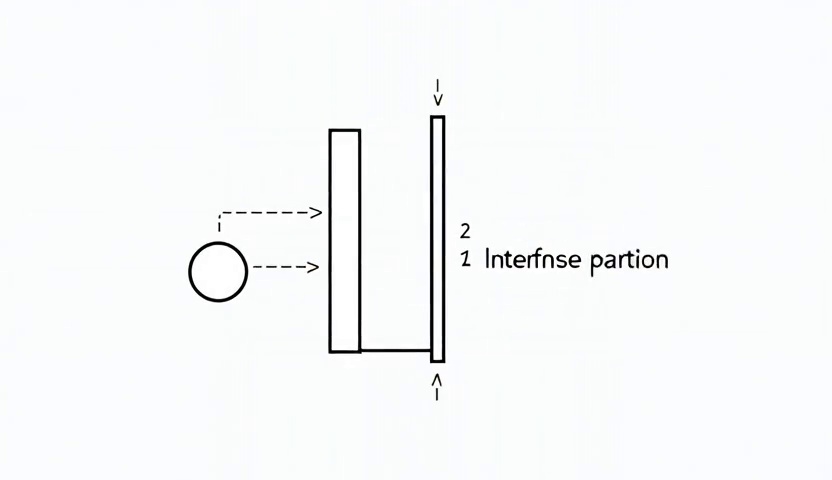} &
        \includegraphics[trim=4cm 0 4.5cm 0, clip, height=0.12\textwidth]{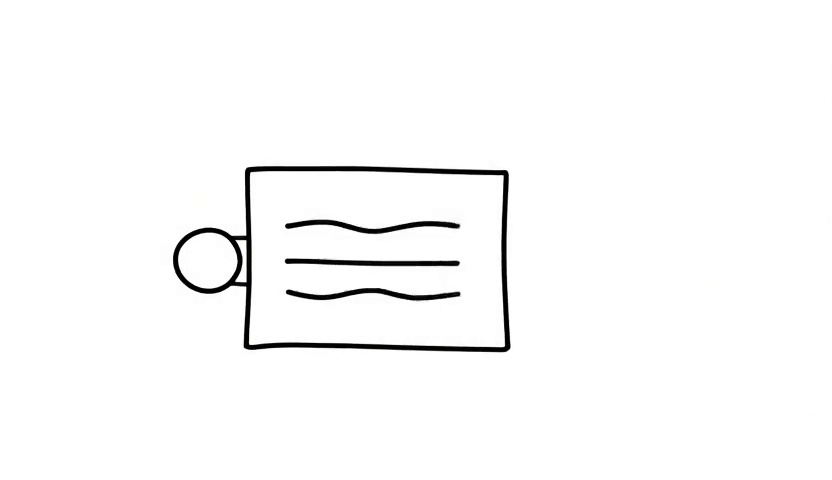} \\
        
        \lefttxt{``Quantum \\ entanglement''} & 
        \includegraphics[height=0.12\textwidth]{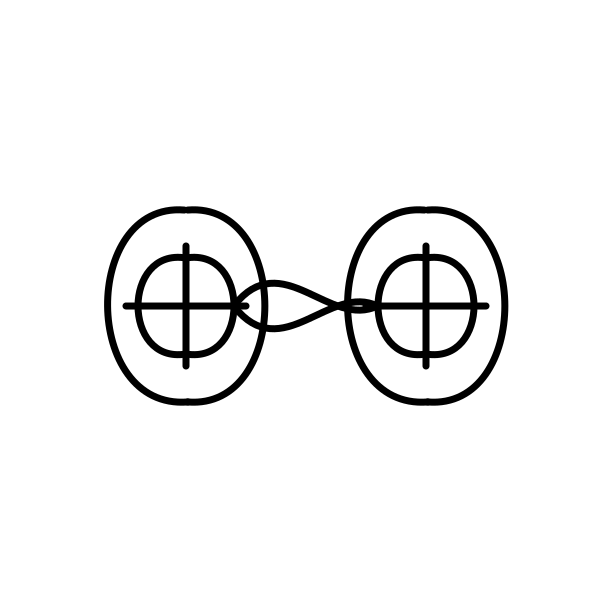} &
        \includegraphics[trim=4cm 0 4.5cm 0, clip, height=0.12\textwidth]{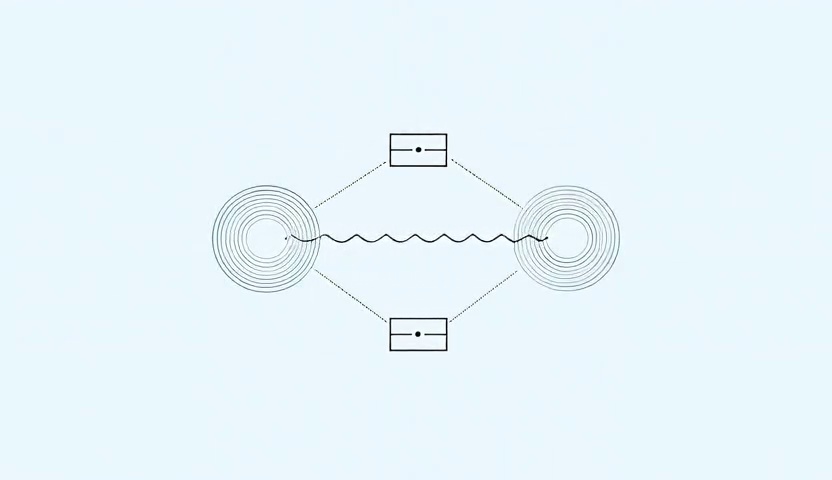} &
        \includegraphics[trim=2cm 0 4cm 0, clip, height=0.11\textwidth]{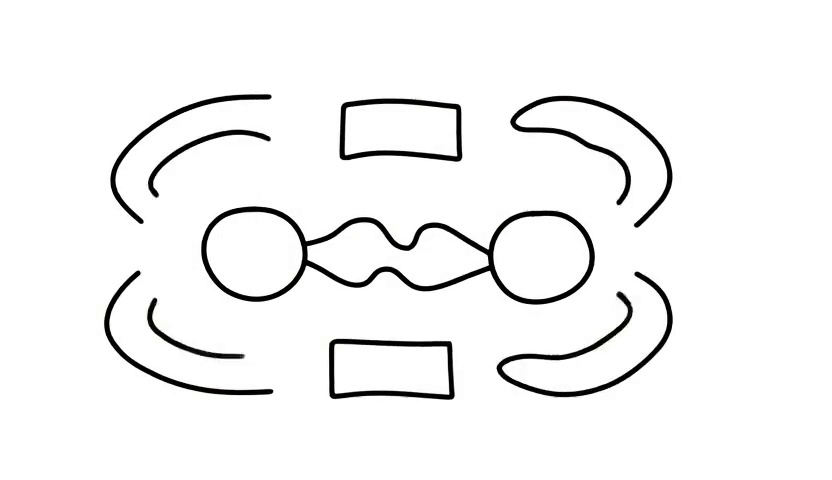} \\
        
        \lefttxt{``Pendulum  \\ motion''} &
        \includegraphics[height=0.12\textwidth]{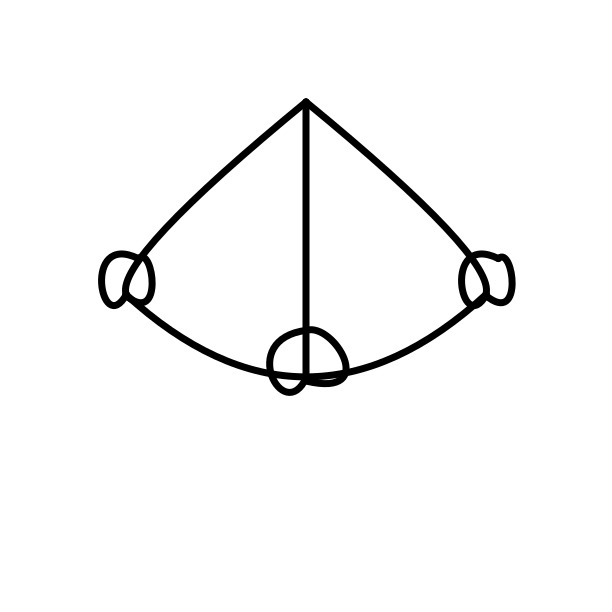} &
        \includegraphics[height=0.12\textwidth, width=0.15\textwidth]{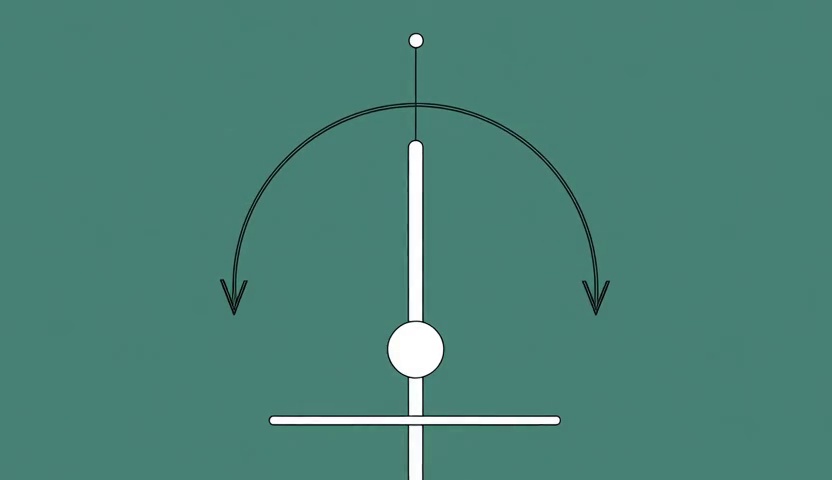} &
        \includegraphics[trim=4cm 0 4.5cm 0, clip, height=0.12\textwidth]{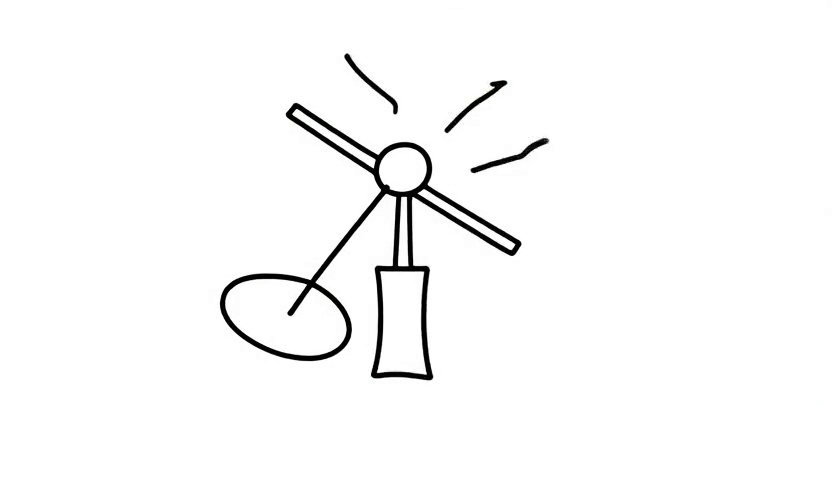} \\
        
         \lefttxt{``Photo-\\synthesis''} &
        \includegraphics[height=0.12\textwidth]{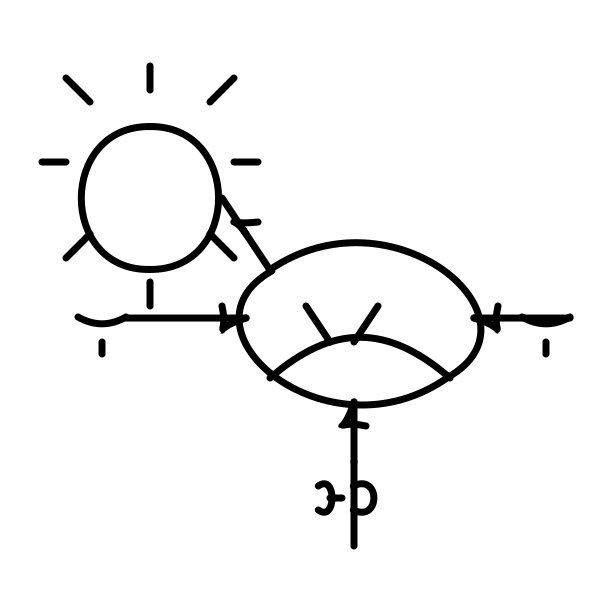} &
        \includegraphics[height=0.12\textwidth, width=0.15\textwidth]{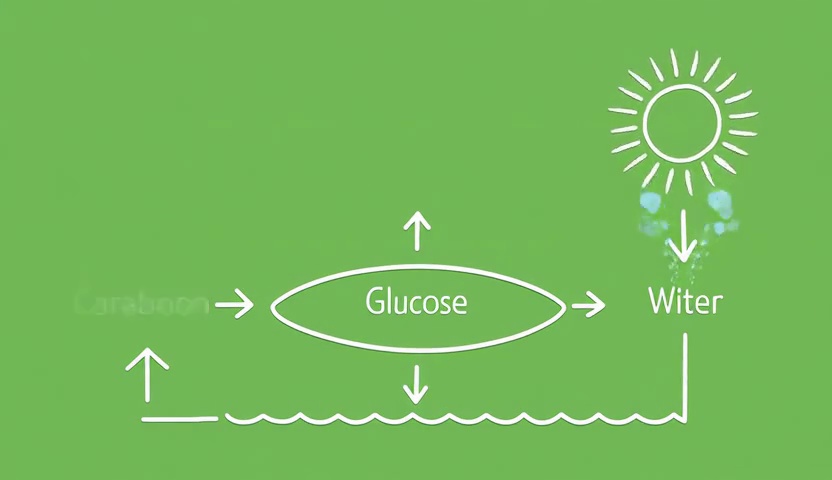} &
        \includegraphics[trim=4cm 0 4.5cm 0, clip, height=0.12\textwidth]{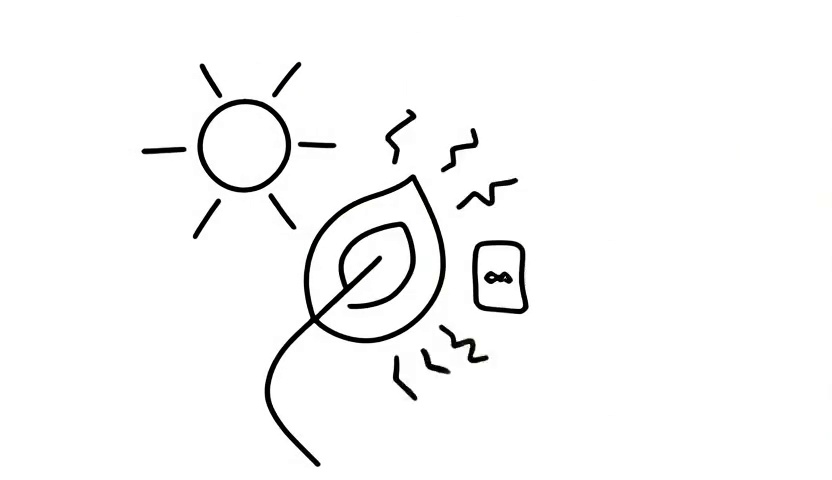} \\
        
       \lefttxt{``Newton's\\laws of\\motion''} &
        \includegraphics[height=0.12\textwidth]{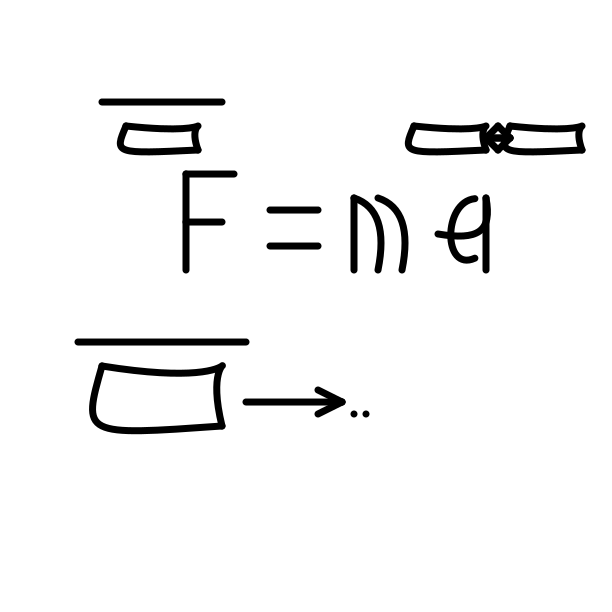} &
        \includegraphics[height=0.12\textwidth, width=0.15\textwidth]{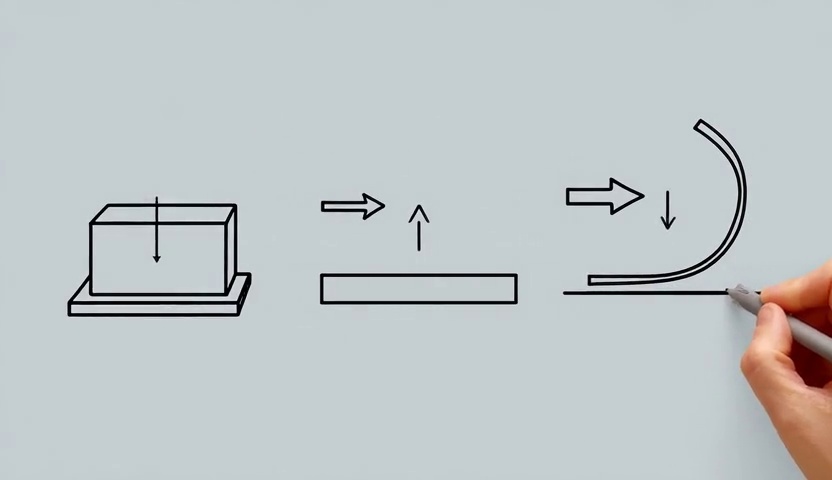} &
        \includegraphics[trim=4cm 0 4.5cm 0, clip, height=0.12\textwidth]{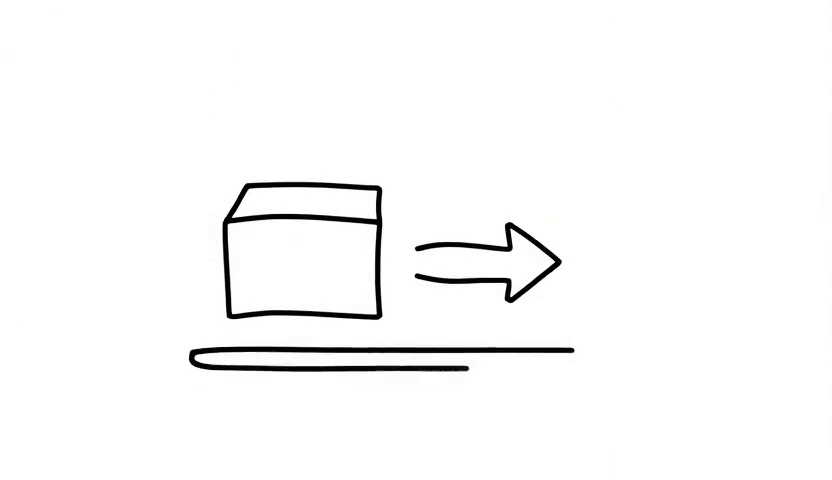} \\
    \end{tabular}
    \caption{\textbf{Scientific Concepts.} Representative results for scientific concepts across methods. SketchAgent leverages LLM knowledge to capture conceptual structure (e.g., the interference pattern in double-slit, F=ma in Newton's laws), though with limited visual detail. Wan 2.1 occasionally produces informative diagrams but often includes colored backgrounds and text labels rather than sketch-style outputs. Our method produces visually coherent sketches, inheriting both the strengths and limitations of the base model—performing well on concepts like photosynthesis while struggling with more abstract ones like pendulum motion.}
    \Description{Comparison grid of final sketches from Ours, SketchAgent, and Wan~2.1 across four general knowledge categories with four concepts each.}
    \label{fig:scientific}
\end{figure}

\section{Additional Experiments}

\vspace{0.5cm}
\subsection{Out-of-Distribution Concept Generation}
In this section we evaluate our method on concepts requiring specialized knowledge. We follow the experimental setup of SketchAgent~\cite{SketchAgent_Vinker2025}, where three categories requiring general knowledge are defined: Scientific Concepts, Diagrams, and Notable Landmarks, with ChatGPT used to produce 10 random textual concepts per category. We extend this setup by adding a Functions category, as functions can be thought of as drawings requiring specialized knowledge while being easier to evaluate for correctness. In summary, we use the following categories and concepts:\\

\begin{itemize}
    \item \textbf{Scientific Concepts:} Double-slit experiment, Pendulum motion, Photosynthesis, DNA replication, Newton's laws of motion, Electromagnetic spectrum, Plate tectonics, Quantum entanglement, Cell division (mitosis), Black hole formation. \vspace{0.2cm}
    
 \item \textbf{Diagrams:} Circuit diagram, Flowchart, Organizational chart, ER diagram (Entity-Relationship), Venn diagram, Mind map, Gantt chart, Network topology diagram, Pie chart, Decision tree. \vspace{0.2cm}
 \item \textbf{Notable Landmarks:} Taj Mahal, Eiffel Tower, Great Wall of China, Pyramids of Giza, Statue of Liberty, Colosseum, Sydney Opera House, Big Ben, Mount Fuji, Machu Picchu. \vspace{0.2cm}
 \item \textbf{Functions:} $y = x^2 y = \sqrt{x}, y = x^3, y = log(x), y = e^x, y = \frac{1}{x}, y = sin(x), y = |x|, y = 2x, y = x$.
\end{itemize}

\begin{figure}[h]
    \centering
    \small
    \setlength{\tabcolsep}{0pt}
    \begin{tabular}{cccc}
         & SketchAgent & Wan~2.1& Ours \\
        \lefttxt{``Circuit \\ diagram''} & 
        \includegraphics[height=0.12\textwidth]{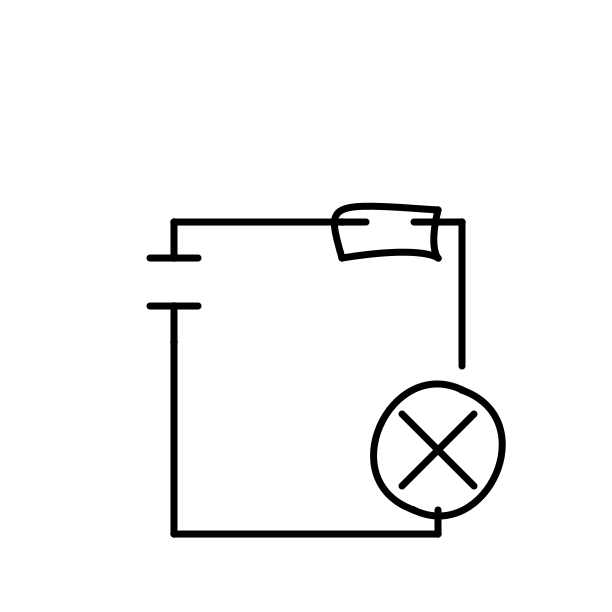} &
        \includegraphics[trim=4cm 0 4.5cm 0, clip, height=0.12\textwidth]{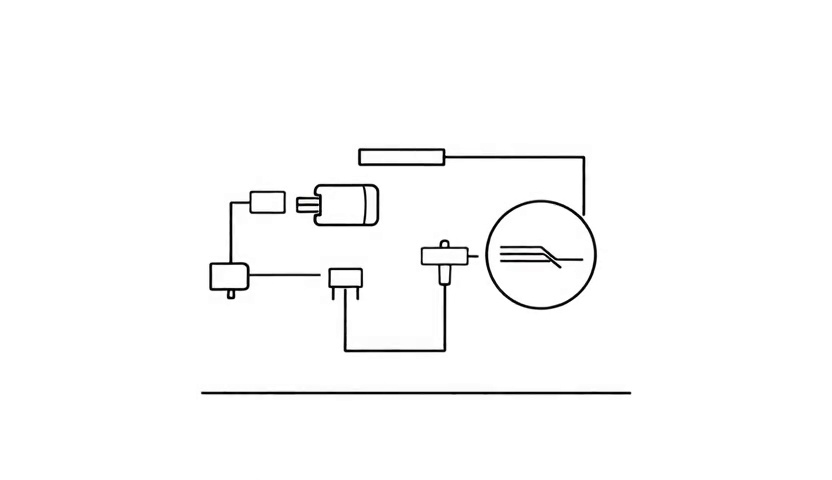} &
        \includegraphics[trim=2.5cm 0 6cm 0, clip, height=0.12\textwidth]{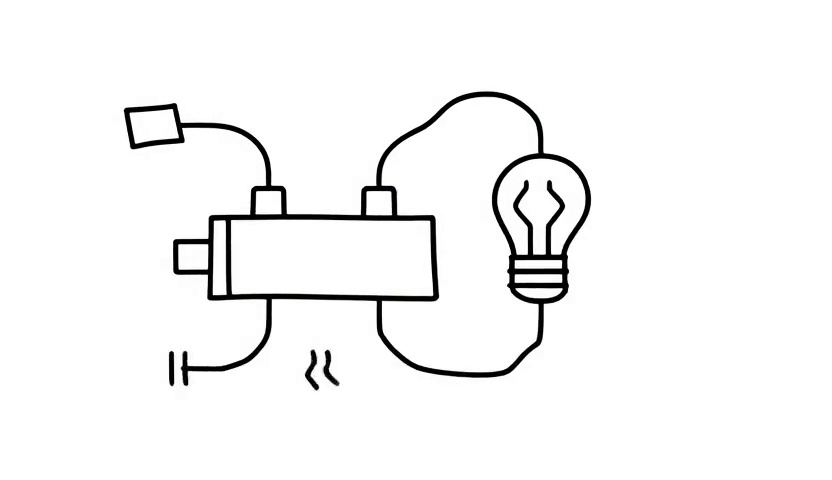} \\
        
        \lefttxt{``Flow\\-chart''} & 
        \includegraphics[height=0.12\textwidth]{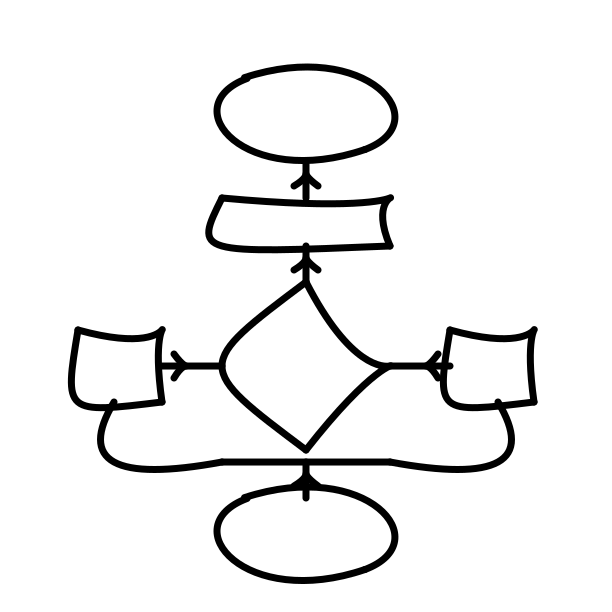} &
        \includegraphics[height=0.12\textwidth, width=0.15\textwidth]{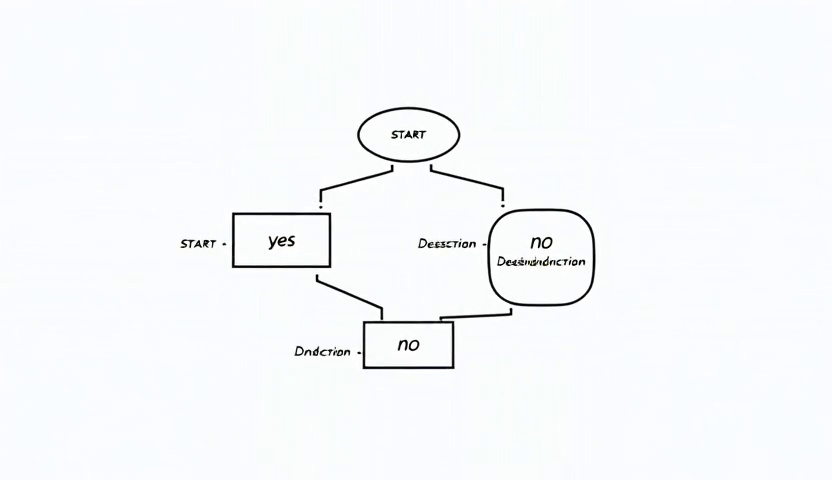} &
        \includegraphics[trim=4cm 0 4.5cm 0, clip, height=0.12\textwidth]{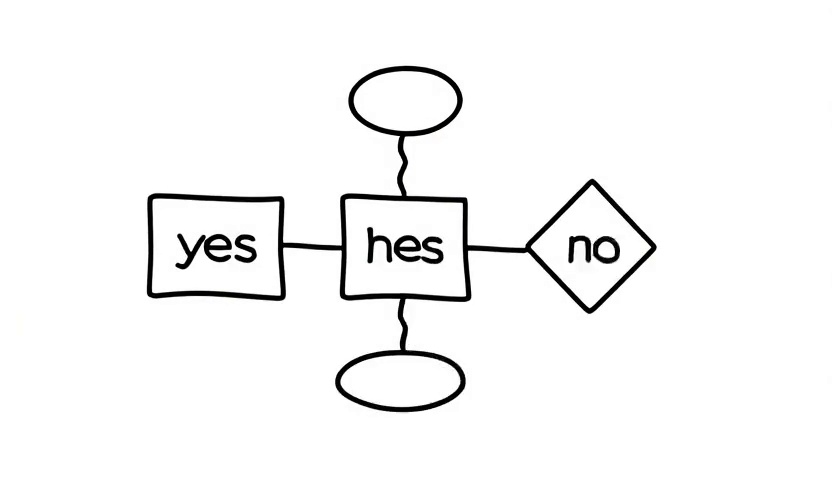} \\
        
        \lefttxt{``Venn \\ diagram''} &
        \includegraphics[height=0.12\textwidth]{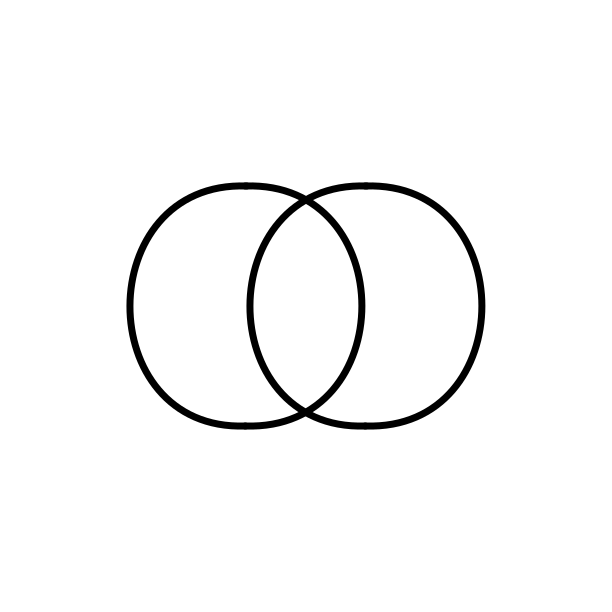} &
        \includegraphics[height=0.12\textwidth, width=0.15\textwidth]{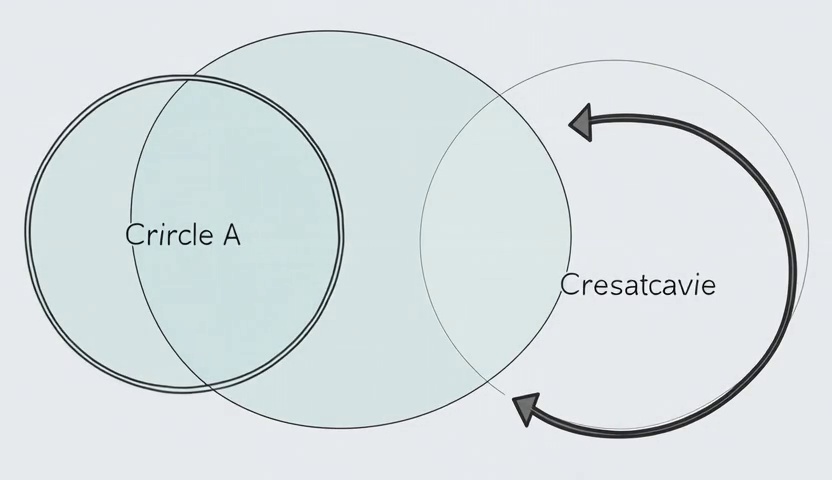} &
        \includegraphics[trim=3cm 0 5cm 0, clip, height=0.12\textwidth]{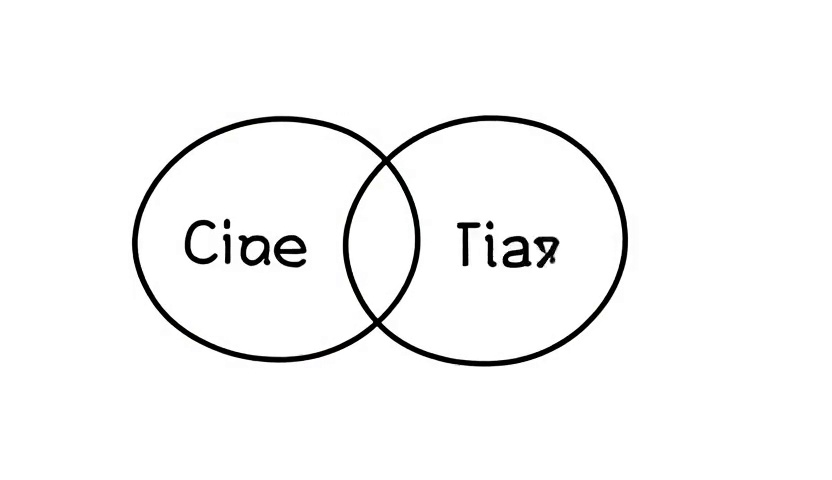} \\
        
         \lefttxt{``Mind  \\ map''} &
        \includegraphics[height=0.12\textwidth]{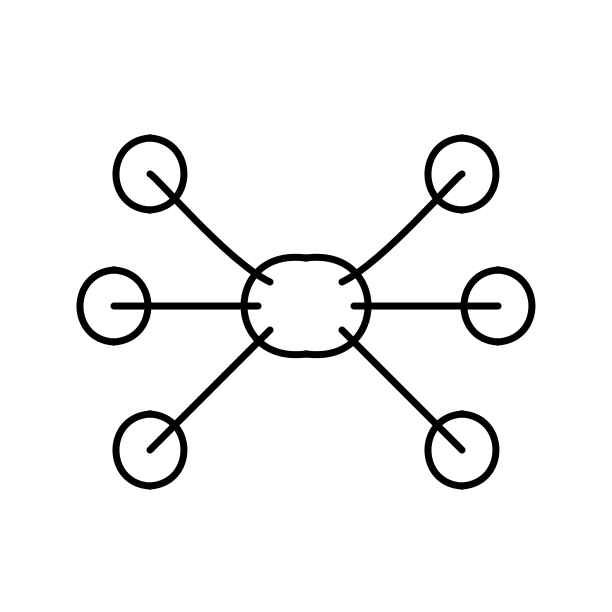} &
        \includegraphics[height=0.12\textwidth, width=0.15\textwidth]{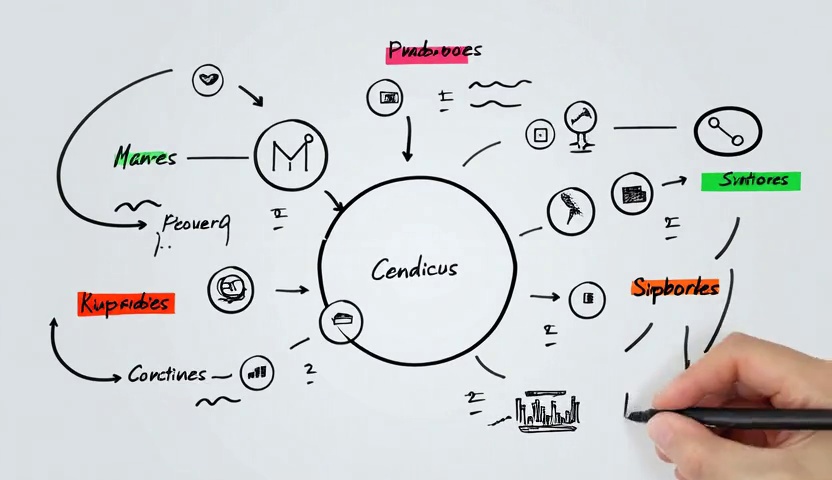} &
        \includegraphics[trim=3cm 0 5cm 0, clip, height=0.12\textwidth]{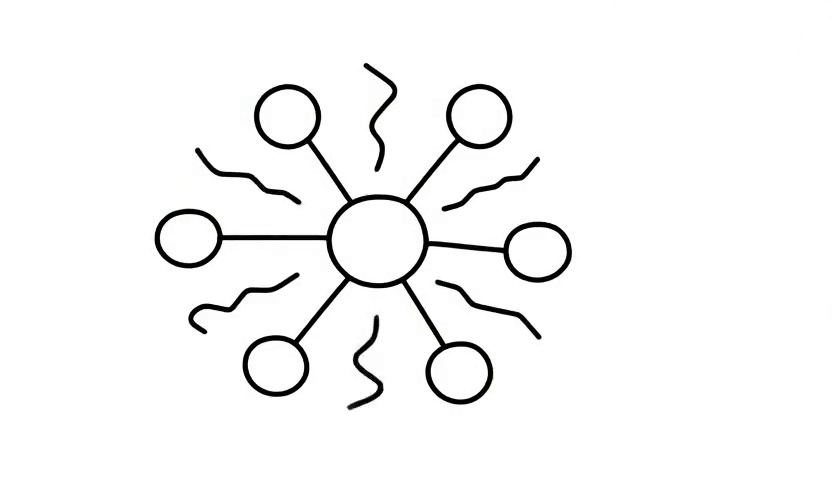} \\
        
        \lefttxt{``Pie  \\ chart''} &
        \includegraphics[height=0.12\textwidth]{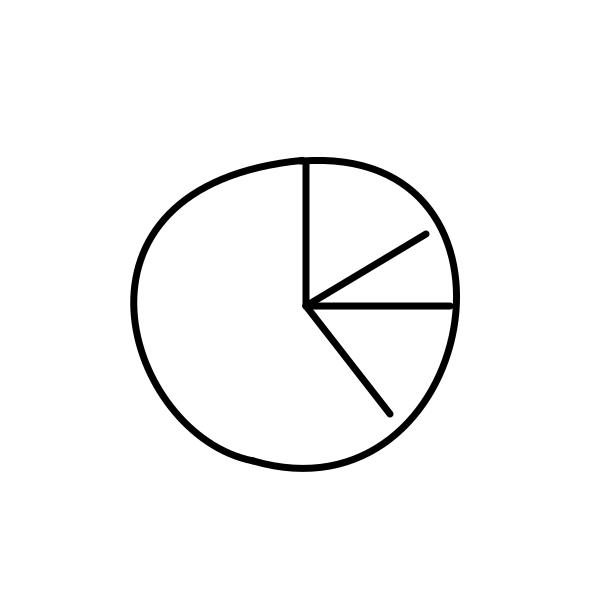} &
        \includegraphics[trim=4cm 0 4.5cm 0, clip, height=0.12\textwidth]{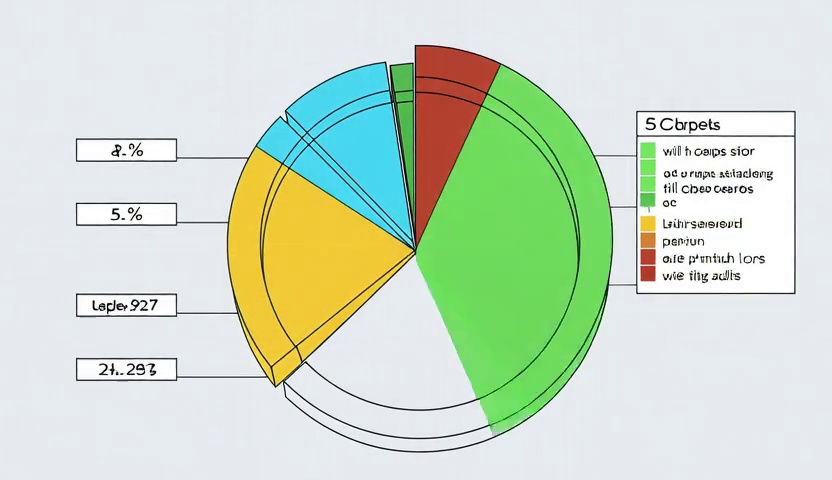} &
        \includegraphics[trim=1cm 0 6cm 0, clip, height=0.12\textwidth]{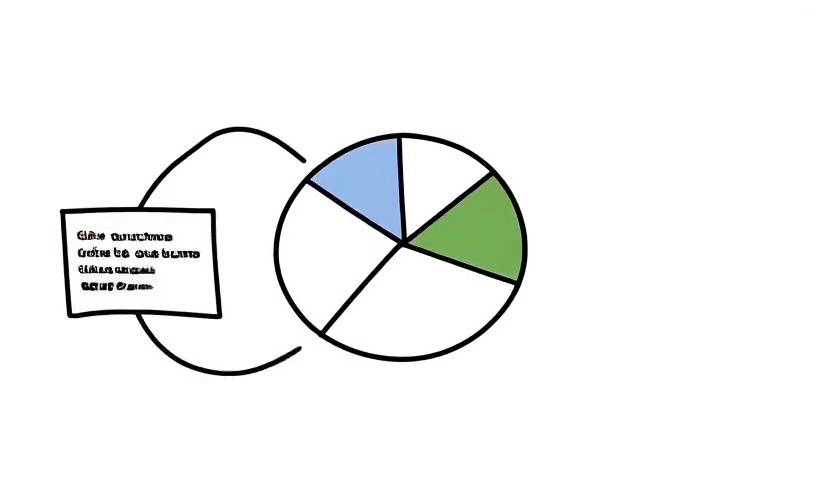} \\
        
    \end{tabular}
    \caption{\textbf{Diagrams.} Representative results for diagram concepts. SketchAgent produces structurally correct but visually minimal outputs. Wan 2.1 generates detailed diagrams with text and color, deviating from a sketch aesthetic. Our method captures the visual structure of diagrams (e.g., connected nodes in flowcharts and mind maps, overlapping circles in Venn diagrams) with a cleaner sketch style, though text elements are often garbled or nonsensical.}
    \Description{Comparison grid of final sketches from Ours, SketchAgent, and Wan~2.1 across four general knowledge categories with four concepts each.}
    \label{fig:diagrams}
\end{figure}

\begin{figure}[H]
    \centering
    \small
    \setlength{\tabcolsep}{0pt}
    \begin{tabular}{cccc}
        & SketchAgent & Wan~2.1 & Ours  \\
        \lefttxt{``Eiffel \\ Tower''} & 
        \includegraphics[height=0.12\textwidth]{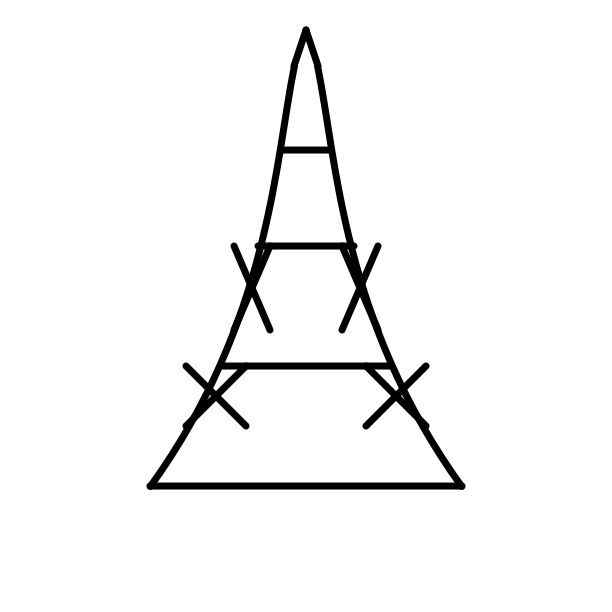} &
        \includegraphics[trim=4cm 0 4.5cm 0, clip, height=0.12\textwidth]{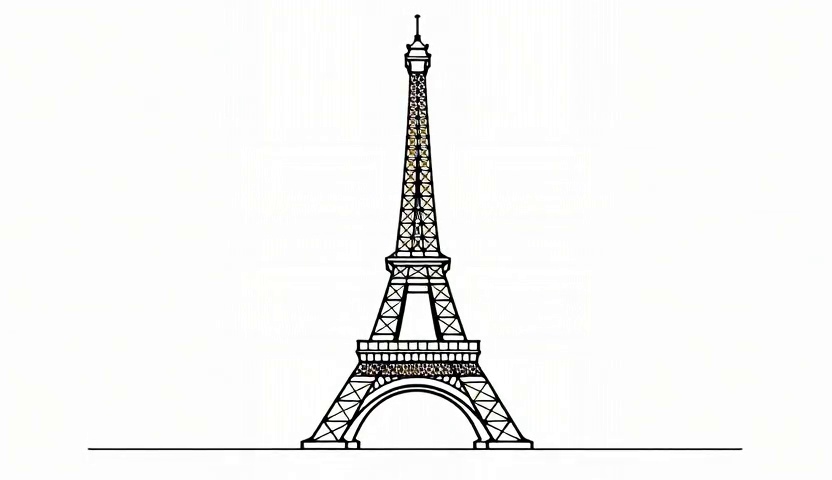} &
        \includegraphics[trim=4cm 0 4.5cm 0, clip, height=0.12\textwidth]{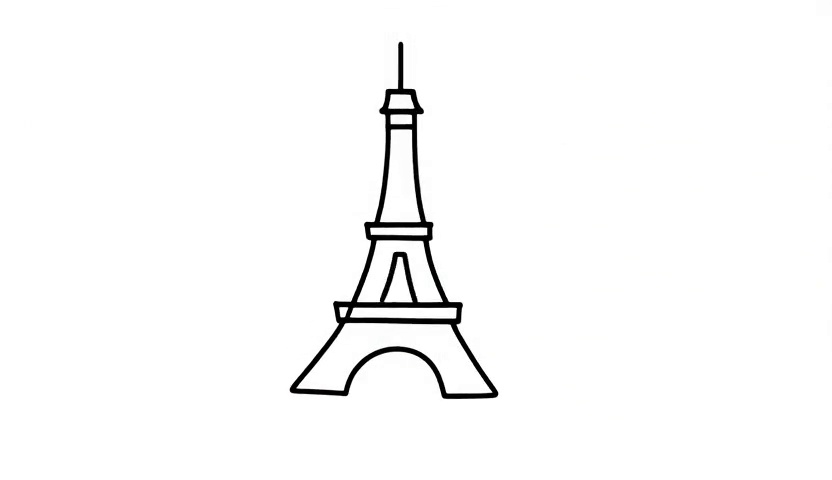} \\
        
        \lefttxt{``Statue of \\ Liberty''} & 
        \includegraphics[height=0.12\textwidth]{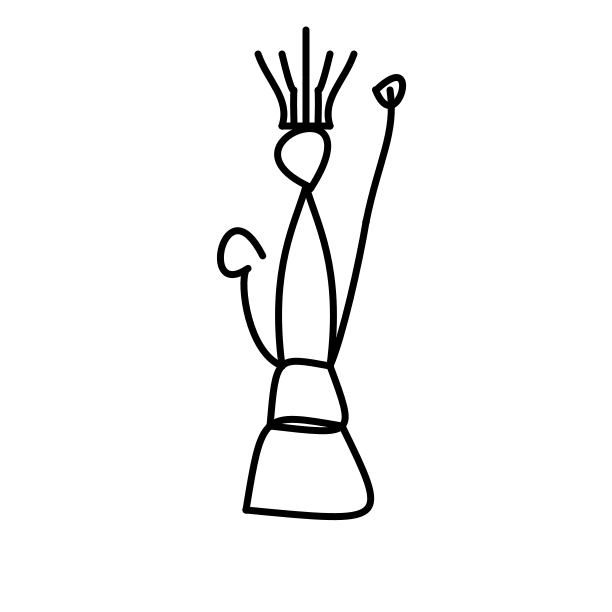} &
        \includegraphics[height=0.12\textwidth, width=0.15\textwidth]{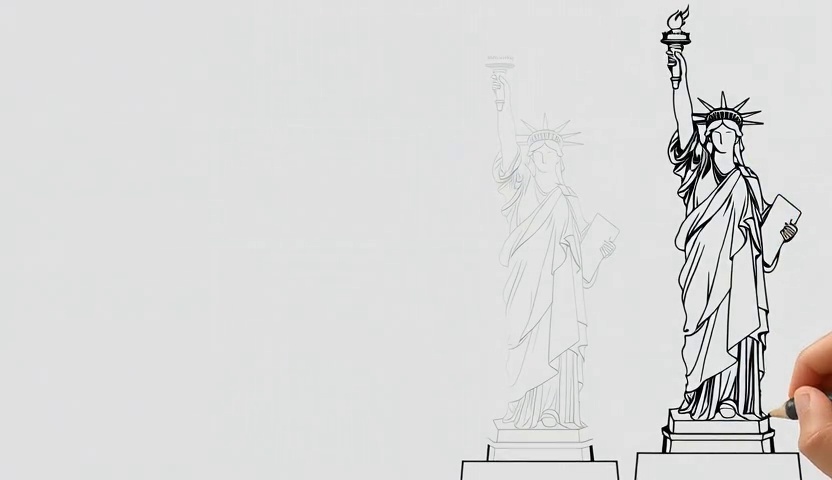} &
        \includegraphics[trim=4cm 0 4.5cm 0, clip, height=0.12\textwidth]{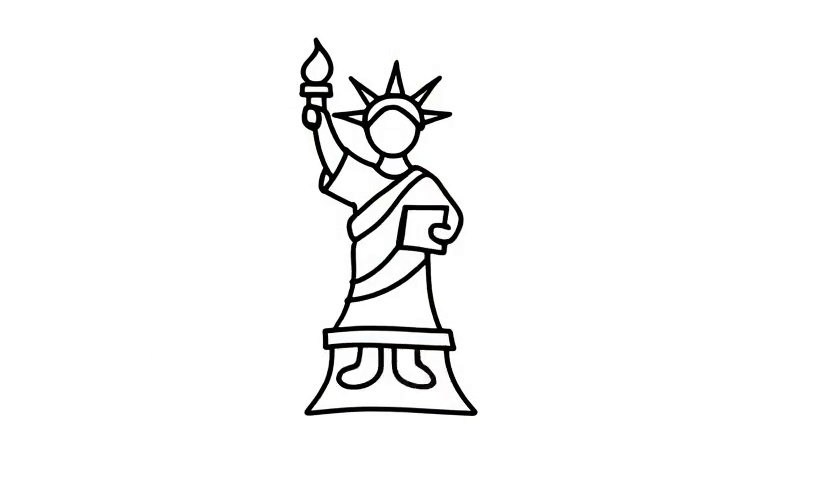} \\

        \lefttxt{``Sydney Opera  \\ House''} &
        \includegraphics[height=0.12\textwidth]{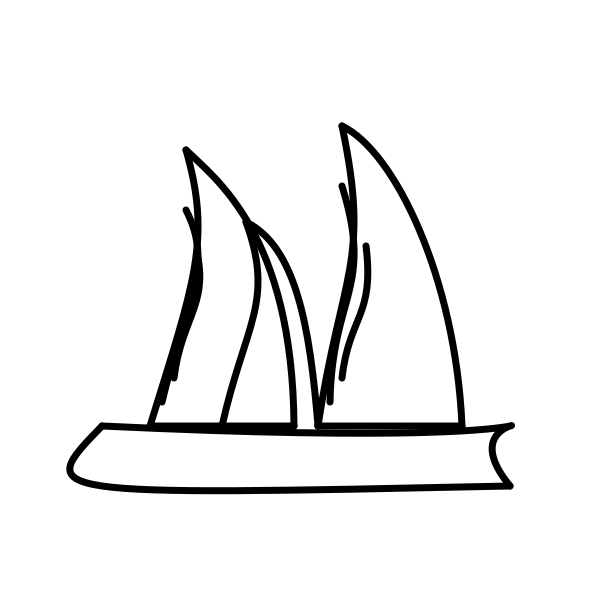} &
        \includegraphics[height=0.12\textwidth, width=0.15\textwidth]{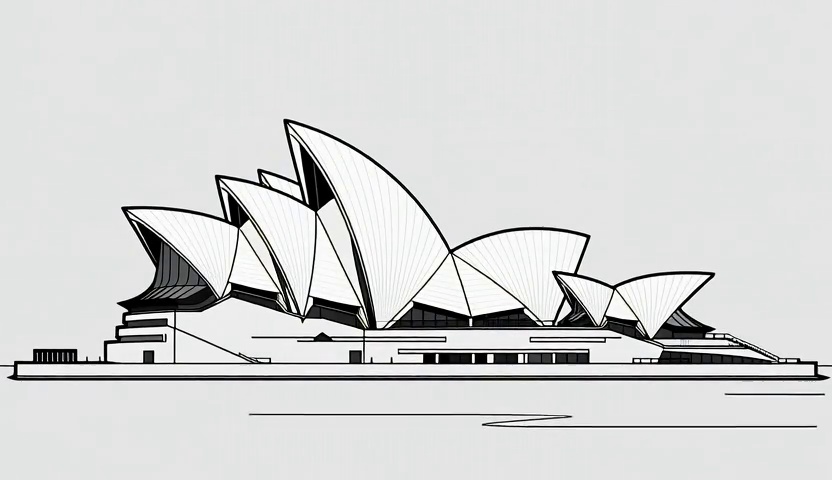} &
        \includegraphics[trim=3cm 0 5cm 0, clip, height=0.12\textwidth]{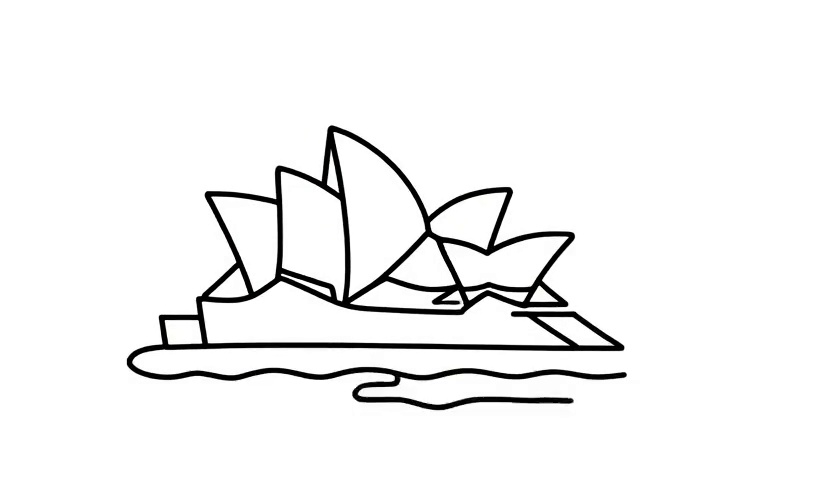} \\
        
         \lefttxt{``Taj  \\ Mahal''} &
        \includegraphics[height=0.12\textwidth]{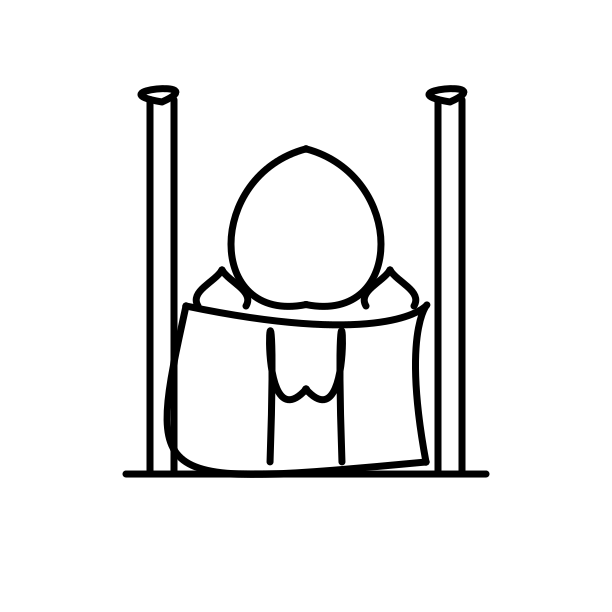} &
        \includegraphics[height=0.12\textwidth, width=0.15\textwidth]{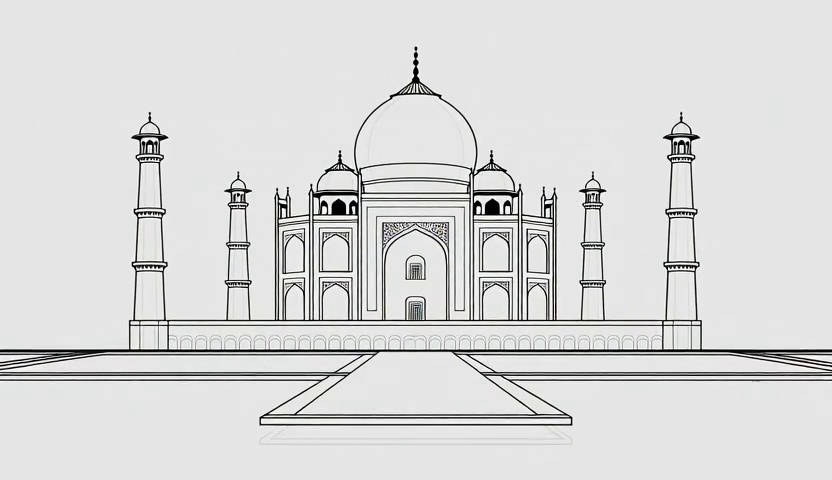} &
        \includegraphics[trim=4cm 0 4.5cm 0, clip, height=0.12\textwidth]{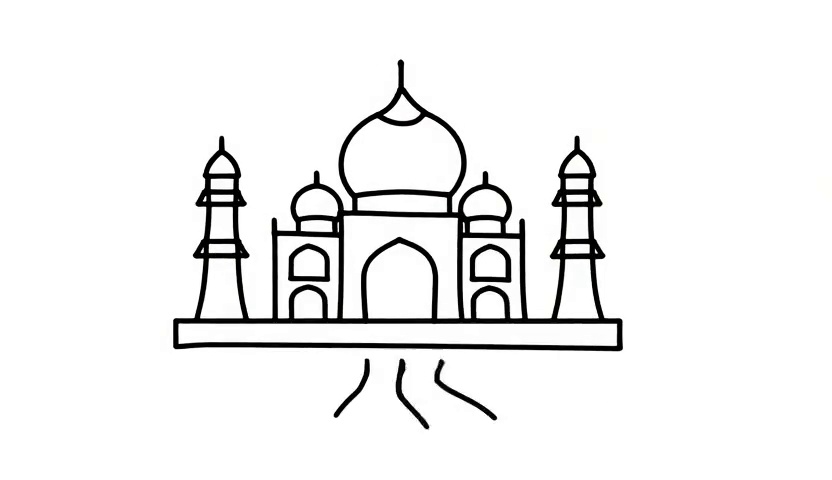} \\

        \lefttxt{``Great Wall \\ of China''} &
        \includegraphics[height=0.12\textwidth]{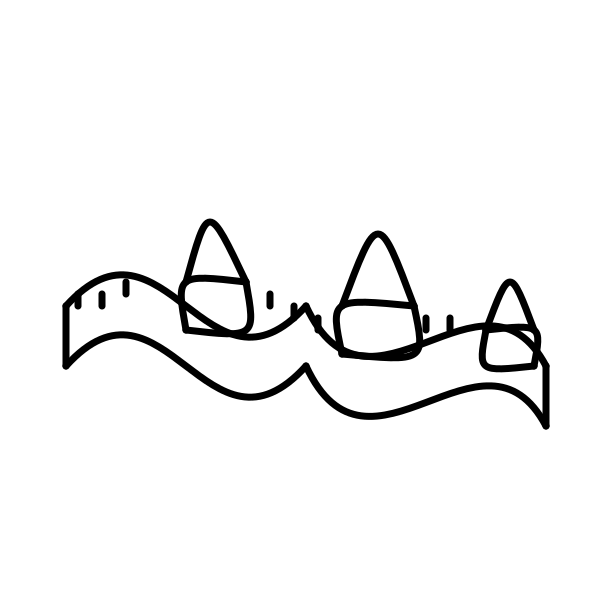} &
        \includegraphics[trim=4cm 0 4.5cm 0, clip, height=0.12\textwidth]{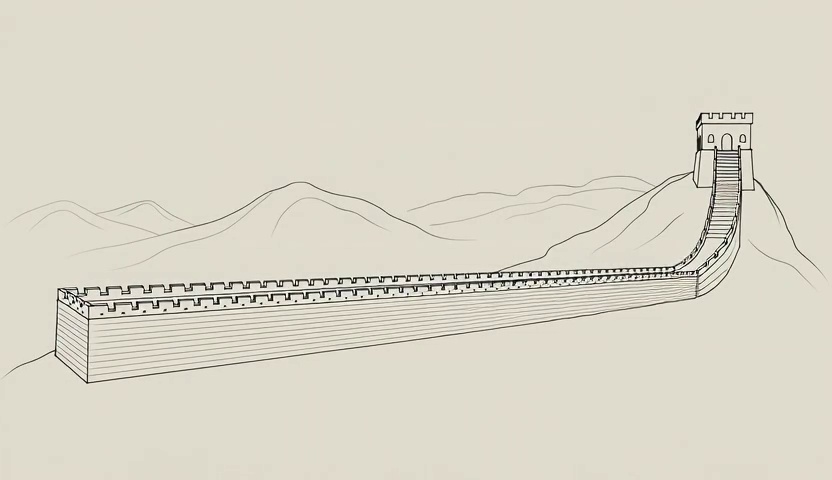} &
        \includegraphics[trim=4cm 0 4.5cm 0, clip, height=0.12\textwidth]{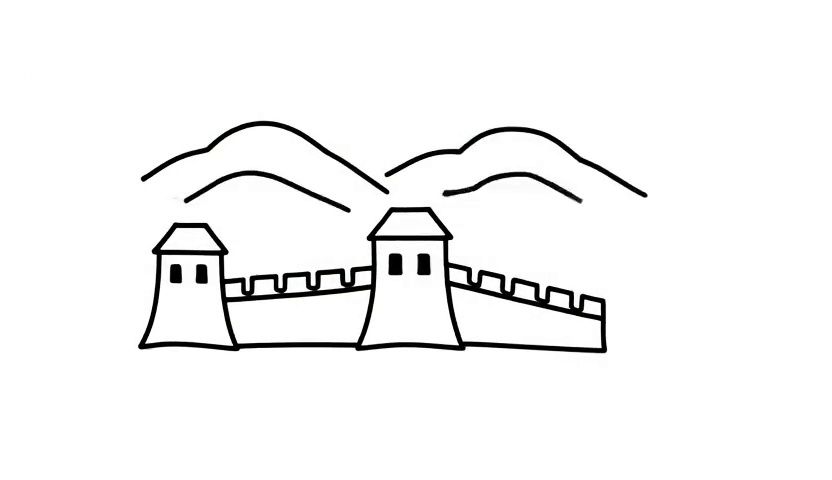} \\
        
    \end{tabular}
    
    \caption{\textbf{Notable Landmarks.} Representative results for landmark concepts. This category highlights the strength of video-based approaches: both Wan 2.1 and our method produce detailed, highly recognizable depictions, reflecting the strong prior knowledge of landmarks in video training data. In contrast, SketchAgent's outputs are overly simplistic and often fail to capture the iconic features (e.g., the Eiffel Tower reduced to basic triangles, the Statue of Liberty barely recognizable).}
    \Description{Comparison grid of final sketches from Ours, SketchAgent, and Wan~2.1 across four general knowledge categories with four concepts each.}
    \label{fig:landmarks}
\end{figure}

\begin{figure}[H]
    \centering
    \small
    \setlength{\tabcolsep}{0pt}
    \begin{tabular}{cccc}
         & SketchAgent & Wan~2.1& Ours \\
        \lefttxt{``$y=x^3$''} & 
        \includegraphics[height=0.12\textwidth]{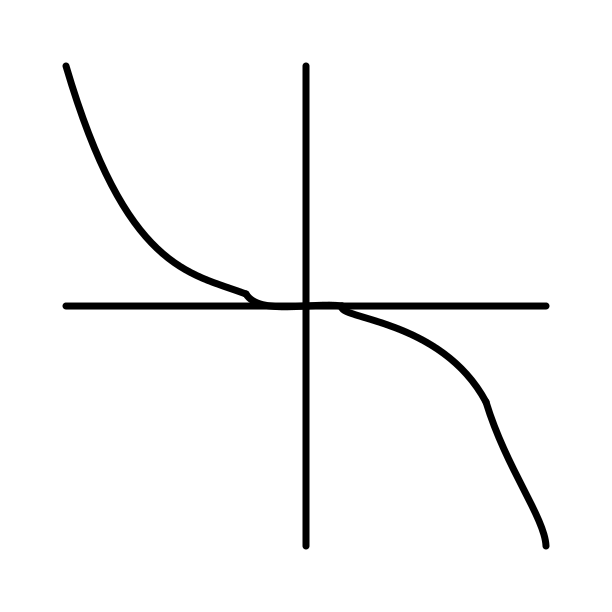} &
        \includegraphics[trim=4cm 0 4.5cm 0, clip, height=0.12\textwidth]{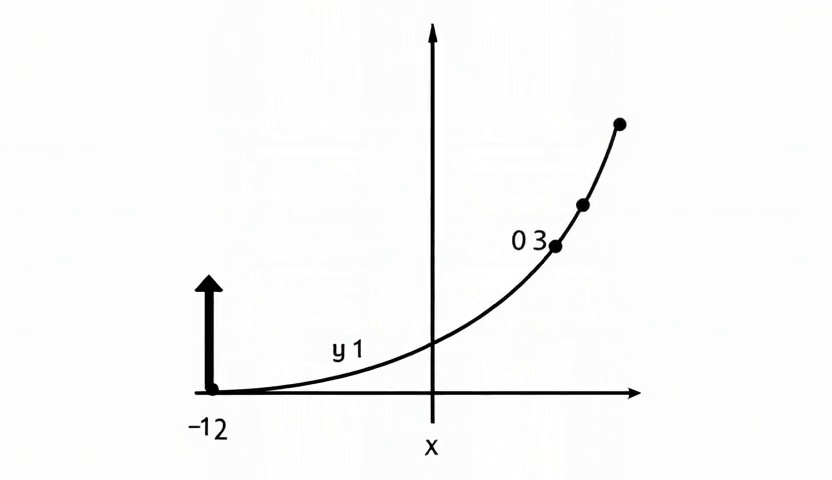} &
        \includegraphics[trim=1cm 0 6cm 0, clip, height=0.12\textwidth]{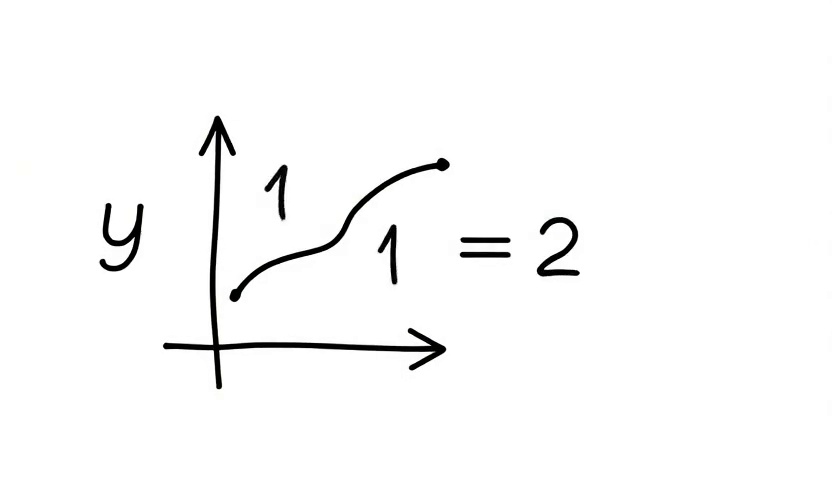} \\
        
        \lefttxt{``$y=\log(x)$''} & 
        \includegraphics[height=0.12\textwidth]{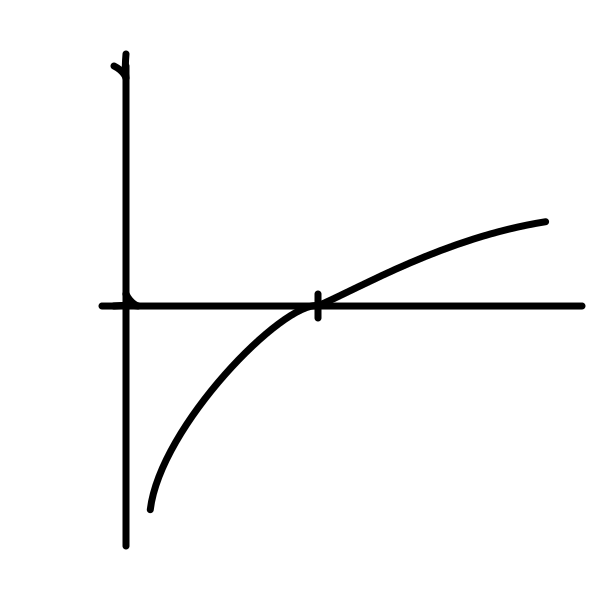} &
        \includegraphics[trim=4cm 0 4.5cm 0, clip, height=0.12\textwidth]{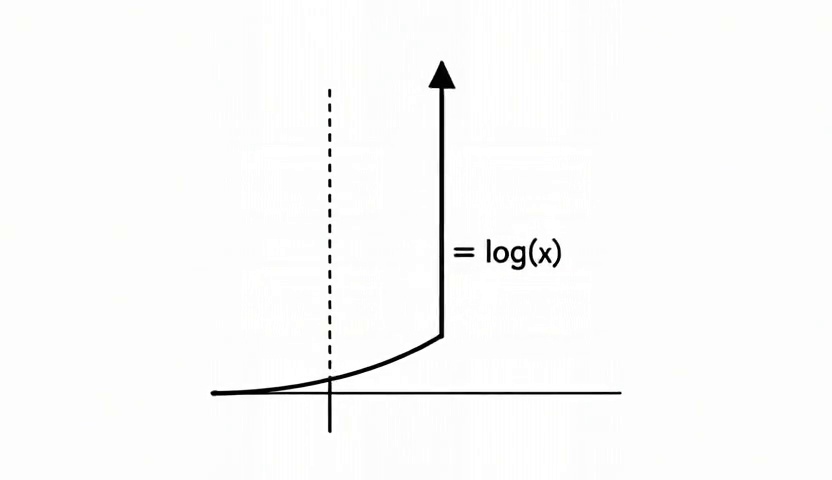} &
         \includegraphics[trim=3cm 0 4.5cm 0, clip, height=0.12\textwidth]{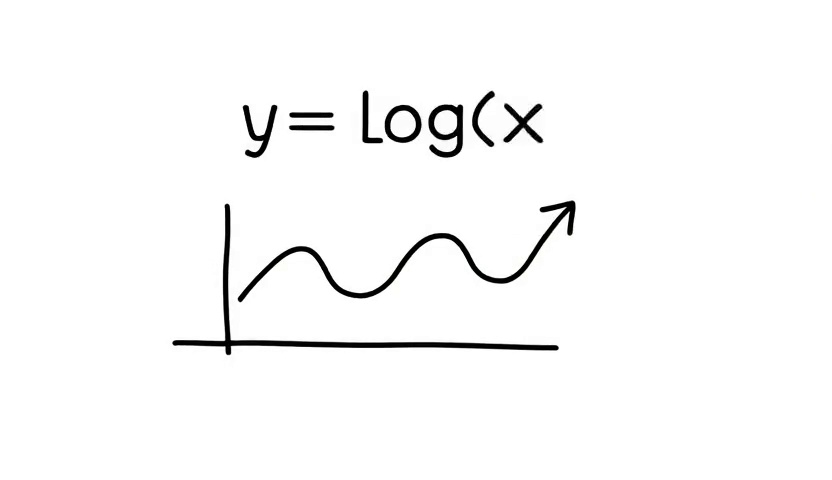} \\
        
        \lefttxt{``$y=x^2$''} &
        \includegraphics[height=0.12\textwidth]{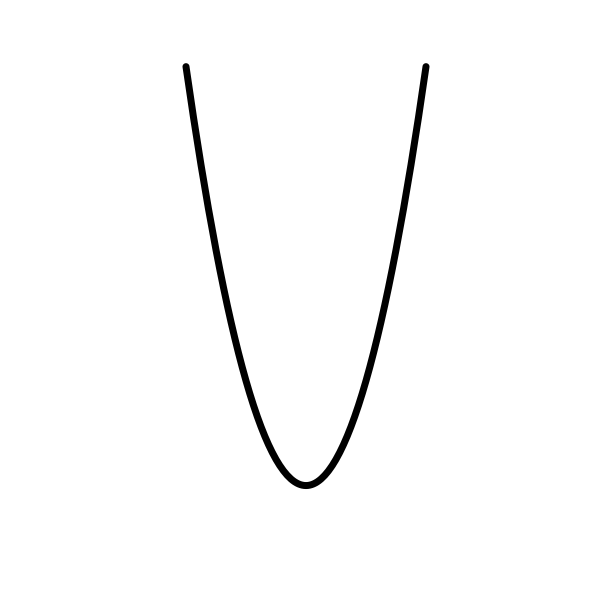} &
        \includegraphics[trim=4cm 0 4.5cm 0, clip, height=0.12\textwidth]{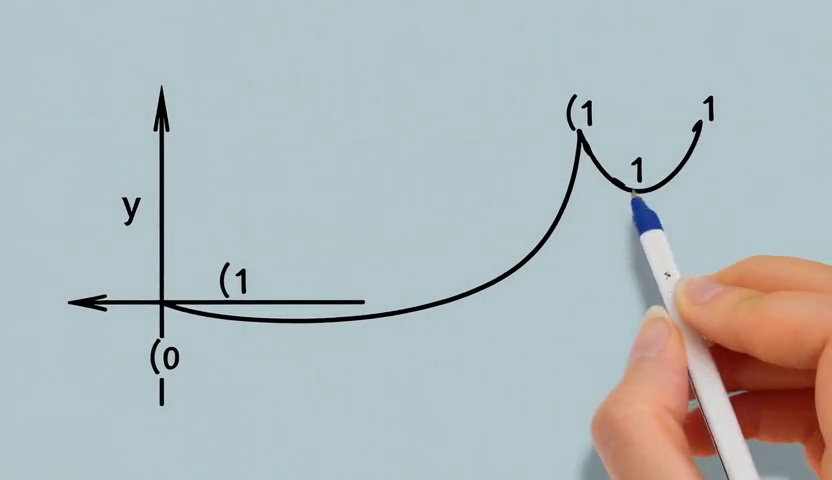} &
       \includegraphics[trim=2cm 0 4.5cm 0, clip, height=0.12\textwidth]{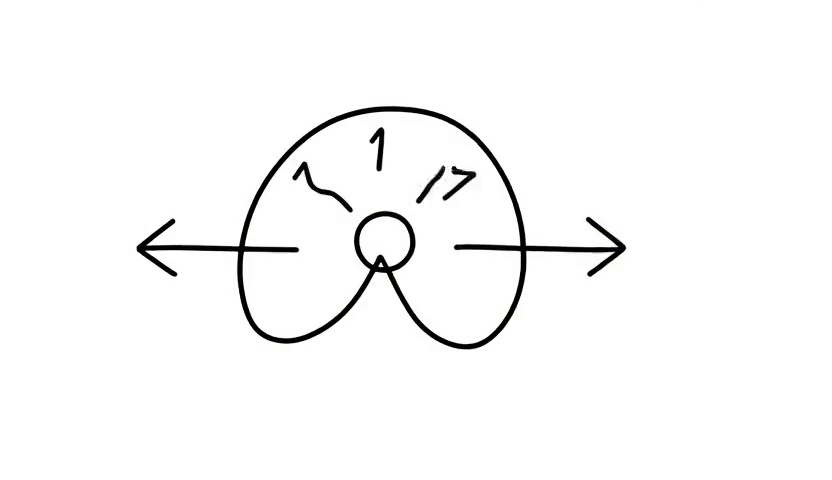} \\
        
         \lefttxt{``$y=\sin(x)$''} &
        \includegraphics[height=0.12\textwidth]{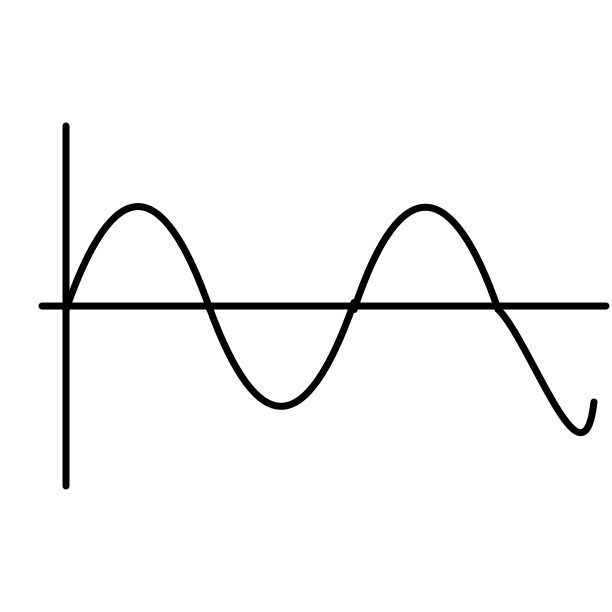} &
        \includegraphics[trim=4cm 0 4.5cm 0, clip, height=0.12\textwidth]{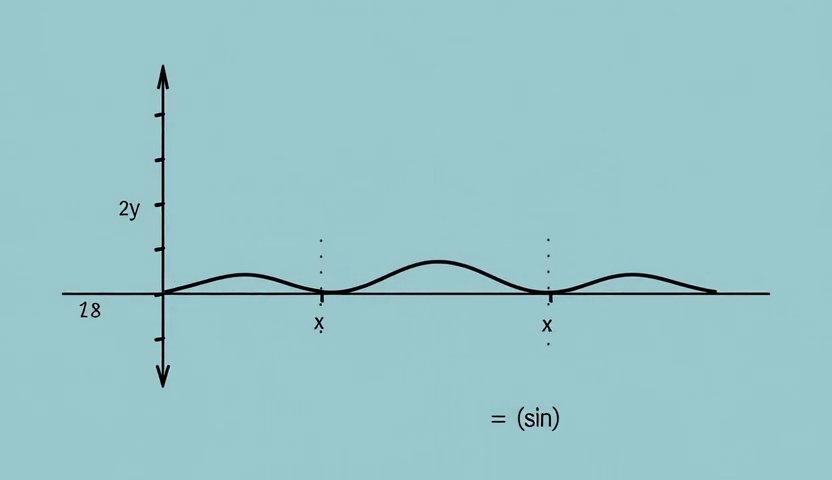} &
        \includegraphics[trim=1.5cm 0 4.5cm 0, clip, height=0.12\textwidth]{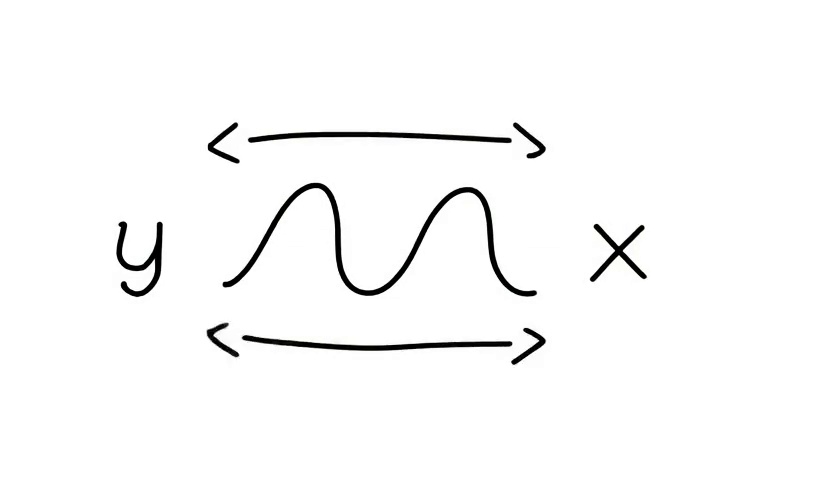} \\
        
       \lefttxt{``$y=e^x$''} &
        \includegraphics[height=0.12\textwidth]{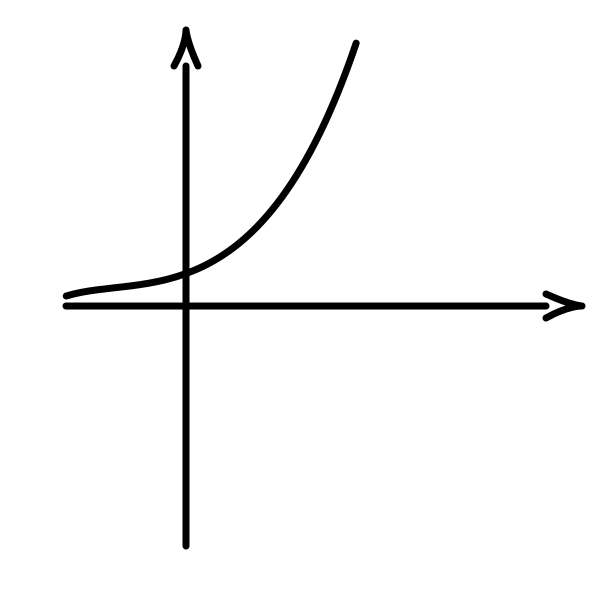} &
        \includegraphics[trim=4cm 0 4.5cm 0, clip, height=0.12\textwidth]{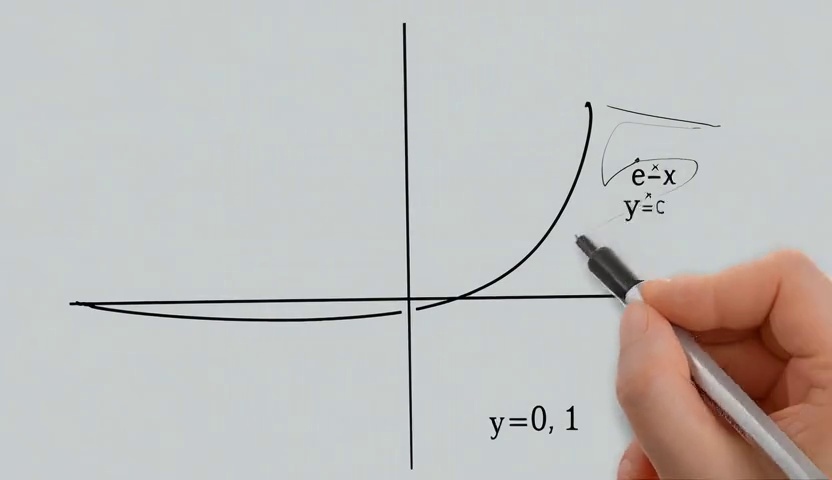} &
        \includegraphics[trim=4cm 0 4.5cm 0, clip, height=0.12\textwidth]{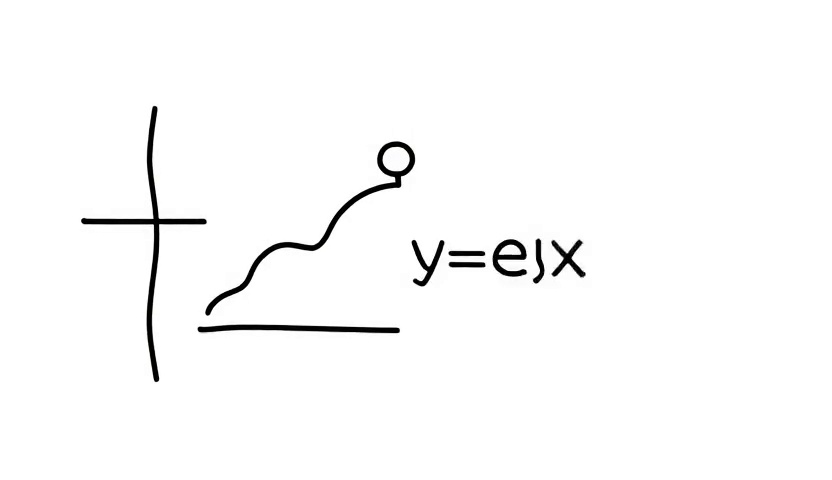} \\
        
    \end{tabular}
    \caption{\textbf{Functions.} Representative results for mathematical functions. This category most clearly demonstrates the advantage of LLM-based approaches: SketchAgent produces precise, mathematically correct curves due to its language model backbone. Both Wan 2.1 and our method struggle—curves are often incorrect or unrecognizable, and text/equations are garbled (e.g., $y = log(x)$ in our output). This reveals a fundamental limitation of video models for concepts requiring symbolic or mathematical knowledge.}
    \Description{Comparison grid of final sketches from Ours, SketchAgent, and Wan~2.1 across four general knowledge categories with four concepts each.}
    \label{fig:functions}
\end{figure}

For each concept, we generate three random sketch sequences using our method and compare with SketchAgent and Wan2.1 (as a baseline, without any fine-tuning). Representative results showing the last frame of each produced video are presented in \Cref{fig:scientific,fig:diagrams,fig:landmarks,fig:functions}.

The Wan2.1 results reveal what the video model already knows prior to fine-tuning. Concepts familiar to the base model will more naturally transfer to our fine-tuned model. This is evident in the Landmarks category (\Cref{fig:landmarks}), where our results are highly detailed and recognizable, reflecting the model's prior knowledge. For Scientific Concepts, the pattern is more nuanced: where the base model is limited (e.g., pendulum motion), our model inherits these limitations, while for concepts like Newton's laws and photosynthesis, our model performs well.

This experiment also highlights the complementary strengths of video-based and LLM-based approaches. For Landmarks (\Cref{fig:landmarks}), SketchAgent produces overly simplistic outputs with low visual quality (e.g., Eiffel Tower, Statue of Liberty), while ours are detailed and recognizable. However, for Scientific Concepts (\Cref{fig:scientific}), the strong priors of LLMs enable SketchAgent to capture the correct structure and rules — achieving correctness despite lower visual aesthetics. This contrast is most apparent in the Functions category: both Wan2.1 and our method fail to draw functions properly, while SketchAgent's LLM backbone enables more precise results.

\begin{figure*}[t]
    \centering
    \includegraphics[width=0.46\linewidth]{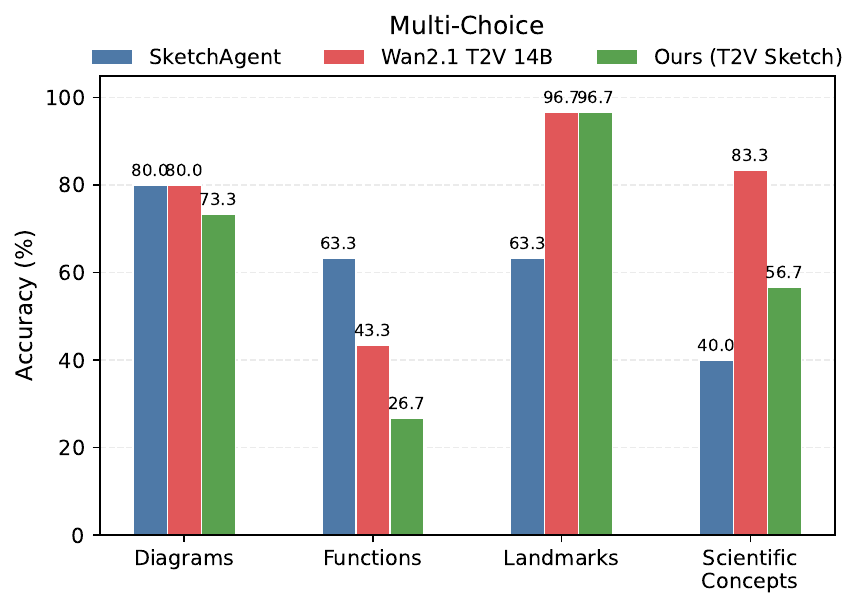}
    \includegraphics[width=0.46\linewidth]{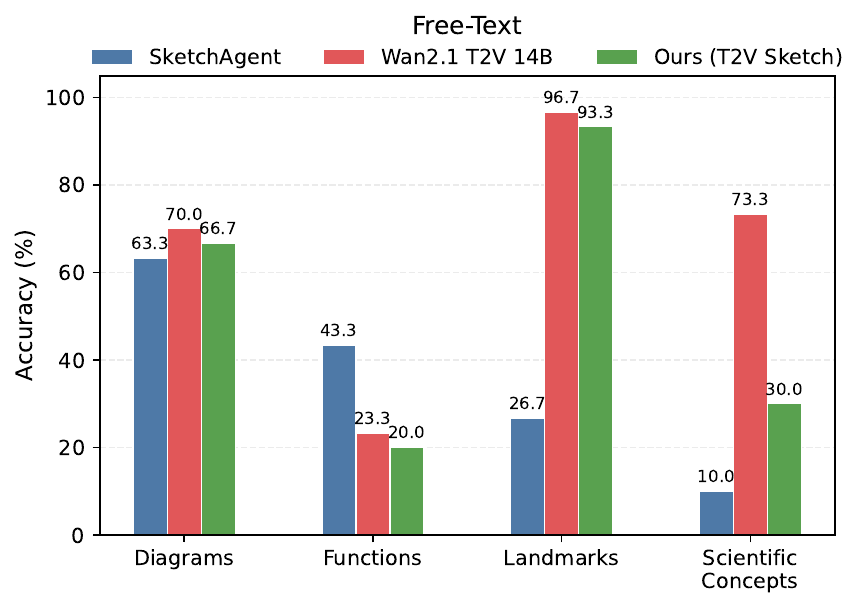}
    \vspace{-0.5cm}
    \caption{\textbf{Classification accuracy across categories.} Video-based methods excel on Landmarks ($~97\%$ vs. $63.3\%$ for SketchAgent), while SketchAgent dominates Functions due to its LLM backbone. Our method's accuracy closely tracks Wan 2.1, confirming that both knowledge and limitations transfer from the base model.}
    \label{fig:gk_per_category_accuracy}
\end{figure*}

We additionally assess recognizability quantitatively via classification with GPT-4o under two settings: (1) \emph{multi-choice}, where the model selects among the 10 category concepts or ``none'' (easier setting), and (2) \emph{free-text}, where the model describes the sketch without provided options and we analyze whether the output matches the class (stricter setting). Results are reported in \Cref{fig:gk_per_category_accuracy}. The bar chart reflects the patterns observed visually: there is clear correlation between base model success (red) and our model (green), where knowledge can be lost (as in Scientific Concepts), while SketchAgent struggles on Landmarks and concepts requiring high detail and visual quality.

\subsection{Sketch Progression} 
\begin{figure}[t]
    \centering
    \includegraphics[width=1\linewidth]{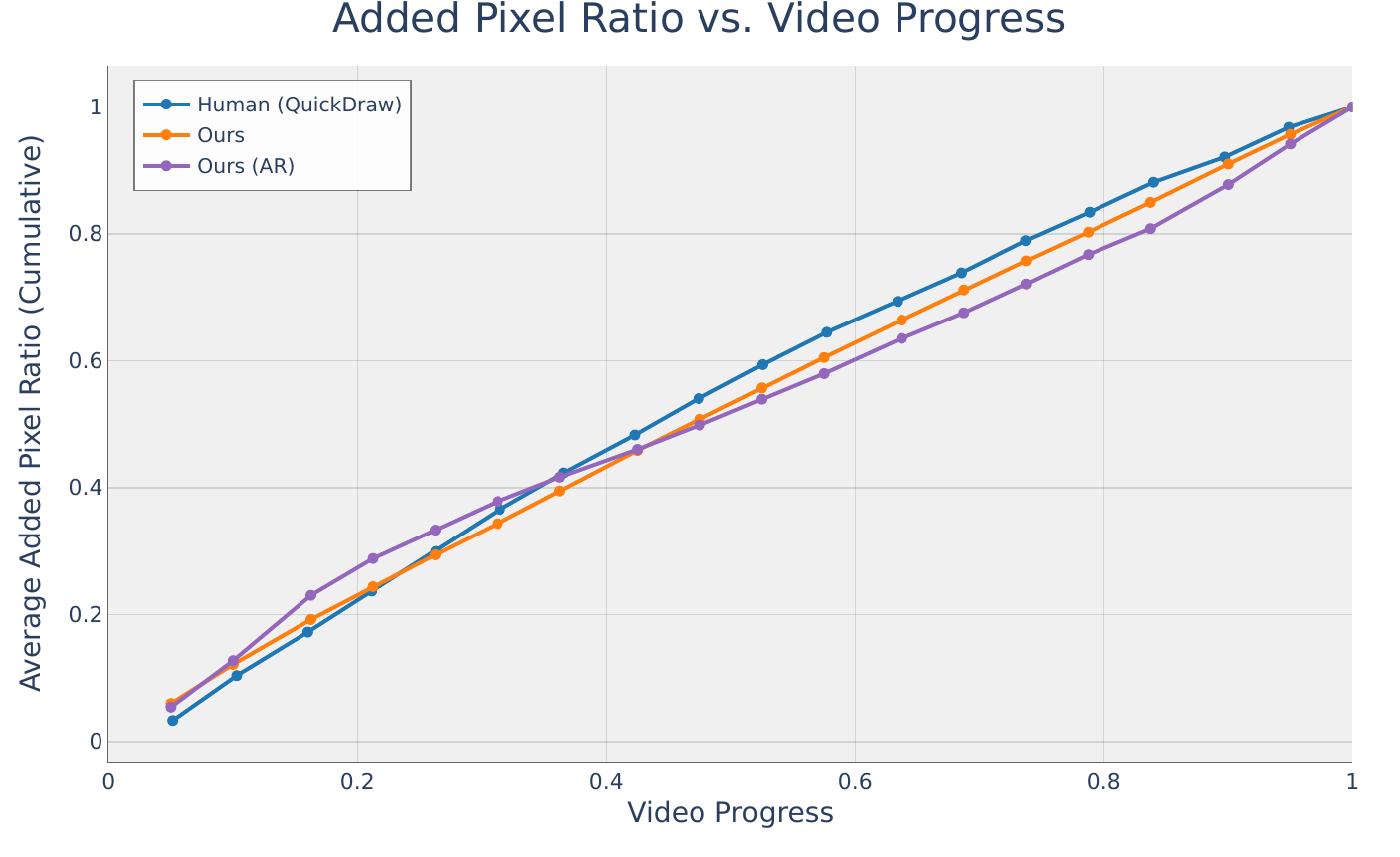} \\[-0.35cm]
    \caption{\textbf{Accumulated ratio of newly added pixels as a function of video progress.} Values are normalized such that the final frame equals 1. Our method exhibits a smooth and steady accumulation curve, reflecting incremental stroke additions over time and closely mirroring human drawing behavior.}
    \label{fig:pixels_added_over_time}
\end{figure}

To verify that generated sketches unfold sequentially rather than collapsing temporally, we measure the number of newly added pixels at each frame throughout the sketching process. Specifically, we compute the cumulative ratio of added pixels as a function of video progress, normalized so that the final frame equals 1 (see \Cref{fig:pixels_added_over_time}).
Both our diffusion-based and autoregressive models exhibit smooth, gradual accumulation curves, indicating that strokes are introduced incrementally across frames in a manner consistent with human drawing behavior. In contrast, baseline methods tend to introduce a large fraction of pixels early in the sequence, as also reflected in the qualitative results.

\subsection{Multi-Stroke Emergence Evaluation}
Because our models generate sketches directly in pixel space, they do not explicitly enforce stroke continuity. As a result, a single frame may contain multiple disjoint strokes that emerge simultaneously. We quantify this effect by computing the percent of frames that contain multiple strokes, reported in \Cref{tab:stroke_continuity}.

We evaluate this behavior on two sets: 100 QuickDraw object concepts, which can be considered relatively simple (e.g., cake, ice cream, pants), and 40 scene-level prompts, which are more complex (e.g., a Paris street at dusk). For each concept, we generate two sketch videos with different random seeds, resulting in 200 QuickDraw videos and 80 scene videos in total. As shown in \Cref{tab:stroke_continuity}, multi-stroke behavior occurs less frequently for QuickDraw concepts than for scene-level prompts, suggesting that increased concept complexity --- typically requiring a larger number of strokes --- makes this phenomenon more pronounced. Exploring models that operate over longer video lengths, as well as mechanisms that more strongly encourage stroke continuity in pixel space, may help distribute stroke generation more evenly over time and mitigate this effect, particularly for complex scenes.

\begin{table}[t]
\centering
\small
\caption{\textbf{Stroke continuity evaluation.} We measure the frequency of frames containing multiple disjoint strokes on QuickDraw object concepts (typically simpler concepts) and scene-level prompts (typically more complex concepts).\\[-0.65cm]}
\begin{tabular}{lccccc}
\toprule
Evaluation Set & Num. & Total & Multi-Stroke & Ratio $\downarrow$ \\
 & Samples & Frames & Frames &  \\
\midrule
QuickDraw Concepts & 200 & 15196 & 2975 & 19.58\% \\
Scenes & 80 & 6361 & 2351 & 36.96\% \\
\bottomrule
\end{tabular}
\label{tab:stroke_continuity}
\end{table}

\subsection{Interactive Sketching Interface}
\begin{figure}
\centering
    \includegraphics[width=1\linewidth]{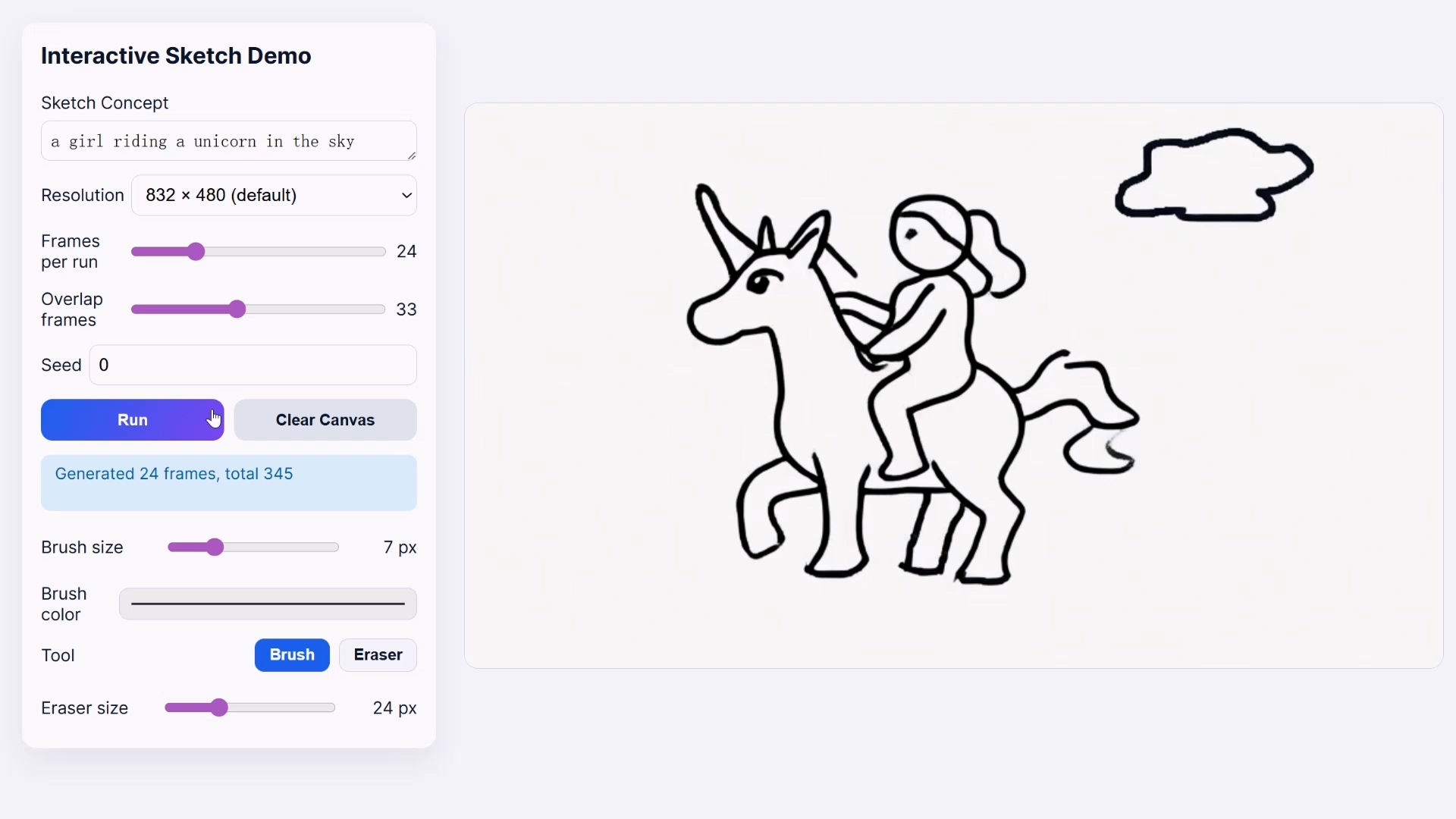} \\[-0.2cm]
    \caption{\textbf{Demo of our interactive sketching interface.} Users can co-draw with the model on the shared canvas for a concept.}
    \label{fig:interactive_demo}
\end{figure}

\Cref{fig:interactive_demo} shows a demo of our interactive sketching interface. 
We built a prototype collaborative interface for the autoregressive model, where the user and the model co-draw on a shared canvas, given a text prompt. The interaction is turn-based: the user can add (or erase) strokes, then the model continues the sketch by predicting the next sequence of strokes conditioned on the current canvas. This enables real-time, incremental refinement and allows the model to adapt to user edits on-the-fly.

The interface exposes core generation parameters such as resolution, number of frames per run, overlap between consecutive runs, and random seed. At first run, the concept from the user input will be refined to a detailed prompt automatically by an LLM. Each press of the \textit{Run} button generates a short continuation segment; overlap frames refers to context frames obtained from the previous run, which are used to stitch segments smoothly while preserving existing content. Users can switch between brush and eraser tools to modify the canvas between runs, enabling iterative co-creation with the model.

\subsection{Generalization Across Drawing Styles}\label{sec:supp_style_generalization}
As discussed in the main paper, our two-stage approach can generalize to new drawing styles beyond the single artist data demonstrated in the main paper. Here, we demonstrate this capability across two distinct sketching styles.

\begin{figure}[t]
    \centering
    \includegraphics[width=1\linewidth]{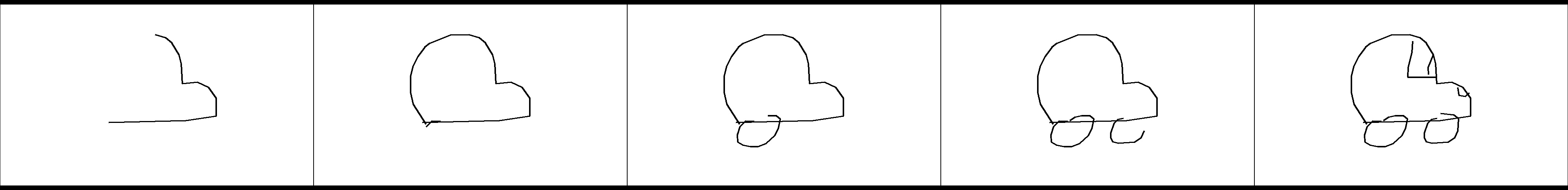}
    \includegraphics[width=1\linewidth]{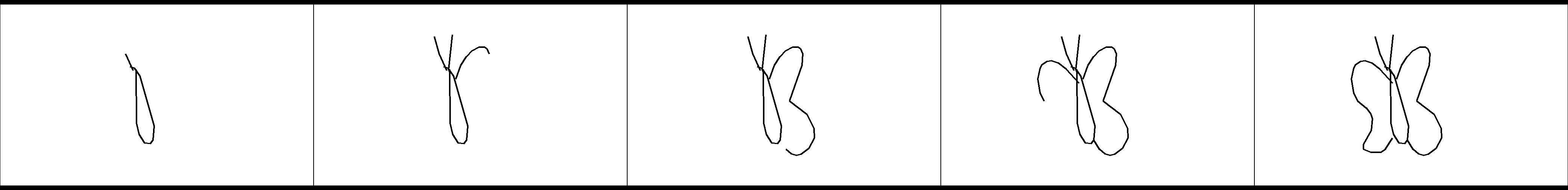}
    \includegraphics[width=1\linewidth]{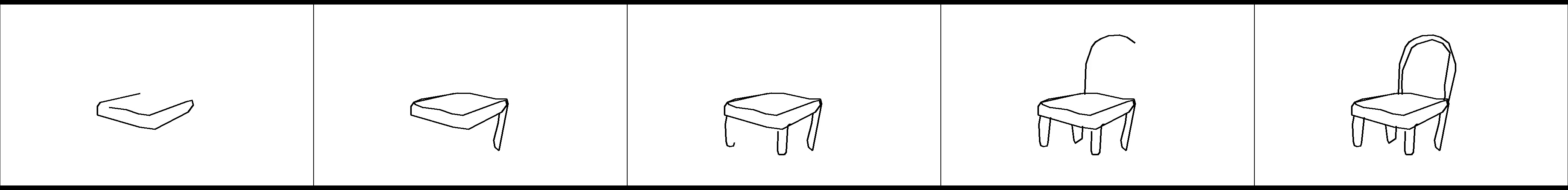}
    \includegraphics[width=1\linewidth]{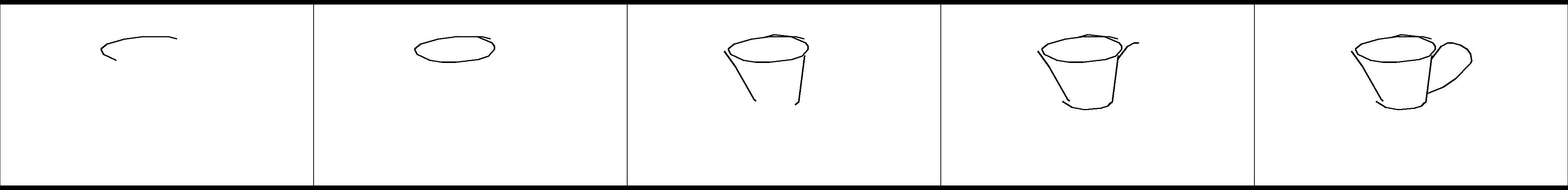}
    \includegraphics[width=1\linewidth]{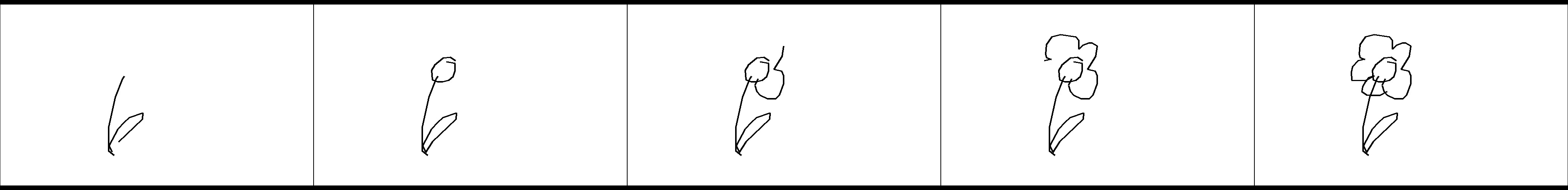}
    \includegraphics[width=1\linewidth]{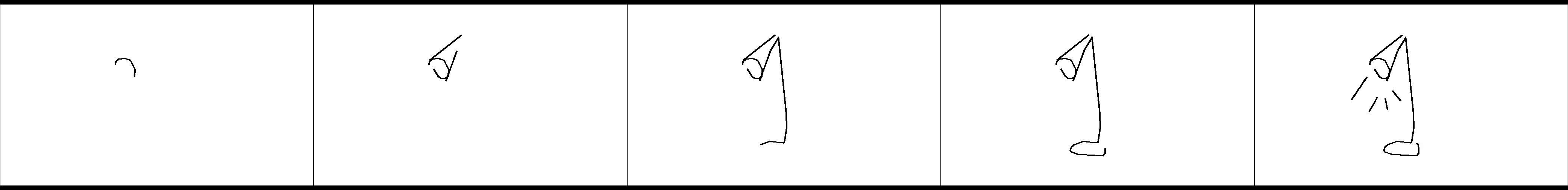}
    \includegraphics[width=1\linewidth]{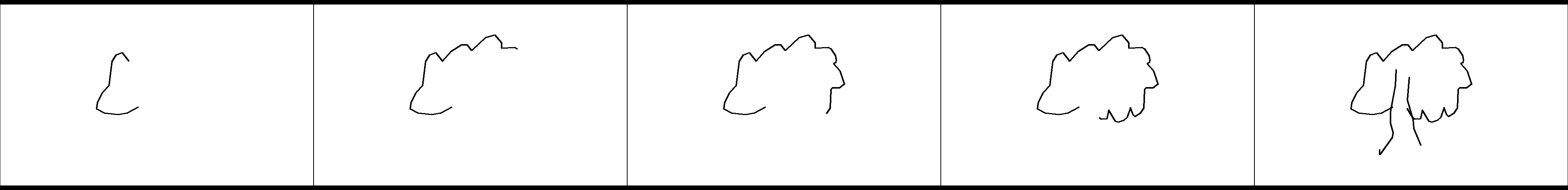}
\\
    \caption{Seven sketching videos derived from the QuickDraw Dataset~\cite{quickDrawData}, illustrating a doodle-like, rapid, drawing style from crowd-sourced participants.}
    \label{fig:7quickdraw_train}
\end{figure}

\begin{figure}
    \centering
    \includegraphics[width=1\linewidth]{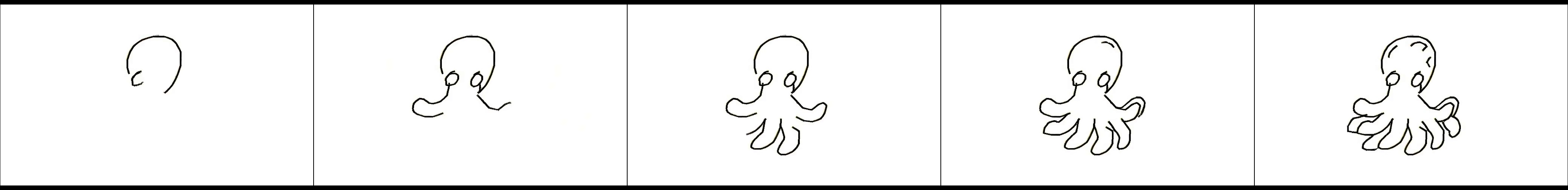} \\

    \includegraphics[width=1\linewidth]{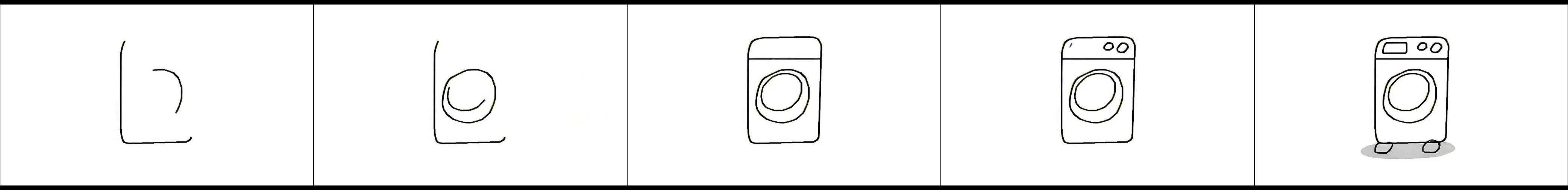} \\

    \includegraphics[width=1\linewidth]{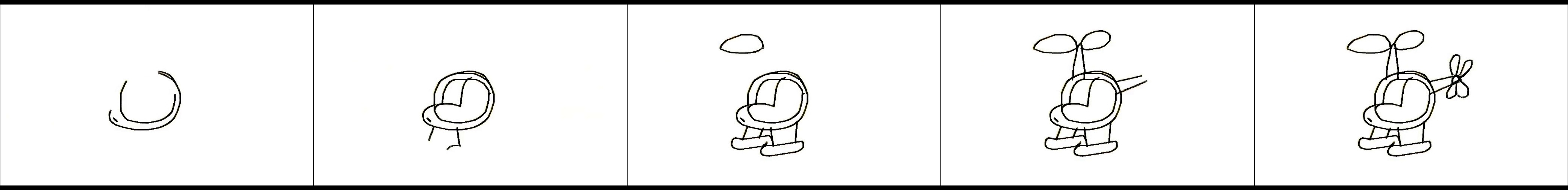} \\
     \includegraphics[width=1\linewidth]{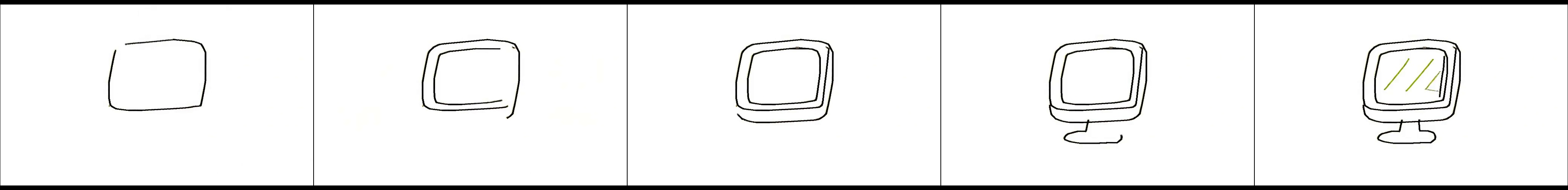} \\

     \includegraphics[width=1\linewidth]{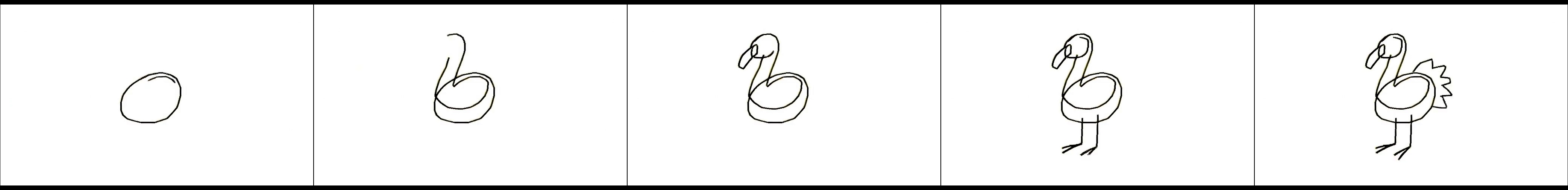} \\

    \includegraphics[width=1\linewidth]{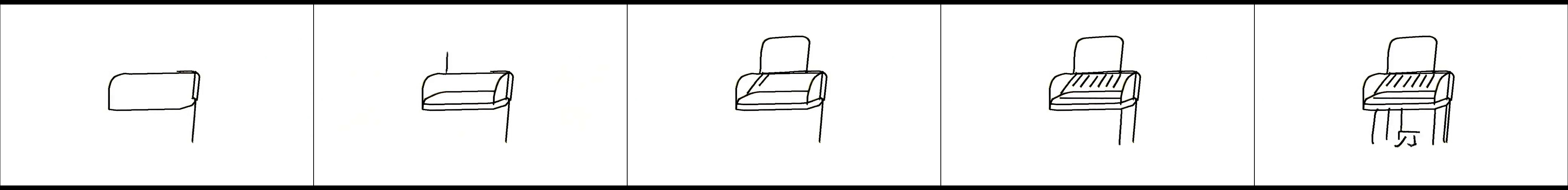} \\

    \includegraphics[width=1\linewidth]{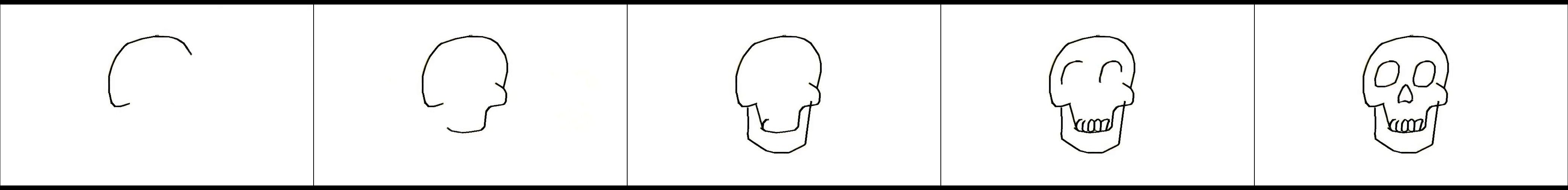} \\

    \includegraphics[width=1\linewidth]{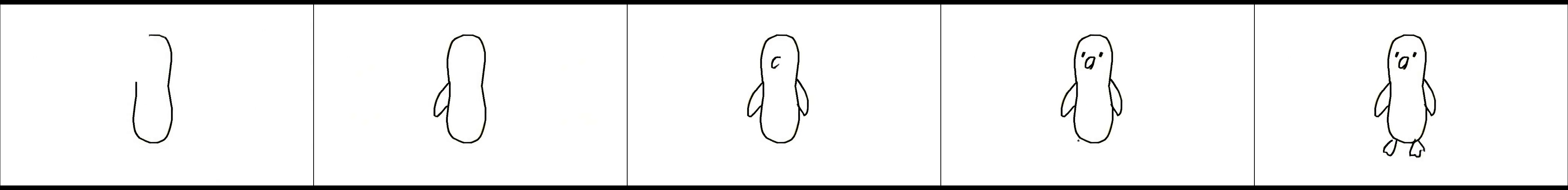} \\
    
    \caption{Generated sketches from the model fine-tuned on the QuickDraw samples, reflecting a flat, simple, and imprecise doodle-like style.}
    \label{fig:7quickdraw_results}
\end{figure}

\begin{figure}[t]
    \centering
    \includegraphics[width=1\linewidth]{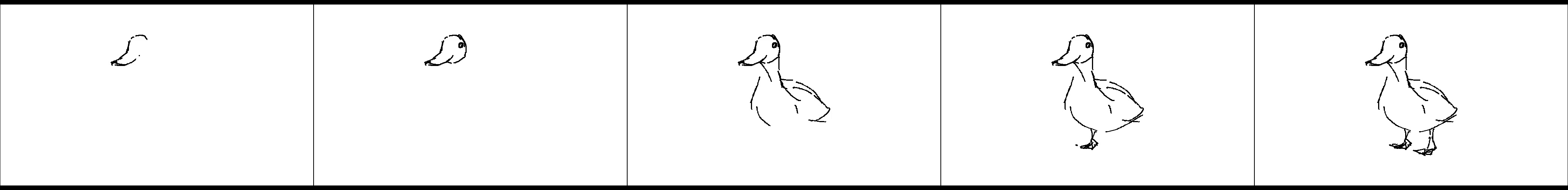}
    \includegraphics[width=1\linewidth]{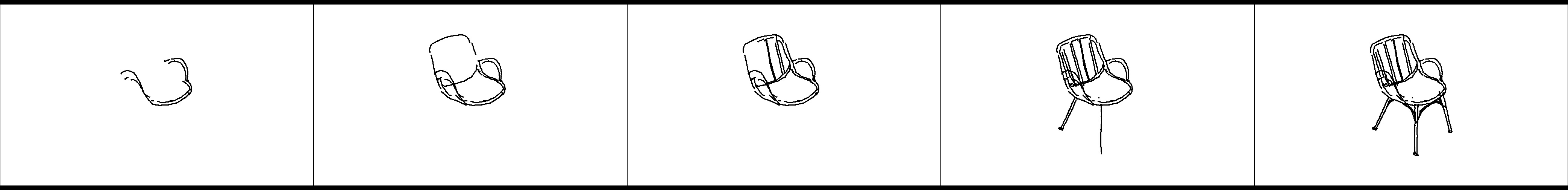}
    \includegraphics[width=1\linewidth]{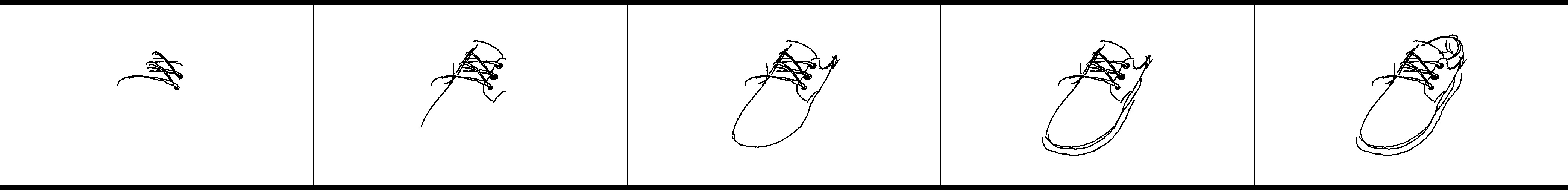}
    \includegraphics[width=1\linewidth]{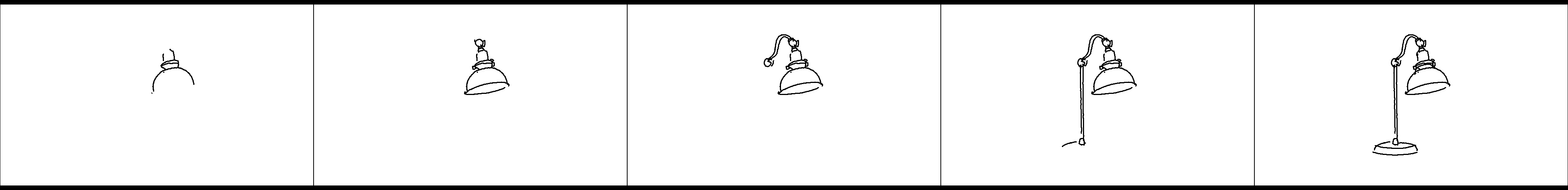}
    \includegraphics[width=1\linewidth]{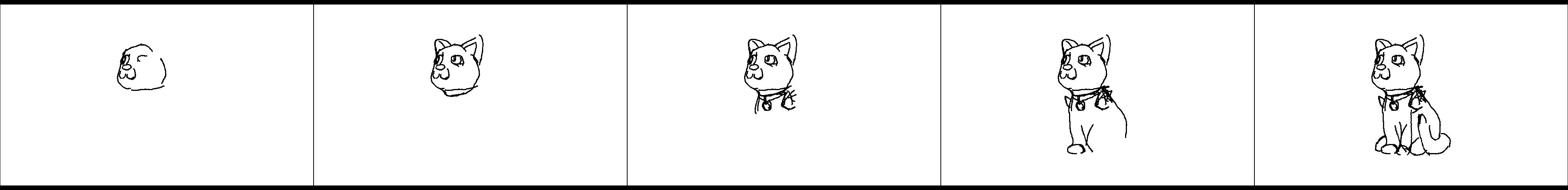}
    \includegraphics[width=1\linewidth]{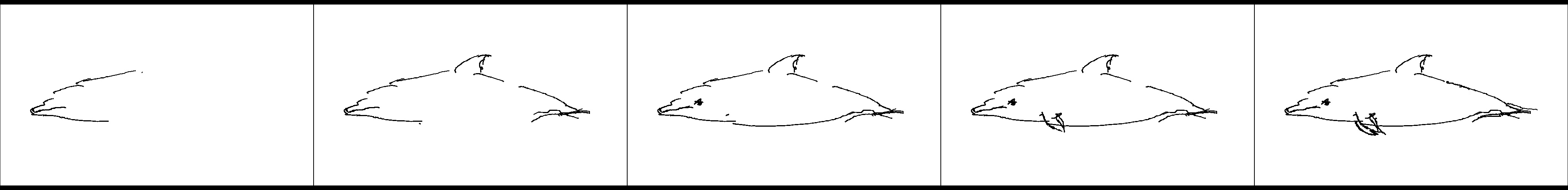}
    \includegraphics[width=1\linewidth]{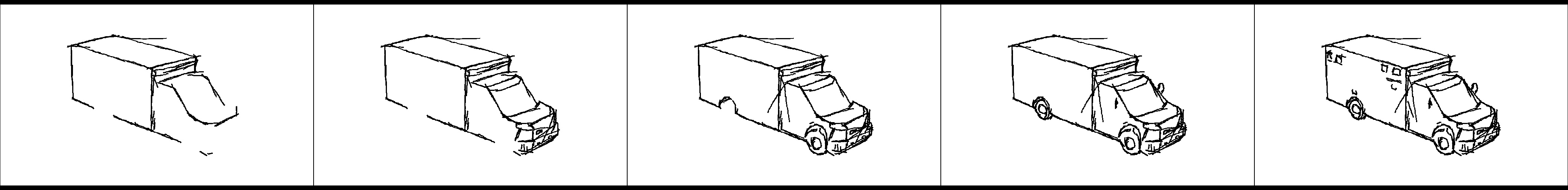}
    \caption{Seven sketching videos derived from the DifferSketching Dataset~\cite{xiao2022differsketching}, showcasing more structured and detailed sketches with 3D viewpoints from professional participants.}
    \label{fig:7differ_data}
\end{figure}

\begin{figure}[h]
    \centering

    \includegraphics[width=1\linewidth]{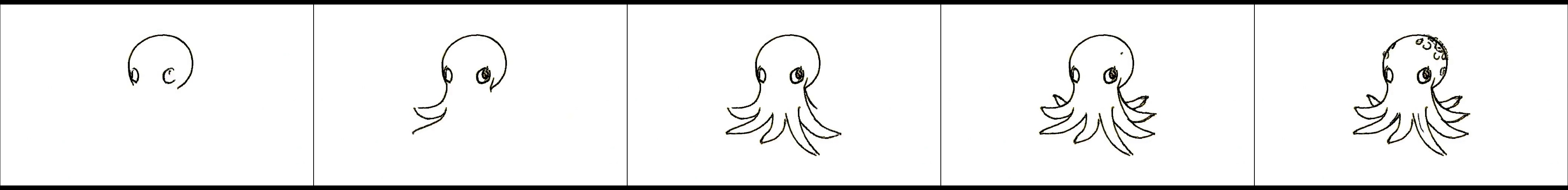} \\

    \includegraphics[width=1\linewidth]{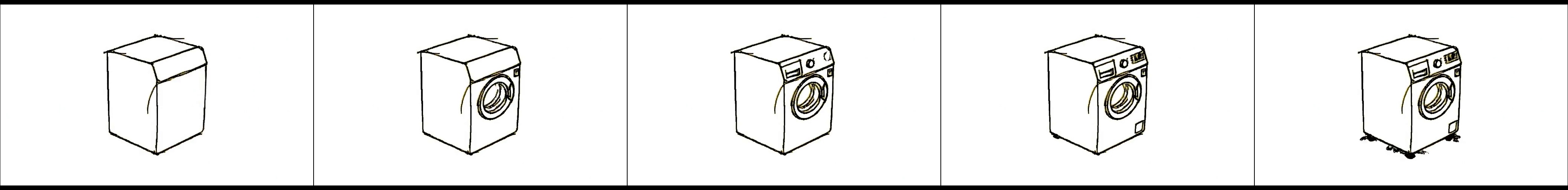} \\

     \includegraphics[width=1\linewidth]{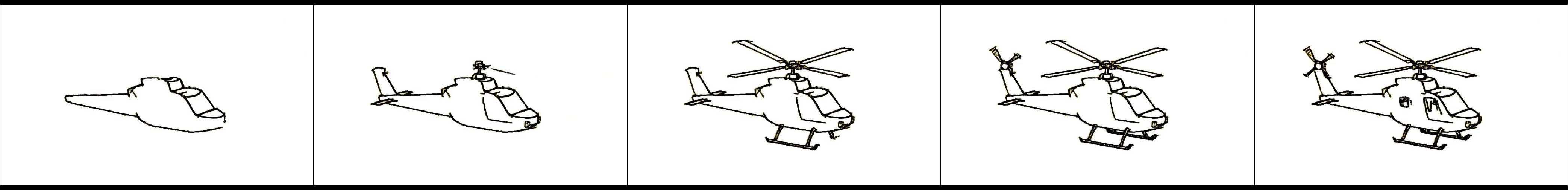} \\

    \includegraphics[width=1\linewidth]{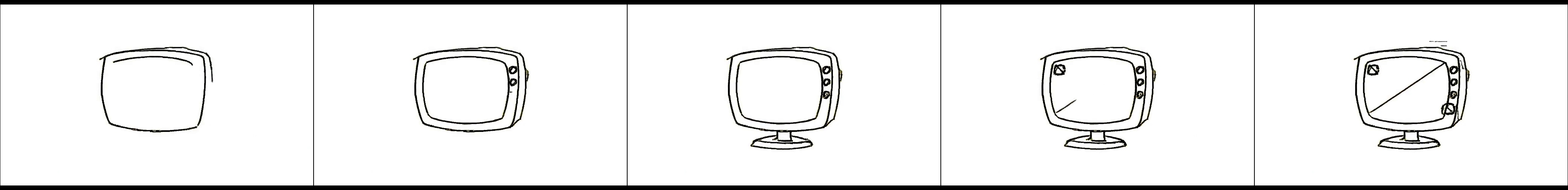} \\

    \includegraphics[width=1\linewidth]{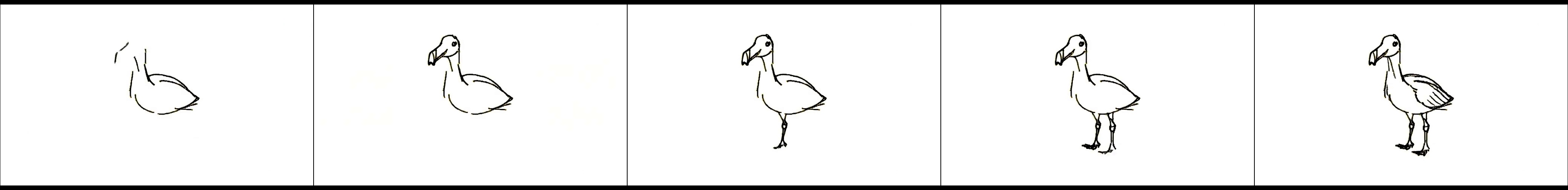} \\

    \includegraphics[width=1\linewidth]{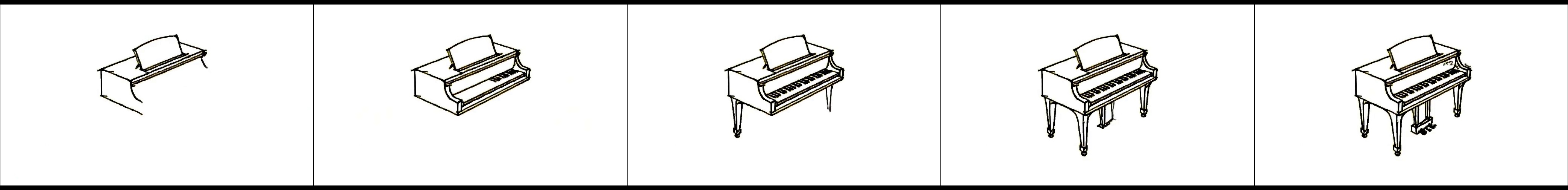} \\

    \includegraphics[width=1\linewidth]{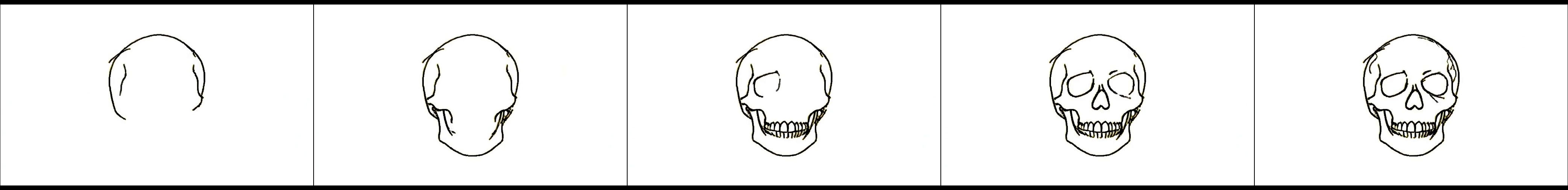} \\

     \includegraphics[width=1\linewidth]{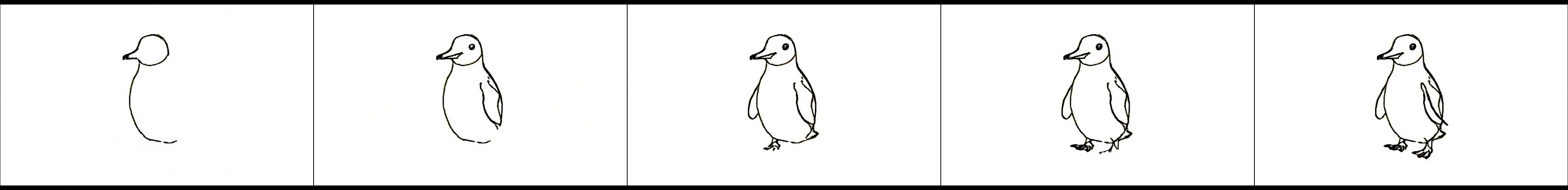} \\
    
    \caption{Generated sketches from the model fine-tuned on samples from the DifferSketching data, capturing a more detailed and structured style with 3D perspective cues.}
    \label{fig:7differ_results}
\end{figure}

We select seven sketches from the QuickDraw Dataset~\cite{quickDrawData}, which capture casual, rapid, crowd-sourced drawings, and seven sketches from the DifferSketching Dataset~\cite{xiao2022differsketching}, which consist of professionally drawn 3D object sketches. We convert the available parametric data in both datasets into sketching videos, resulting in seven videos per dataset, shown in \Cref{fig:7quickdraw_train,fig:7differ_data} for QuickDraw and DifferSketching, respectively. The QuickDraw samples are typically simplistic and imprecise, with a doodle-like, flat appearance, whereas the selected DifferSketching examples are more structured and detailed, often incorporating 3D viewpoints to convey depth.
Each set is used as our Stage 2 fine-tuning data. We then apply the resulting models to randomly sampled categories from the QuickDraw dataset. The results are shown in \Cref{fig:7quickdraw_results,fig:7differ_results} for the QuickDraw- and DifferSketching-trained models respectively.
As shown, the generated sketches reflect the style of their respective training data. The model trained on QuickDraw produces flat, simple, doodle-like outputs, while the model trained on DifferSketching captures a more structured, detailed style, often exhibiting 3D viewpoints. For example, the penguin and the skull in the generated DifferSketching style, appear to have depth cues, resembling the training samples.

These results highlight the versatility of our approach in capturing intrinsic stylistic properties, demonstrating the strength of the underlying video prior.

\subsection{Perceptual Quality - User Study}\label{sec:supp_user_study}

\begin{figure}[t]
    \centering
    \includegraphics[width=1\linewidth]{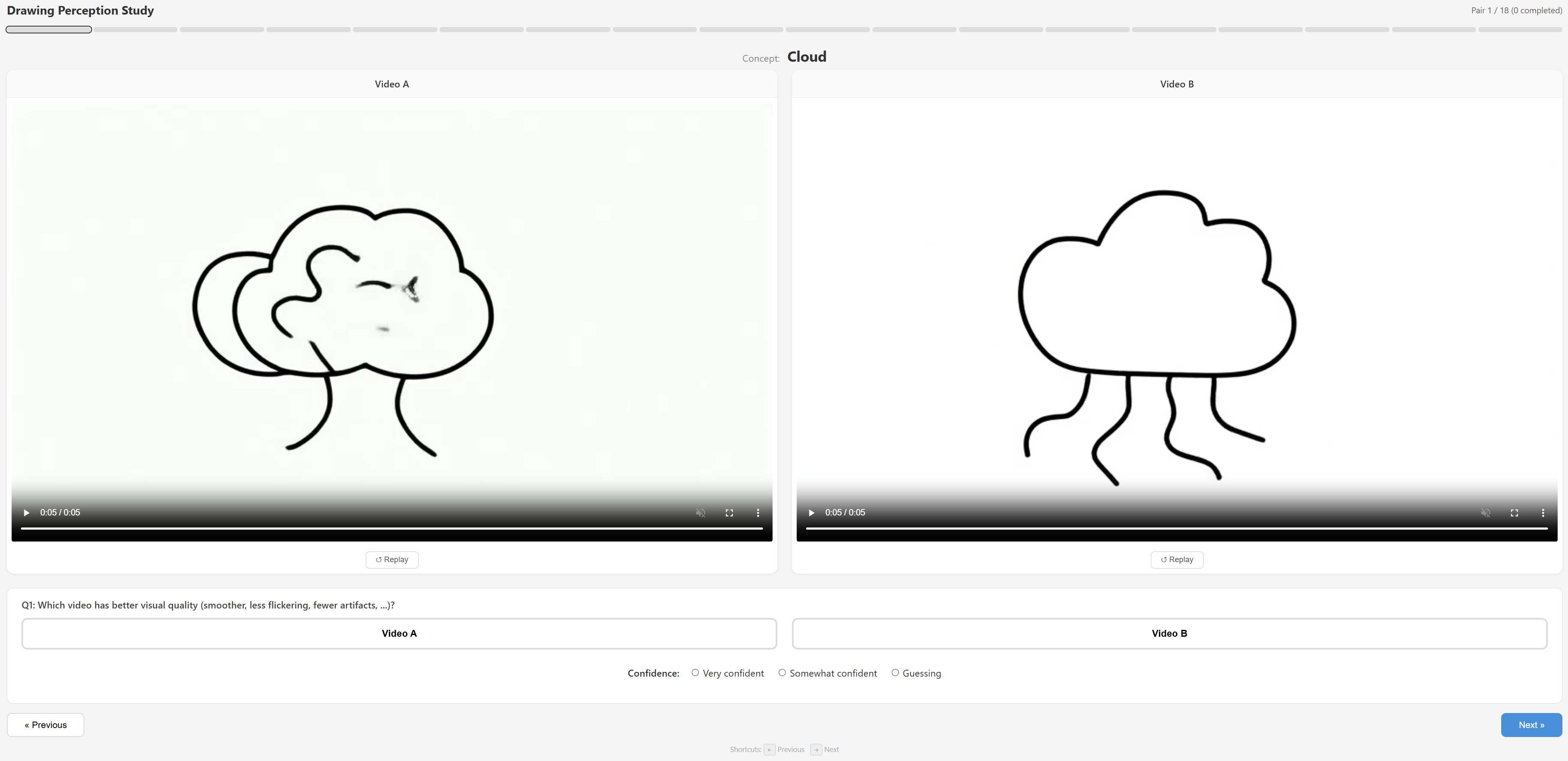}
    \caption{\textbf{User study interface.} Participants view two anonymous sketch videos side by side and answer the question: ``Which video has better visual quality (smoother, less flickering, fewer artifacts)?'' with a confidence level. The interface displays 18 pairs per session with progress tracking and navigation controls.}
    \label{fig:user_study_interface}
\end{figure}

To evaluate perceptual quality, we designed a web-based user study platform for pairwise comparison of sketch drawing videos. Below we describe the study design, and \Cref{fig:user_study_interface} shows the evaluation interface.
We compare six methods: QuickDraw~\cite{quickDrawData}, SketchAgent~\cite{SketchAgent_Vinker2025}, CausVid~\cite{Causvid}, Paints-UNDO~\cite{paintsundo}, Wan~2.1~\cite{Wan2.1}, and Ours. Instead of evaluating all $\binom{6}{2}=15$ method pairs, we adopt an \emph{anchor-based} design using CausVid (our autoregressive variant) and Ours (our fine-tuned diffusion model) as dual anchors. Each anchor is compared against all other methods plus against each other, yielding 9 pairs total. This keeps the comparison graph fully connected, allowing Bradley-Terry analysis~\cite{bradley1952rank} to infer a global ranking from partial pairwise data.

We evaluate 50 unique sketch concepts. For each comparison, two videos depicting the same concept are shown side by side, labeled anonymously as ``Video~A'' and ``Video~B'' (method identities are hidden). The left/right assignment is randomized per question. For each pair, participants answer the question: ``Which video has better visual quality (smoother, less flickering, fewer artifacts)?''
Each response is accompanied by a confidence rating (very confident / somewhat confident / guessing).
Each questionnaire contains 18 video pairs (exactly 2 per model pair). Videos for each pair are randomly sampled from the available results per method and concept.

We collected answers from 41 participants.
We aggregate preferences using Bradley-Terry analysis~\cite{bradley1952rank}, yielding normalized scores in which our method scores 1.00, outperforming PaintsUndo (0.60), Ours AR (0.40), Wan 2.1 (0.39), SketchAgent (0.32), and QuickDraw (0.09). Our method is consistently preferred over all baselines, confirming the advantage of combining video diffusion priors with structured stroke ordering.

\subsection{Prompt Adherence}
\begin{figure*}[t]
\centering
\setlength{\tabcolsep}{0pt}        %
\setlength{\fboxsep}{0pt}          %
\setlength{\fboxrule}{0.4pt}       %
\renewcommand{\arraystretch}{0.9}
\small

\begin{tabular}{c c c c c}
    \fbox{\includegraphics[width=0.19\linewidth]{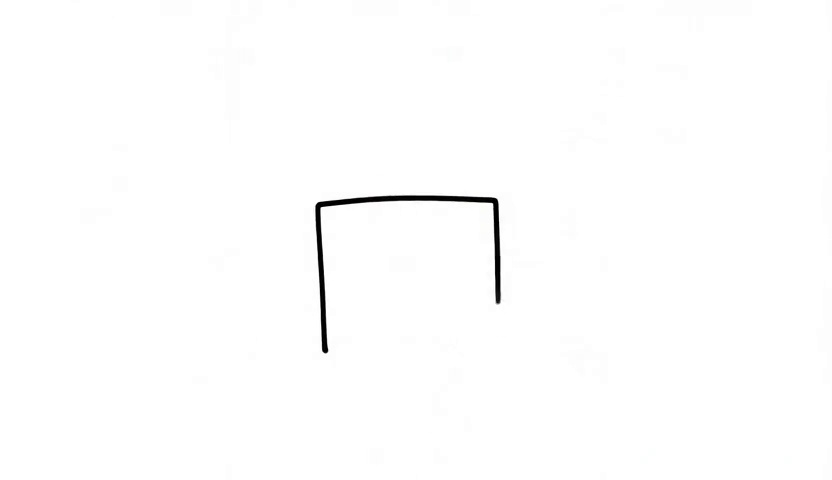}} &
    \fbox{\includegraphics[width=0.19\linewidth]{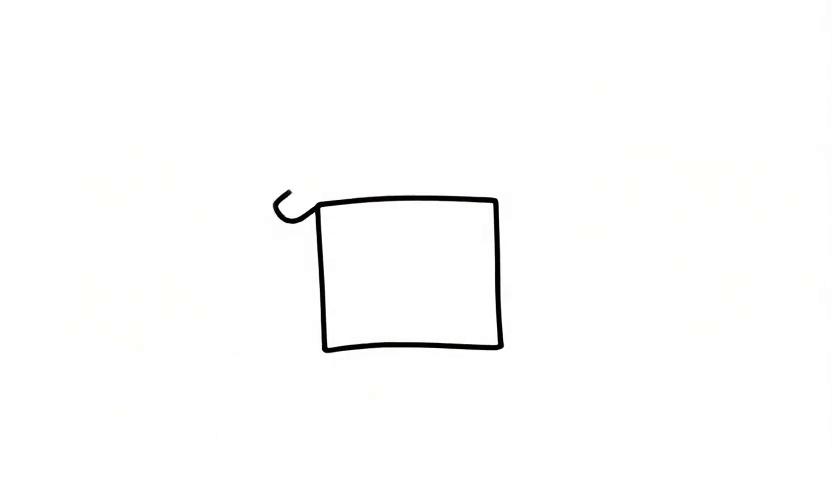}} &
    \fbox{\includegraphics[width=0.19\linewidth]{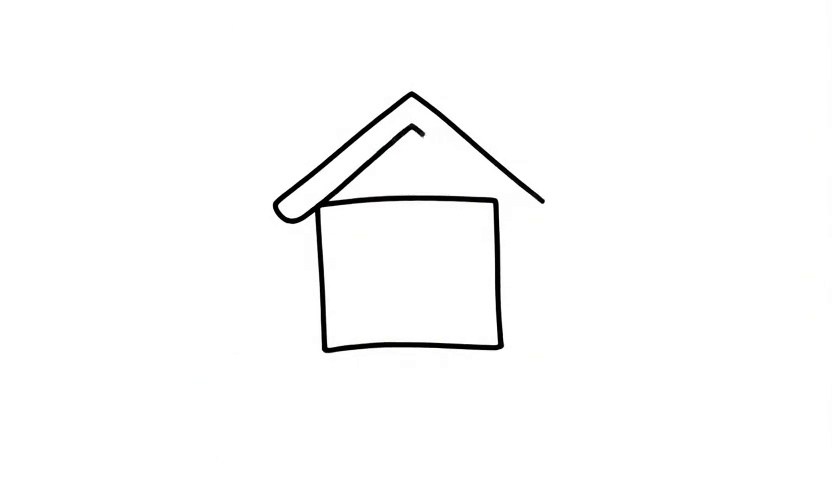}} &
    \fbox{\includegraphics[width=0.19\linewidth]{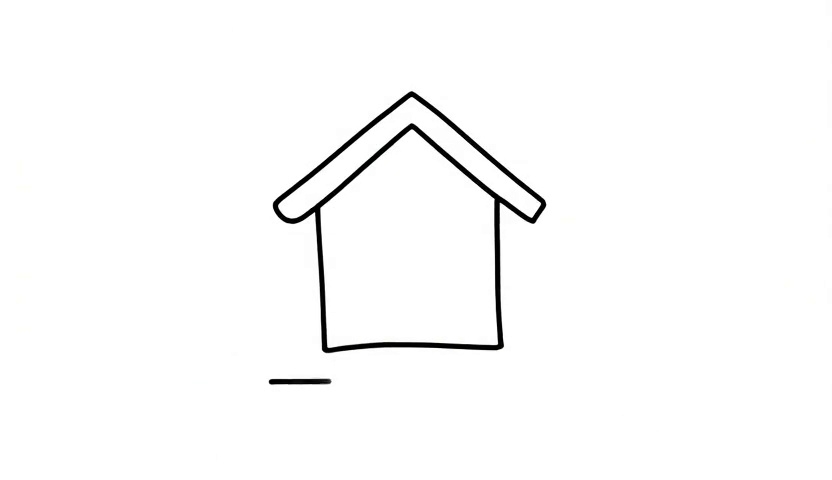}} &
    \fbox{\includegraphics[width=0.19\linewidth]{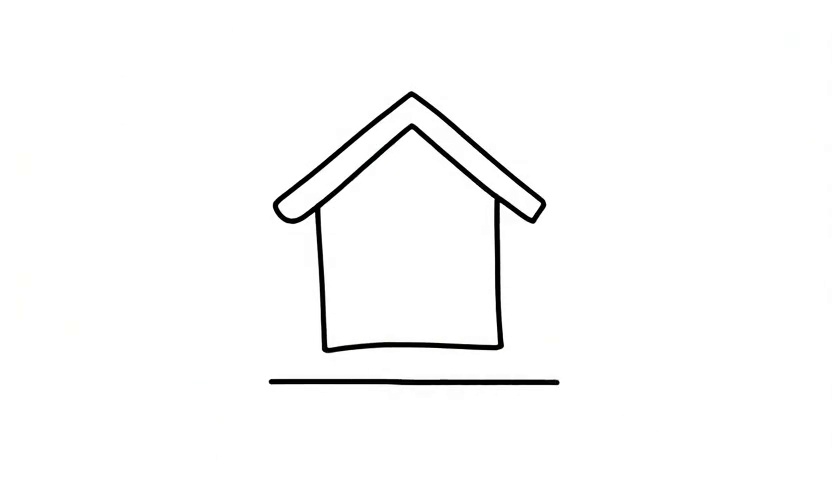}} \\
    \multicolumn{5}{c}{\emph{``A house''}} \\[6pt]

    \fbox{\includegraphics[width=0.19\linewidth]{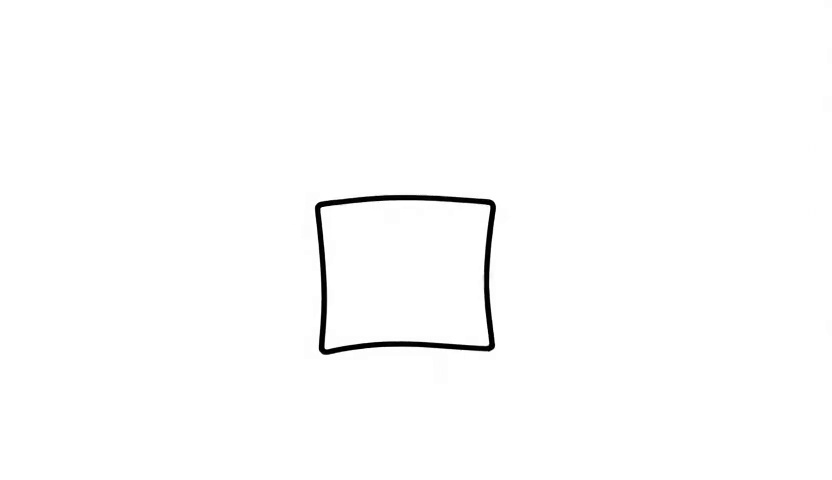}} &
    \fbox{\includegraphics[width=0.19\linewidth]{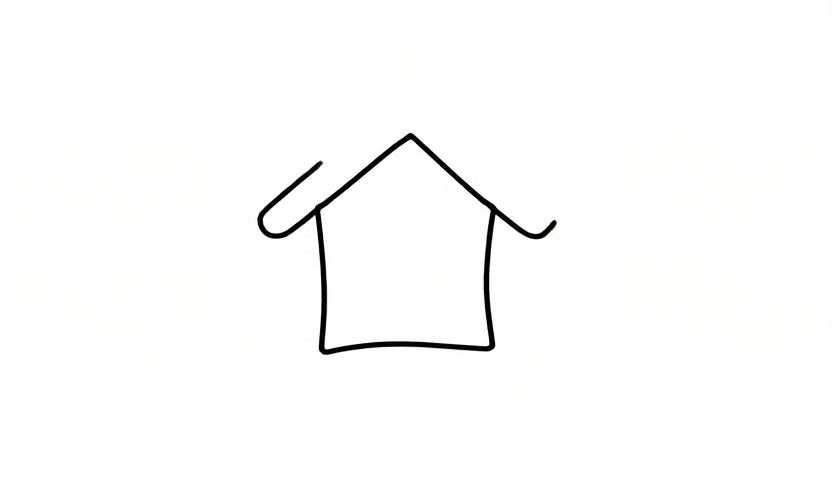}} &
    \fbox{\includegraphics[width=0.19\linewidth]{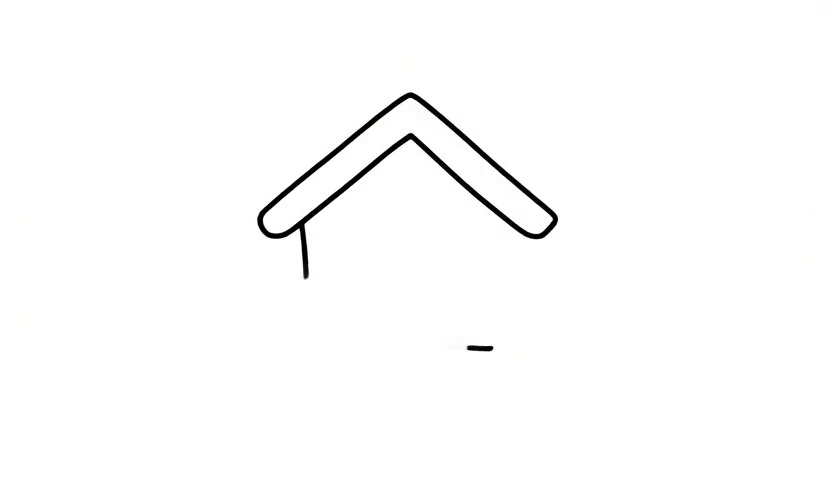}} &
    \fbox{\includegraphics[width=0.19\linewidth]{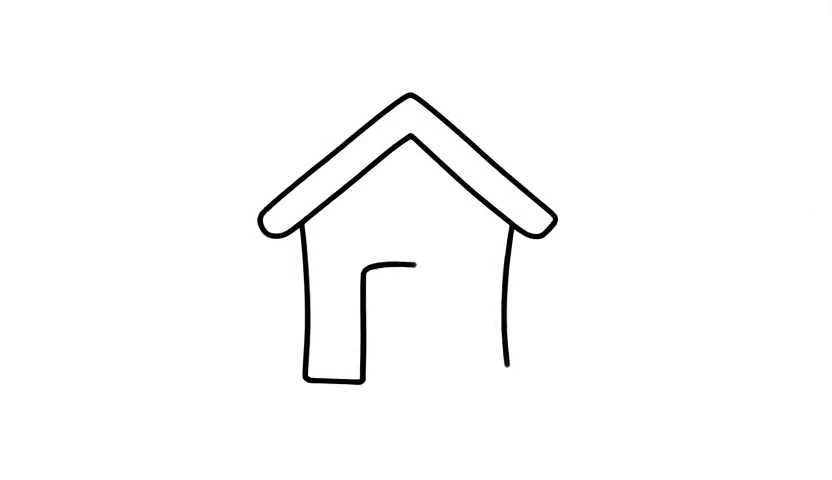}} &
    \fbox{\includegraphics[width=0.19\linewidth]{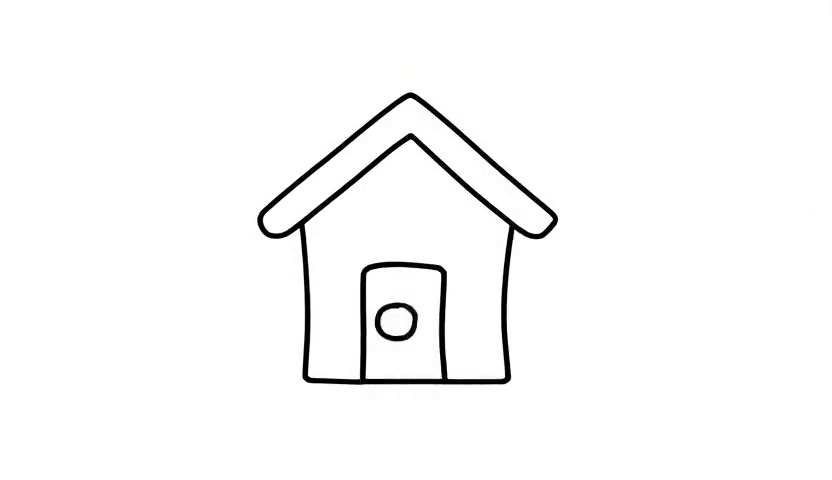}}\\
    \multicolumn{5}{c}{\emph{``A house with doors''}} \\[6pt]

   \fbox{\includegraphics[width=0.19\linewidth]{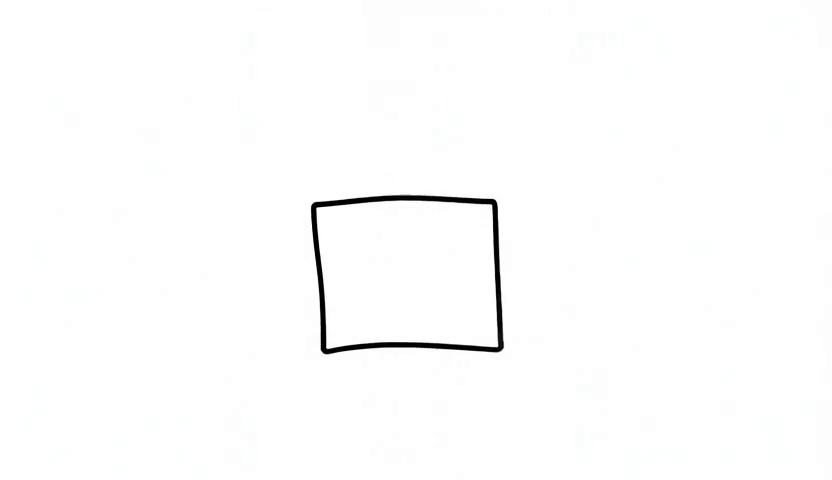}} &
   \fbox{\includegraphics[width=0.19\linewidth]{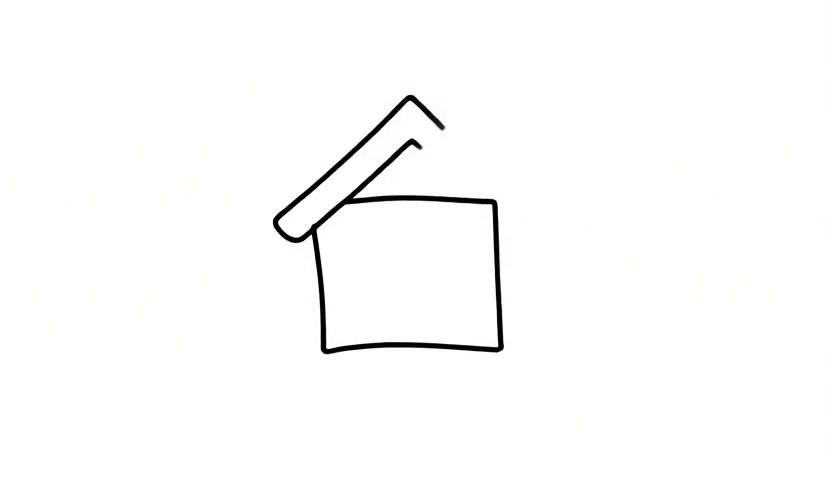}} &
   \fbox{\includegraphics[width=0.19\linewidth]{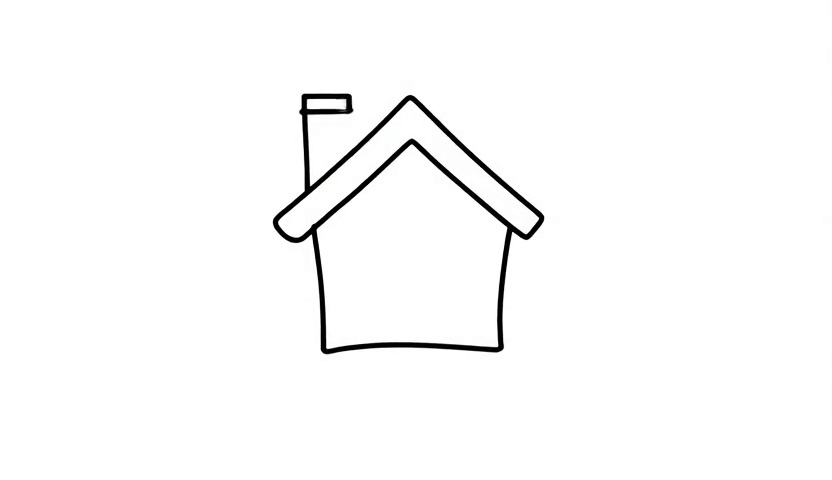}} &
   \fbox{\includegraphics[width=0.19\linewidth]{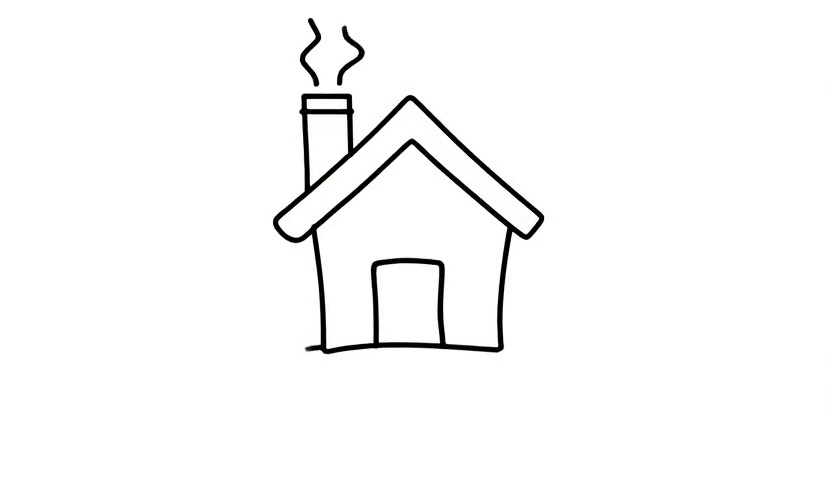}} &
   \fbox{\includegraphics[width=0.19\linewidth]{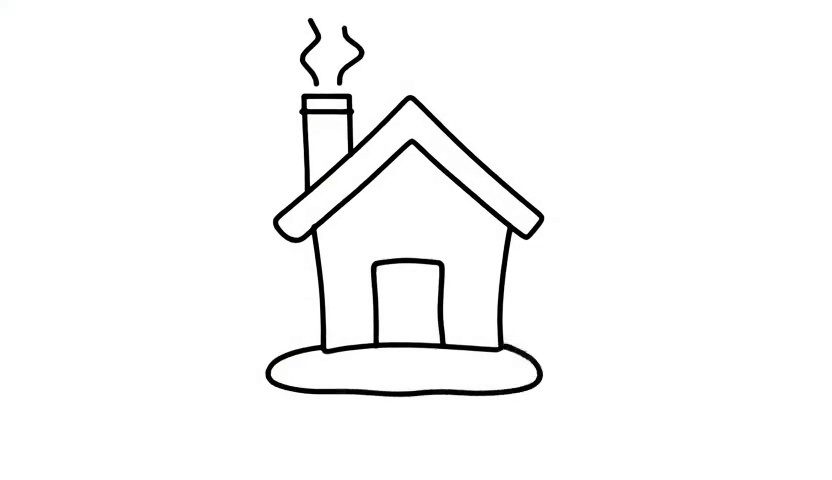}}\\
    \multicolumn{5}{c}{\emph{``A house with chimney and doors''}} \\[6pt]

    \fbox{\includegraphics[width=0.19\linewidth]{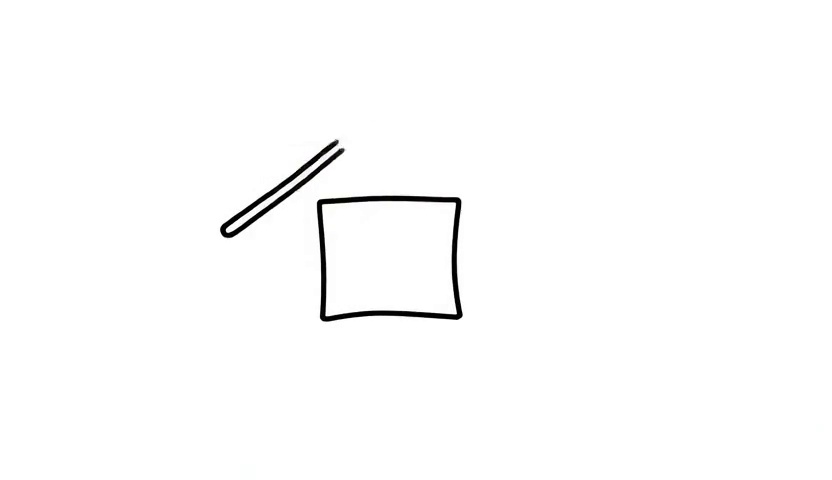}} &
    \fbox{\includegraphics[width=0.19\linewidth]{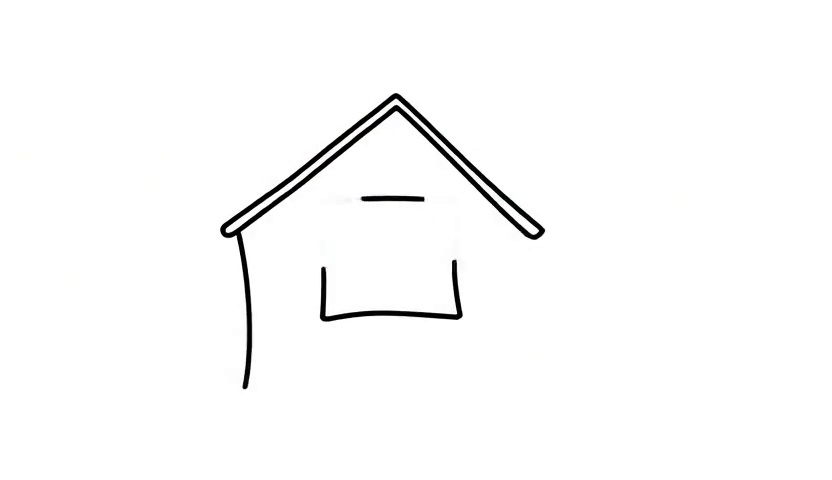}} &
    \fbox{\includegraphics[width=0.19\linewidth]{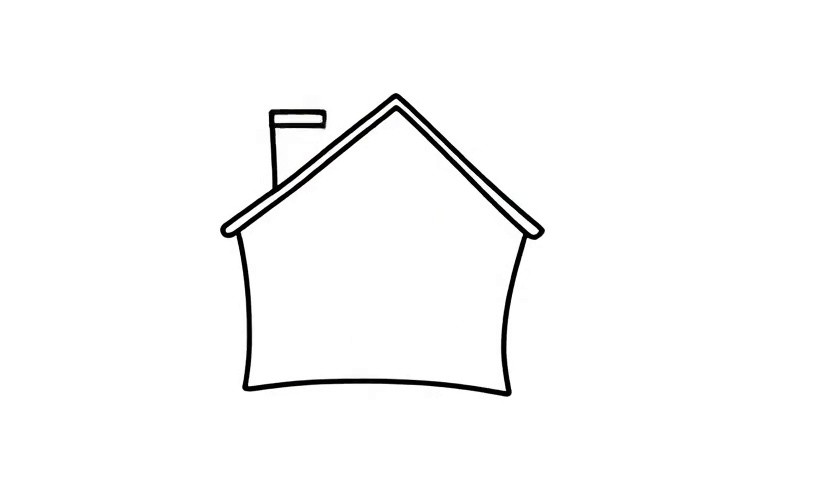}} &
    \fbox{\includegraphics[width=0.19\linewidth]{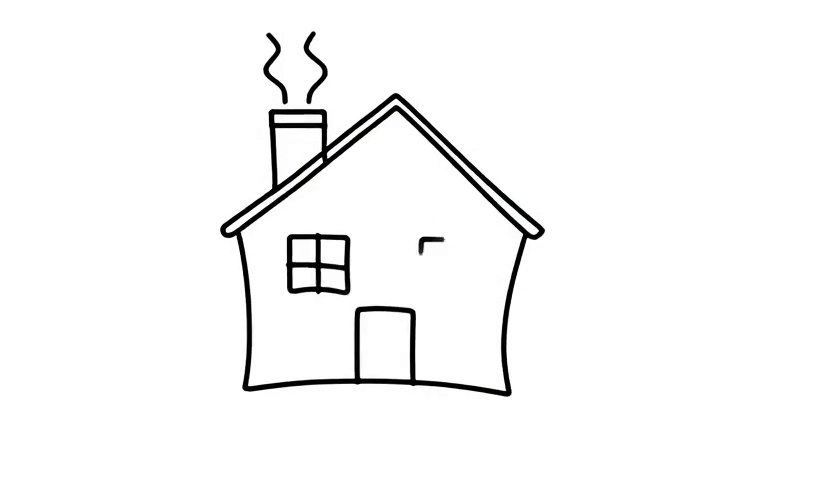}} &
    \fbox{\includegraphics[width=0.19\linewidth]{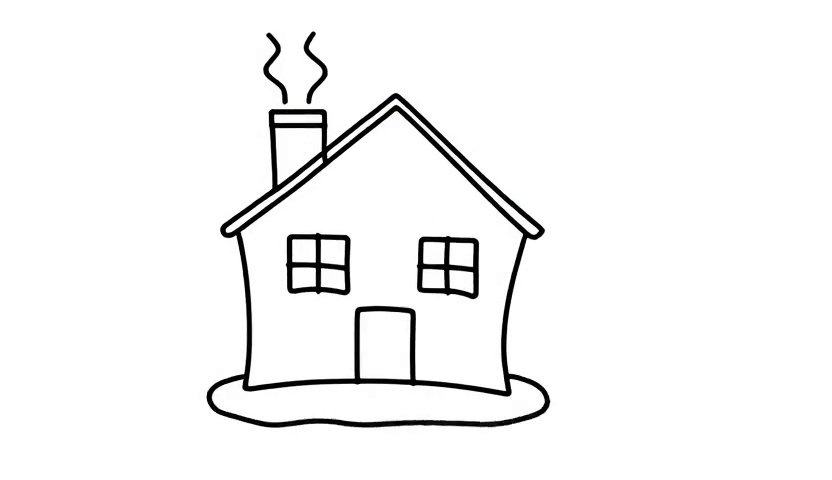}}\\
    \multicolumn{5}{c}{\emph{``A house with chimney, doors, and windows''}} \\[6pt]

    \fbox{\includegraphics[width=0.19\linewidth]{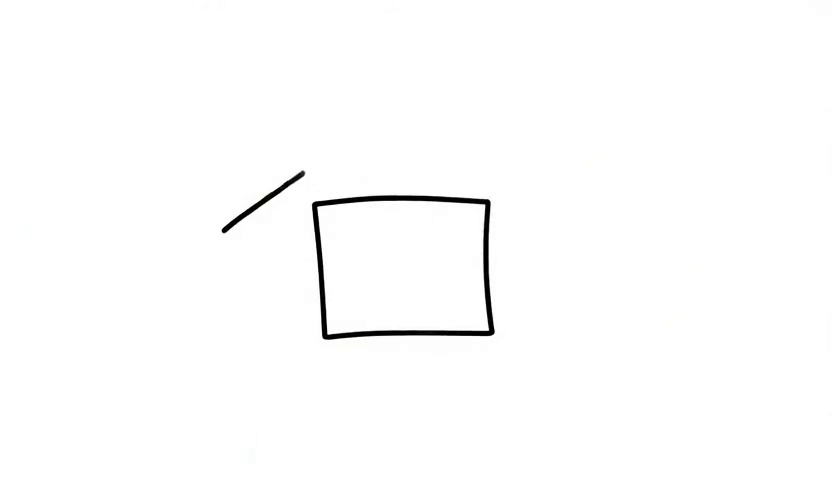}} &
    \fbox{\includegraphics[width=0.19\linewidth]{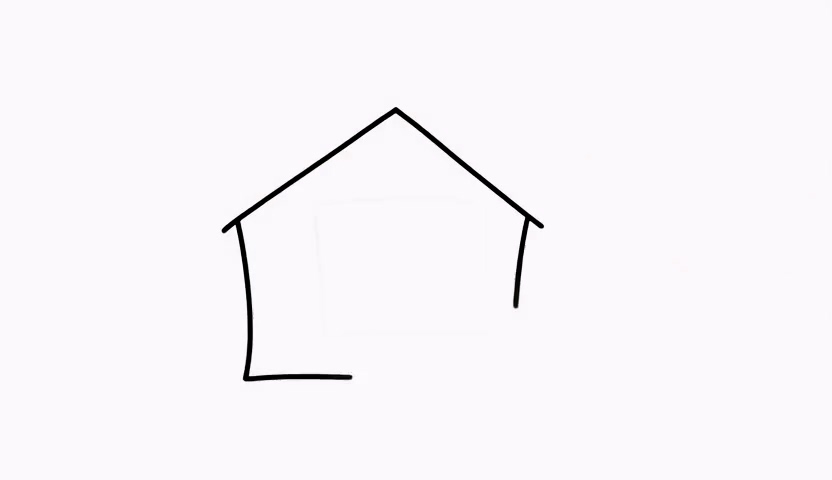}} &
    \fbox{\includegraphics[width=0.19\linewidth]{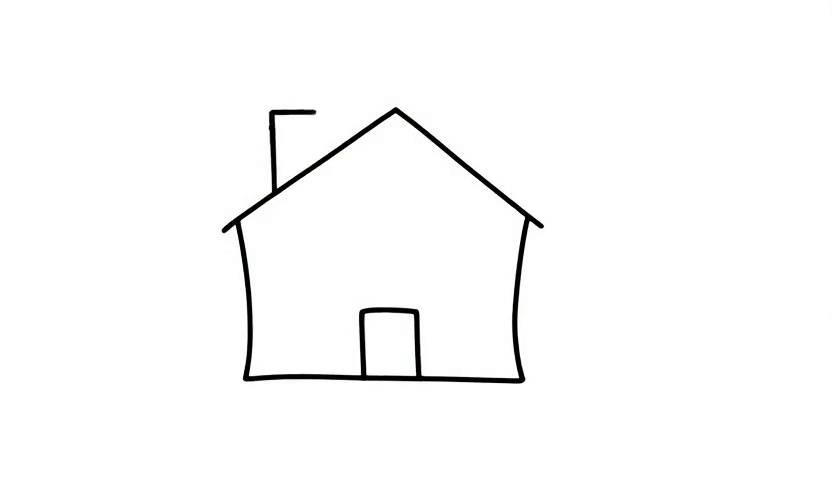}} &
    \fbox{\includegraphics[width=0.19\linewidth]{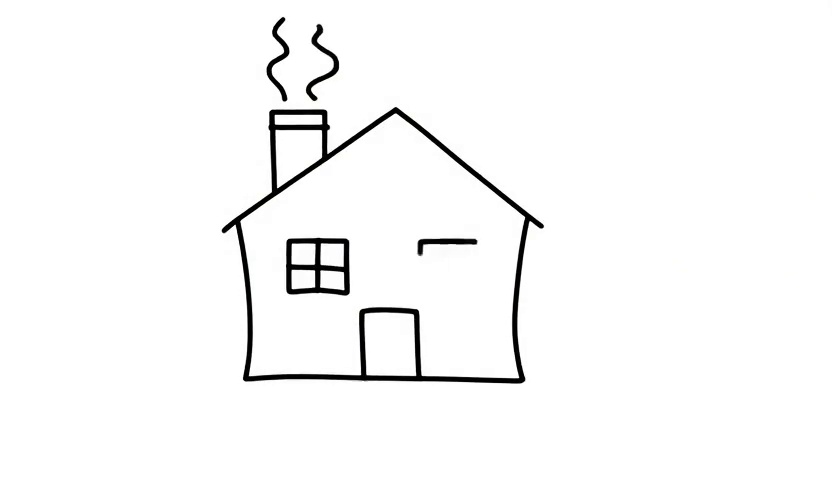}} &
    \fbox{\includegraphics[width=0.19\linewidth]{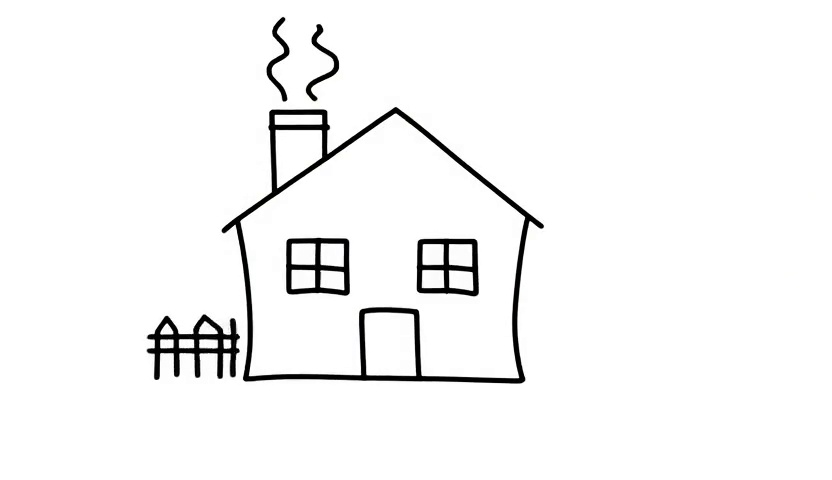}}\\
    \multicolumn{5}{c}{\emph{``A house with chimney, doors, windows, and fence''}}
\end{tabular}
\vspace{-6pt}
\caption{\textbf{Prompt adherence with incremental details.} Given a base concept (\emph{a house}), we progressively add details in a prompt (e.g., \emph{with doors}, \emph{with chimney and doors}, \emph{with chimney, doors, and windows}, and \emph{with chimney, doors, windows, and fence}). The generated sketch sequences consistently follow the prompt and incorporate the newly requested elements.}
\vspace{-0.15cm}
\label{fig:prompt_adherence}
\end{figure*}

We demonstrate prompt adherence by progressively enriching an initially simple text instruction with additional details.
As shown in \Cref{fig:prompt_adherence}, the model follows the updated prompt by adding the newly requested elements (e.g., doors, chimney, windows, fence) while maintaining the previously drawn structure, demonstrating compositional control over the sketching process.

\subsection{Per-Brush-Style Metric Breakdown}\label{sec:supp_brush_per_style}
To complement the aggregate brush-style alignment result reported in the main paper, we provide a per-style breakdown in \Cref{tab:brush_per_style}. For each of the 25 brush-color combinations (5 brushes $\times$ 5 colors), we manually draw a reference sketch in the target style and use it as the ground-truth style exemplar. We then report the average Gram-matrix distance between each generated sketch and its corresponding reference (``Score''), together with a baseline capturing the expected distance to mismatched brush-color pairs (mean distance to the other 24 references). Lower Score indicates stronger alignment to the target brush style. Each per-style entry is averaged over 30 videos; for the Average row we additionally report the standard deviation taken across the 25 style means.

Across all 25 styles, our Score is consistently lower than the mismatched-pair baseline, confirming that the model adapts to the specified brush rather than producing generic textures. The gap is largest for visually distinctive combinations (e.g., \ap{pixel-tiles\_black}, \ap{hard-large-dots-fixed\_black}, \ap{caligraphy-vertical\_pink}), where the target texture differs most strongly from the other exemplars. For calibration, \Cref{sec:supp_brush_per_style_trainset} reports the same metric on the brush-color combinations seen during training, which provides an empirical lower-bound reference.

\begin{table}[t]
\centering
\small
\caption{\textbf{Per-brush-style metric breakdown.} Gram-matrix distance between generated frames and a manually drawn reference sketch in the target style (Score $\downarrow$), compared to the expected distance to mismatched brush-color references (mean distance to the other 24 references). Each per-style entry reports mean $\pm$ standard deviation over 30 videos; the Average row reports mean $\pm$ standard deviation taken across the 25 style means.\\[-0.3cm]}
\label{tab:brush_per_style}
\setlength{\tabcolsep}{4pt}
\begin{tabular}{llcc}
\toprule
Brush Style & Color & Score $\downarrow$ & Baseline \\
\midrule
\multirow{5}{*}{bubbles-medium}
  & black         & $0.118 \pm 0.005$ & $0.287 \pm 0.005$ \\
  & indigo-blue   & $0.083 \pm 0.009$ & $0.265 \pm 0.004$ \\
  & mocha-brown   & $0.077 \pm 0.005$ & $0.265 \pm 0.003$ \\
  & mustard-olive & $0.058 \pm 0.003$ & $0.272 \pm 0.002$ \\
  & pink          & $0.081 \pm 0.007$ & $0.284 \pm 0.003$ \\
\midrule
\multirow{5}{*}{caligraphy-vertical}
  & black         & $0.230 \pm 0.017$ & $0.415 \pm 0.022$ \\
  & indigo-blue   & $0.159 \pm 0.022$ & $0.315 \pm 0.011$ \\
  & mocha-brown   & $0.153 \pm 0.018$ & $0.303 \pm 0.007$ \\
  & mustard-olive & $0.110 \pm 0.017$ & $0.282 \pm 0.002$ \\
  & pink          & $0.137 \pm 0.023$ & $0.345 \pm 0.016$ \\
\midrule
\multirow{5}{*}{hard-large-dots-fixed}
  & black         & $0.254 \pm 0.030$ & $0.504 \pm 0.017$ \\
  & indigo-blue   & $0.180 \pm 0.029$ & $0.402 \pm 0.016$ \\
  & mocha-brown   & $0.163 \pm 0.021$ & $0.350 \pm 0.006$ \\
  & mustard-olive & $0.102 \pm 0.014$ & $0.280 \pm 0.003$ \\
  & pink          & $0.118 \pm 0.020$ & $0.315 \pm 0.004$ \\
\midrule
\multirow{5}{*}{pixel-tiles}
  & black         & $0.276 \pm 0.028$ & $0.506 \pm 0.014$ \\
  & indigo-blue   & $0.215 \pm 0.033$ & $0.393 \pm 0.025$ \\
  & mocha-brown   & $0.168 \pm 0.012$ & $0.315 \pm 0.005$ \\
  & mustard-olive & $0.114 \pm 0.011$ & $0.289 \pm 0.003$ \\
  & pink          & $0.134 \pm 0.024$ & $0.339 \pm 0.016$ \\
\midrule
\multirow{5}{*}{versatile-ink}
  & black         & $0.230 \pm 0.015$ & $0.413 \pm 0.023$ \\
  & indigo-blue   & $0.092 \pm 0.010$ & $0.270 \pm 0.003$ \\
  & mocha-brown   & $0.091 \pm 0.010$ & $0.273 \pm 0.002$ \\
  & mustard-olive & $0.075 \pm 0.012$ & $0.276 \pm 0.003$ \\
  & pink          & $0.113 \pm 0.014$ & $0.312 \pm 0.002$ \\
\midrule
\multicolumn{2}{l}{\textit{Average}} & $0.141 \pm 0.061$ & $0.331 \pm 0.070$ \\
\bottomrule
\end{tabular}
\end{table}

\subsubsection{Per-Brush-Style Metric Breakdown on Training-Distribution Styles}\label{sec:supp_brush_per_style_trainset}
To establish a quantitative reference point for the unseen-styles result reported above, we additionally evaluate brush-style alignment on the brush-color combinations seen during training (6 brushes $\times$ 8 colors, 48 combinations in total; see \Cref{fig:brushes}). For each brush-color combination, we directly select a single concept sample from the training set as the ground-truth style reference, and evaluate the model on 7 object categories not used to construct that reference (one random seed per category), yielding 336 generations in total. We follow exactly the same evaluation protocol as the unseen-styles experiment: for each generated sketch we report the average Gram-matrix distance to its corresponding reference (``Score''), together with a baseline capturing the expected distance to mismatched brush-color pairs (mean distance to the other 47 references). The overall average is $0.079 \pm 0.029$ for our method, compared to $0.201 \pm 0.033$ for the mismatched-pair baseline, a relative reduction of $60.7\%$. Per-style results are listed in \Cref{tab:brush_per_style_trainset}.

Because these brush-color combinations fall within the training distribution, the resulting Score represents the level of brush alignment that the model has demonstrably learned to achieve, and therefore serves as an empirical lower-bound reference for the unseen-styles evaluation. The unseen-styles average reported in \Cref{tab:brush_per_style} ($0.141 \pm 0.061$) is higher than this reference, as expected when extrapolating to brush-color combinations not seen during training, yet it remains well below the corresponding mismatched-pair baseline of that setting ($0.331 \pm 0.070$). Reading the two settings jointly thus calibrates the Score: $0.079$ approximates the in-distribution floor of the metric for our model, while the unseen-styles Score of $0.141$ lies roughly halfway between this floor and the mismatched-pair baseline, quantifying how much of the alignment is preserved under brush-color generalization.

\begin{table*}[t]
\centering
\small
\caption{\textbf{Per-brush-style metric breakdown on training-distribution styles.} Gram-matrix distance between generated frames and a training-set reference sketch in the target style (Score $\downarrow$), compared to the expected distance to mismatched brush-color references (mean distance to the other 47 references). Each per-style entry reports mean $\pm$ standard deviation over 7 videos; the Average row reports mean $\pm$ standard deviation taken across the 48 style means. The Average serves as an empirical lower-bound reference for the unseen-styles evaluation in \Cref{tab:brush_per_style}.\\[-0.3cm]}
\label{tab:brush_per_style_trainset}
\setlength{\tabcolsep}{6pt}
\begin{tabular}{llcc@{\hspace{3em}}llcc}
\toprule
Brush & Color & Score $\downarrow$ & Baseline & Brush & Color & Score $\downarrow$ & Baseline \\
\midrule
\multirow{8}{*}{charcoal}
  & blue          & $0.088 \pm 0.053$ & $0.211 \pm 0.035$ & \multirow{8}{*}{marker}     & blue          & $0.038 \pm 0.022$ & $0.162 \pm 0.007$ \\
  & carrot        & $0.109 \pm 0.062$ & $0.235 \pm 0.043$ &                              & carrot        & $0.044 \pm 0.024$ & $0.167 \pm 0.006$ \\
  & dark-red      & $0.115 \pm 0.069$ & $0.252 \pm 0.052$ &                              & dark-red      & $0.045 \pm 0.025$ & $0.163 \pm 0.007$ \\
  & midnight-blue & $0.102 \pm 0.059$ & $0.229 \pm 0.039$ &                              & midnight-blue & $0.046 \pm 0.026$ & $0.166 \pm 0.007$ \\
  & nephritis     & $0.082 \pm 0.048$ & $0.207 \pm 0.032$ &                              & nephritis     & $0.040 \pm 0.023$ & $0.161 \pm 0.007$ \\
  & turquoise     & $0.087 \pm 0.051$ & $0.215 \pm 0.036$ &                              & turquoise     & $0.039 \pm 0.022$ & $0.159 \pm 0.007$ \\
  & wisteria      & $0.083 \pm 0.048$ & $0.203 \pm 0.030$ &                              & wisteria      & $0.038 \pm 0.022$ & $0.162 \pm 0.007$ \\
  & yellow        & $0.093 \pm 0.054$ & $0.221 \pm 0.036$ &                              & yellow        & $0.039 \pm 0.022$ & $0.165 \pm 0.005$ \\
\midrule
\multirow{8}{*}{gouache}
  & blue          & $0.081 \pm 0.051$ & $0.188 \pm 0.036$ & \multirow{8}{*}{pen}        & blue          & $0.108 \pm 0.065$ & $0.229 \pm 0.049$ \\
  & carrot        & $0.097 \pm 0.059$ & $0.209 \pm 0.049$ &                              & carrot        & $0.126 \pm 0.074$ & $0.260 \pm 0.061$ \\
  & dark-red      & $0.108 \pm 0.067$ & $0.226 \pm 0.063$ &                              & dark-red      & $0.139 \pm 0.083$ & $0.286 \pm 0.072$ \\
  & midnight-blue & $0.100 \pm 0.061$ & $0.216 \pm 0.052$ &                              & midnight-blue & $0.134 \pm 0.076$ & $0.269 \pm 0.062$ \\
  & nephritis     & $0.083 \pm 0.050$ & $0.192 \pm 0.038$ &                              & nephritis     & $0.105 \pm 0.062$ & $0.226 \pm 0.048$ \\
  & turquoise     & $0.083 \pm 0.051$ & $0.192 \pm 0.039$ &                              & turquoise     & $0.107 \pm 0.063$ & $0.228 \pm 0.050$ \\
  & wisteria      & $0.087 \pm 0.052$ & $0.197 \pm 0.041$ &                              & wisteria      & $0.108 \pm 0.063$ & $0.228 \pm 0.049$ \\
  & yellow        & $0.093 \pm 0.057$ & $0.211 \pm 0.048$ &                              & yellow        & $0.115 \pm 0.069$ & $0.258 \pm 0.059$ \\
\midrule
\multirow{8}{*}{hb-pencil}
  & blue          & $0.071 \pm 0.043$ & $0.184 \pm 0.019$ & \multirow{8}{*}{watercolor} & blue          & $0.040 \pm 0.023$ & $0.170 \pm 0.006$ \\
  & carrot        & $0.085 \pm 0.051$ & $0.198 \pm 0.024$ &                              & carrot        & $0.055 \pm 0.030$ & $0.178 \pm 0.013$ \\
  & dark-red      & $0.101 \pm 0.061$ & $0.229 \pm 0.038$ &                              & dark-red      & $0.055 \pm 0.031$ & $0.179 \pm 0.013$ \\
  & midnight-blue & $0.103 \pm 0.061$ & $0.238 \pm 0.039$ &                              & midnight-blue & $0.050 \pm 0.027$ & $0.176 \pm 0.009$ \\
  & nephritis     & $0.059 \pm 0.035$ & $0.167 \pm 0.012$ &                              & nephritis     & $0.044 \pm 0.025$ & $0.170 \pm 0.007$ \\
  & turquoise     & $0.061 \pm 0.037$ & $0.171 \pm 0.014$ &                              & turquoise     & $0.045 \pm 0.024$ & $0.170 \pm 0.007$ \\
  & wisteria      & $0.076 \pm 0.046$ & $0.191 \pm 0.022$ &                              & wisteria      & $0.051 \pm 0.028$ & $0.178 \pm 0.011$ \\
  & yellow        & $0.067 \pm 0.041$ & $0.179 \pm 0.017$ &                              & yellow        & $0.055 \pm 0.032$ & $0.183 \pm 0.015$ \\
\midrule
\multicolumn{4}{c}{} & \multicolumn{2}{l}{\textit{Average}} & $0.079 \pm 0.029$ & $0.201 \pm 0.033$ \\
\bottomrule
\end{tabular}
\end{table*}

\section{Full Prompts}\label{sec:prompts}
\begin{figure*}[t]
\centering
\setlength{\tabcolsep}{2pt}
\renewcommand{\arraystretch}{0.2}
\small

\begin{tabular}{c}
  \parbox{0.95\linewidth}{\scriptsize\emph{``Step by step sketch process of a large oval with smaller shapes inside, following this drawing order: 1. A large oval in the center. 2. A small even-sided triangle inside the large oval at the top left. 3. A small oval inside the large oval at the top right. 4. A square with slightly curved sides inside the large oval in the center. 5. Three small circles grouped together inside the large oval at the bottom.''}} \\[6pt]
  \includegraphics[width=0.95\linewidth]{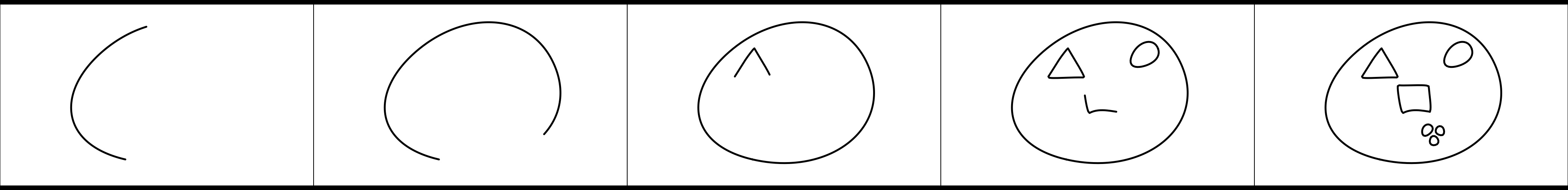} \\[6pt]

  \parbox{0.95\linewidth}{\scriptsize\emph{``Step by step sketch process of a large oval with smaller shapes inside, following this drawing order: 1. A square with slightly curved sides at the center. 2. A small even-sided triangle at the top left. 3. A small oval at the top right. 4. Three small circles grouped together at the bottom. 5. A large oval surrounding all the shapes.''}} \\[6pt]
  \includegraphics[width=0.95\linewidth]{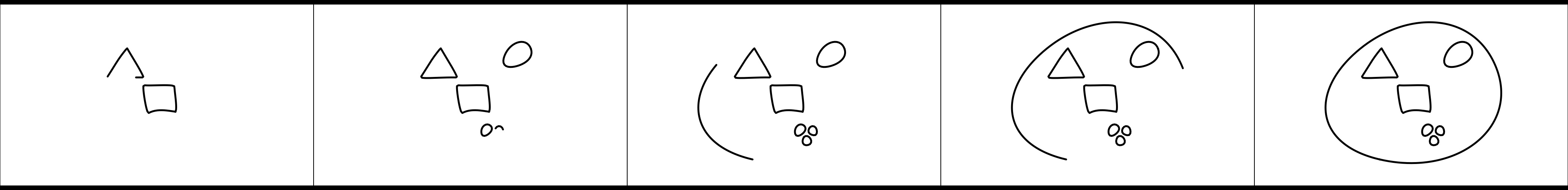} \\[6pt]

  \parbox{0.95\linewidth}{\scriptsize\emph{``Step by step sketch process of a large oval with smaller shapes inside, following this drawing order: 1. A square with slightly curved sides at the center. 2. A small even-sided triangle at the top left. 3. A small oval at the top right. 4. A large oval surrounding all the shapes. 5. Three small circles grouped together inside the large oval at the bottom.''}} \\[6pt]
  \includegraphics[width=0.95\linewidth]{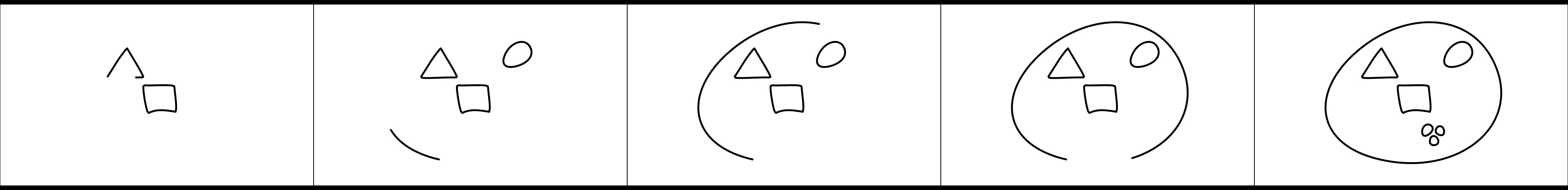} \\[6pt]

  \parbox{0.95\linewidth}{\scriptsize\emph{``Step by step sketch process of three polygonal shapes, following this drawing order: 1. A small triangle with slightly curved edges on the left. 2. An uneven six-sided shape with straight edges in the middle. 3. A tilted, trapezoid-like shape on the right.''}} \\[6pt]
  \includegraphics[width=0.95\linewidth]{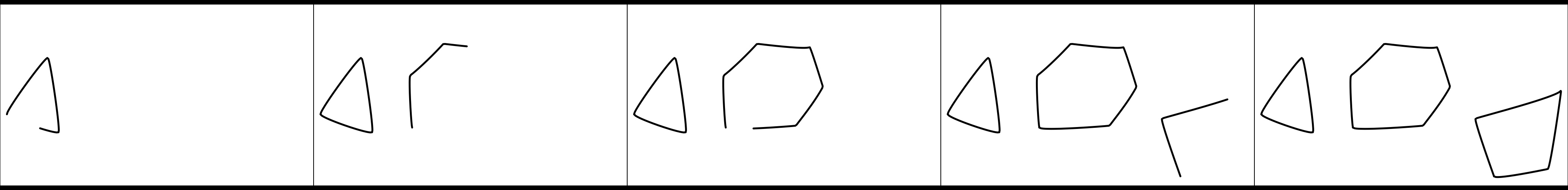} \\[6pt]

  \parbox{0.95\linewidth}{\scriptsize\emph{``Step by step sketch process of three polygonal shapes, following this drawing order: 1. A tilted, trapezoid-like shape on the right. 2. An uneven six-sided shape with straight edges in the middle. 3. A small triangle with curved edges on the left.''}} \\[6pt]
  \includegraphics[width=0.95\linewidth]{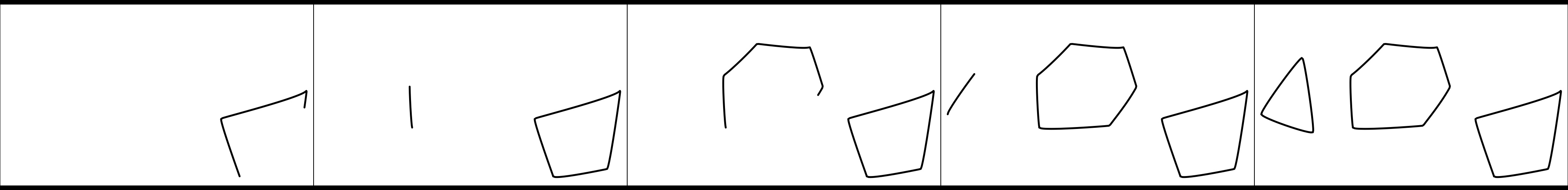} \\[6pt]

  \parbox{0.95\linewidth}{\scriptsize\emph{``Step by step sketch process of three polygonal shapes, following this drawing order: 1. An uneven six-sided shape with straight edges in the middle. 2. A small triangle with curved edges on the left. 3. A tilted, trapezoid-like shape on the right.''}} \\[6pt]
  \includegraphics[width=0.95\linewidth]{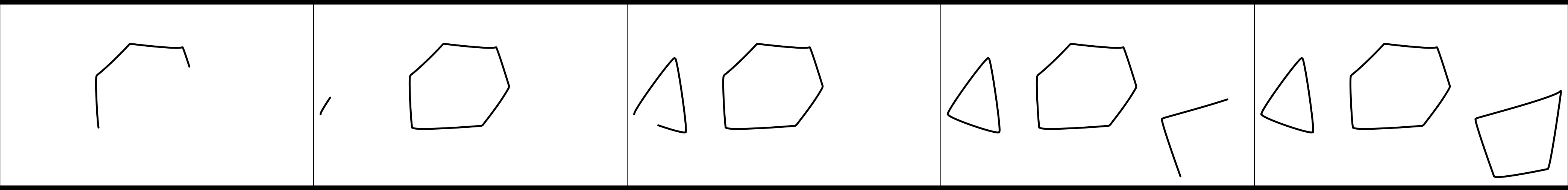} \\

  \parbox{0.95\linewidth}{\scriptsize\emph{``Step by step sketch process of two shapes and several loose lines, following this drawing order: 1. A horizontal ellipse at the top left. 2. A shape with a flat bottom and rounded sides below the ellipse. 3. On the right, two loose diagonal lines, one slanting downward and one slanting upward. 4. Slightly below and left to the lines, a straight horizontal line. 5. On the bottom right, three short, straight vertical lines grouped together.''}} \\[6pt]
  \includegraphics[width=0.95\linewidth]{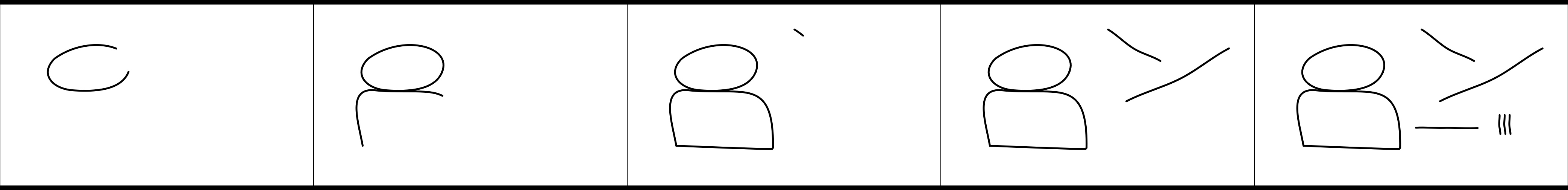} \\[6pt]

\end{tabular}

\vspace{-8pt}
\caption{\textbf{Training data exemplars of our simple geometric primitives.} Examples of the simple geometric primitives used in the first stage of our fine-tuning pipeline, focused on teaching the model basic drawing ``grammar''. Each composition is rendered with multiple drawing orders, as specified by the text prompt.}
\label{fig:supp_trainset_shapes_1}
\end{figure*}

\begin{figure*}[t]
\centering
\setlength{\tabcolsep}{2pt}
\renewcommand{\arraystretch}{0.2}
\small

\begin{tabular}{c}
 \parbox{0.95\linewidth}{\scriptsize\emph{``Step by step sketch process of two shapes and several loose lines, following this drawing order: 1. A shape with a flat bottom and rounded sides at the bottom left. 2. A horizontal ellipse on top of it. 3. A straight horizontal line to the right of the flat bottom shape. 4. Three short, straight vertical lines grouped together on the bottom right, to the right of the horizontal line. 5. On the top right, two loose diagonal lines, one slanting downward and one slanting upward.''}} \\[6pt]
  \includegraphics[width=0.95\linewidth]{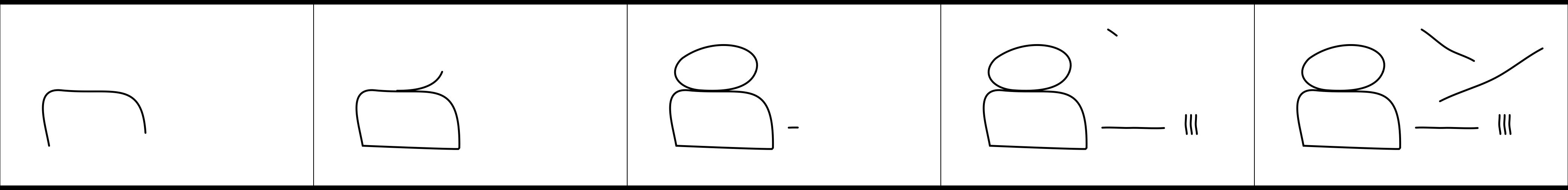} \\[6pt]
  
   \parbox{0.95\linewidth}{\scriptsize\emph{``Step by step sketch process of two shapes and several loose lines, following this drawing order: 1. Two loose diagonal lines at the top right, one slanting downward and one slanting upward. 2. At the bottom left, a shape with a flat bottom and rounded sides. 3. A horizontal ellipse on top of it. 4. Slightly below and left to the lines, a straight horizontal line. 5. On the bottom right, three short, straight vertical lines grouped together.''}} \\
  \includegraphics[width=0.95\linewidth]{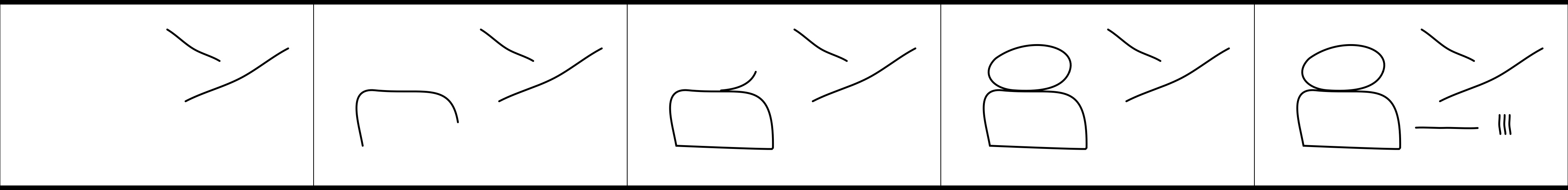} \\[6pt]
  
  \parbox{0.95\linewidth}{\scriptsize\emph{``Step by step sketch process of three stacked shapes, following this drawing order: 1.A wide triangle with a curved base at the bottom. 2.A slightly curved tall rectangle in the middle. 3.A slightly oval circle on top.''}} \\
  \includegraphics[width=0.95\linewidth]{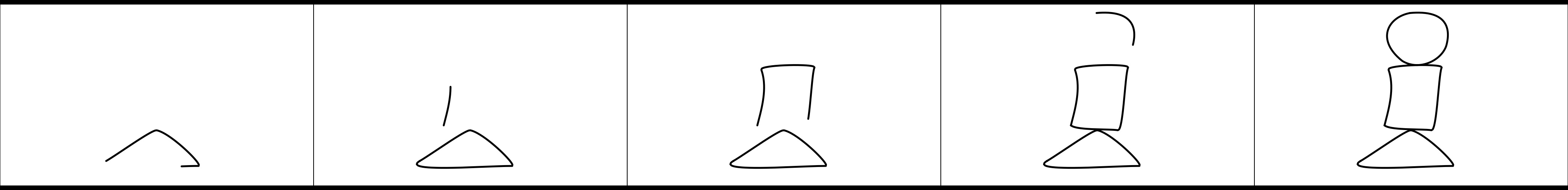} \\[6pt]

  \parbox{0.95\linewidth}{\scriptsize\emph{``Step by step sketch process of three stacked shapes, following this drawing order: 1. A slightly oval circle on top. 2. A wide triangle with a curved base at the bottom. 3. A slightly curved tall rectangle in the middle.''}} \\[6pt]
  \includegraphics[width=0.95\linewidth]{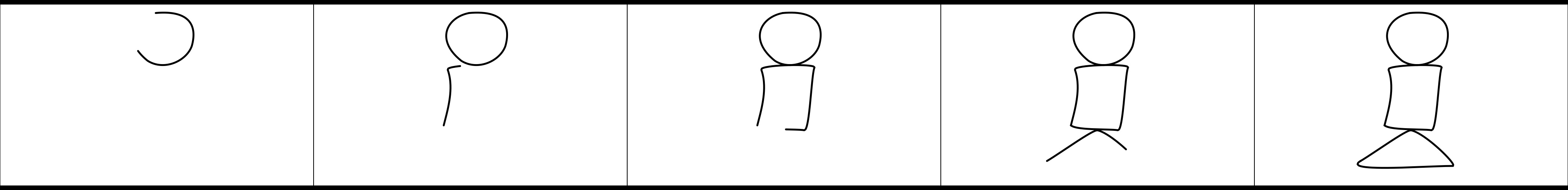} \\[6pt]

  \parbox{0.95\linewidth}{\scriptsize\emph{``Step by step sketch process of three stacked shapes, following this drawing order: 1. A slightly curved tall rectangle in the middle. 2. A slightly oval circle on top. 3. A wide triangle with a curved base at the bottom.''}} \\[6pt]
  \includegraphics[width=0.95\linewidth]{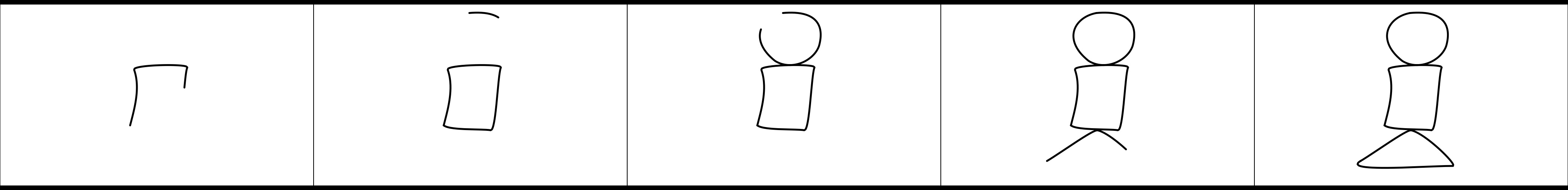} \\[6pt]

  \parbox{0.95\linewidth}{\scriptsize\emph{``Step by step sketch process of several free-form shapes and loose lines, following this drawing order: 1. Top right -- a large, irregular shape with smooth, flowing curves. 2. A smaller, irregular shape with rounded edges at the bottom left. 3. A short, loose curved line at the top left. 4. A longer, loose curved line at the bottom center. 5. A loose spiral line at the bottom right.''}} \\[6pt]
  \includegraphics[width=0.95\linewidth]{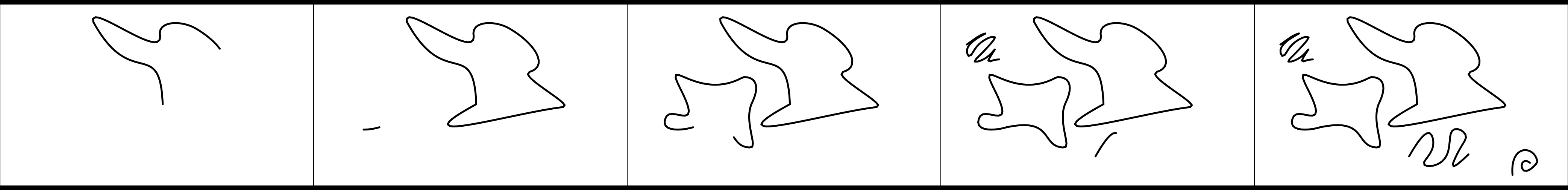} \\[6pt]

  \parbox{0.95\linewidth}{\scriptsize\emph{``Step by step sketch process of several free-form shapes and loose lines, following this drawing order: 1. A smaller, irregular shape with rounded edges at the bottom left. 2. Top right -- a large, irregular shape with smooth, flowing curves. 3. A short, loose curved line at the top left. 4. A longer, loose curved line at the bottom center. 5. A loose spiral line at the bottom right.''}} \\[6pt]
  \includegraphics[width=0.95\linewidth]{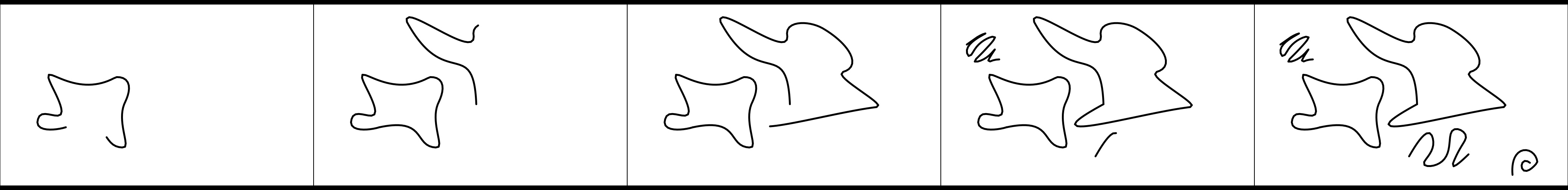} \\[6pt]

  \parbox{0.95\linewidth}{\scriptsize\emph{``Step by step sketch process of several free-form shapes and loose lines, following this drawing order: 1. Top right -- a large, irregular shape with smooth, flowing curves. 2. A smaller, irregular shape with rounded edges at the bottom left. 3. A loose spiral line at the bottom right. 4. A longer, loose curved line at the bottom center. 5. A short, loose curved line at the top left.''}} \\[6pt]
  \includegraphics[width=0.95\linewidth]{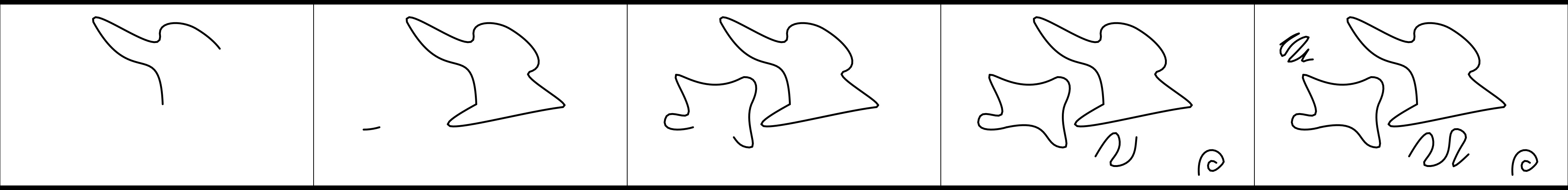} \\[6pt]

\end{tabular}

\vspace{-8pt}
\caption{\textbf{Training data exemplars of our simple geometric primitives (continued).} Additional stacked-shape compositions with varied drawing orders.}
\label{fig:supp_trainset_shapes_2}
\end{figure*}

\begin{figure*}[t]
\centering
\setlength{\tabcolsep}{2pt}
\renewcommand{\arraystretch}{0.2}
\small

\begin{tabular}{c}

  \parbox{0.95\linewidth}{\scriptsize\emph{``Step by step sketch process of a butterfly, following this drawing order: 1. Body -- long oval shape running vertically. 2. Right wing. 3. Left wing. 4. Head -- a small circle on top of the body. 5. Antennae -- two long curved lines extending upward from the head, ending in small curls.''}} \\[6pt]
  \includegraphics[width=0.95\linewidth]{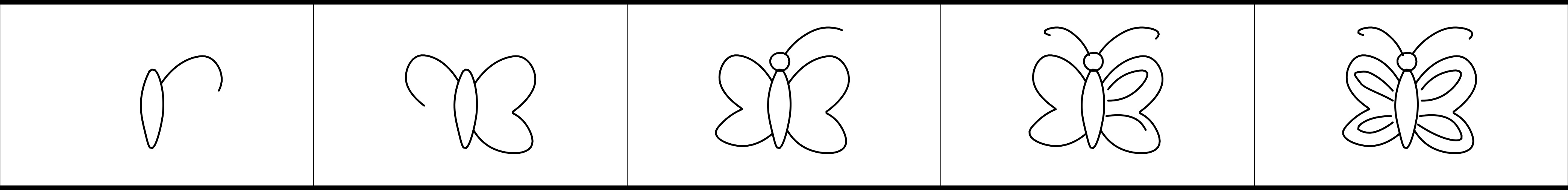} \\[6pt]

  \parbox{0.95\linewidth}{\scriptsize\emph{``Step by step sketch process of a car, following this drawing order: 1. Two circles for the wheels. 2. The car's body. 3. The side windows. 4. The front light. 5. The doors.''}} \\[6pt]
  \includegraphics[width=0.95\linewidth]{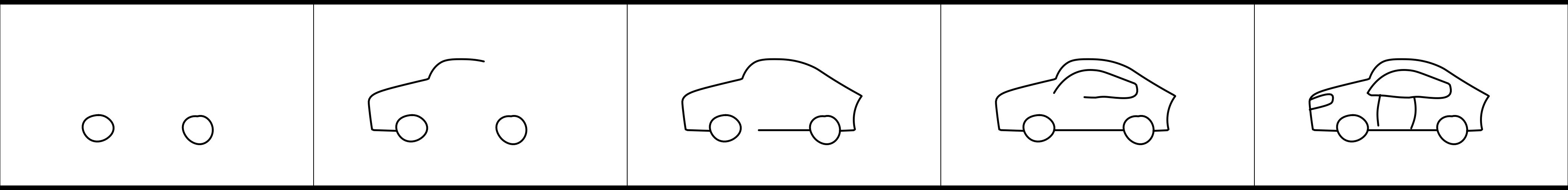} \\[6pt]

  \parbox{0.95\linewidth}{\scriptsize\emph{``Step by step sketch process of a chair, following this drawing order: 1. Seat -- a flat horizontal surface, shape like a trapezoid. 2. Backrest Slats -- two horizontal bars. 3. Backrest Supports -- vertical section connecting the seat and the backrest slats. 4. Front Legs -- two rectangles at the front. 5. Back Legs -- two rectangles at the back.''}} \\[6pt]
  \includegraphics[width=0.95\linewidth]{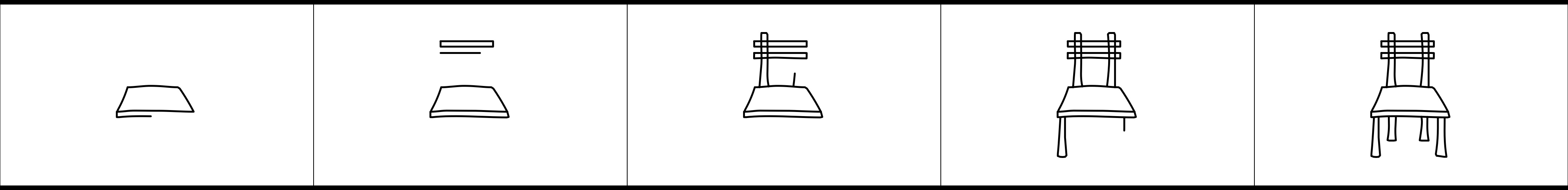} \\[6pt]

  \parbox{0.95\linewidth}{\scriptsize\emph{``Step by step sketch process of a coffee cup, following this drawing order: 1. The cup's body in the center, as a U shape. 2. The top edge of the cup as an oval on top. 3. The coffee surface as a curved line. 4. The handle on the right. 5. The cup's base. 6. Three wavy lines above the cup to indicate steam.''}} \\[6pt]
  \includegraphics[width=0.95\linewidth]{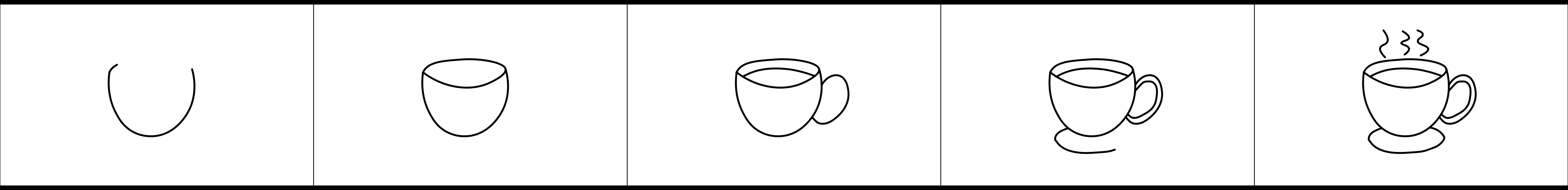} \\[6pt]

  \parbox{0.95\linewidth}{\scriptsize\emph{``Step by step sketch process of a flower, following this drawing order: 1. A circle indicating the flower's head. 2. Six petals around the circle drawn clockwise. 3. The stem, a curved line extends downward from the flower head. 4. Leaf on the left side of the stem.''}} \\[6pt]
  \includegraphics[width=0.95\linewidth]{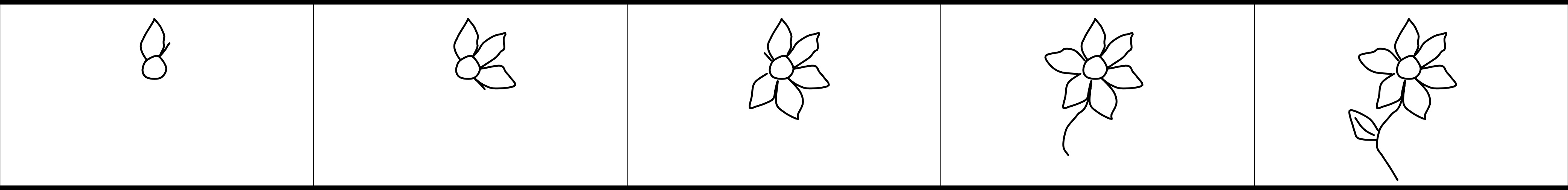} \\[6pt]

  \parbox{0.95\linewidth}{\scriptsize\emph{``Step by step sketch process of a desk lamp, following this drawing order: 1. Lampshade -- a cone-shaped top part that directs the light downward. 2. Light bulb inside the lampshade, an inner oval line. 3. The upper arm. 4. The joint connecting the arm segments -- a small circle. 5. The lower arm. 6. The base -- a rounded slightly flattened structure at the bottom. 7. Light beam emanating from the bulb, represented as four short lines.''}} \\[6pt]
  \includegraphics[width=0.95\linewidth]{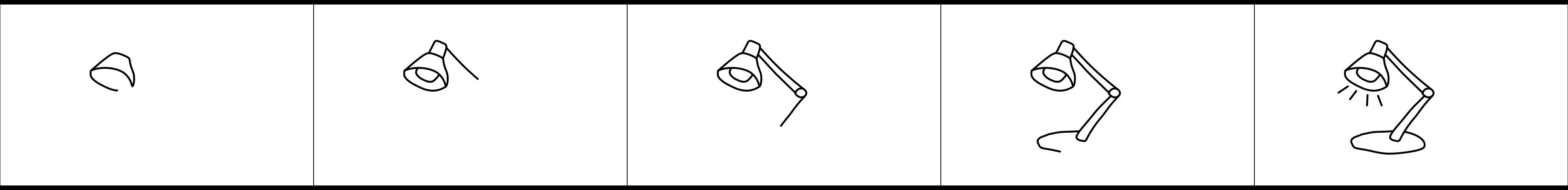} \\[6pt]

  \parbox{0.95\linewidth}{\scriptsize\emph{``Step by step sketch process of a tree, following this drawing order: 1. The trunk -- two vertical lines. 2. The canopy from the two sides of the trunk. 3. The branches inside the canopy.''}} \\[6pt]
  \includegraphics[width=0.95\linewidth]{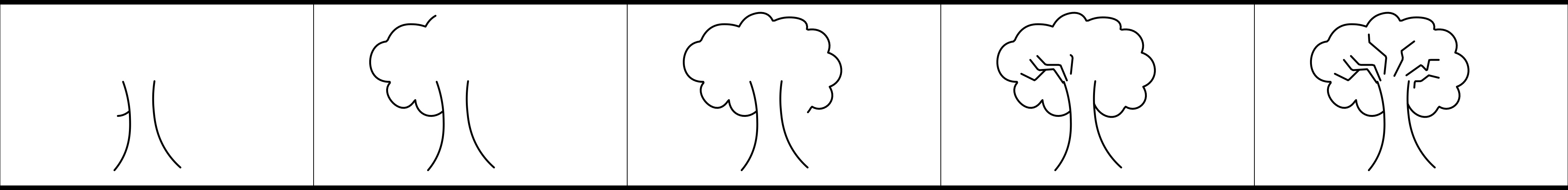}
\end{tabular}

\vspace{-8pt}
\caption{\textbf{Our seven real human sketches used for training.} We provide the seven human-drawn sketches used in the second stage of our fine-tuning pipeline, along with their full text prompts.}
\label{fig:supp_trainset_sketches}
\end{figure*}

We provide the full prompts used throughout our pipeline. The prompts for our trainset videos are presented alongside the visualizations in~\Cref{fig:supp_trainset_shapes_1,fig:supp_trainset_shapes_2,fig:supp_trainset_sketches}.

\subsection{Drawing-Order Planning Prompt}\label{sec:prompt_order_plan}
At inference time, given a user-provided concept (\eg, ``giraffe''), we prompt an LLM to produce a structured, step-by-step drawing-order plan. The user message follows the template: \emph{``Convert this concept into drawing instructions: \{concept\}''}. The full system prompt and few-shot examples are shown below.

\begin{promptbox}[Drawing-Order Planning -- System Prompt]
You are a drawing instruction generator. Convert any given concept into clear, step-by-step sketch instructions.

\medskip
\textbf{Output Format:}\\
``Step by step sketch process of a [concept], following this drawing order: 1.~[Part name] -- [simple shape description and position]. 2.~[Part name] -- [simple shape description and position]. \ldots''

\medskip
\textbf{Ordering Principles (in priority):}
\begin{enumerate}[leftmargin=1.5em, itemsep=1pt, parsep=0pt, topsep=2pt]
\item \textbf{Iconic features first} -- Start with the parts that make this thing instantly recognizable (cat $\rightarrow$ ears + face shape; elephant $\rightarrow$ trunk; Eiffel Tower $\rightarrow$ tapered lattice frame).
\item \textbf{Anchoring structure} -- Include the core element that other parts attach to, but only if needed for positioning.
\item \textbf{Distinctive before generic} -- Unique characteristics before common shapes.
\item \textbf{Connected parts in sequence} -- Parts that attach to each other should be drawn in logical connection order.
\end{enumerate}

\medskip
\textbf{Rules:}
\begin{itemize}[leftmargin=1.5em, itemsep=1pt, parsep=0pt, topsep=2pt]
\item Output exactly ONE line. Do not include any newline characters.
\item Use this exact template: ``Step by step sketch process of a [concept], following this drawing order: 1. 2. 3. \ldots''
\item Describe each part with a simple shape (oval, circle, rectangle, curved line, etc.).
\item Include position relative to other parts: ``above,'' ``attached to,'' ``extending from.''
\item Keep descriptions concise (under 15 words per step).
\item Provide 5--10 steps for most subjects.
\item Keep the language consistent with the user input.
\item Do not add numbering or bullet points outside the required format.
\end{itemize}

\medskip
\textbf{Examples:}

\medskip
\textit{Concept:} butterfly\\
\textit{Output:} Step by step sketch process of a butterfly, following this drawing order: 1.~Body -- long vertical oval. 2.~Right wing -- large rounded shape attached to body. 3.~Left wing -- mirror of right wing. 4.~Head -- small circle atop body. 5.~Antennae -- two curved lines from head, ending in curls.

\medskip
\textit{Concept:} car\\
\textit{Output:} Step by step sketch process of a car, following this drawing order: 1.~Wheels -- two circles spaced apart. 2.~Body -- rounded rectangle connecting wheels. 3.~Windows -- smaller rectangles on top. 4.~Headlight -- small circle at front. 5.~Door -- vertical line on body.

\medskip
\textit{Concept:} tree\\
\textit{Output:} Step by step sketch process of a tree, following this drawing order: 1.~Trunk -- two vertical lines. 2.~Canopy -- large cloud shape from top of trunk. 3.~Branches -- smaller lines visible within canopy. 4.~Roots -- short angled lines at trunk base.
\end{promptbox}

\subsection{Ordering Fidelity Evaluation Prompts}\label{sec:prompt_eval}
Our ordering fidelity metric (described in~\Cref{sec:implementation}) uses an LLM in two stages. In the first stage, a video captioning prompt instructs the LLM to observe a sketch video and extract the sequence of drawn semantic parts. In the second stage, a head-to-head comparison prompt evaluates which of two model outputs better preserves the target ordering.

\begin{promptbox}[Stage 1: Video Captioning Prompt]
You are an AI assistant specialized in analyzing sketching videos and extracting the ordered sequence of semantic parts that are drawn.

\medskip
\textbf{Task.}
You will receive: (1) a video showing a step-by-step sketching process on a canvas, (2) the name of the object being drawn, and (3) a numbered list describing what was supposed to be drawn and in what order.
Your task is to OBSERVE the video and report the ACTUAL order in which semantic parts are drawn.
The numbered list is provided ONLY as a semantic reference for part names --- NOT as an ordering constraint.

\medskip
\textbf{How to Observe the Video.}
Watch the video carefully and identify drawing steps by looking for: pen/brush touching the canvas (start of a drawing action), pen/brush lifting off (end of a drawing action), clear spatial jumps to different regions of the canvas, and completion of a recognizable semantic part before moving to another.
Focus on: chronological order of appearance, which semantic component is being drawn at each time, and natural grouping of strokes that together form a single part.

\medskip
\textbf{Definition of a ``Step.''}
A step corresponds to ONE semantic part of the object.

\textit{Merge into ONE step when:} multiple strokes together form a single semantic part, and the drawing of that part is continuous or nearly continuous.

\textit{Split into SEPARATE steps when:} the same semantic part is drawn in clearly separated time intervals, or distinct sub-parts are drawn at clearly different times (\eg, front legs early, back legs much later).

\textit{Returning to a part:} if additions occur immediately after $\rightarrow$ treat as the same step; if other parts are drawn in between $\rightarrow$ treat as a NEW step (\eg, ``Additional detail on head'').

\medskip
\textbf{Granularity Constraint (Critical).}
Prefer using the EXACT semantic part names from the numbered list. Match the granularity of the numbered list as closely as possible. Do NOT introduce abstract or generic steps such as ``Details,'' ``Cleanup,'' ``Refinement,'' ``Shading,'' or ``Outline cleanup'' unless they explicitly appear in the numbered list. Do NOT merge multiple numbered-list parts into one step unless the video clearly draws them together as a single inseparable action. Do NOT invent parts that are not visibly drawn.

\medskip
\textbf{Critical Rules.}
Report ONLY what is visible in the video. Do NOT assume the video follows the provided list order. Do NOT correct or ``fix'' the order. Do NOT judge correctness.

\medskip
\textbf{Output Format.}
Output a numbered list of parts, one per line (1.~[Part name], 2.~[Part name], \ldots). Part name ONLY --- no descriptions or explanations. Minimum 2 steps (typically 3--7).

\medskip
\textbf{Forbidden Outputs.}
Single-word output (\eg, ``butterfly''); paragraphs or sentences; unnumbered lists; abstract/meta steps not in the list; empty responses.

\medskip
If the video is ambiguous, make the best possible attempt based on visible drawing actions.
\end{promptbox}

\begin{promptbox}[Stage 2: Head-to-Head Comparison Prompt]
You are an AI assistant specialized in comparing two sketch captions to determine which one more closely preserves an original drawing order.

\medskip
\textbf{Task.}
You will receive: (1) an Original numbered list describing the intended drawing order, (2) a Model~1 Caption, and (3) a Model~2 Caption. Your task is to determine which model is MORE SIMILAR to the Original drawing order.

\medskip
\textbf{Evaluation Scope.}
Compare ONLY the ORDER of semantic parts. Ignore wording differences and visual quality. Do NOT reward verbosity or descriptive detail.

\medskip
\textbf{Deviation Types.}
For each model, count the following deviations relative to the Original:
\begin{enumerate}[leftmargin=1.5em, itemsep=1pt, parsep=0pt, topsep=2pt]
\item \texttt{reorder\_count}: any instance where original parts appear in a different relative order.
\item \texttt{middle\_omission\_count}: original parts missing from anywhere except the end.
\item \texttt{middle\_insertion\_count}: new parts inserted between original parts.
\item \texttt{split\_or\_merge\_count}: splitting one original step into multiple steps OR merging multiple original steps.
\item \texttt{end\_omission\_count}: original parts omitted from the end (used ONLY as final tie-breaker).
\end{enumerate}

\medskip
\textbf{Comparison Logic (Strict Lexicographic).}
Compare models in this exact order: (1) lower \texttt{reorder\_count} wins; (2) if tied, lower \texttt{middle\_omission\_count} wins; (3) if tied, lower \texttt{middle\_insertion\_count} wins; (4) if tied, lower \texttt{split\_or\_merge\_count} wins; (5) if tied, lower \texttt{end\_omission\_count} wins; (6) if ALL are equal $\rightarrow$ tie.

\medskip
\textbf{Special Case: No Valid Sequence.}
If a model provides no usable ordered list at all: \texttt{reorder\_count} = 0, \texttt{middle\_omission\_count} = number of original parts, \texttt{middle\_insertion\_count} = 0, \texttt{split\_or\_merge\_count} = 0, \texttt{end\_omission\_count} = number of original parts.

\medskip
\textbf{Output Format.}
Respond with a JSON object containing: \texttt{decision} (integer: 0 = tie, 1 = Model~1 is closer, 2 = Model~2 is closer) and \texttt{explanation} (concise reasoning tied to deviation counts).

\medskip
\textbf{Critical Constraints.}
You MUST NOT introduce additional comparison criteria. You MUST NOT prefer a model for being more detailed or descriptive. You MUST follow the lexicographic comparison exactly.
\end{promptbox}

\begin{figure*}[t]
\centering
\setlength{\tabcolsep}{2pt}
\renewcommand{\arraystretch}{0.5}
\small
\begin{tabular}{r c c}
\raisebox{0.6cm}{Wan 2.1} &
\includegraphics[width=0.45\linewidth]{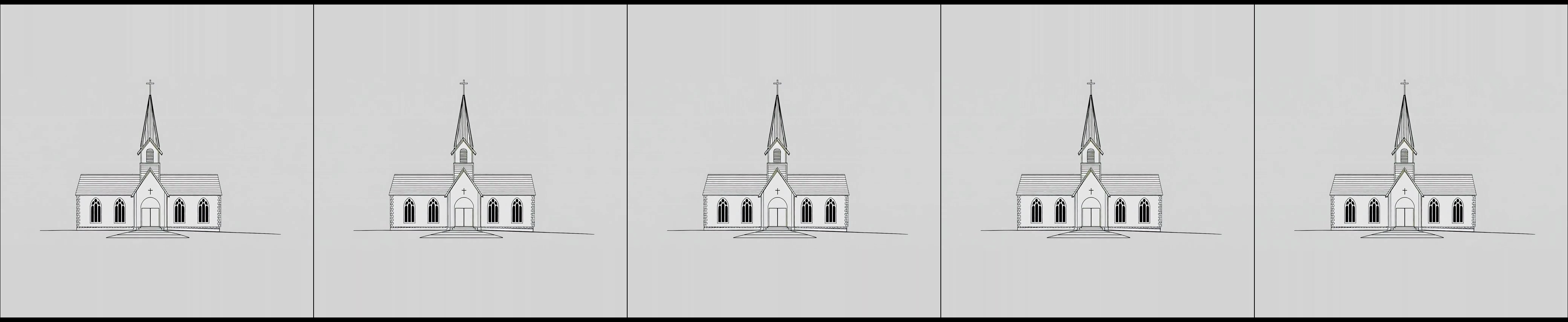} & \includegraphics[width=0.45\linewidth]{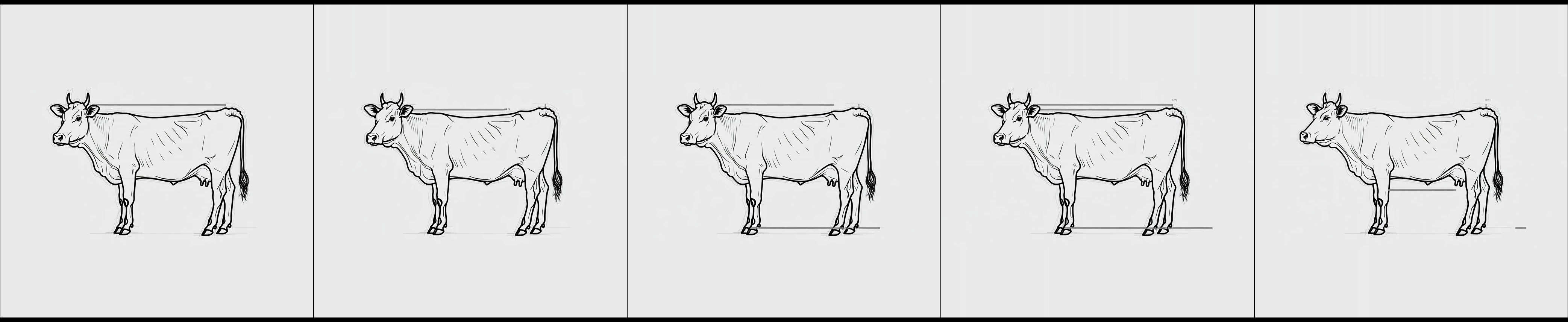} \\

\raisebox{0.6cm}{PaintsUndo} &
\includegraphics[width=0.45\linewidth]{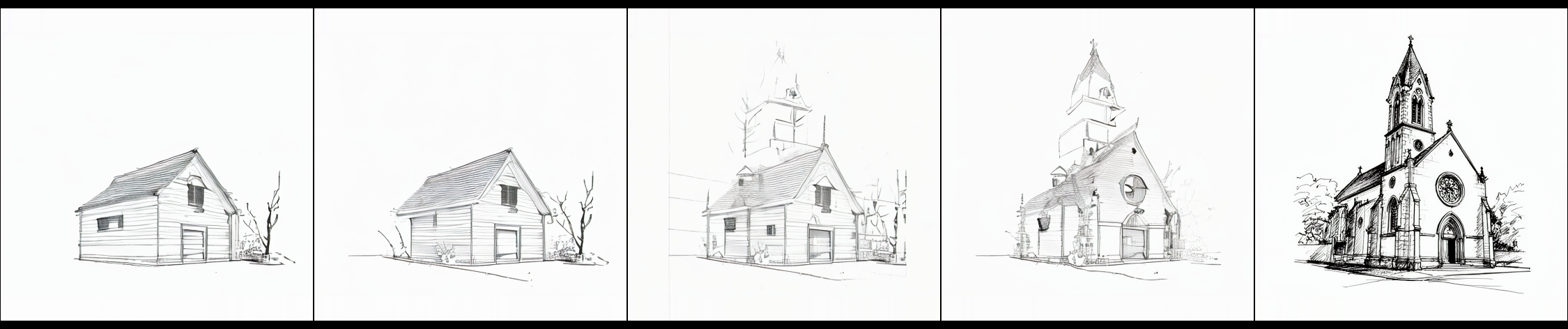} & \includegraphics[width=0.45\linewidth]{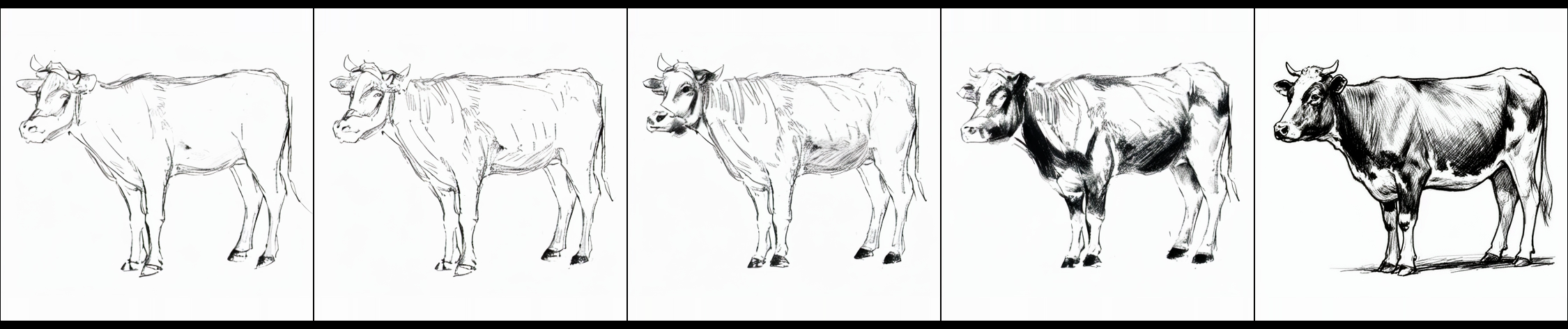} \\

\raisebox{0.6cm}{SketchAgent} &
\includegraphics[width=0.45\linewidth]{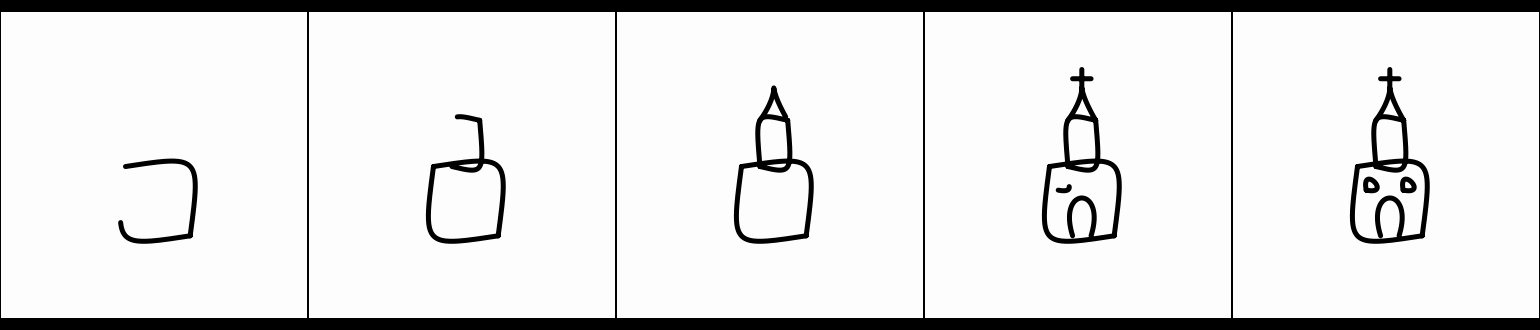} & \includegraphics[width=0.45\linewidth]{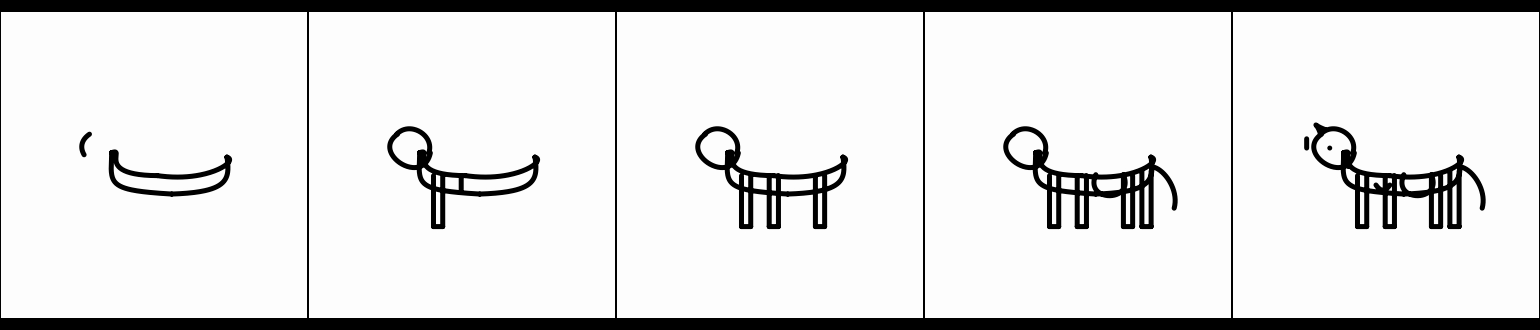} \\

\raisebox{0.6cm}{Human} &
\includegraphics[width=0.45\linewidth]{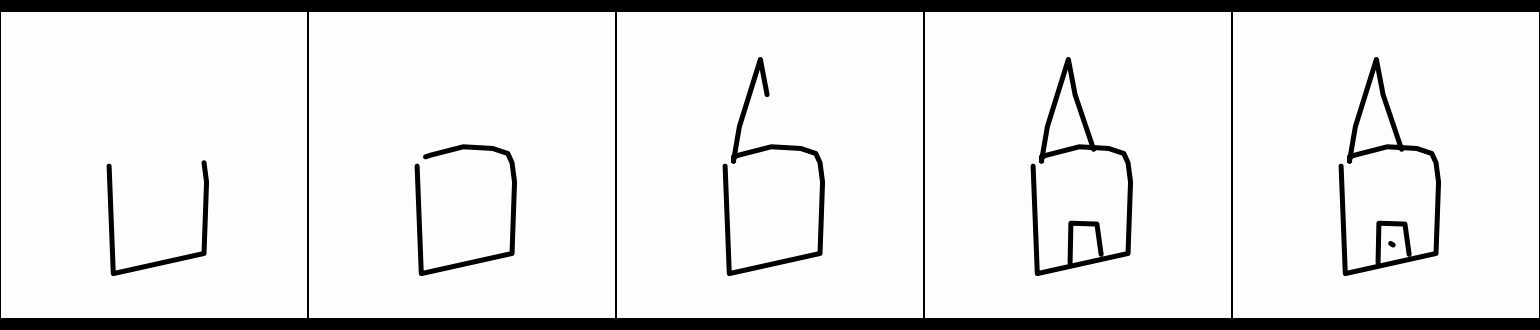} & \includegraphics[width=0.45\linewidth]{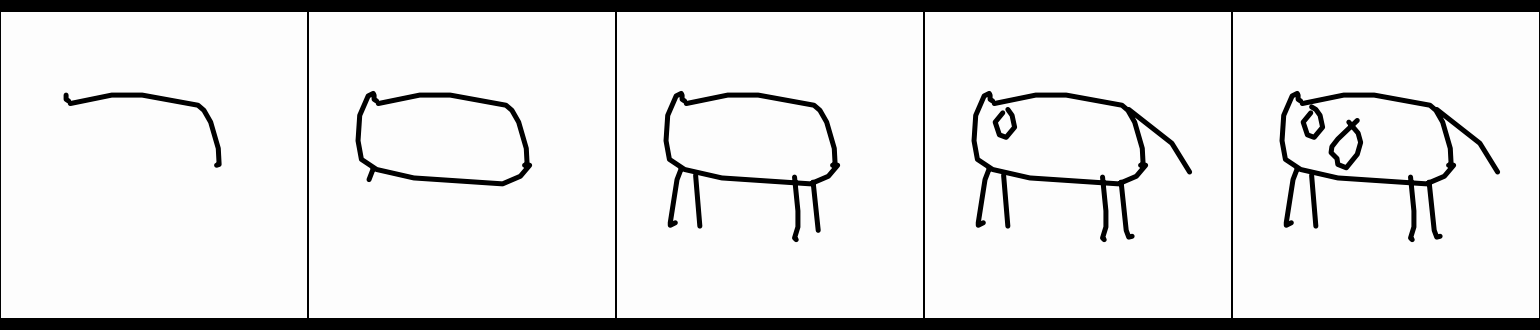} \\

\raisebox{0.6cm}{\textbf{Ours}} &
\includegraphics[width=0.45\linewidth]{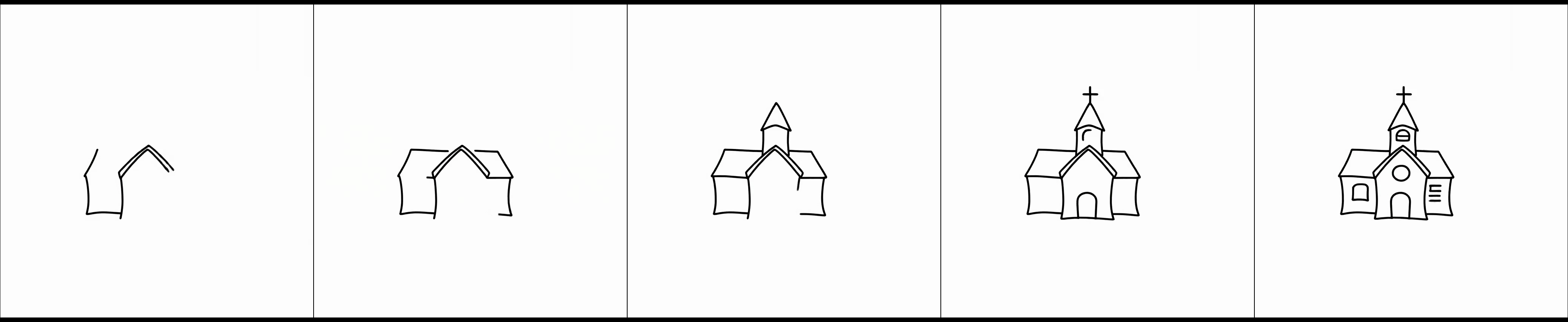} & \includegraphics[width=0.45\linewidth]{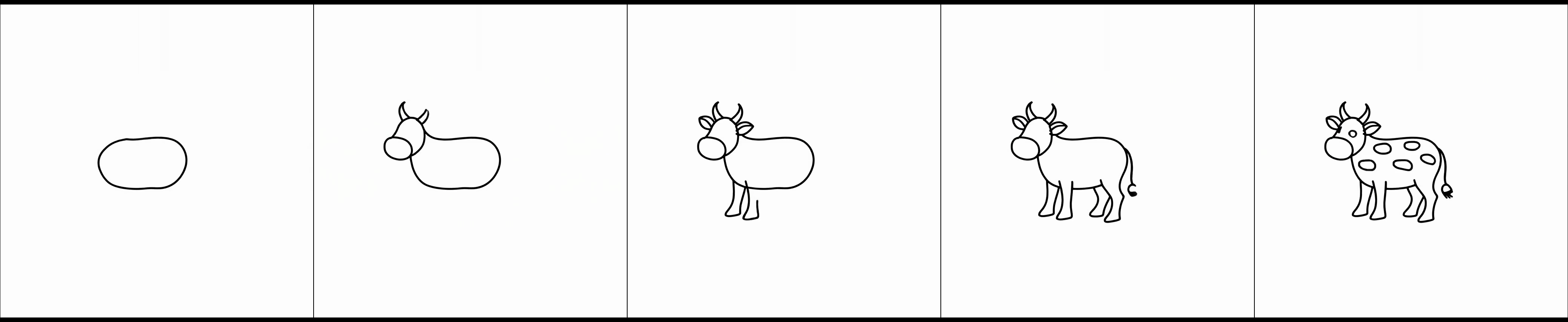}\\

& \emph{``A church''} & \emph{``A cow''}\\

\end{tabular}
\vspace{-4pt}
\caption{\textbf{Additional qualitative comparison of sequential sketch generation across methods.} The Human drawings are taken from QuickDraw~\cite{quickDrawData}. Full videos, and many more results are provided in the supplementary html.}
\label{fig:additional_qual_comparison_church}
\vspace{-6pt}
\end{figure*}

\newpage
\section{Additional Qualitative Results}
\Cref{fig:additional_qual_comparison_church} show comparisons to alternative methods on two additional concepts. The full set and videos are available in the supplamntary html.

\begin{figure*}[t]
\centering
\setlength{\tabcolsep}{2pt}
\renewcommand{\arraystretch}{0.2}
\small

\begin{tabular}{c}
  \parbox{1\linewidth}{\scriptsize\emph{``Step by step sketch process of an astronaut watering a garden on the moon, following this drawing order: 1. Astronaut body – rounded spacesuit figure. 2. Helmet and backpack. 3. Watering can in hand. 4. Small plants growing from the ground. 5. Moon surface with craters. 6. Earth floating in the sky behind.''}} \\[6pt]
  \includegraphics[width=1\linewidth]{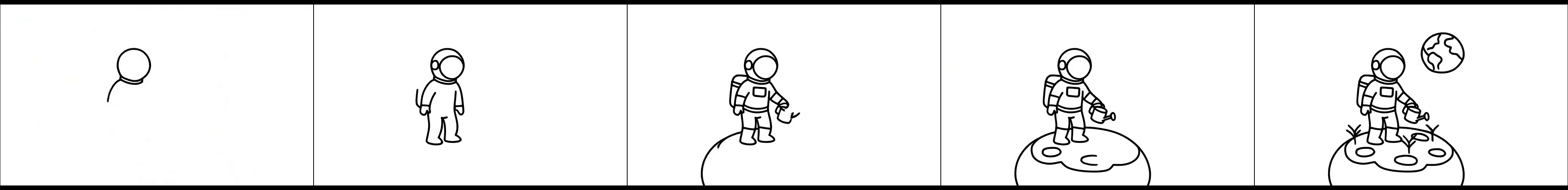} \\[2pt]
  \includegraphics[width=1\linewidth]{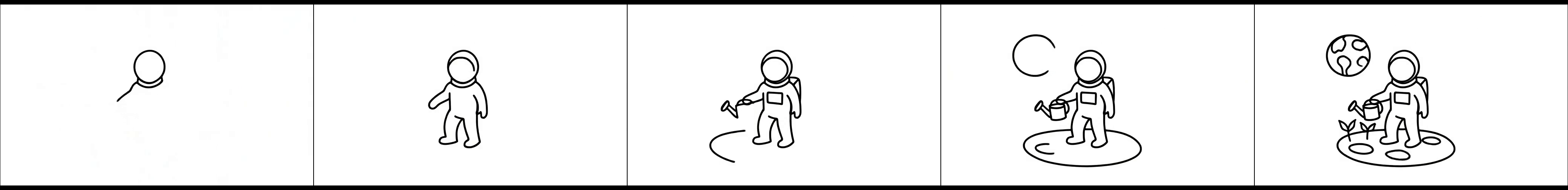} \\[2pt]

\includegraphics[width=1\linewidth]{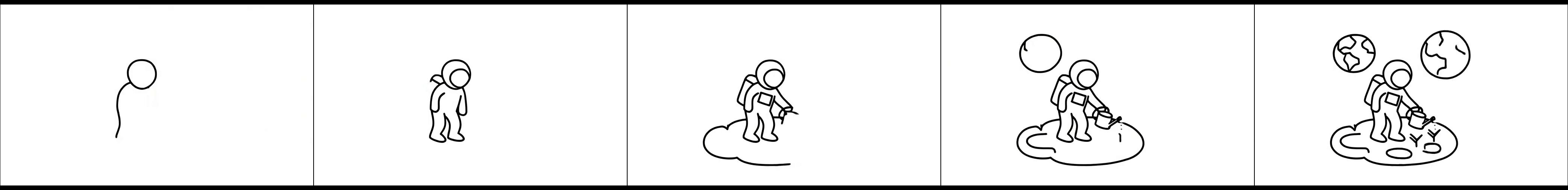} \\[2pt]

\includegraphics[width=1\linewidth]{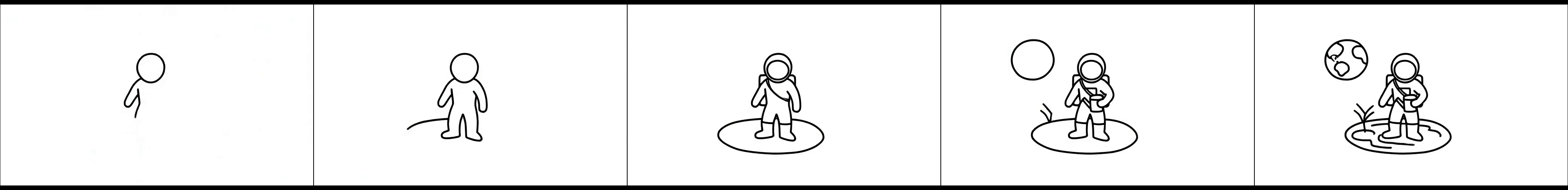} \\[8pt]  

  \parbox{1\linewidth}{\scriptsize\emph{``Step by step sketch process of a girl opening a doorway into outer space from her bedroom wall, following this drawing order: 1. Girl figure – standing pose. 2. Door frame on the wall. 3. Open doorway revealing stars and planets. 4. Bedroom furniture - bed or desk nearby. 5. Floating objects pulled toward the doorway. 6. Moon and planets beyond.''}} \\[6pt]
  \includegraphics[width=1\linewidth]{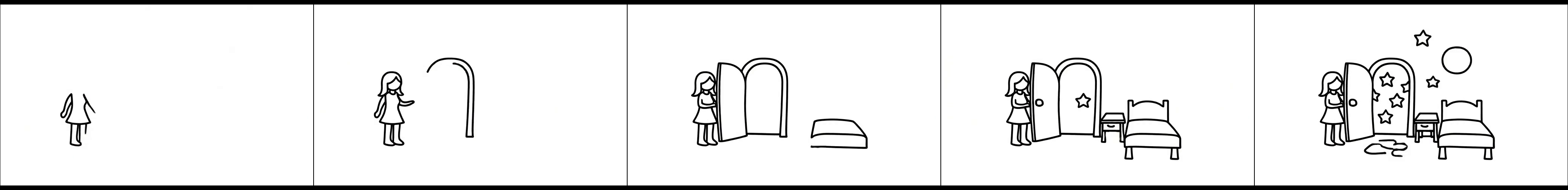} \\[2pt]
  \includegraphics[width=1\linewidth]{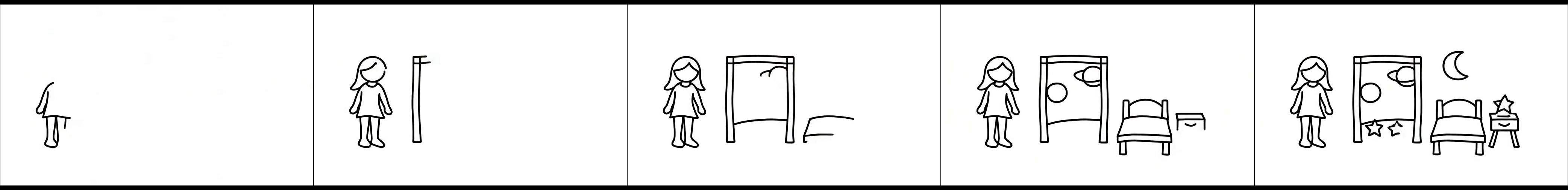} \\[2pt]

\includegraphics[width=1\linewidth]{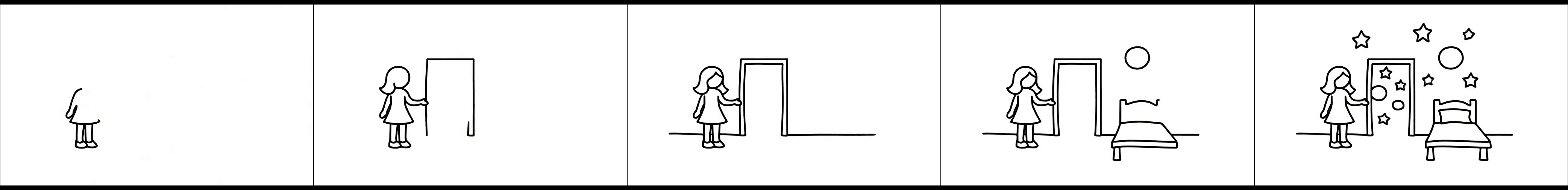} \\[2pt]

\includegraphics[width=1\linewidth]{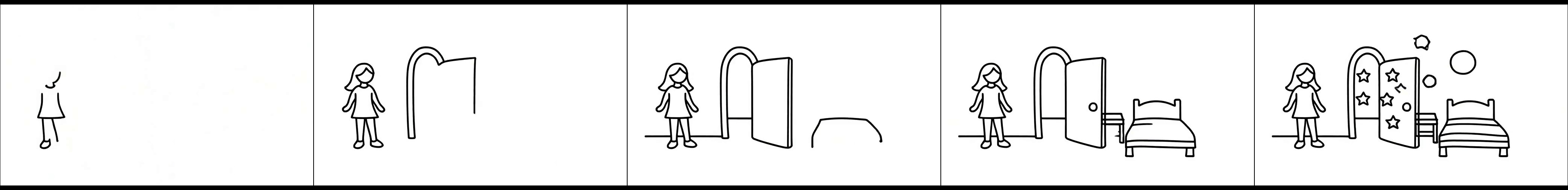} \\[2pt]  

\end{tabular}

\vspace{-8pt}
\caption{\textbf{Diversity of sequential sketch generation.} Using four differnt seeds for the same prompt result in four differnt sketches.}
\label{fig:seeds_diversity1}
\end{figure*}

\begin{figure*}[t]
\centering
\setlength{\tabcolsep}{2pt}
\renewcommand{\arraystretch}{0.2}
\small

\begin{tabular}{c}
  \parbox{1\linewidth}{\scriptsize\emph{``Step by step sketch process of a whale flying above a city with birds on its back, following this drawing order: 1. Whale body – large curved shape. 2. Tail and fins. 3. Head and eye. 4. Small birds perched on its back. 5. Buildings below. 6. Clouds surrounding the whale.''}} \\[6pt]
  \includegraphics[width=1\linewidth]{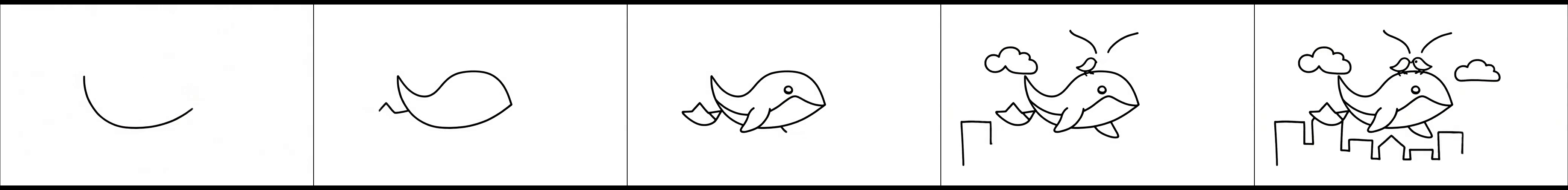} \\[2pt]
  \includegraphics[width=1\linewidth]{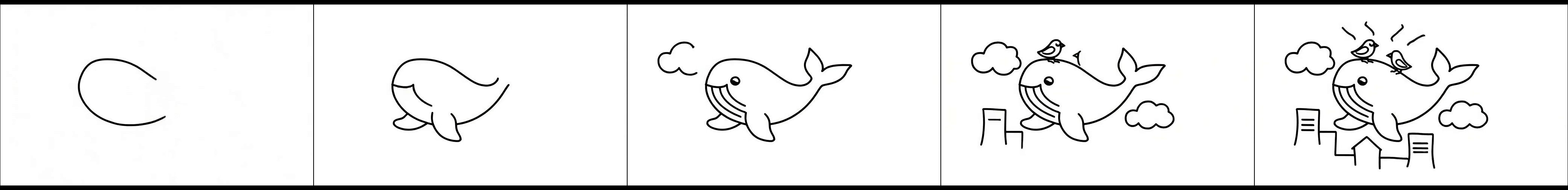} \\[2pt]

\includegraphics[width=1\linewidth]{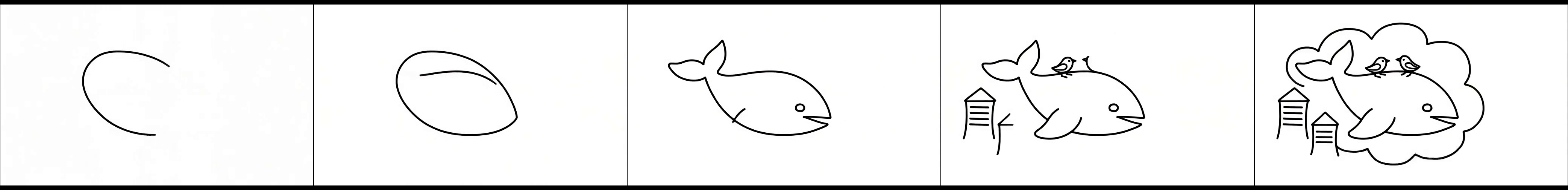} \\[2pt]

\includegraphics[width=1\linewidth]{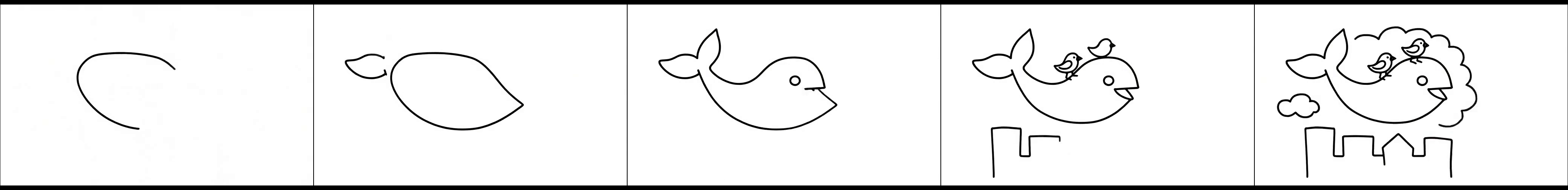} \\
[8pt]  

  \parbox{1\linewidth}{\scriptsize\emph{``Step by step sketch process of a fox exploring an ancient library, following this drawing order: 1. Fox body – small standing animal shape. 2. Head and ears. 3. Tail – large and fluffy. 4. Open book in front of the fox. 5. Tall bookshelves around it. 6. Floating candles or soft magical lights above.''}} \\[6pt]
  \includegraphics[width=1\linewidth]{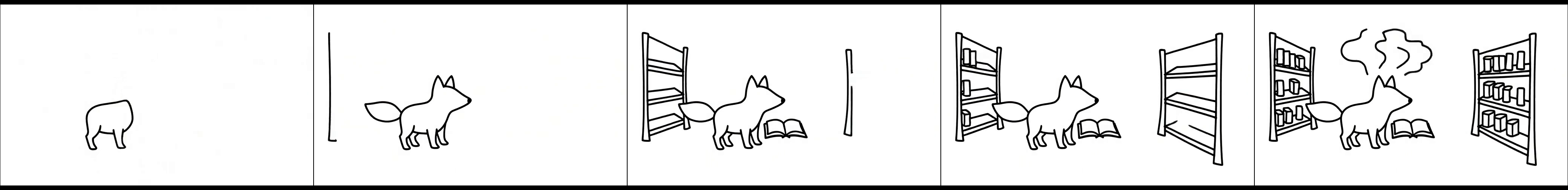} \\[2pt]
  \includegraphics[width=1\linewidth]{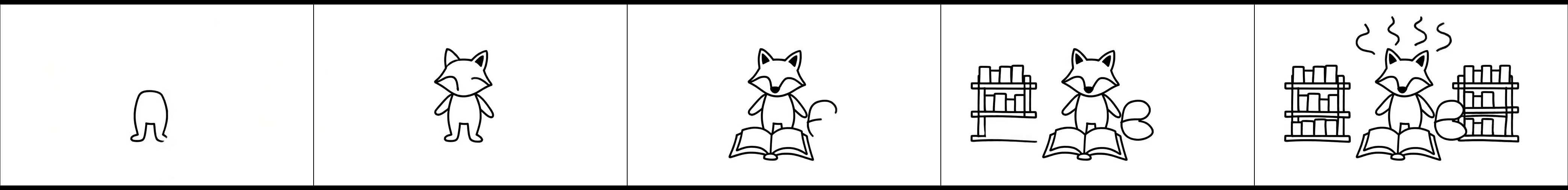} \\[2pt]

\includegraphics[width=1\linewidth]{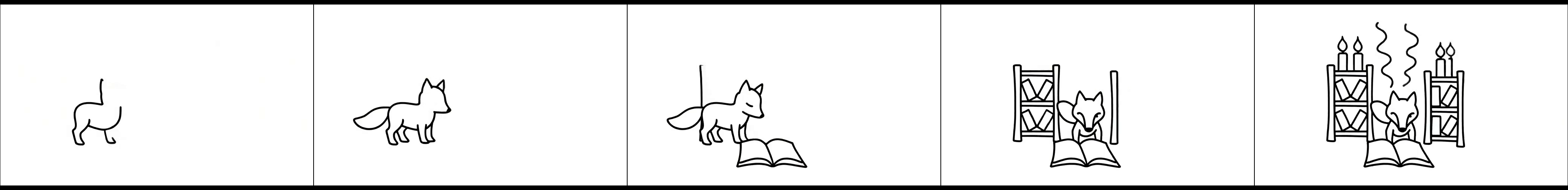} \\[2pt]

\includegraphics[width=1\linewidth]{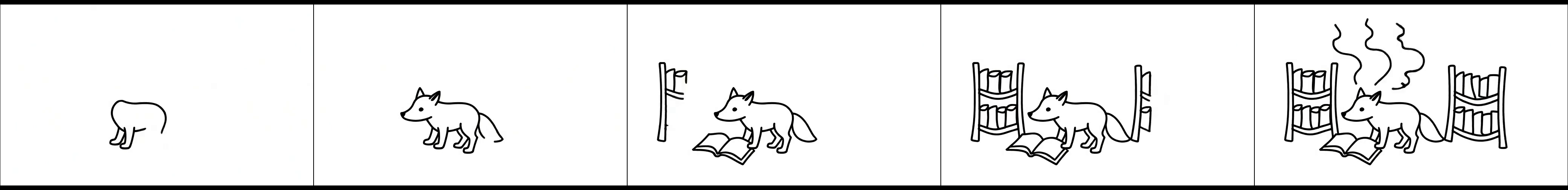} \\[2pt]  

\end{tabular}

\vspace{-8pt}
\caption{\textbf{Diversity of sequential sketch generation.} Using four differnt seeds for the same prompt result in four differnt sketches.}
\label{fig:seeds_diversity2}
\end{figure*}

\Cref{fig:seeds_diversity1,fig:seeds_diversity2} show the effect of varying the random seed while keeping the input prompt fixed. As can be seen, across different prompts and scenes, the generated sequences consistently follow the same high-level drawing plan, while exhibiting noticeable variation in composition, pose, and stroke execution. 
For example, in \Cref{fig:seeds_diversity1}, bottom, all sequences follow the same drawing plan: the girl is drawn first, followed by the doorway, then the interior of the doorway revealing space, the surrounding bedroom elements, and lastly the stars and planets.
While the specific realization varies, such as the shape and orientation of the door differ (e.g., arched vs. rectangular, partially vs. fully open), the placement of the bed and furniture changes, and the stars and planets are arranged differently across samples. 
This highlights that the model captures a stable drawing procedure while allowing flexibility in how each element is visually instantiated.

\begin{figure*}[t]
\centering
\setlength{\tabcolsep}{2pt}
\renewcommand{\arraystretch}{0.2}
\small

 \parbox{1\linewidth}{\scriptsize\emph{``Step by step sketch process of a polar bear fishing under the northern lights, using the color and style of the brush shown in the top-left corner, following this drawing order: 1. Polar bear body – large seated animal shape. 2. Head and paws. 3. Ice hole in the ground. 4. Fishing rod held by the bear. 5. Fish jumping from the water. 6. Northern lights sweeping across the sky.''}} \\
\includegraphics[width=\linewidth]{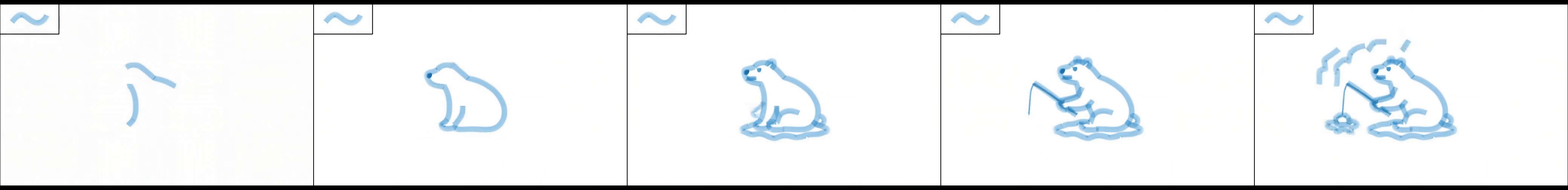} \\[2pt]
\includegraphics[width=\linewidth]{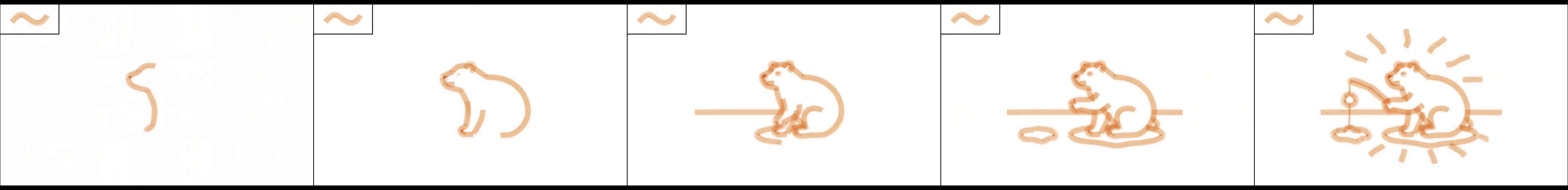} \\[2pt]
\includegraphics[width=\linewidth]{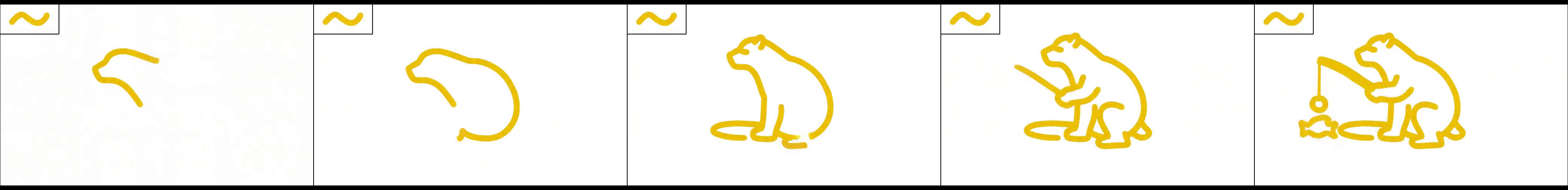} \\[2pt]
\includegraphics[width=\linewidth]{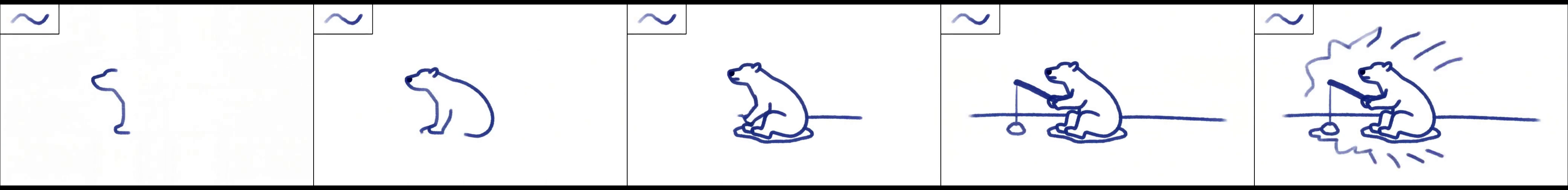}
\\[2pt]
\includegraphics[width=\linewidth]{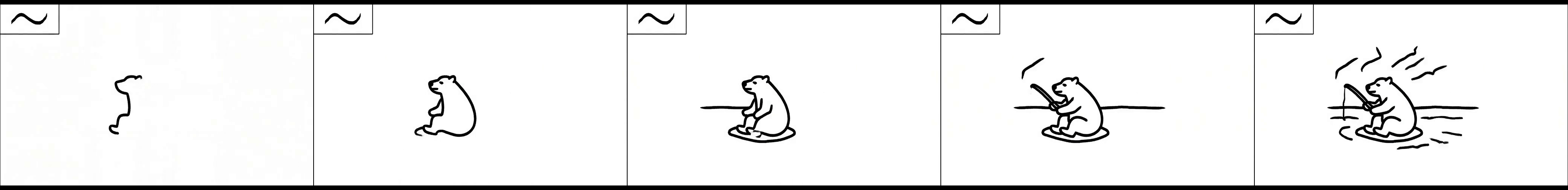}\\[10pt]

 \parbox{1\linewidth}{\scriptsize\emph{``Step by step sketch process of a girl riding a unicorn in the sky, using the color and style of the brush shown in the top-left corner, following this drawing order: 1. Unicorn body – large horse-like shape. 2. Legs – running or floating pose. 3. Head and horn. 4. Rider – seated figure on the back. 5. Mane and tail flowing. 6. Clouds around them.''}} \\
\includegraphics[width=\linewidth]{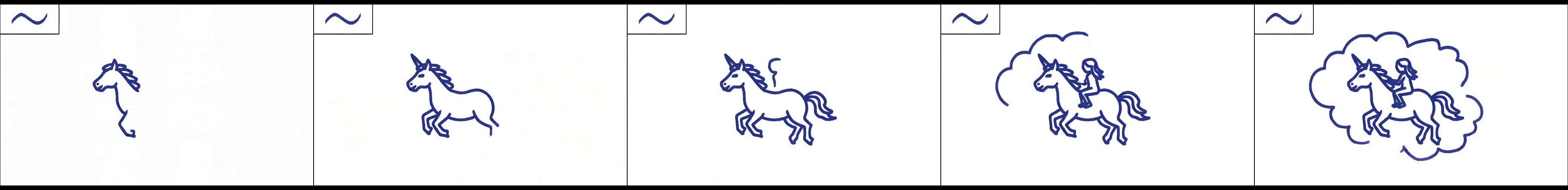} \\[2pt]
\includegraphics[width=\linewidth]{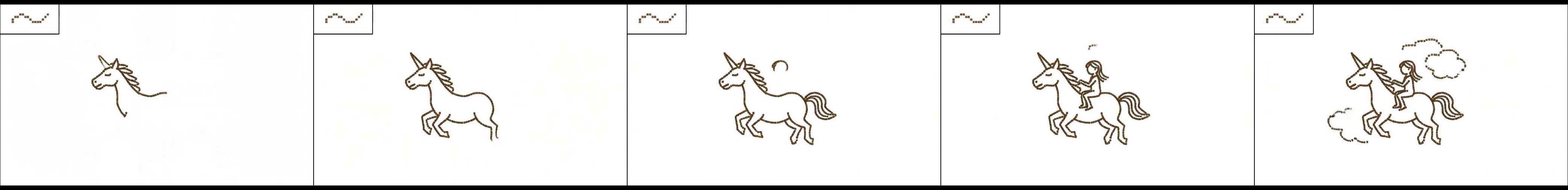} \\[2pt]
\includegraphics[width=\linewidth]{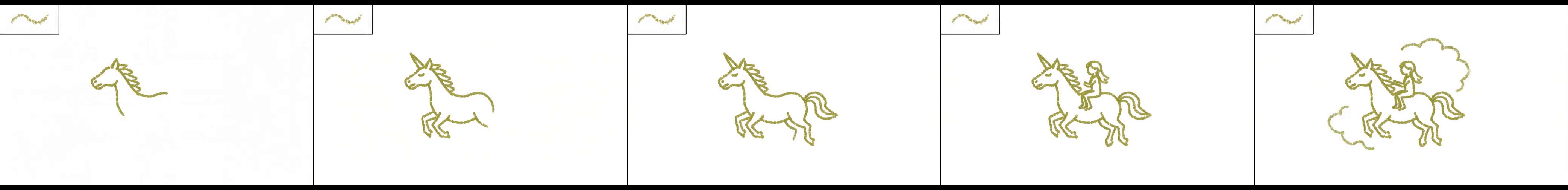} \\[2pt]
\includegraphics[width=\linewidth]{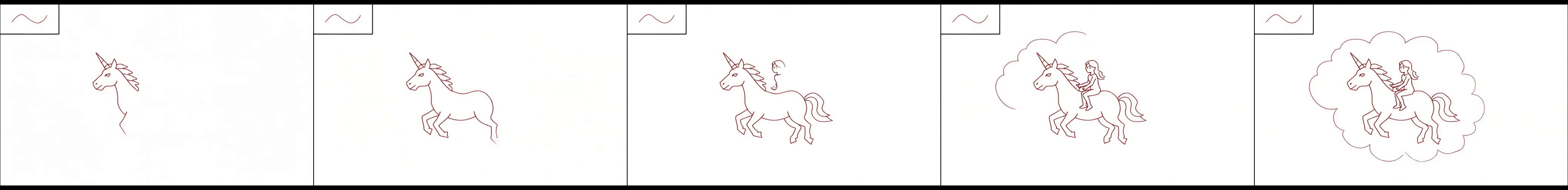}

\vspace{-8pt}
\caption{Brush control results for four unseen brush style and color combinations.}
\label{fig:supp_ours_i2v_unicorn}
\end{figure*}

\begin{figure*}[t]
\centering
\setlength{\tabcolsep}{2pt}
\renewcommand{\arraystretch}{0.2}
\small
\parbox{1\linewidth}{\scriptsize\emph{``Step by step sketch process of an Amsterdam canal, using the color and style of the brush shown in the top-left corner, following this drawing order: 1. Canal water shape running through the scene. 2. Narrow buildings along the canal edge. 3. A bridge arch crossing the water. 4. Bicycles along the railing. 5. A small boat floating in the canal. }} \\[2pt]
\includegraphics[width=\linewidth]{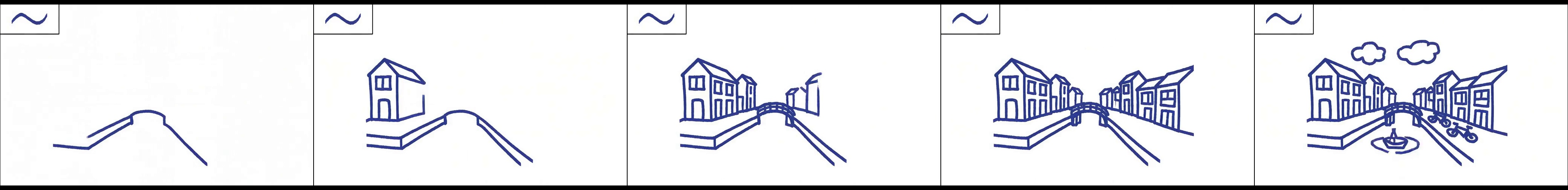} \\[2pt]
\includegraphics[width=\linewidth]{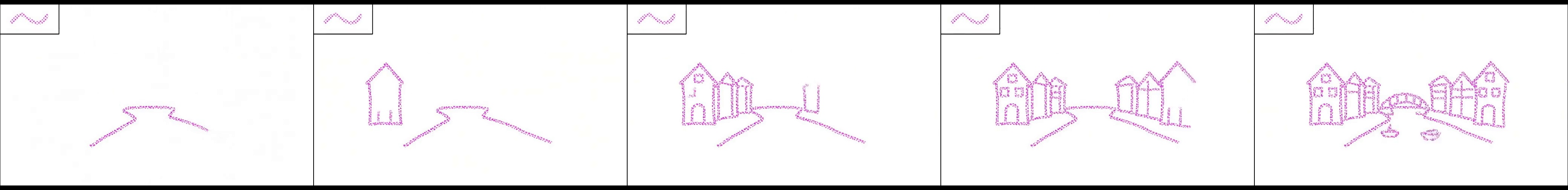} \\[2pt]
\includegraphics[width=\linewidth]{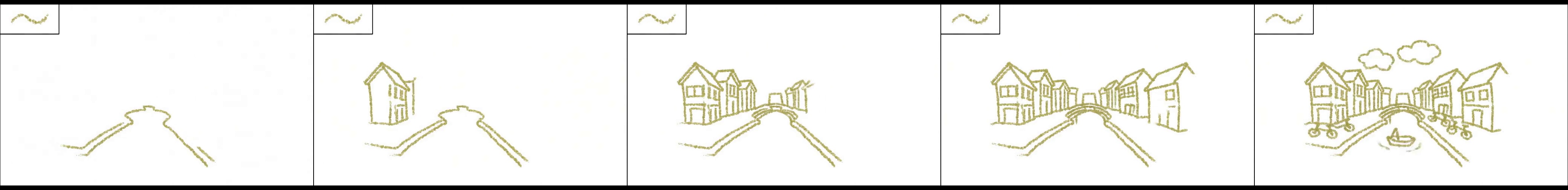} \\[2pt]
\includegraphics[width=\linewidth]{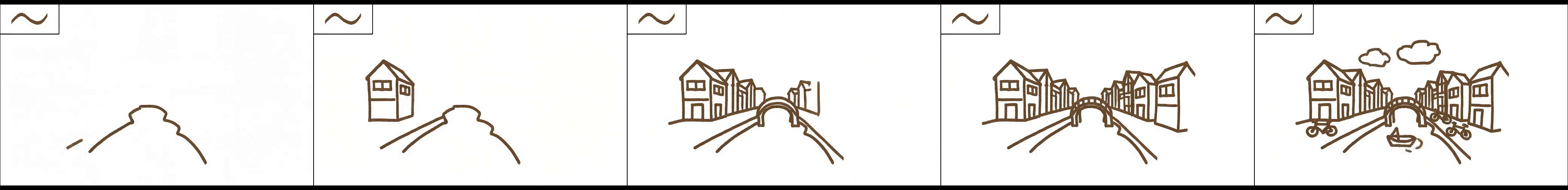} \\[10pt]

\parbox{1\linewidth}{\scriptsize\emph{``Step by step sketch process of a snail carrying a tiny village on its shell, using the color and style of the brush shown in the top-left corner, following this drawing order: 1. Snail body – soft curved shape. 2. Shell – large spiral form. 3. Tiny houses on top of the shell. 4. Small trees and fences around the houses. 5. Antennae and face of the snail. 6. Grass and flowers below.}} \\[2pt]
\includegraphics[width=\linewidth]{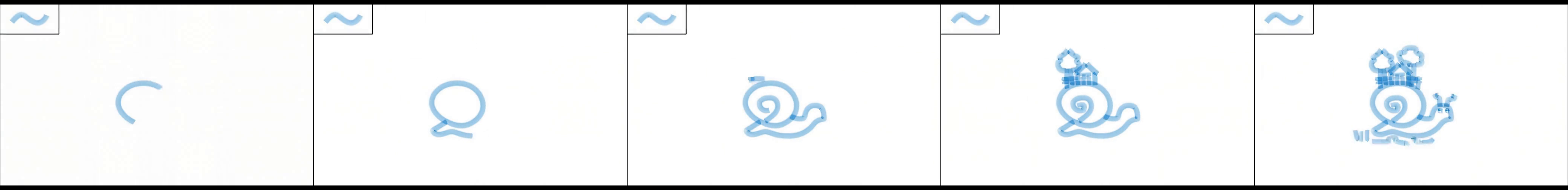} \\[2pt]
\includegraphics[width=\linewidth]{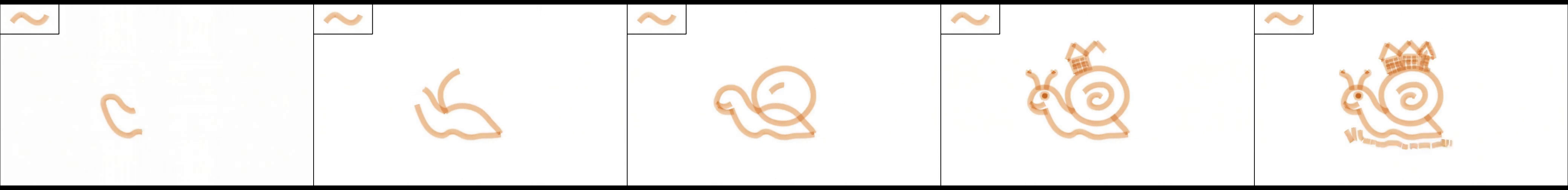} \\[2pt]
\includegraphics[width=\linewidth]{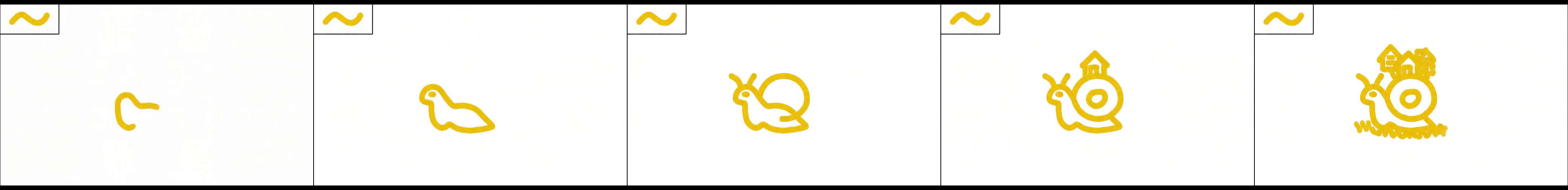} \\[2pt]
\includegraphics[width=\linewidth]{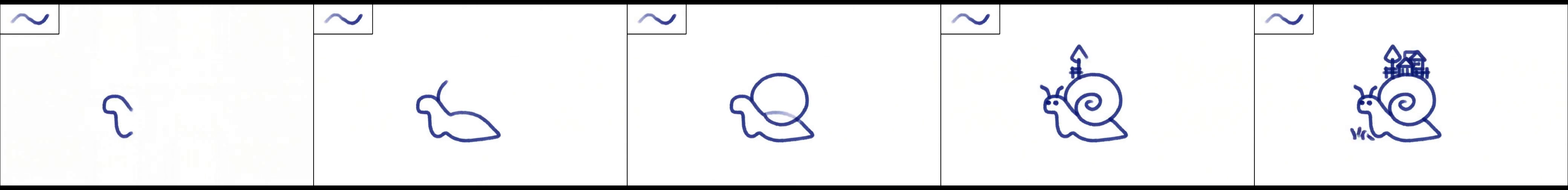}\\[2pt]
\includegraphics[width=\linewidth]{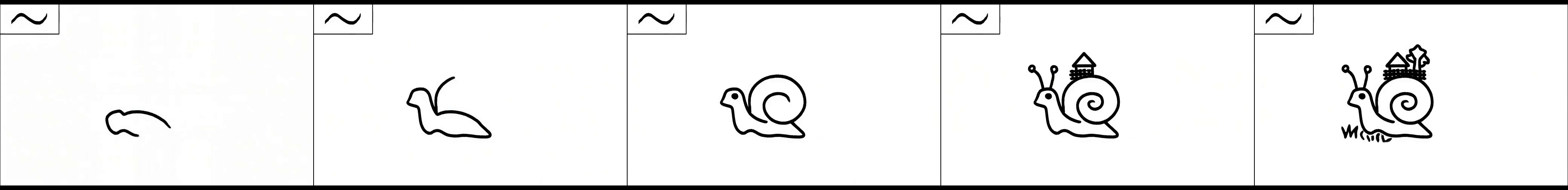}

\vspace{-8pt}
\caption{\textbf{Brush control I2V results for ``Amsterdam canal.''} Results shown for four unseen brush style and color combinations.}
\label{fig:supp_ours_i2v_amsterdam}
\end{figure*}

We show additional brush control results in~\Cref{fig:supp_ours_i2v_unicorn,fig:supp_ours_i2v_amsterdam}.
Conditioning on the first-frame brush exemplar enables faithful transfer of color and texture across the full sequence, including brush styles not seen during training. For example, in the polar bear scene, distinct brush styles (e.g., smooth pastel strokes, thick marker-like lines, or calligraphic brush) are consistently preserved across all drawing steps.
Importantly, while the appearance of strokes varies significantly with the chosen brush, the underlying drawing process and structure faithfully follow the prompt.

\begin{figure*}[t]
\centering
\setlength{\tabcolsep}{2pt}
\renewcommand{\arraystretch}{0.2}
\small

\begin{tabular}{c}

\parbox{1\linewidth}{\scriptsize\emph{``Step by step sketch process of a snail carrying a tiny village on its shell, following this drawing order: 1. Snail body – soft curved shape. 2. Shell – large spiral form. 3. Tiny houses on top of the shell. 4. Small trees and fences around the houses. 5. Antennae and face of the snail. 6. Grass and flowers below.}} \\[4pt]

  \includegraphics[width=1\linewidth]{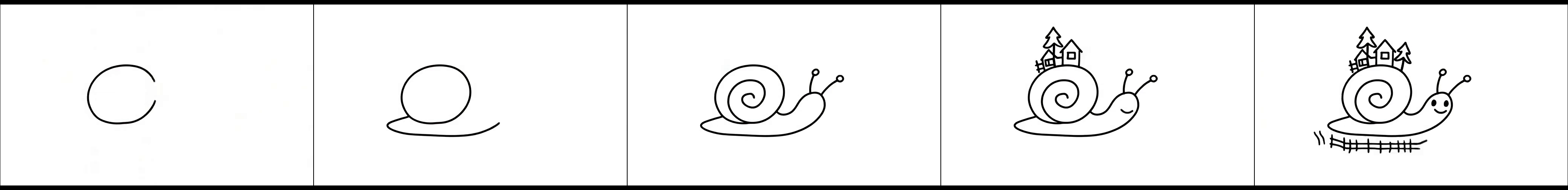} \\[6pt]

\parbox{1\linewidth}{\scriptsize\emph{``Step by step sketch process of a witch cooking soup inside a pumpkin house, following this drawing order: 1. Pumpkin house – large round pumpkin shape. 2. Door and windows carved into it. 3. Witch figure standing outside or inside the doorway. 4. Cauldron with bubbling soup. 5. Broom leaning nearby. 6. Twisted trees and stars around.''}} \\[4pt]
  \includegraphics[width=1\linewidth]{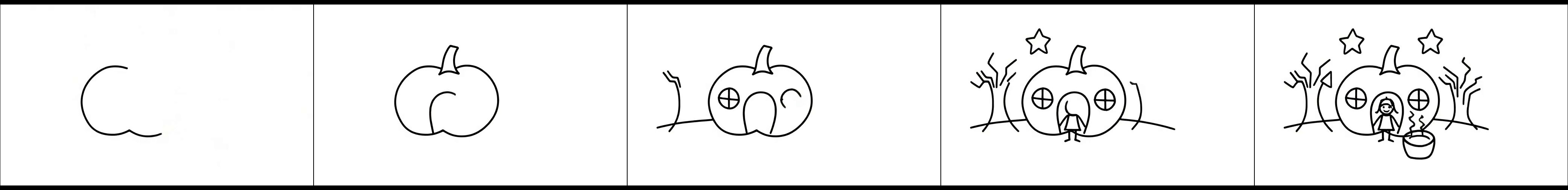} \\[6pt]

\parbox{1\linewidth}{\scriptsize\emph{``Step by step sketch process of a child flying with a bundle of balloons over rooftops, following this drawing order: 1. Child’s body – small floating figure. 2. Head, arms, and legs. 3. Balloon strings in hand. 4. Cluster of large balloons above. 5. Rooftops below. 6. Birds or clouds in the sky.''}} \\[4pt]
  \includegraphics[width=1\linewidth]{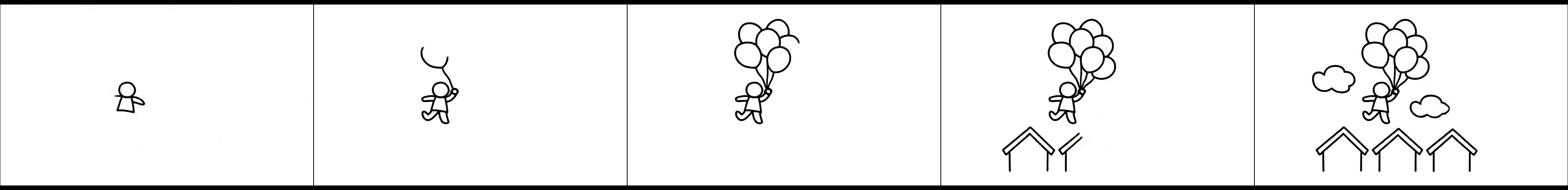} \\[6pt]

\parbox{1\linewidth}{\scriptsize\emph{``Step by step sketch process of an ancient observatory, following this drawing order: 1. Hill silhouette. 2. Observatory dome. 3. Base structure. 4. Telescope outline. 5. Person observing.''}} \\[4pt]

  \includegraphics[width=1\linewidth]{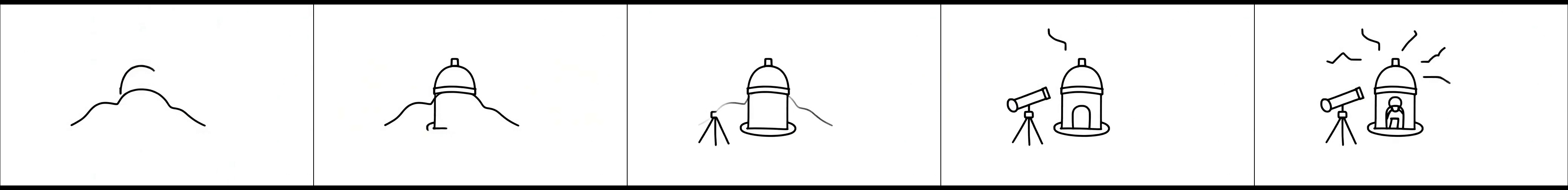} \\[6pt]

\parbox{1\linewidth}{\scriptsize\emph{``Step by step sketch process of a bakery opening at dawn, following this drawing order: 1. Shop front rectangle. 2. Door and window. 3. Counter inside. 4. Baker figure. 5. Bread shapes.''}} \\[4pt]

  \includegraphics[width=1\linewidth]{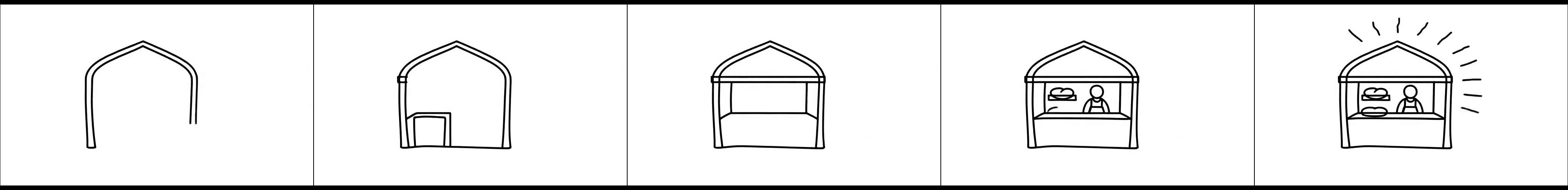} \\[6pt]

\parbox{1\linewidth}{\scriptsize\emph{``Step by step sketch process of a floating city, following this drawing order: 1. Sky horizon. 2. City mass. 3. Supporting structures. 4. Building details. 5. Clouds and airships.''}} \\[4pt]
  \includegraphics[width=1\linewidth]{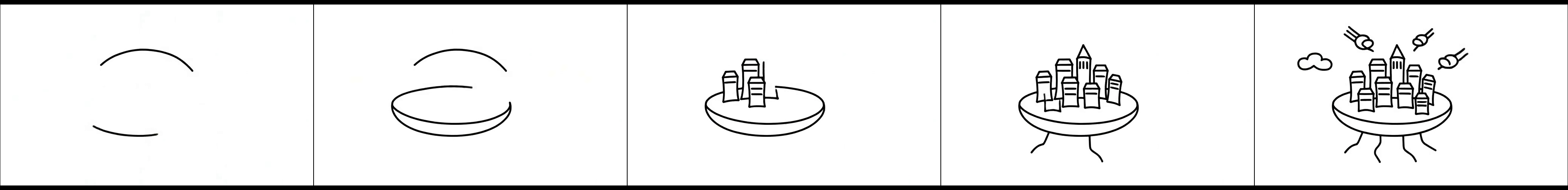} \\[6pt]

  \parbox{1\linewidth}{\scriptsize\emph{``Step by step sketch process of a rocket launch, following this drawing order: 1. Launchpad base. 2. Rocket body. 3. Nose cone and fins. 4. Smoke clouds. 5. Flames and spectators.''}} \\[4pt]
  \includegraphics[width=1\linewidth]{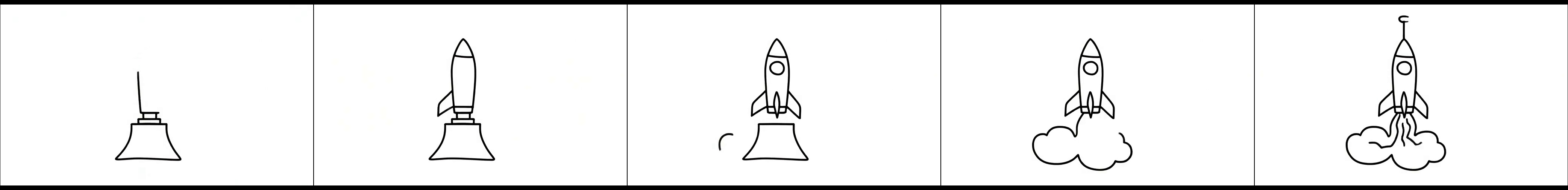} \\[6pt]

\end{tabular}

\vspace{-8pt}
\caption{\textbf{Additional text-to-video results.} Additional qualitative results generated with our text-to-video model.}
\label{fig:supp_ours_t2v_1}
\end{figure*}

We provide additional sketch results of our text-to-video model in \Cref{fig:supp_ours_t2v_1}, covering diverse prompts and instructions. Full videos and additional examples are provided in the supplamntary html.

\begin{figure*}[t]
\centering
\setlength{\tabcolsep}{2pt}
\renewcommand{\arraystretch}{0.2}
\small

\begin{tabular}{c}

\parbox{1\linewidth}{\scriptsize\emph{``Step by step sketch process of a cake, following this drawing order: 1. Base shape -- a wide rectangle or oval. 2. Top layer line to show thickness. 3. Frosting edge with small drips. 4. Side details like stripes or shading lines. 5. Candles on top.''}} \\[4pt]
  \includegraphics[width=1\linewidth]{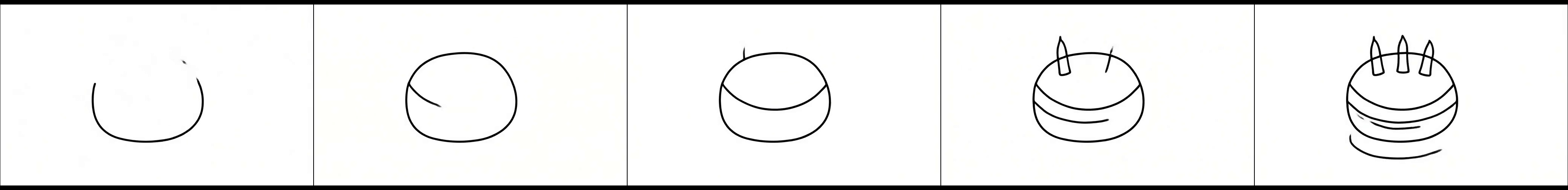} \\[6pt]

\parbox{1\linewidth}{\scriptsize\emph{``Step by step sketch process of pants, following this drawing order: 1. Waistband -- a long curved rectangle. 2. Outer pant legs dropping down. 3. Inner leg seam and the gap between legs. 4. Pockets and zipper area. 5. Cuffs and a few fold lines.''}} \\[4pt]
  \includegraphics[width=1\linewidth]{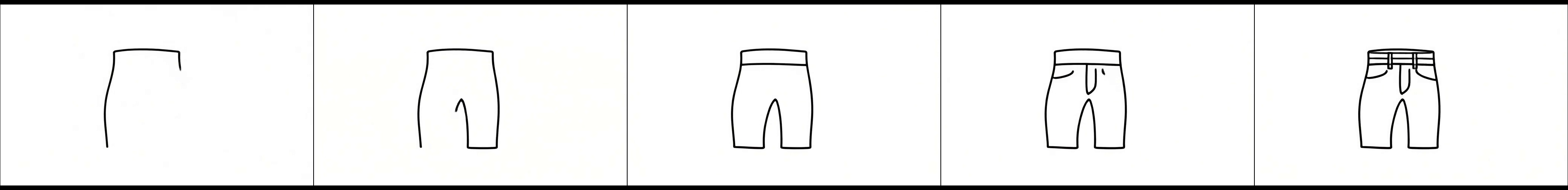} \\[6pt]

\parbox{1\linewidth}{\scriptsize\emph{``Step by step sketch process of a teapot, following this drawing order: 1. Body -- a rounded oval shape. 2. Lid -- a small dome on top. 3. Spout extending from one side. 4. Handle curving from the other side. 5. Base ring and simple decoration lines.''}} \\[4pt]
  \includegraphics[width=1\linewidth]{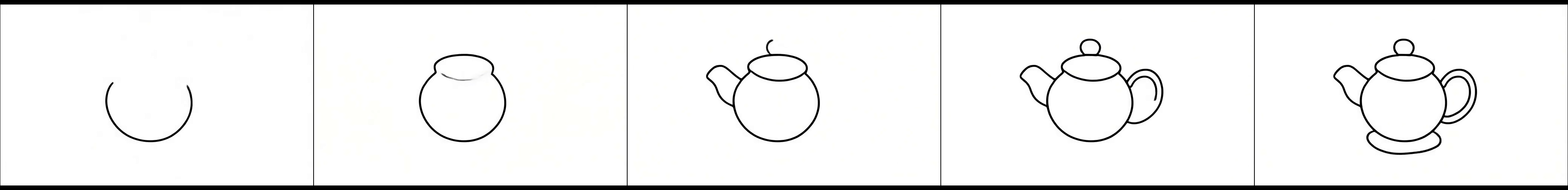} \\[6pt]

\parbox{1\linewidth}{\scriptsize\emph{``Step by step sketch process of a squirrel, following this drawing order: 1. Body -- oval shape. 2. Head -- smaller circle at the front. 3. Tail -- big fluffy curve rising behind the body. 4. Legs and paws -- small bent shapes under the body. 5. Face and ears -- pointed ears, eye, nose, and mouth line. 6. Fur details -- short strokes along the tail and chest.''}} \\[4pt]
  \includegraphics[width=1\linewidth]{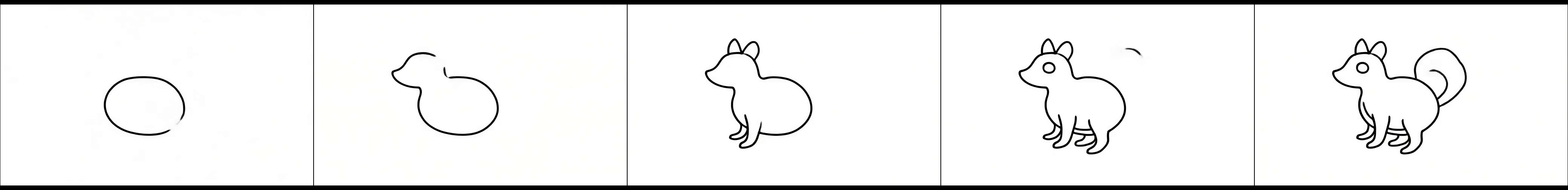} \\[6pt]

\parbox{1\linewidth}{\scriptsize\emph{``Step by step sketch process of a piano, following this drawing order: 1. The main body -- a long rectangle. 2. The top lid -- a slightly angled rectangle above. 3. The keyboard area -- a thin long rectangle. 4. The keys -- repeating short vertical lines. 5. The legs -- three or four straight supports. 6. The pedals -- small shapes under the center.''}} \\[4pt]
  \includegraphics[width=1\linewidth]{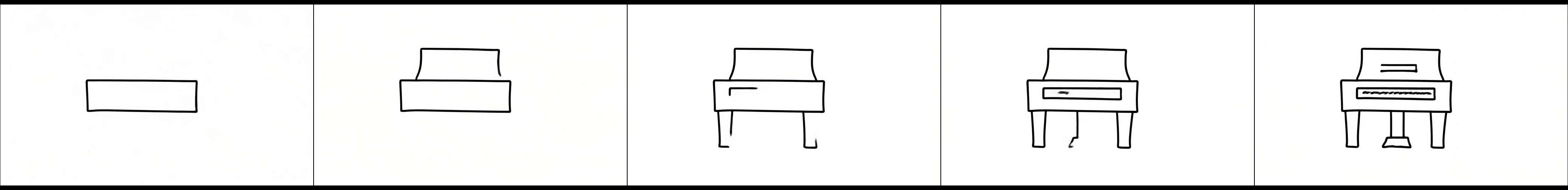} \\[6pt]

\parbox{1\linewidth}{\scriptsize\emph{``Step by step sketch process of a table, following this drawing order: 1. The tabletop -- a wide rectangle. 2. The thickness -- a second line under the tabletop. 3. The front legs -- two vertical lines down. 4. The back legs -- two slightly offset lines down. 5. The cross supports -- short connecting lines between legs.''}} \\[4pt]
  \includegraphics[width=1\linewidth]{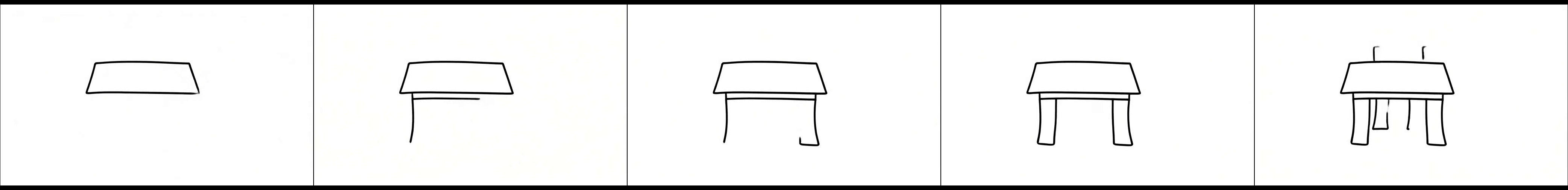} \\[6pt]

\parbox{1\linewidth}{\scriptsize\emph{``Step by step sketch process of a television, following this drawing order: 1. The screen -- a wide rectangle with rounded corners. 2. The frame -- a slightly larger outline around the screen. 3. The stand -- a small base under the TV. 4. The buttons -- tiny circles or rectangles on one side. 5. Screen shine -- a light diagonal highlight line.''}} \\[4pt]
  \includegraphics[width=1\linewidth]{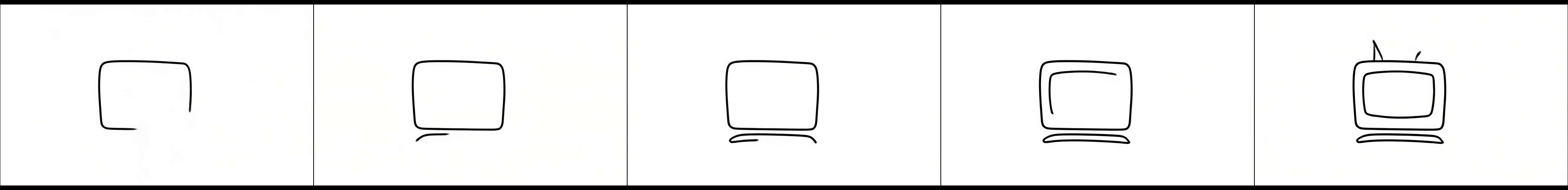}

\end{tabular}

\vspace{-8pt}
\caption{\textbf{Autoregressive results (1/3).} Additional autoregressive model results on QuickDraw prompts.}
\label{fig:supp_ar_quickdraw_merged_0}
\end{figure*}

\begin{figure*}[t]
\centering
\setlength{\tabcolsep}{2pt}
\renewcommand{\arraystretch}{0.2}
\small

\begin{tabular}{c}

\parbox{1\linewidth}{\scriptsize\emph{``Step by step sketch process of a bee, following this drawing order: 1. The body -- an oval with a pointed end. 2. The head -- a small circle attached to the front. 3. The wings -- two teardrop shapes on top. 4. The stripes -- curved bands across the body. 5. The legs -- short bent lines underneath. 6. The antennae -- two small curved lines on the head.''}} \\[4pt]
  \includegraphics[width=1\linewidth]{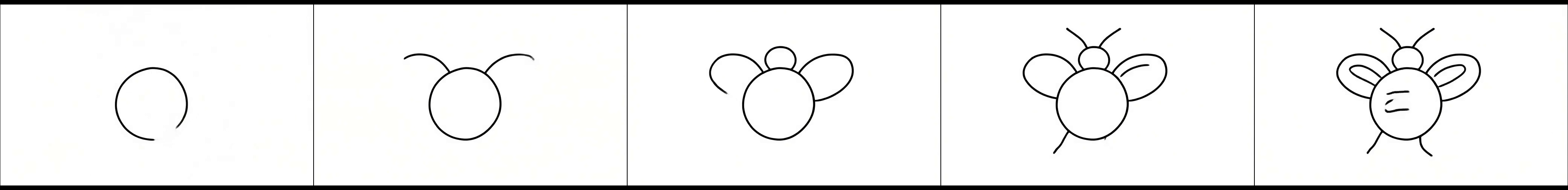} \\[6pt]

\parbox{1\linewidth}{\scriptsize\emph{``Step by step sketch process of a dishwasher, following this drawing order: 1. Main box -- a tall rectangle. 2. Inner opening -- a smaller rectangle inset. 3. Control panel -- a thin strip along the top front. 4. Door edge and handle -- define the door line and add a handle. 5. Racks -- simple horizontal lines inside the opening.''}} \\[4pt]
  \includegraphics[width=1\linewidth]{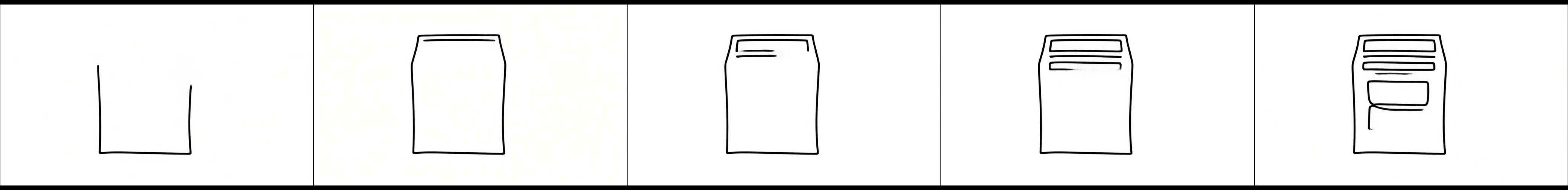} \\[6pt]

\parbox{1\linewidth}{\scriptsize\emph{``Step by step sketch process of a sink, following this drawing order: 1. Basin -- a wide rounded rectangle. 2. Inner bowl -- a smaller oval inside. 3. Drain -- a small circle near the bottom center. 4. Rim and thickness lines. 5. Faucet -- a short spout above the back edge. 6. Handles -- two small knobs beside the faucet.''}} \\[4pt]
  \includegraphics[width=1\linewidth]{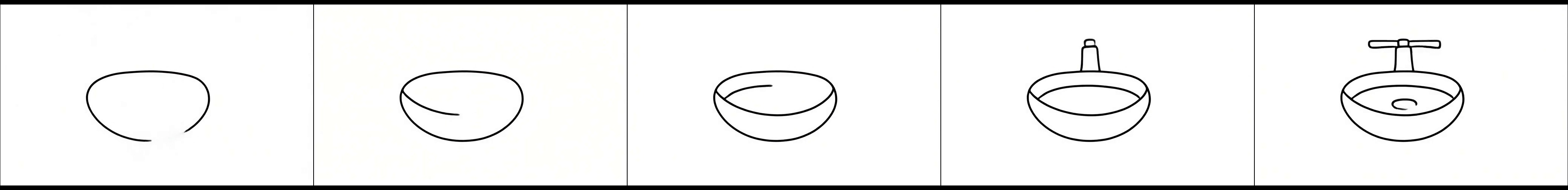} \\[6pt]

\parbox{1\linewidth}{\scriptsize\emph{``Step by step sketch process of a helicopter, following this drawing order: 1. Fuselage -- a long rounded oval. 2. Cockpit window -- a curved shape at the front. 3. Tail boom -- a long narrow extension. 4. Landing skids -- two curved rails below. 5. Main rotor mast and blades on top. 6. Tail rotor -- a small propeller at the tail end.''}} \\[4pt]
  \includegraphics[width=1\linewidth]{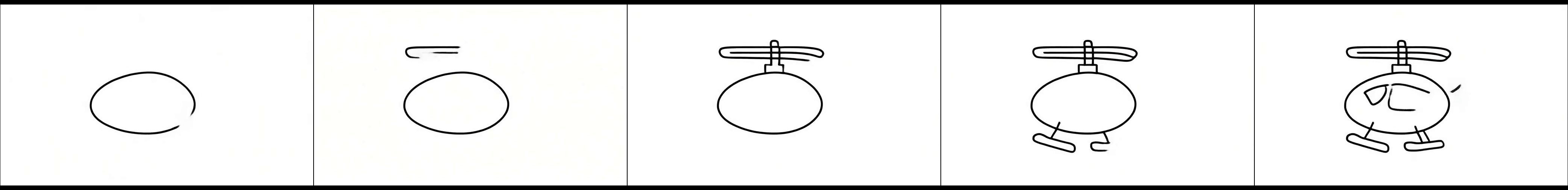} \\[6pt]

\parbox{1\linewidth}{\scriptsize\emph{``Step by step sketch process of a mug, following this drawing order: 1. Cup body -- a tall rounded rectangle. 2. Rim -- an oval at the top. 3. Base curve at the bottom. 4. Handle -- a C-shape on the side. 5. Inner rim line for thickness. 6. Simple shading line on one side.''}} \\[4pt]
  \includegraphics[width=1\linewidth]{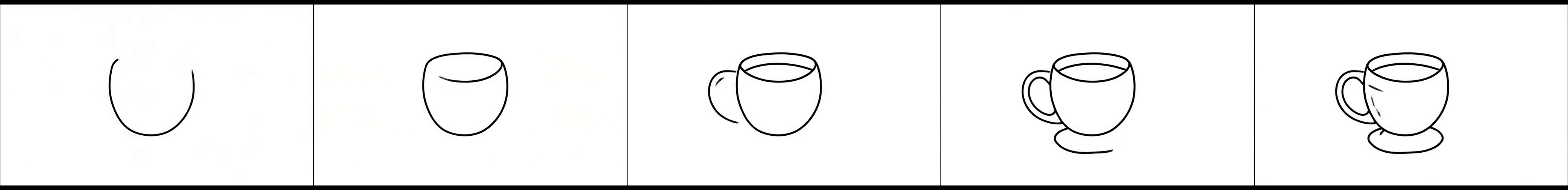} \\[6pt]

\parbox{1\linewidth}{\scriptsize\emph{``Step by step sketch process of a bulldozer, following this drawing order: 1. Tracks -- a long rounded rectangle with two circles inside. 2. Main body -- a boxy shape above the tracks. 3. Cab -- a smaller box on top with a window. 4. Blade -- a wide curved rectangle in front. 5. Arms -- two bars connecting body to blade. 6. Details -- vents, lights, and track treads.''}} \\[4pt]
  \includegraphics[width=1\linewidth]{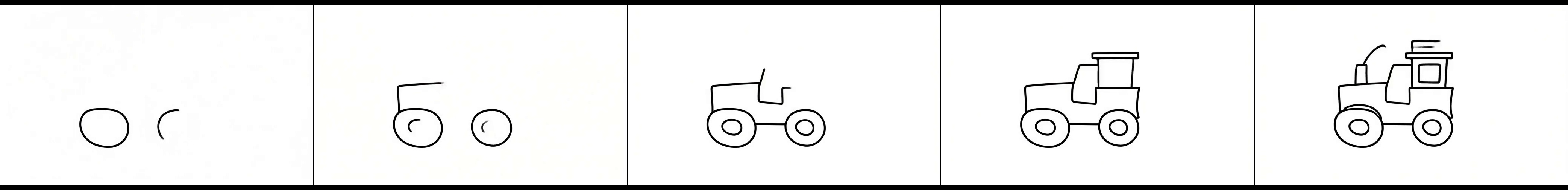} \\[6pt]

\parbox{1\linewidth}{\scriptsize\emph{``Step by step sketch process of a washing machine, following this drawing order: 1. Outer body -- a tall rectangle. 2. Door -- a large circle on the front. 3. Inner drum ring inside the door. 4. Control panel -- a thin rectangle at the top. 5. Buttons and dial shapes. 6. Bottom shadow line and small feet.''}} \\[4pt]
  \includegraphics[width=1\linewidth]{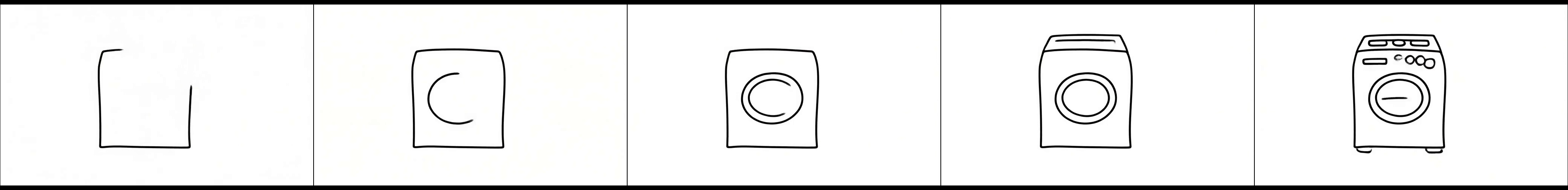}

\end{tabular}

\vspace{-8pt}
\caption{\textbf{Autoregressive results (3/3).} Additional autoregressive model results on QuickDraw prompts.}
\label{fig:supp_ar_quickdraw_merged_2}
\end{figure*}

We include additional sketches of our autoregressive model in~\Cref{fig:supp_ar_quickdraw_merged_2,fig:supp_ar_quickdraw_merged_0}.

\clearpage

\clearpage

\clearpage

\clearpage

\clearpage

\clearpage
\begin{figure*}[t]
\centering
\setlength{\tabcolsep}{2pt}
\renewcommand{\arraystretch}{0.2}
\small

\begin{tabular}{c}

\parbox{1\linewidth}{\scriptsize\emph{``Step by step sketch process of a bowtie, following this drawing order: 1. The center knot -- a small square. 2. The right bow -- a wide triangle/leaf shape attached to the knot. 3. The left bow -- match the shape on the other side. 4. The folds -- a few inner lines pointing toward the knot. 5. The strap ends -- short lines extending behind the knot.''}} \\[4pt]
  \includegraphics[width=1\linewidth]{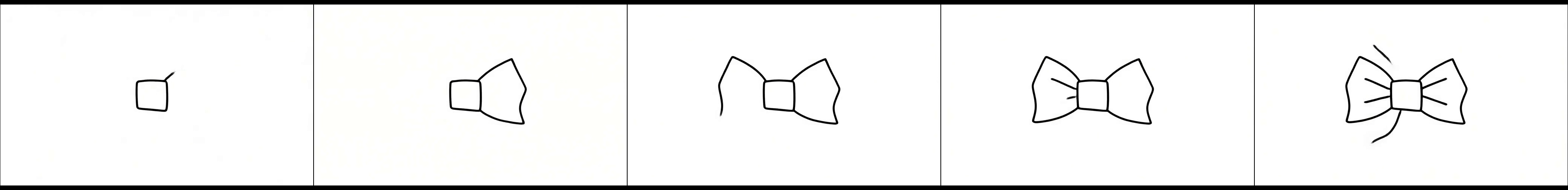} \\[6pt]

\parbox{1\linewidth}{\scriptsize\emph{``Step by step sketch process of a cell phone, following this drawing order: 1. The phone outline -- a tall rounded rectangle. 2. The screen -- an inner rectangle. 3. The top speaker -- a thin slot near the top. 4. The button or home bar -- a small circle or short line at the bottom. 5. The camera -- a tiny circle on the back corner or front top. 6. The side buttons -- small bumps along one edge.''}} \\[4pt]
  \includegraphics[width=1\linewidth]{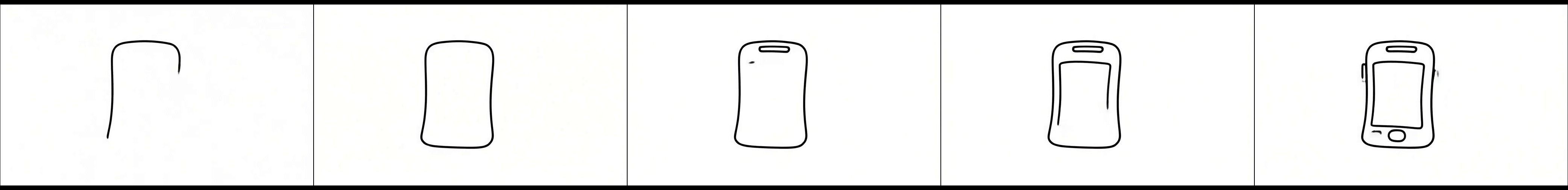} \\[6pt]

\parbox{1\linewidth}{\scriptsize\emph{``Step by step sketch process of an octopus, following this drawing order: 1. The head -- a large rounded dome. 2. The face -- two eyes and a small mouth under them. 3. The front tentacles -- draw two long curved arms flowing downward. 4. The remaining tentacles -- add six more arms around, varying the curves. 5. The suckers -- small circles along the undersides. 6. The texture -- light spots and contour lines on the head.''}} \\[4pt]
  \includegraphics[width=1\linewidth]{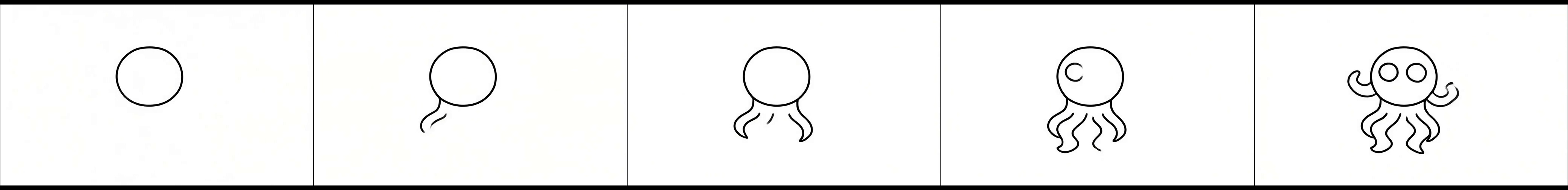} \\[6pt]

\parbox{1\linewidth}{\scriptsize\emph{``Step by step sketch process of a couch, following this drawing order: 1. Draw a long rectangle for the seat. 2. Add a taller rectangle behind for the backrest. 3. Sketch armrests on both sides as rounded blocks. 4. Add cushions with a few curved seam lines. 5. Finish with short legs under the base.''}} \\[4pt]
  \includegraphics[width=1\linewidth]{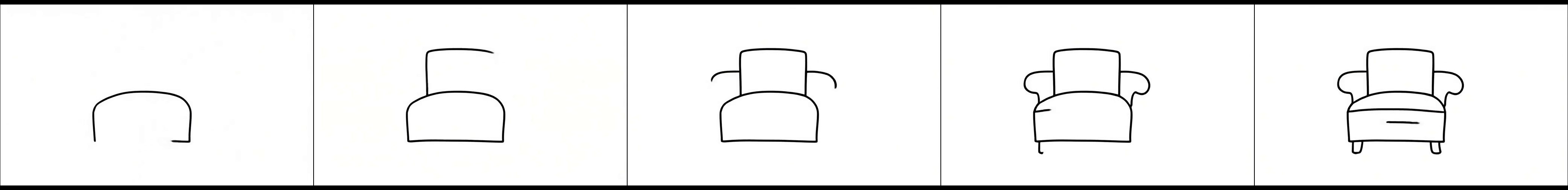} \\[6pt]

\parbox{1\linewidth}{\scriptsize\emph{``Step by step sketch process of a flamingo, following this drawing order: 1. The body -- a large oval. 2. The neck -- a long S-curve rising from the body. 3. The head and beak -- a small circle and a hooked triangle. 4. The legs -- two long thin lines, one bent. 5. The wing -- a curved inner shape with a few feather lines. 6. The feet -- small angled strokes at the ends.''}} \\[4pt]
  \includegraphics[width=1\linewidth]{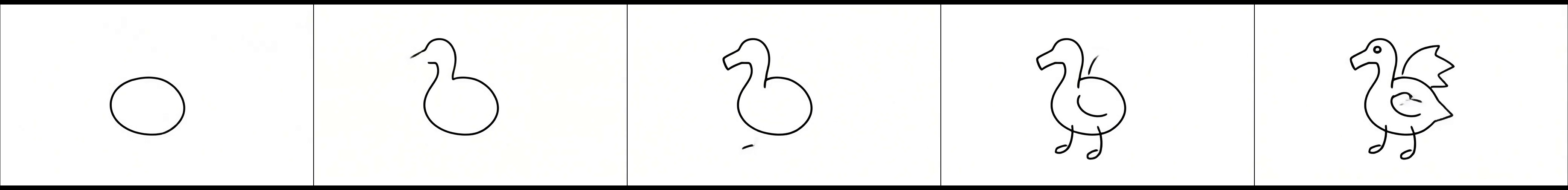} \\[6pt]

\parbox{1\linewidth}{\scriptsize\emph{``Step by step sketch process of a shoe, following this drawing order: 1. The sole -- a long curved base line. 2. The upper outline -- a rounded shape above the sole. 3. The heel and toe cap contours. 4. The lace area -- a small tongue shape and eyelets. 5. The laces -- crisscross lines across the top.''}} \\[4pt]
  \includegraphics[width=1\linewidth]{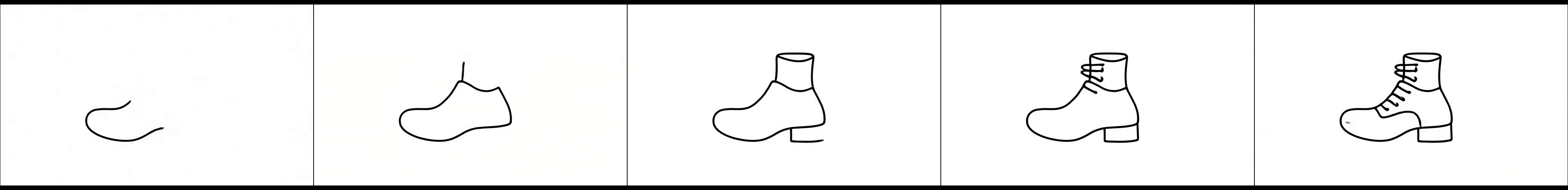} \\[6pt]

\parbox{1\linewidth}{\scriptsize\emph{``Step by step sketch process of a skull, following this drawing order: 1. The cranium -- a large rounded oval. 2. The jaw -- a smaller U-shape attached below. 3. The eye sockets -- two large ovals. 4. The nose cavity -- an upside-down heart or triangle. 5. The teeth -- a row of small vertical rectangles.''}} \\[4pt]
  \includegraphics[width=1\linewidth]{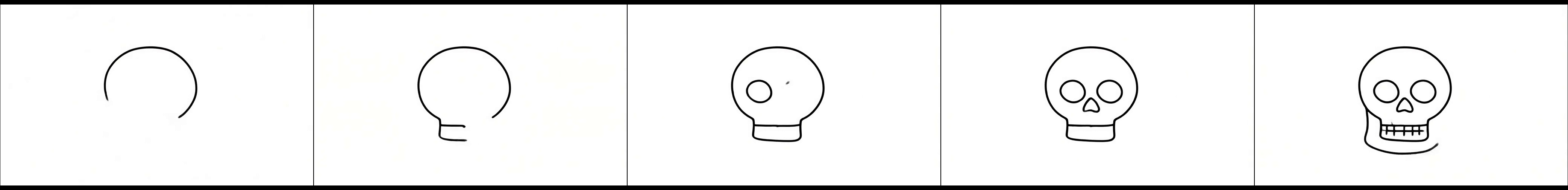}

\end{tabular}

\vspace{-8pt}
\caption{\textbf{Autoregressive results (2/3).} Additional autoregressive model results on QuickDraw prompts.}
\label{fig:supp_ar_quickdraw_merged_1}
\end{figure*}

\clearpage

\end{document}